%% file: main.tex
\documentclass[12pt]{graphicsclass}
\pdfoutput=1
\usepackage[utf8]{inputenc}
\RequirePackage[T1]{fontenc}
\RequirePackage[tt=false, type1=true]{libertine}
\RequirePackage[varqu]{zi4}
\usepackage{nameref}
\usepackage{amssymb,amsmath,bm,amsthm,mathtools}
\usepackage{graphicx}
\graphicspath{ {figures/} }
\usepackage[caption=false,font=scriptsize]{subfig}
\usepackage[ruled,vlined]{algorithm2e}

\usepackage[format=plain,
            labelfont={bf},
            textfont=it]{caption}
\usepackage{setspace}
\usepackage[margin=0.6in]{geometry}
\usepackage{hyperref}
\hypersetup{colorlinks,linkcolor={blue},citecolor={blue},urlcolor={blue}}  
\usepackage[round]{natbib}
\usepackage{xcolor}
\usepackage{overpic}
\usepackage[all]{nowidow}
\usepackage{enumerate}
\usepackage{color,colortbl}
\usepackage{epstopdf}
\usepackage{booktabs, multirow}
\usepackage{enumitem}
\usepackage{mathtools}
\usepackage{cleveref}
\usepackage{nicematrix}
\usepackage[super]{nth}
\usepackage{nicefrac}
\usepackage{wrapfig,sidecap}
\RequirePackage{fancyhdr}
\usepackage{minitoc}
\usepackage[breakable]{tcolorbox}
\usepackage{longtable}
\usepackage{breakurl}

\NewDocumentCommand{\codeword}{v}{%
\texttt{\textcolor{purple}{#1}}%
}

\tcbuselibrary{theorems}
\newtcbtheorem
  []
  {coloreddefinition}
  {Definition}
  {%
    colback=green!5,
    colframe=green!35!black,
    fonttitle=\bfseries,
  }
  {def}

\newtcbtheorem
  []
  {coloredfact}
  {Fact}
  {%
    colback=teal!5,
    colframe=teal!35!black,
    fonttitle=\bfseries,
  }
  {fact}

\newtcbtheorem
  []
  {coloredtheorem}
  {Theorem}
  {%
    colback=blue!5,
    colframe=blue!35!black,
    fonttitle=\bfseries,
  }
  {thm}

\newtcbtheorem
  []
  {coloredroutine}
  {Routine}
  {%
    colback=purple!5,
    colframe=purple!35!black,
    fonttitle=\bfseries,
  }
  {rtn}

\input{definitions}

\title{
{\textit{Brain-Inspired Planning for Better  Generalization in Reinforcement Learning}}\\~\\~\\
{\large Mingde ``Harry'' Zhao, School of Computer Science \\ 
	McGill University, Montr\'eal \\ 
	Nov, 2025 \\~\\~\\
	An \textbf{updated version} of the thesis submitted to McGill University in partial fulfillment of the requirements of the degree of \\~\\ Ph.D. in Computer Science}\\~\\
}

\renewcommand{\thepage}{\roman{page}}

\author{\textcopyright Mingde Zhao, 2025}
\date{}

\begin{document}
\dominitoc
\maketitle

\begin{spacing}{1.2}
\chapter*{Abstract}
\label{cha:engAbstract}
\addcontentsline{toc}{chapter}{\nameref{cha:engAbstract}}
Existing Reinforcement Learning (RL) systems encounter significant challenges when applied to real-world scenarios, primarily due to poor generalization across environments that differ from their training conditions. This thesis explores the direction of enhancing agents' zero-shot systematic generalization abilities by granting RL agents reasoning behaviors that are found to help systematic generalization in the human brain. Inspired by human conscious planning behaviors, we first introduced a top-down attention mechanism, which allows a decision-time planning agent to dynamically focus its reasoning on the most relevant aspects of the environmental state given its instantaneous intentions, a process we call ``spatial abstraction''. This approach significantly improves systematic generalization outside the training tasks. Subsequently, building on spatial abstraction, we developed the \Skipper{} framework to automatically decompose complex tasks into simpler, more manageable sub-tasks. \Skipper{} provides robustness against distributional shifts and efficacy in long-term, compositional planning by focusing on pertinent spatial and temporal elements of the environment. Finally, we identified a common failure mode and safety risk in planning agents that rely on generative models to generate state targets during planning. It is revealed that most agents blindly trust the targets they hallucinate, resulting in delusional planning behaviors. Inspired by how the human brain rejects delusional intentions, we propose learning a feasibility evaluator to enable rejecting hallucinated infeasible targets, which led to significant performance improvements in various kinds of planning agents. Finally, we suggest directions for future research, aimed at achieving general task abstraction and fully enabling abstract planning.
\end{spacing}

\begin{spacing}{1.2}
\tableofcontents
\end{spacing}

\begin{spacing}{1.2}
\input{chapter_0_nomenclature}
\end{spacing}

\newpage

\pagenumbering{arabic} 


\begin{spacing}{1.2}
\input{chapter_1_intro}
\end{spacing}

\setcounter{mtc}{5}
\begin{spacing}{1.2}
\input{chapter_2_basics}

\end{spacing}

\setcounter{mtc}{6}
\begin{spacing}{1.2}
\input{chapter_3_preliminary}
\end{spacing}

\setcounter{mtc}{7}
\begin{spacing}{1.2}
\input{chapter_4_CP}
\end{spacing}

\setcounter{mtc}{8}
\begin{spacing}{1.2}
\input{chapter_5_Skipper_main}
\end{spacing}

\setcounter{mtc}{9}
\begin{spacing}{1.2}
\input{chapter_6_Delusions}
\end{spacing}

\setcounter{mtc}{10}
\begin{spacing}{1.2}
\input{chapter_7_conclusion}
\end{spacing}

\begin{spacing}{1.0}
\bibliography{references}
\bibliographystyle{abbrvnat}
\end{spacing}

\setcounter{mtc}{11}
\begin{spacing}{1.2}
\input{chapter_8_appendices}
\end{spacing}

\end{document}

%% file: definitions.tex
\definecolor{darkgreen}{rgb}{0, 0.5, 0}
\definecolor{orange}{rgb}{1, 0.7, 0}
\definecolor{red}{rgb}{1, 0, 0}
\definecolor{purple}{rgb}{0.5, 0, 0.5}
\definecolor{gray}{rgb}{0.4, 0.4, 0.4}

\newcommand\blue{\textcolor{blue}}
\newcommand\red{\textcolor{red}}
\newcommand\orange{\textcolor{orange}}
\newcommand\darkgreen{\textcolor{darkgreen}}
\newcommand\purple{\textcolor{purple}}
\newcommand\gray{\textcolor{gray}}

\DeclareMathOperator*{\argmax}{argmax}

\newcommand\doubleE{\mathbb{E}}
\newcommand\doubleP{\mathbb{P}}

\newcommand\doubleR{\mathbb{R}}
\newcommand\scriptR{\mathcal{R}}
\newcommand\scriptS{\mathcal{S}}
\newcommand\scriptE{\mathcal{E}}
\newcommand\scriptO{\mathcal{O}}
\newcommand\scriptT{\mathcal{T}}
\newcommand\scriptA{\mathcal{A}}
\newcommand\scriptB{\mathcal{B}}
\newcommand\scriptU{\mathcal{U}}
\newcommand\scriptX{\mathcal{X}}

\newcommand\ie{\textit{i.e.}}
\newcommand\eg{\textit{e.g.}}
\newcommand\vs{\textit{v.s.}}
\newcommand\st{\textit{s.t.}}
\newcommand\wrt{\textit{w.r.t.}}
\newcommand\aka{\textit{a.k.a.}}
\newcommand\etc{\textit{etc.}}

\newcommand\scriptM{\mathcal{M}}
\newcommand\scriptQ{\mathcal{Q}}
\newcommand\scriptG{\mathcal{G}}
\newcommand\scriptL{\mathcal{L}}

\newtheorem*{theorem*}{Theorem}

\newtheorem*{proposition*}{Proposition}

\newtheorem*{corollary*}{Corollary}

\newcommand{\softmax}{{\fontfamily{qcr}\selectfont softmax}}
\newcommand{\MuZero}{{\fontfamily{cmss}\selectfont MuZero}}
\newcommand{\AlphaGo}{{\fontfamily{cmss}\selectfont AlphaGo}}
\newcommand{\PlaNet}{{\fontfamily{cmss}\selectfont PlaNet}}
\newcommand{\DQN}{{\fontfamily{cmss}\selectfont DQN}}
\newcommand{\DDQN}{{\fontfamily{cmss}\selectfont DDQN}}

\usepackage{titlecaps}

\newcommand\decompress{Integrate}

\newcommand\compressor{Selector}
\newcommand\decompressor{Integrator}

\newcommand{\Dyna}{{\fontfamily{cmss}\selectfont D\lowercase{yna}}}
\newcommand{\NOSET}{{\fontfamily{cmss}\selectfont NOSET}}

\newcommand{\Dreamer}{{\fontfamily{cmss}\selectfont Dreamer}}
\newcommand{\UNREAL}{{\fontfamily{cmss}\selectfont UNREAL}}

\newcommand{\Skipper}{{\fontfamily{cmss}\selectfont Skipper}}
\newcommand{\LEAP}{{\fontfamily{cmss}\selectfont LEAP}}
\newcommand{\Director}{{\fontfamily{cmss}\selectfont Director}}
\newcommand{\GSP}{{\fontfamily{cmss}\selectfont GSP}}
\newcommand{\SimPLe}{{\fontfamily{cmss}\selectfont SimPLe}}

\newcommand{\SSMfull}{{\fontfamily{qcr}\selectfont SwordShieldMonster}}
\newcommand{\RDSfull}{{\fontfamily{qcr}\selectfont RandDistShift}}
\newcommand{\SSM}{{\fontfamily{qcr}\selectfont SSM}}
\newcommand{\RDS}{{\fontfamily{qcr}\selectfont RDS}}
\newcommand{\SkipperPlus}{\textcolor[rgb]{1.0, 0.5, 0.0}{\fontfamily{cmss}\selectfont S\lowercase{kipper+}}}
\newcommand{\DynaPlus}{\textcolor[rgb]{1.0, 0.5, 0.0}{\fontfamily{cmss}\selectfont D\lowercase{yna+}}}

\newcommand{\futurestr}{{\textsc{\fontfamily{LinuxLibertineT-LF}\selectfont \textcolor[rgb]{0, 0.7, 0}{\underline{future}}}}}

\newcommand{\episodestr}{{\textsc{\fontfamily{LinuxLibertineT-LF}\selectfont \textcolor[rgb]{0, 0, 0.7}{\underline{episode}}}}}

\newcommand{\pertaskstr}{{\textsc{\fontfamily{LinuxLibertineT-LF}\selectfont \textcolor[rgb]{0.7, 0.0, 0.0}{\underline{pertask}}}}}
\newcommand{\generatestr}{{\textsc{\fontfamily{LinuxLibertineT-LF}\selectfont \textcolor[rgb]{0.4, 0.4, 0.4}{\underline{generate}}}}}

\newcounter{Gdelusions}
\newcommand{\gdelusion}[1]{\refstepcounter{Gdelusions}\label{#1}}
\newcommand{\Gzero}[0]{G.\ref{delusion:g0}}
\newcommand{\Gone}[0]{G.\ref{delusion:g1}}
\newcommand{\Gtwo}[0]{G.\ref{delusion:g2}}

\newcounter{Edelusions}
\newcommand{\edelusion}[1]{\refstepcounter{Edelusions}\label{#1}}
\newcommand{\Ezero}[0]{E.\ref{delusion:e0}}
\newcommand{\Eone}[0]{E.\ref{delusion:e1}}
\newcommand{\Etwo}[0]{E.\ref{delusion:e2}}

\theoremstyle{plain}
\newtheorem{result}{Result}[section]

\newcommand{\FE}[0]{\textcolor[rgb]{0, 0, 0.7}{\underline{E}}}
\newcommand{\FP}[0]{\textcolor[rgb]{0.7, 0.0, 0.0}{\underline{P}}}

\newcommand{\FEG}[0]{\textcolor[rgb]{0.0, 0.5, 1.0}{\underline{(E+G)}}}
\newcommand{\FEP}[0]{\textcolor[rgb]{0.8, 0.0, 0.8}{\underline{(E+P)}}}
\newcommand{\FEPG}[0]{\textcolor[rgb]{1.0, 0.5, 0.0}{\underline{(E+P+G)}}}



%% file: chapter_0_nomenclature.tex
\chapter*{Nomenclature \& Indices}
\label{cha:nomenclature}
\addcontentsline{toc}{chapter}{\nameref{cha:nomenclature}}

\section*{Terminologies \& Abbreviations}
\label{cha:nomen_terminologies}

For the readers' convenience in cross-referencing, the rows are sorted alphabetically and some terms defined in the following list are highlighted in \textbf{bold}.

\addcontentsline{toc}{section}{\nameref{cha:nomen_terminologies}}

\begin{longtable*}{m{0.2\textwidth} m{0.8\textwidth}}
\textbf{Agent} & \parbox[t]{0.8\textwidth}{An entity that makes decisions and takes actions in an environment. Often refers to the computational methods.} \\

\textbf{Actor-Critic} & \parbox[t]{0.8\textwidth}{A reinforcement learning algorithm with two components: one for taking action (actor) and one for value estimation (critic). See Sec.~\ref{sec:actor_critic} on Page.~\pageref{sec:actor_critic}.} \\

\textbf{Attention} & \parbox[t]{0.8\textwidth}{A mechanism allowing computations to focus on parts of the input. See Sec.~\ref{sec:attention} on Page.~\pageref{sec:attention}.} \\

\textbf{Auxiliary} & \parbox[t]{0.8\textwidth}{Additional tasks or information used to assist a model. We use the term ``auxiliary learners'' to describe components in a reinforcement learning agent that do not conduct \textbf{value estimation}. See Sec.~\ref{sec:RL_agent_components} on Page.~\pageref{sec:RL_agent_components} and Sec.~\ref{sec:auxiliary_learners} on Page.~\pageref{sec:auxiliary_learners}.} \\

\textbf{Background Planning} & \parbox[t]{0.8\textwidth}{A planning behavior that does not immediately improve an agent's next decision \citep{alver2022understanding}.} \\

\textbf{Baseline} & \parbox[t]{0.8\textwidth}{An existing method used as a reference point for the performance of new methods. When a new agent is proposed, existing agents can serve as baselines. When a variant of an agent is proposed, the agent (in its original form) should serve as the baseline.} \\

\textbf{Behavior Policy} & \parbox[t]{0.8\textwidth}{An agent's adopted policy when interacting with the environment. Introduced in  Sec.~\ref{sec:off_policy_evaluation} on Page.~\pageref{sec:off_policy_evaluation} to differentiate with the target policy.} \\

\textbf{Bottleneck} & \parbox[t]{0.8\textwidth}{A point in a system that limits overall performance. Chap.~\ref{cha:CP} contributes a bottleneck mechanism limiting the number of objects that the model could \textbf{reason} with at decision time (Chap.~\ref{cha:CP} on Page.~\pageref{cha:CP}).} \\

\textbf{Checkpoints} & \parbox[t]{0.8\textwidth}{Saved snapshots of model parameters, agent or environment states, containing complete information with which the previous behaviors could be resumed. Chap.~\ref{cha:skipper} uses checkpoints to differentiate \textit{the subset of states imagined by the generative model} from regular states, while also emphasizing their info-completeness (Sec.~\ref{sec:skipper_proxy_problem}, Page.~\pageref{sec:skipper_proxy_problem}).} \\

\textbf{Consciousness} & \parbox[t]{0.8\textwidth}{A capability that refers to agents that can act with awareness of the states of the world and the states of the self.} \\

\textbf{Consciousness in the \nth{1} Sense (C1)} & \parbox[t]{0.8\textwidth}{A capability allowing humans to act with awareness of the relevant environmental entities \citep{dehane2017consciousness}. C2 refers to the awareness of the self.} \\

\textbf{Decision Point} & \parbox[t]{0.8\textwidth}{A specific state or point in time where an agent needs to make a decision.} \\

\textbf{Decision-Time} & \parbox[t]{0.8\textwidth}{The moment when an agent must make a decision about how to act next} \\

\textbf{Decision-Time Planning} & \parbox[t]{0.8\textwidth}{The behavior of planning at \textbf{decision-time} to reason about the decision that is to be made imminently \citep{alver2022understanding}.} \\

\textbf{Deep Learning} & \parbox[t]{0.8\textwidth}{A subset of machine learning using more-than-one layer of artificial neural networks to establish learning systems \citep{goodfellow2016deep}.} \\

\textbf{Delusions} & \parbox[t]{0.8\textwidth}{Obviously false ideas / beliefs which an agent is unable to reject. Reflects a pathology in the learning system \citep{kiran2009understanding}.} \\

\textbf{Delusional Planning Behaviors} & \parbox[t]{0.8\textwidth}{Behaviors triggered by an agent's delusions. Describes agents' often erratic actions while seeking to achieve infeasible targets (Chap.~\ref{cha:delusions}, Page.~\pageref{cha:delusions}).} \\

\textbf{Discrete Action Space} & \parbox[t]{0.8\textwidth}{A set of distinct, countable actions decision-making agents can choose from.} \\

\textbf{Dynamics} & \parbox[t]{0.8\textwidth}{The rules and processes governing the state transitions of an environment.} \\

\textbf{Episode} & \parbox[t]{0.8\textwidth}{A single run or sequence of interactions between an agent and an environment, from start to finish. Often marked with an initial and a terminal state.} \\

\textbf{Environment} & \parbox[t]{0.8\textwidth}{The external world with which decision-making agents interact and draw observations, perhaps rewards. See Fig.~\ref{fig:agent_environment_interaction} on Page.~\pageref{fig:agent_environment_interaction}.} \\

\textbf{Estimator} & \parbox[t]{0.8\textwidth}{A model or method used to estimate certain quantities, like \textbf{value estimators} that estimate \textbf{value functions} in reinforcement learning.} \\

\textbf{Evaluator} & \parbox[t]{0.8\textwidth}{In Chap.~\ref{cha:delusions}, this refers to an estimator used to produce an estimate indicating whether a target is feasible, named after the belief evaluation system in the human brain \citep{kiran2009understanding}. See Sec.~\ref{sec:delusions_intro} on Page.~\pageref{sec:delusions_intro}.} \\

\textbf{Generator} & \parbox[t]{0.8\textwidth}{A model or process that produces data. In Chap.~\ref{cha:delusions}, it refers to the component that proposes targets for the agent to reason with, named after the brain's belief formation system \citep{kiran2009understanding}. See Sec.~\ref{sec:delusions_intro} on Page.~\pageref{sec:delusions_intro}.} \\

\textbf{Goal} & \parbox[t]{0.8\textwidth}{A desired outcome an agent seeks to achieve. Discussed in Chap.~\ref{cha:basics} on Page.~\pageref{cha:basics}.} \\

\textbf{Gridworld} & \parbox[t]{0.8\textwidth}{A grid-based environment used for evaluating agents. Agents move from one cell to another by taking navigation actions, and possibly interacts with objects located in certain cells.} \\

\textbf{Hallucination} & \parbox[t]{0.8\textwidth}{The behavior of generating nonfactual beliefs.} \\

\textbf{Intuition} & \parbox[t]{0.8\textwidth}{Describes \textbf{estimators}' inexplicable knowledge, also the ability to make fast predictions without reasoning. Correspond to System-1 \citep{daniel2017thinking}.} \\

\textbf{KL-Divergence} & \parbox[t]{0.8\textwidth}{The Kullback–Leibler (KL) divergence, \aka{} relative entropy, is a measure of the difference between two probability distributions. Formally, it is defined as: $D_{\mathrm{KL}}(P \| Q) = \sum_{x} P(x) \log \frac{P(x)}{Q(x)}$. It can be used as a loss function to make the distributional output of a neural network closer to its \textbf{update targets}.} \\

\textbf{Loss Function} & \parbox[t]{0.8\textwidth}{A function used to measure the quality of the output of a model, which can then  guide model training.} \\

\textbf{Lookup Table} & \parbox[t]{0.8\textwidth}{A table used for quick retrieval of precomputed values.} \\

\textbf{Macro Actions} & \parbox[t]{0.8\textwidth}{High-level actions that are sequences of smaller, simpler actions. See Sec.~\ref{sec:macro_actions} on Page.~\pageref{sec:macro_actions}.} \\

\textbf{Model-Based RL} & \parbox[t]{0.8\textwidth}{A type of reinforcement learning assisted by predictive / generative models. See Sec.~\ref{sec:model_basedL_drl} on Page.~\pageref{sec:model_basedL_drl}.} \\

\textbf{Optimizer} & \parbox[t]{0.8\textwidth}{In the context of training parametric models with gradient-based methods, an algorithm used to optimize an objective function.} \\

\textbf{Option} & \parbox[t]{0.8\textwidth}{A structured temporally extended action, sometimes representing a \textbf{macro-action}. See Sec.~\ref{sec:options} on Page.~\pageref{sec:options}.} \\

\textbf{Oracle} & \parbox[t]{0.8\textwidth}{A model established over ground truths that learning agents should no access to.} \\

\textbf{Planning} & \parbox[t]{0.8\textwidth}{The reasoning process of deciding what actions to take in the relative future. See Sec.~\ref{sec:planning} on Page.~\pageref{sec:planning}.} \\

\textbf{Policy} & \parbox[t]{0.8\textwidth}{The mechanism that defines how an agent would act given different environmental states. See Sec.~\ref{sec:policies_values} on Page.~\pageref{sec:policies_values}.} \\

\textbf{Priority} & \parbox[t]{0.8\textwidth}{An ordering determining which work is to be done first. Tree search algorithms use priority queues to determine which branch of the search tree is to be simulated next. See Sec.~\ref{sec:MPC} on Page.~\pageref{sec:MPC}.} \\

\textbf{Proxy Problem} & \parbox[t]{0.8\textwidth}{A simplified version of a problem used as a proxy of the full problem. Chap.~\ref{cha:skipper} uses proxy problems as a way to decompose Markov decision processes. See Sec.~\ref{sec:skipper_proxy_problem} on Page.~\pageref{sec:skipper_proxy_problem}.} \\

\textbf{Reasoning} & \parbox[t]{0.8\textwidth}{The ability of an AI system to infer new information based on available knowledge. \textbf{Planning} is a form of reasoning.} \\

\textbf{Regularization} & \parbox[t]{0.8\textwidth}{Techniques used in machine learning to prevent overfitting \citep{goodfellow2016deep}.} \\

\textbf{Scalability} & \parbox[t]{0.8\textwidth}{The ability of a model or algorithm to handle increased amounts of data or complexity effectively. In Chap.~\ref{cha:skipper}, we tested the agents' generalization abilities with increasing numbers of training task instances. See Sec.~\ref{sec:skipper_exp} on Page.~\pageref{sec:skipper_exp}.} \\

\textbf{Semi-Hard Attention} & \parbox[t]{0.8\textwidth}{A variant of \textbf{attention} mechanism, where the attention weights are continuous but can take the discrete value $0$ to fully rule out some choices.  See Sec.~\ref{sec:semihard_attention} on Page.~\pageref{sec:semihard_attention}.} \\

\textbf{Source-Target Pairs} & \parbox[t]{0.8\textwidth}{Pairs of data where one element is used as the source (input) and the other as the target (desired output) in learning. In Chap.~\ref{cha:delusions}, we use \textbf{source-target pairs} to differentiate how \textbf{TAP agents} learn based on two decision points, to learn the relationship between a source state and a target. See Sec.~\ref{sec:source_target_pair_hindsight_relabeling} on Page.~\pageref{sec:source_target_pair_hindsight_relabeling} and Chap.~\ref{cha:delusions} on Page.~\pageref{cha:delusions}.} \\

\textbf{Spatial Abstraction} & \parbox[t]{0.8\textwidth}{A computational description of \textbf{consciousness in the \nth{1} sense}. Refers to an agent's ability to focus on partial aspects of the state for decision-making. Spatial abstraction is a special, dynamic, intention-dependent form of state abstraction. Named intuitively to differentiate from generic state abstraction and to rhyme with temporal abstraction.} \\

\textbf{State Representation} & \parbox[t]{0.8\textwidth}{An encoding of the state of the \textbf{environment}, used by agents' policies for decision-making. See Sec.~\ref{sec:state_representations} on Page.~\pageref{sec:state_representations}.} \\

\textbf{System-2} & \parbox[t]{0.8\textwidth}{A mode of computation in the human brain that involves deliberate, slow, and logical \textbf{reasoning} on-the-fly, as opposed to \textbf{intuition}-based System-1 thinking \citep{daniel2017thinking}.} \\

\textbf{Target} & \parbox[t]{0.8\textwidth}{An abbreviation of a ``state target''. Chap.~\ref{cha:delusions} uses targets to refer to the observations or states or sets of states generated by the model of a planning agent that can be used to guide its behaviors. Targets correspond to a set of states that agents seek to achieve. Note that a \textbf{target state} instead corresponds to a singleton target, \ie{}, a target of a single state. See Sec.~\ref{sec:source_target_pair_hindsight_relabeling} on Page.~\pageref{sec:source_target_pair_hindsight_relabeling} and Chap.~\ref{cha:delusions} on Page.~\pageref{cha:delusions}.} \\

\textbf{Target-Assisted Planning (TAP)} & \parbox[t]{0.8\textwidth}{Target-Assisted Planning (TAP) is a methodology of decision-making agents that generate \textbf{targets} during planning. See Sec.~\ref{sec:source_target_pair_hindsight_relabeling} on Page.~\pageref{sec:source_target_pair_hindsight_relabeling} and Chap.~\ref{cha:delusions} on Page.~\pageref{cha:delusions}.} \\

\textbf{Top-$k$} & \parbox[t]{0.8\textwidth}{A technique often used in machine learning to focus on the top $k$ results of a search or model, based on certain criteria. A top-$k$ mechanism is used to implement \textbf{semi-hard attention}. See Sec.~\ref{sec:semihard_attention} on Page.~\pageref{sec:semihard_attention}.} \\

\textbf{Transition} & \parbox[t]{0.8\textwidth}{A change in the environment's state due to an agent's action. Specifically in reinforcement learning, it refers to a data structure resembling $\langle s, a, r, s' \rangle$, where $s$ and $s'$ are two consecutive environmental states, where the transition between them is triggered by the agent taking action $a$, and  $r$ is the immediate reward received. See Sec.~\ref{sec:MDP} on Page.~\pageref{sec:MDP}.} \\

\textbf{Trajectory} & \parbox[t]{0.8\textwidth}{A sequence of states, actions, and environmental feedbacks an agent experiences during an episode. See Sec.~\ref{sec:MDP} on Page.~\pageref{sec:MDP}.} \\

\textbf{Update Target} & \parbox[t]{0.8\textwidth}{The target towards which an update of an estimate is made. For example, in Sec.~\ref{sec:TD_learning} (Page.~\pageref{sec:TD_learning}), we discussed how temporal difference learning constructs its update targets.}\\

\textbf{Value} & \parbox[t]{0.8\textwidth}{A measure of the desirability of a state or state-action pair in reinforcement learning, defined as expected return. See Sec.~\ref{sec:policies_values} on Page.~\pageref{sec:policies_values}.} \\

\textbf{Value Estimate} & \parbox[t]{0.8\textwidth}{An approximation of the \textbf{value}, learned by a \textbf{value estimator}. See Sec.~\ref{sec:RL_agent_components} on Page.~\pageref{sec:RL_agent_components}.} \\

\textbf{Value Function} & \parbox[t]{0.8\textwidth}{A function that outputs the \textbf{value} of a given state in RL. Defined in Sec.~\ref{sec:policies_values} on Page.~\pageref{sec:policies_values}.} \\

\textbf{Vanilla} & \parbox[t]{0.8\textwidth}{A term often used to describe basic versions of ice-creams and algorithms.} \\

\textbf{World Model} & \parbox[t]{0.8\textwidth}{A model that represents the agent's understanding of the environment, used for planning and decision-making. Specifically, in Chap.~\ref{cha:CP}, it is used to refer to a model-based planning methodology proposed in \citet{ha2018world}, which trains a model of the environment independently of any rewards, through an unsupervised exploration stage. See Sec.~\ref{sec:world_models} on Page.~\pageref{sec:world_models}.} \\

\end{longtable*}

\listoffigures %
\addcontentsline{toc}{section}{\listfigurename}
\listoftables
\addcontentsline{toc}{section}{\listtablename}

%% file: chapter_1_intro.tex
\chapter{Introduction}
\label{sec:intro}
Reinforcement Learning (RL) is a methodology for learning through trial-and-error, aiming to reinforce behaviors that yield long-term benefits in sequential decision-making scenarios. Recent advances have been driven by integrating RL with artificial Neural Networks (NNs), leading to successes like mastering strategic games such as Chess and Go \citep{silver2016mastering}, achieving superhuman performance in pixel-based Atari games \citep{schrittwieser2019mastering}, etc. However, current RL systems still struggle when deployed in real-world contexts, primarily due to challenges in generalizing their learned capabilities to environments different from those in which they were trained \citep{igl2019generalization,quinonero2022dataset}. This ``generalization gap'' significantly restricts both the application and academic interest in RL \citep{ada2024diffusion}.

Recent studies suggest that this limitation stems from RL agents' inadequate reasoning capabilities when dealing with Out-Of-Distribution (OOD) scenarios \citep{alver2022understanding,langosco2022goal}. Many agents rely solely on intuition-based decision-making (akin to system-1 thinking from \citet{daniel2017thinking}) (including both model-free and background planning model-based methods), or cannot make necessary longer-term plans \citep{daniel2017thinking,zhao2024consciousness}. These insights have spurred the development of agents capable of adaptive reasoning in novel situations, which is the central theme of this thesis.

\section{Thesis Overview}

This thesis explores enhancing generalization in RL agents by granting them reasoning capabilities inspired by human higher-level cognitive functions. The structure of this thesis is as follows:

\textbf{Part I: Literature Review on Background Knowledge}:

\begin{itemize}[leftmargin=*]
\item
In Chap.~\ref{cha:basics} (Page.~\pageref{cha:basics}), I provide an overview of the literature and introduce foundational concepts of reinforcement learning that are pertinent to the subsequent chapters.
\item
In Chap.~\ref{cha:prelim} (Page.~\pageref{cha:prelim}), I discuss more advanced concepts, covering deep reinforcement learning, generative modeling with deep learning, attention mechanisms, model-based reinforcement learning, and related experimental methodologies. This chapter prepares readers who have a basic understanding of reinforcement learning for the content in Part II, by establishing necessary background knowledge.
\end{itemize}

\textbf{Part II: Methodology \& Original Research Findings}:

\begin{itemize}[leftmargin=*]
\item
Chap.~\ref{cha:CP} (Page.~\pageref{cha:CP}), inspired by the OOD generalization abilities facilitated by consciousness, introduces bottleneck mechanism, allowing a decision-time planning agent to dynamically focus its reasoning on the relevant aspects of the state based on its instantaneous intent. This bottleneck mechanism, which achieves ``spatial abstraction'', enables significant Out-Of-Distribution (OOD) systematic generalization abilities.
\item
Chap.~\ref{cha:skipper} (Page.~\pageref{cha:skipper}), inspired by spatial and temporal abstraction abilities in human planning, builds upon the bottleneck mechanism and proposes a framework named \Skipper{} that automatically decomposes an overall given task into smaller and more manageable steps. This framework shows potential to be robust in distributional shifts and compositional long-term planning, by focusing attention on relevant parts of the environment (spatial) and aspects of the future (temporal);
\item
Chap.~\ref{cha:delusions} (Page.~\pageref{cha:delusions}) discovers a commonly shared failure mode / safety risk of planning agents which rely on the generated observations / states / goals. This failure mode resembles hallucinations and the resulting delusional behaviors in the human brain. Inspired by understanding of how human brains address delusions, we propose general solutions that enable agents to autonomously and preemptively avoid issues such as blindly trusting hallucinated targets.
\end{itemize}

\begin{figure}
\centering
\includegraphics[width=0.6\textwidth]{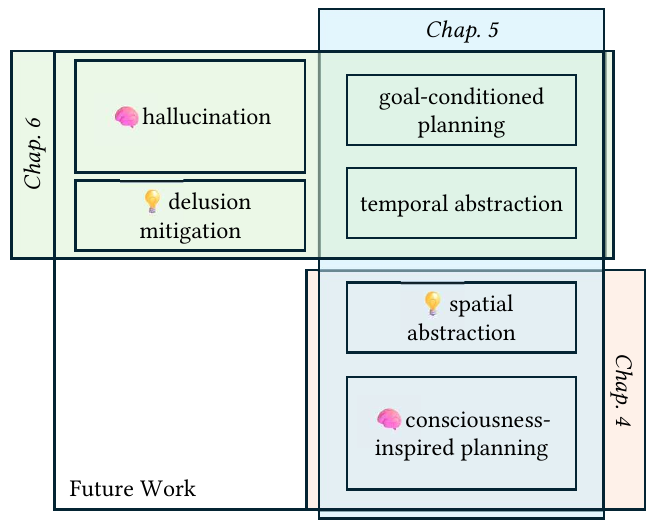}
\caption[Main Chapters in this Thesis (Research Methodologies \& Original Findings)]{\textbf{Main Chapters in this Thesis (Research Methodologies \& Original Findings)}: Chap.~\ref{cha:CP} (\nameref{cha:CP}) is heavily influenced by human \textbf{conscious planning}, and proposes a \textbf{spatial abstraction} process, which is later combined organically with \textbf{temporal abstraction}, to create the \Skipper{} framework in Chap.~\ref{cha:skipper} (\nameref{cha:skipper}). The \Skipper{} framework embraces \textbf{goal-conditioned planning}, which utilizes \textbf{temporal abstraction}. \textbf{Goal-conditioned planning} agents, \Skipper{} included, suffer from delusions in planning. Chap.~\ref{cha:delusions} (\nameref{cha:delusions}) takes inspiration from how the human brain addresses delusional behaviors, to propose \textbf{delusion mitigation} strategies that allow \textbf{target-assisted planning} agents to preemptively and autonomously address delusional planning behaviors during training. The proposed future work encompasses all the key inspirations and contributions (Sec.~\ref{sec:future_work}, Page.~\pageref{sec:future_work}).}
\label{fig:fig_thesis_projects}
\end{figure}

The 3 main chapters on the methodologies and original research findings are in lock-steps to serve as the milestones for the thesis topic. Each main chapter serves as the basis for the upcoming chapters, \ie{}, Chap.~\ref{cha:skipper} is based on the findings of Chap.~\ref{cha:CP} and Chap.~\ref{cha:delusions} is based on both Chap.~\ref{cha:CP} \& Chap.~\ref{cha:skipper}. We present a chart for the relationship among the contributions in Part II, illustrated with key ideas and methodologies, in Fig.~\ref{fig:fig_thesis_projects}.

\textbf{Part III: Discussions \& Conclusions}:

\begin{itemize}[leftmargin=*]
\item
Chap.~\ref{cha:conclusion} (Page.~\pageref{cha:conclusion}) provides a comprehensive scholarly discussion of all findings, including how the contributions met the objectives of the doctoral study, the impact of the contributions, their limitations, along with directions on future work.
\end{itemize}

\section{Summary of Original Contributions to Knowledge}

The following are the \textit{short} summaries of the original contributions of this thesis. Chap.~\ref{cha:conclusion} expands on the detailed contributions of each main chapter.

Chap.~\ref{cha:CP} describes a decision-time planning agent that can dynamically focus on interesting partial aspects of the state for better OOD generalization, a first in the literature. The core bottleneck mechanism is a top-down attention computation inspired by conscious reasoning in humans \citep{dehane2017consciousness}. This work is one of the first works utilizing transformer-based architectures in computational decision-making. This work also opens up discussions about the ways in which ideas from higher-level cognitive functions in humans can be used to improve the generalization abilities of computational decision-making agents.

Chap.~\ref{cha:skipper} proposes a framework that automatically decomposes an overall task into smaller and more manageable steps. \Skipper{} utilizes a constrained form of option-based planning, which builds on the consciousness-inspired spatial abstraction mechanisms in Chap.~\ref{cha:CP} when considering each steps. Novel mechanisms are proposed to learn a problem decomposition consistent with the agent's own capabilities of handling each decomposed step. We also prove that the performance of the approach is guaranteed under practical conditions. This work shows that spatially and temporally abstract planning, like that of humans, is not only viable, but its performance can also be guaranteed RL agents, showing a promising direction of option-based planning.

Chap.~\ref{cha:delusions} points out an important flaw in planning agents that utilize generated state targets: they blindly trust hallucinated targets. Many agents do not fully understand the targets that they can propose during planning. Inspired by the belief evaluation system in the human brain, we propose a method to reject hallucinated targets by properly learning a target feasibility evaluator. To be able to learn such an evaluator effectively, we analyze the categories of delusions that can appear in a model generating state targets, and propose a comprehensive solution that can reliably address delusional behaviors resulting from such targets. The solution is a combination of update rules, model architectures as well as hindsight relabeling strategies to solve the mismatch between the planning agents' training and behaviors. In experiments, we find that our solution significantly reduces feasibility errors and the frequency of delusional behaviors, and boosts OOD generalization performance compared to existing methods. Instead of blindly optimizing for sample efficiency, this is the first work that systematically discusses the failure modes of relevant planning agents, introducing the hallucination-delusion perspective. These ideas could be used to save future research efforts from continuing todevelop delusional / unsafe agents.

\section{Collaborator Contributions Breakdown}
For work presented in the 3 main chapters of this thesis, I assumed the primary role in formulating the methodological ideas, mathematical components and proofs, writing, implementation and experiments. My co-authors mostly participated in discussions and brainstorming sessions, providing feedback, contributing to proofreading, and improving communication.

\subsection{Chap.~\ref{cha:CP} - \nameref{cha:CP}}
The collaborators of work presented in this chapter include myself, Zhen Liu, Sitao Luan, Shuyuan Zhang, Doina Precup and Yoshua Bengio. Some contents of this chapter are published as a conference paper at the Conference on Neural Information Processing Systems (NeurIPS) 2021 \citep{zhao2021consciousness}.

Yoshua and I brainstormed and formulated the original abstract idea of the chapter. Then, Sitao, Doina and I discussed the high-level design to implement the ideas. I supervised Zhen and Shuyuan in developing detailed algorithmic designs, implementations as well as experiments. All collaborators proofread the accepted manuscript and contributed to its communication.

\subsection{Chap.~\ref{cha:skipper} - \nameref{cha:skipper}}
The collaborators of work presented in this chapter include myself, Safa Alver, Harm van Seijen, Romain Laroche, Doina Precup and Yoshua Bengio. Some contents of this chapter are published as a conference paper at the International Conference on Learning Representations (ICLR) 2024 \citep{zhao2024consciousness}.

Based on earlier brainstorming with Doina and Yoshua, Harm, Romain and I brainstormed and formulated the ideas of the chapter and identified the milestones needed. I conducted all the detailed algorithmic designs, implementations as well as experiments. Safa helped me implement a baseline in the experiments. I collaborated with Romain closely on proving the theoretical results. All collaborators proofread the accepted manuscript and contributed to the communications.

\subsection{Chap.~\ref{cha:delusions} - \nameref{cha:delusions}}
The collaborators of work presented in this chapter include myself, Tristan Sylvain, Romain Laroche, Doina Precup and Yoshua Bengio. Some contents of this chapter are published as a conference paper at the International Conference on Machine Learning (ICML) 2025 \citep{zhao2024delusions}.

I conceived the idea of this chapter after realizing that certain delusional behaviors I observed in Chap.~\ref{cha:skipper} are commonly shared among existing methods. Then, Tristan and I developed the ideas for a controlled environment to identify the causes of these behaviors. I implemented the environments, conducted the experiments, and identified the types and root causes of delusions and wrote the manuscript. Romain, Doina and I investigated the theoretical aspects of the chapter and drafted the formal definitions. All collaborators proofread the manuscript and contributed to the communications.

\section{Copyright}

Most figures used in this work are directly created by me. Some figures are original for this thesis, while some others are taken from the related conference papers that I have published as the first author, for which I hold the copyright.

%% file: chapter_2_basics.tex
\chapter{Literature Review: Reinforcement Learning}
\label{cha:basics}
\textit{\small This chapter presents basic background knowledge about  Reinforcement Learning}

\minitoc

\section{What is Reinforcement Learning?}
Reinforcement Learning (RL) is a methodology aimed at addressing decision-making problems through learning from interactions with the environment, without explicit supervision. RL draws significant inspiration from the neural basis of learning and operant conditioning in neuroscience \citep{sutton2018reinforcement}.

\subsection{Learning by Trial-and-Error}

A distinctive feature of RL is that agents learn through trial-and-error. In essence, RL methods autonomously discover both the nature of the situation and the appropriate actions to take post-deployment, highlighting the reduced need for domain-specific knowledge.

The methodology of RL is both powerful and versatile, as trial-and-error learning can be applied to virtually any decision-making problem. It is worth noting that the study of RL extends beyond merely refining learning algorithms: the learning process can also be impeded by challenges associated with environment setup, data collection, and other preparatory activities.

RL is predominantly applied on the reward maximization scenarios in sequential decision-making \citep{sutton2018reinforcement}.

\subsection{Reward Maximization in Sequential Decision-Making}

Sequential Decision-Making is a concept which involves making a series of decisions and actions over time to optimize objective functions, such as maximizing cumulative rewards, or somewhat equivalently, minimizing costs. Each decision influences subsequent choices and system outcomes, taking into account the current status of the system, available actions, and the probabilistic nature of action-induced transitions \citep{puterman2014markov}.

In a sequential decision-making problem, we name the decision-maker an \textit{agent}, and everything else the \textit{environment}. An agent can be an algorithm or a method, while the environment represent the ``world'' that the agent is in.

This thesis is particularly interested in the case of Sequential Decision-Making with the objective of maximizing cumulative scalar rewards, which is the case that RL seeks to tackle. A \textit{reward} signal defines the objective of a sequential decision-making problem and is received after taking each action. The reward signal thus defines what are the good and bad events for the agent. Reward signals can also explicitly correspond to a given ``goal'', \eg{}, a sparse terminal reward if some goal is achieved.

The generality of RL's maximization approach is backed by the so-called reward hypothesis, which posits that, ``all of what we mean by goals and purposes can be well thought of as maximization of the expected value of the cumulative sum of a received scalar signal (reward)''. The readers of this thesis can use this hypothesis to specify the implicit requirements on goals and purposes under which the hypothesis holds \citep{bowling2023settling}.

Particularly, we focus on \textit{discrete-time} scenarios where the timings of decision and corresponding environment change are placed over discrete time intervals named \textbf{timesteps} defined between \textbf{decision points}. These can be formulated as follows: let the decision points (in the flow of time) be $t_0, t_1, \cdots$. At a decision point $t$, the agent receives an \textit{observation} $x_t \in \scriptX$ and chooses an action $a_t \in \scriptA$, where $\scriptX$ is the \textit{observation space}, the collection of possible observations for the agent that is often assumed to be arbitrarily large, and $\scriptA$ is the collection of all possible actions that the agent could take, named the \textit{action space}. Possibly represented in a variety of ways, an observation is what an environment allows the agent to observe during the agent-environment interactions, and can often conceal important information regarding the environmental state. An action is a choice among all the possible ways that the environment allows an agent to interact with itself. Let $t'$ be the next decision point after $t$. For the period from $t$ to $t'$, the agent receives feedback $r_{t \to t'}$, namely \textit{reward}, which depends on the agent's action. The objective of an agent in this scenario is to maximize the cumulative \textbf{return} received over its lifetime, expressed as a sum of rewards, until the agent can no longer interact with the environment.

In addition to the sequential decision-making framework above, RL relies on the notion of \textbf{environment state}, or \textbf{states} for short, to ground itself in the mathematical framework of Markov Decision Processes (MDPs), to be introduced later. In the aforementioned definitions, observations are the system outcomes of the actions that are exposed to the agent, while some changes to the environment that remain hidden to the agent could still influence the future outcomes and decisions. A state is a notion unifying both the exposed parts and the hidden \citep{amortila2024reinforcement}, an identifier of the current situation of the environment, which can be hidden from the agent (and can be exposed fully as well) and may require the agent to infer through its own perceptions. The \textbf{environment state} refers to the complete representation of the current state of the environment in which the agent operates. This state includes all relevant information about the environment that can affect the agent's decision-making and future interactions. Conversely, an agent needs to infer the \textbf{agent state} from its interaction history with the environment. These agent states refer to the internal representations or understandings that the agent has of its current situation based on its observations and prior knowledge, and thus may not necessarily match the complete environment state.

\phantomsection
\label{sec:state_representations}

When we talk about ``state representations'' in the later parts, we are either referring to the observations that are lossless transformations of the real environment states (emitted by the environment) or to the agent states constructed by the agents themselves. We overload the term ``state'' for convenience, and should be able to easily differentiate based on the context.

With the notion of states, in the RL setting, the \textit{environment} becomes an ensemble of a reward function (a scalar feedback for the decisions), state dynamics (transition probabilities of states by actions) and optionally a termination signal (a binary feedback for ending the episode). The environment emits observations to the agent, and the agent needs to predict the current environmental state from the interaction history, which can be used as the basis for decision-making, compared to considering all interaction history.

\begin{figure}
\centering
\includegraphics[width=0.6\textwidth]{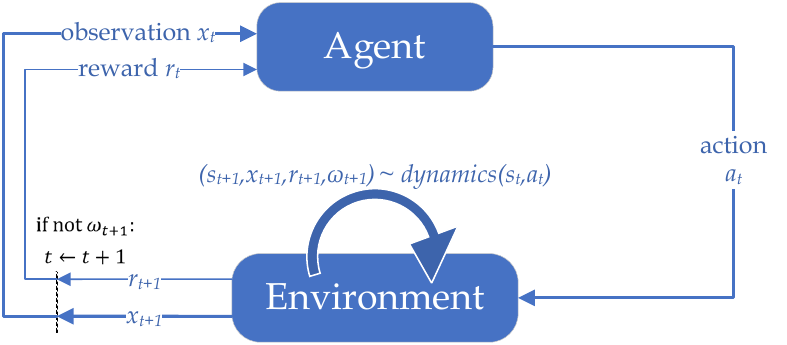}
\caption[Agent-Environment Interaction in RL]{\textbf{Agent-Environment Interaction in an RL Problem (from an agent-centric view)}: $\omega_{t+1}$ is the binary termination signal indicating if $s_{t+1}$ is terminal, \ie{}, if the current episode ends and the agent can take no more actions. The environment can hide the states from the agent, only exposing the observations. The environment dynamics is only decided by the environment state $s_t$ and the action $a_t$.}
\label{fig:agent_environment_interaction}
\end{figure}

\subsection{Markov Decision Processes}
\label{sec:MDP}

Markov Decision Processes (MDPs), an abstraction of the sequential-decision making problems, are classically used as a mathematical formalization to ground RL in mathematical analyses, a connection introduced and popularized by \citet{watkins1989learning}. MDPs can be seen as an environment-centric view of RL problems, contrary to that presented in Fig.~\ref{fig:agent_environment_interaction}.


Coinciding with RL, MDPs are developed through the viewpoints of state transitions, as the states are used to specify the starts and the ends of all transitions. Observations are often ignored in this environment-centric view of the RL problem. In an MDP, an agent is in a state and only one state at any time.

States convey some sense of ``how the environment is'' at a particular time and act as sufficient statistics to make optimal decisions, \st{} the previous interaction histories with the environment needs not be considered for decision-making, giving rise to MDP's Markovian properties. In other words, a state is a compact lossless compression of the agent-environment interaction history. With the help of the notion of states, we establish the framework of MDPs.

Formally, an MDP is a five-tuple $\langle \scriptS, \scriptA, p, R, p_0 \rangle$, defining the properties of the world and the objective within it. The state space $\scriptS$ defines the set of all possible states. Each decision that the agent can make is known as an action $a \in \scriptA$, where the action space $\scriptA$ defines the set of all actions. The function $p: \scriptS \times \scriptR \times \scriptS \times \scriptA \to [0,1]$ defines the \textit{dynamics} of the MDP, and is often recognized as the \textit{transition probability function} or simply \textit{transition function}. In the MDP, the rewards are specified with a function $R$, a marginalized descriptor based on $p$, established over a transition. The states that the agent start from in a finite MDP can be described using a probability distribution $p_0(s) \coloneqq \doubleP\{S_0 = s\}$ for each $s \in \scriptS$, named the \textbf{initial state distribution}. 



In a (discrete-time) MDP, at each timestep $t\in\{ 0, 1, 2, \dots \}$, the agent resides in a state $S_t \in \scriptS{}$ and on that basis selects an action, $A_t \in \scriptA{}$. One timestep later at $t+1$, in part as a consequence of its action, the agent receives a numerical reward, $R_{t+1} \in \scriptR{} \subset \doubleR{}$ and finds itself in a new state $S_{t+1}$. Thus, $p$ explains the joint probabilities of transitions organized in the form $\langle s_t, a_t, r_{t+1}, s_{t+1}\rangle$, where $s_t$, $a_t$ correspond to the state and the action the agent took at time $t$, while $s_{t+1}$ correspond to the state the agent transitioned to following $\langle s_t, a_t \rangle$, and finally $r_{t+1}$ corresponds to the reward for such transition.

Technically, in a finite MDP, where the state set $\scriptS$, the action set $\scriptA$ and the reward set $\scriptR$ are all finite, the random variables $R_t$ and $S_t$ have well-defined discrete probability distributions that only depend on the preceding state and action. That is, for particular values of these random variables, $s' \in \scriptS$ and $r \in \scriptR$, there is a probability of those values occurring at timestep $t$, given particular preceding state and action:
$$p(s',r|s,a) \coloneqq \doubleP\{ S_{t+1} = s', R_{t+1} = r | S_{t} = s, A_{t} = a \}, \forall s, s' \in \scriptS, \forall r \in \scriptR, \forall a \in \scriptA$$

An episode of agent-environment interactions begins with an initial state $s_0$ sampled from an initial state distribution $p_0(s) \coloneqq \doubleP\{S_0 = s\}$ over each state $s$. The following sequence defines the process of an MDP (without termination signals):
\begin{enumerate}[leftmargin=*]
    \item The agent starts from an initial state $s_0$, whose distribution can be described with $p_0$;
    \item The agent selects an action (according to its policy);
    \item A new state and reward are sampled according to the Markovian dynamics function $p$;
    \item The process is repeated from Step 2 if the agent is not in a terminal state.
\end{enumerate}

The MDP formulation describes what happens during agent-environment interactions with all perfect access to all the environmental dynamics. Note that in an MDP formulation, the termination signal is absorbed into the agent's knowledge about the states.

The MDP and the actions (taken by an agent) together thereby give rise to a \textit{trajectory} like:

$$S_t,A_t,R_{t+1},S_{t+1},A_{t+1},R_{t+2},\dots$$

An \textbf{episode} is used to describe a full trajectory from the initial state $S_0$ until termination in $S_T$.

Note that we have used the uppercase letters to denote the random variables since we have not yet observed the states, the rewards or the actions. Yet, if we have already, we would use lowercase to denote their specific instantiation. For example, at timestep $t$, the agent took action $a_t$ based on state $s_t$ and transitioned to the state $s_{t+1}$ while receiving the reward $r_{t+1}$.

\subsubsection{Markov Property}
In an MDP, the probabilities given by $p$ \textit{completely} characterize the environment's dynamics. That is, the probability of each possible value for $S_{t+1}$ and $R_{t+1}$ depends only on the immediately preceding state and action $S_{t}$ and $A_{t}$, not on earlier states and actions. This is best viewed a restriction not on the decision process but on the state, which means the state must include information about all aspects of the past agent-environment interaction that make a difference for the future.

The $4$-argument transition function is a most general form of a transition function defined in an MDP. There are also alternative forms of the transition function, which rely on either additional assumptions of the environment or exist as marginalized expectations.

Marginalizing over rewards yields the $3$-argument state-transition probability function:
$$p(s'|s,a) \coloneqq \doubleP\{ S_t = s' | S_{t-1} = s, A_{t-1} = a \} = \sum_{r \in \scriptR{}}{p(s',r|s,a)}$$
where $p$ is overloaded. However, if we assume that the state transition and the reward are jointly determined, \ie{}, a fixed transition from one state to another always generates the same reward, then the $4$-argument transition function collapse into the $3$-argument version $p: \scriptS \times \scriptS \times \scriptA \to \times [0,1]$.

Many other useful expected statistics can be derived from the general $4$-argument $p$ by marginalizing. These include:

Expected rewards for state-action pairs as a $2$-argument function $r: \scriptS \times \scriptA \to \doubleR$:
$$r(s,a) \coloneqq \doubleE[ R_t | S_{t-1} = s, A_{t-1} = a ] = \sum_{r \in \scriptR{}}{r \sum_{s' \in \scriptS{}}{ p(s',r|s,a) }} $$
Since the only way that the agent could interact with the environment is through the action, there is no way for the agent to optimize the transition and reward by any other means, this $2$-argument expected reward function should be an appropriate choice when the agent tries to model the reward function for decisioning, through agent-environment interactions.

Expected rewards for state-action-next-state triples as a $3$-argument function $r: \scriptS \times \scriptA \times \scriptS \to \doubleR$:

$$r(s,a,s') \coloneqq \doubleE[ R_t | S_{t-1} = s, A_{t-1} = a, S_{t} = s' ] = \sum_{r \in \scriptR{}}{r \frac{p(s',r|s,a)}{p(s'|s,a)}} $$

where $p(s'|s,a)$ is the $3$-argument transition function we derived earlier. This function can be estimated to predict the reward incurred by some certain transition, which is often used in model-based RL.

For many environments, the reward function is a signal consistent with the encouragement of reaching a certain goal, that is, a certain subset of $\scriptS$. Under these circumstances, we could also use goals to represent rewards. However, it is worth noting that not all wanted behaviors can be incentivized by Markovian rewards in the context of MDPs \citep{abel2021expressivity}.

\subsubsection{Limitations to MDP Formulation}

\textbf{Implicit Pause-able Environment Assumption}: MDP assumes that the environment only changes after the agent takes an action, whereas this assumption seems inappropriate in many other sequential decision-making problems, \eg{} when the agent is planning with a complex model \citep{ramstedt2019real}.

\textbf{Stationarity}: MDPs assume a stationary environment, where the transition and reward functions do not change over time. In dynamic environments, the models may need to adapt, which can complicate learning and decision-making.

\textbf{Discrepancy between State and Observations}: the MDP formulation can be used to analyze the behaviors of agents, but often cannot be used to direct RL agents' learning behaviors when the environment is not fully observable, \ie{}, when the observations do not contain full information of the corresponding state, or when it is difficult for information related to making better decisions to be extracted from observations. A way to address such discrepancy is to use the Partially-Observable MDP (POMDP) formulation, whose discussions I will skip because of the low relevance to this thesis.


\subsection{RL Objectives in MDPs (Episodic Setting)}

In MDPs, actions influence more than just the immediate rewards, essentially feedback from the environment, but also subsequent situations, or states, and through the future rewards. Thus, MDPs involve delayed reward and the need to tradeoff immediate and delayed reward.

By connecting RL with an MDP, we can formulate the objectives that RL agents seek to optimize. Note that in this thesis, we only discuss cases where trajectories have finite lengths, often recognized as the \textbf{episodic} setting. The end of trajectory is triggered when an agent transitions into a terminal state. After reaching a terminal state, the agent is taken out of the MDP.

\subsubsection{Undiscounted Episodic Return}

In the form most aligned with sequential decision-making, RL seeks to maximize the expected return $G_t$, which, in the simplest case, is defined as the sum of the rewards:

\begin{equation}
\label{eq:undiscounted_return}
G_t \equiv R_{t+1} + R_{t+2} + \cdots + R_{T}
\end{equation}

where $T$ is a final timestep. The notion of \textit{episodes} is naturally formed as the interaction sequence from the starting timestep $0$ until the terminal time $T$, which is a random variable that normally varies and independent with each other, \ie{}, one episode does not affect the environment dynamics of the next.

In episodic tasks, it is sometimes necessary to distinguish the (sub-)set of all non-terminal states, denoted $\scriptS$ from the set of all states plus the terminal states $\scriptS^{+}$. In the literature, sometimes ``trajectory'' is used loosely to denote a segment of an episode. For instance, in some control tasks, a Maximum Episode Length (MEL) would be set to make sure that the agent does not get stuck for an unacceptable amount of time in uninteresting regions of the state space \citep{erraqabi2022temporal}. These trajectories are segments of full episodes, since they are not terminated by transitioning into a terminal state.

\subsubsection{Discounted Episodic Return}

In the episodic setting, which is of interest of this thesis, RL agents would often employ an alternative objective additionally defined by a notion of discounting. This means, instead of RL agents trying to maximize the simple sum of episodic rewards, some try to maximize a ``discounted'' version of the original objective. Such discounting is often introduced to improve the convergence of policy evaluation methods, to be introduced later.

According to the discounting approach, the agent tries to select actions so that the sum of the discounted rewards it receives over the future is maximized. In particular, it chooses to maximize the \textit{expected discounted return}:

\begin{equation}
\label{eq:discounted_return}
G_t \equiv R_{t+1} + \gamma R_{t+2} + \gamma^2 R_{t+3} + \cdots = \sum_{k=0}^{\infty}{\gamma^{k}R_{t+k+1}}
\end{equation}

where $\gamma \in [0, 1]$ is the \textit{discount parameter}, also recognized as the \textit{discount rate}. Note that here the discount parameter is constant throughout the states and the episodes, but it can be a state or observation-based function as well \citep{zhao2020meta}.

Let us see an example of how discounting could influence the behavior of a return-maximizing agent. Imagine a navigation task, where an agent receives a terminal reward of $+1$ upon success. By maximizing the original undiscounted objective, the agent would have no incentive to reach the goal more quickly, as it just needs to reach the goal in the end to get the same amount of reward. While, for an agent with the discounted objective, every timestep wasted is a penalty towards the objective, thus it would be incentivized to reach the goal quickly.

The fact that the discount parameters can be set to $1$ gives a unified formulation for the discounted episodic tasks as well as the undiscounted. Ideally, the discount factors should come from the task itself, as its value reflects the objective of the task. However, for complicated tasks with Deep Reinforcement Learning (DRL), which is essentially using artificial neural networks for RL, it is generally observed that lowering the discount factor yields significantly more stable performance rather than using $\gamma = 1$, even if the objective includes no discounting \citep{mnih2015human}. These blur the line how we should see the discount factor, which classically should be seen as some kind of built-in characteristics of the environment yet now a parameter that could be set or learned for some purposes.

Discounting schemes, other than a constant $\gamma$-controlled exponential weighting presented above, are also investigated in the literature \citep{schultheis2022reinforcement}, \eg{} state-based discounting \citep{white2016greedy,zhao2020meta}.

In this thesis, we focus on the discounted episodic setting.

\subsection{Policies \& Value Functions}
\label{sec:policies_values}

A \textit{policy} is a component used by an agent to decide what to do. In an MDP, a policy is a function which takes in the (environmental) state the agent is in as input and outputs decisions about actions. Formally,

\begin{coloreddefinition}{policy}{policy}
In an MDP, a policy $\pi$ is a mapping from states to probabilities of selecting each possible action, \ie{}, $\pi: \scriptS{} \times \scriptA{} \to [0, 1]$.
\end{coloreddefinition}

In an MDP, if the agent is following policy $\pi$ at timestep $t$, then $\pi(a|s)$ is the probability that $A_t = a$ if $S_t = s$. If all probability mass is concentrated at a single action, then we call the policy \textbf{deterministic}, otherwise \textbf{stochastic}, in the sense that an action would be sampled according to the probabilities. Note that it is also common to define state-based action sets, however we will stick loyal to the simple setting of \citet{sutton2018reinforcement} for this thesis, \ie{}, assume that all actions are available at all times. The policies in RL are essentially stationary decision rules defined for more general Markov chains \citep{puterman2014markov}, where ``stationary'' means that the decision rules are consistent for every possible states, as in we are only interested in the stationary MDPs in this thesis.

The \textit{value function} of a state $s$ under a policy $\pi$, denoted $v_{\pi}(s)$, is the expected (discounted) return if an agent starts in $s$ and following $\pi$ thereafter. For historical reasons, we often call it ``V-value'' to contrast with ``Q-value'' given by the state-action value function. Formally,

\begin{coloreddefinition}{state-value function (V-values)}{vfunction}
In an MDP, the \textit{state-value function} for policy $\pi$ or simply \textit{value function} $v_\pi(s)$, given discount function $\gamma(s)$ and policy $\pi(s)$, is defined as
$$v_{\pi}(s) \coloneqq \doubleE_{\pi}[ G_t | S_t = s ] = \doubleE_{\pi} \left[ {\sum_{k=0}^{\infty}{\gamma^k \cdot  R_{t+k+1}}} \mid {S_t = s} \right]$$

where $\doubleE_{\pi}$ denotes the expected value of the random variable given that the agent follows policy $\pi$ and the values of the terminal states are defined as $0$.
\end{coloreddefinition}

We are also interested in the state-action value function, which is more useful for control cases, \eg{} when searching for better policies with a value estimator.

\begin{coloreddefinition}{state-action-value function (Q-values)}{qfunction}
In an MDP, the \textit{state-action-value for policy $\pi$} $q_\pi(s)$, given discount function $\gamma$ and policy $\pi(s)$, is defined as the expected return starting from $s$, taking the action $a$ and thereafter following $\pi$:
$$q_{\pi}(s, a) \coloneqq \doubleE_{\pi}[ G_t | S_t = s, A_t = a ] = \doubleE_{\pi} \left[ {\sum_{k=0}^{\infty}{\gamma^k \cdot  R_{t+k+1}}} \mid {S_t = s, A_t = a} \right]$$
\end{coloreddefinition}

One of the key subroutines of RL is to estimate $v_\pi$ or $q_\pi$ from experience, as $\hat{v}_\pi$ or $\hat{q}_\pi$. This estimation can also sometimes be recognized as \textit{policy evaluation} or \textit{prediction}. This concept is to be introduced in detail in Sec.~\ref{sec:PEPI}.

With the recursive definition in Def.~\ref{def:vfunction}, we can obtain the following equation for the state-value function $v_\pi$, given a policy $\pi$:

\begin{equation}
\label{eq:bellman_v_general}
\begin{aligned}
v_{\pi}(s) &  \coloneqq \doubleE_{\pi}[ G_t | S_t = s ] = \doubleE_{\pi}[ R_{t+1} + \gamma G_{t+1}]\\
& = \sum_{a}{\pi(a|s) \sum_{s',r}{p(s',r|s,a)\left[ r + \gamma \cdot \doubleE_{\pi}[G_{t+1}|S_{t+1}=s'] \right]}}\\
& = \sum_{a}{\pi(a|s) \sum_{s',r}{p(s',r|s,a)\left[ r + \gamma \cdot v_{\pi}(s')\right]}}
\end{aligned}
\end{equation}

The equation is essentially one basic form of the \textit{Bellman equation} for $v_{\pi}$. Serving as a core equation of RL, it expresses the relationship between the value of one state and its successor states. With this equation and access to environment dynamics (defined as an MDP), the method of dynamic programming can be used to exactly compute the value function $v_{\pi}$ or $q_{\pi}$ given any valid policy $\pi$, to be introduced later.

\subsection{Optimal Policies \& Optimal Value Functions}

To find the optimal policy, we must define what it means for one policy to be better than another. This is done via the value functions, which define a partial ordering over policies in an MDP.

\begin{coloreddefinition}{Partial Order of Policies}{partialorderpolicy}
In an MDP with certain discount $\gamma$, a policy $\pi$ is defined to be \textit{better than or equal to} a policy $\pi'$ if its expected return is greater than or equal to that of $\pi'$ for all states, \ie{}, $\pi \geq \pi'$ \textit{\textbf{iff}} $\forall s \in \scriptS{}, v_{\pi}(s) \geq v_{\pi'}(s)$.
\end{coloreddefinition}

There always exists at least one policy that is better than or equal to all other policies, which is identified as the \textbf{optimal policy} $\pi_{*}$. Intuitively, the optimal policy is the policy with the largest value. However, less intuitively,

\begin{coloredfact}{Existence of Optimal Deterministic Markovian Policies}{optdeterpols}
Given a finite MDP, there always exists a deterministic Markovian policy $\pi_{*}$ that achieves the optimal values.
\end{coloredfact}

Together with the policy improvement theorem \citep{bellman1957dynamic}, to be discussed later, the fact serves as the reason why we could confidently focus on the space of deterministic Markovian policies, since an optimization trajectory towards an optimal policy always exist inside. This also justifies some RL methods without explicit policies (as a separate component), \eg{} value-based methods that use $\epsilon$-greedy upon deterministic Q-value estimates as policies, such as Q-learning (Sec.~\ref{sec:q_learning}, Page.~\pageref{sec:q_learning}) \citep{sutton2018reinforcement}. Also, despite that there can be more than one optimal policies, we denote all of them by $\pi_{*}$ for simplicity. In accordance with Def.~\ref{def:partialorderpolicy}, all optimal policies must share the same state-value function, called the \textit{optimal state-value function} $v_{*}$, which is defined as:

\begin{coloreddefinition}{optimal state-value (V-value) function}{optVvalue}
The optimal state-value function is shared by all optimal policies $\pi_{*}$ and defined as:

$$v_{*}(s) \equiv \max_{\pi}{v_{\pi} (s)}, \forall s \in \scriptS{}$$

where the $\max$ operator is defined upon the policy partial orders.
\end{coloreddefinition}

Optimal policies also share the same optimal action-value function $q_{*}$, which is defined as

\begin{coloreddefinition}{optimal state-action-value (Q-value) function}{optQvalue}
The optimal state-action-value function is shared by all optimal policies $\pi_{*}$ and defined as:

$$q_{*}(s,a) \equiv \max_{\pi}{q_{\pi} (s, a)}, \forall s \in \scriptS{}, \forall a \in \scriptA{}$$

where the $\max$ operator is defined upon the policy partial orders.
\end{coloreddefinition}

$q_{*}$ gives the expected return for taking action $a$ in state $s$ and thereafter following an optimal policy. Naturally, we can establish the following connections between Def.~\ref{def:optVvalue} and Def.~\ref{def:optQvalue}:

$$q_{*}(s,a) = \doubleE \left[ R_{t+1} + \gamma v_{*}(S_{t+1}) | S_t = s, A_t = a \right]$$

There are also \textit{Bellman optimality equations}, which are specialized for optimal policies\footnote{Historically, Bellman optimality equation is often abbreviated as ``Bellman Equation''. While we explicitly differentiate it with the Bellman (policy evaluation) equation in Eq.~\ref{eq:bellman_v_general}}.

The methods for finding better policies, are recognized as \textit{policy improvement} methods. The Bellman optimality equation conducts policy evaluation and policy improvement at the same time. Later, such equation would be used to support value iteration, a classic dynamic programming method for computing the optimal policy, to be discussed in Sec.~\ref{sec:VI} on Page.~\pageref{sec:VI}.

\section{Policy Evaluation \& Policy Improvement}
\label{sec:PEPI}
Central to the search for better policies are two important synergetic mechanisms: \textbf{policy evaluation} and \textbf{policy improvement}, which refers to understanding how it is doing and improving behaviors to do better, respectively. These are central to both dynamic programming and (approximate) RL methods, to be introduced later.

\textbf{Policy Evaluation} enables the understanding of how good a policy is, and hence to help find better policies \citep{sutton2018reinforcement}. Policy evaluations produce either an estimate of state value function (``V'') or an estimate of state-action value function (``Q''). Representative methods of policy evaluation include Monte Carlo estimation or (approximate) dynamic programming (temporal-difference methods), to be introduced later.

\textbf{Policy Improvement} is the process of enhancing a policy based on its evaluation, with the aim of maximizing the expected return or cumulative rewards. This step typically follows policy evaluation and is essential for iterative approaches like policy iteration. 

Policy evaluation points the direction of policy improvement, \ie{}, improving values equals to improving policies. This is supported by the policy improvement theorem \citep{bellman1957dynamic}, which states that from any given policy, a superior policy can be obtained by choosing actions with a higher state value for any state applicable. For instance, if conveniently we have for a state $s$ that $Q(s,a^{'})$ is larger than the current policy's action $a$'s $Q(s,a)$, then the policy can be improved by learning to choose $a^{'}$ at $s$ instead of $a$.

\phantomsection
\label{sec:actor_critic}

If a policy is parameterized, \ie{}, not a deterministic function over the value estimates (\eg{}, greedy policy based on the value estimates) and the policy can be itself improved by certain optimization methods, the resulting RL agent architecture is commonly named \textbf{actor-critic}, with the \textbf{actor} being the policy and the \textbf{critic} being the value estimator. When an agent with an actor-critic architecture employs a parameterized policy optimized by gradient-based optimization methods, \textbf{policy gradients} methods are often used,  where the actor is the policy improved via estimated gradients established over the estimates from the critic \citep{konda1999actor,sutton2018reinforcement}. Evidently, the idea behind actor-critic\footnote{I will not introduce policy gradient methods in this chapter because of its low relevance to this thesis.} is essentially alternating policy evaluation and policy improvement, which is recognized as \textbf{policy iteration}. Policy iteration is a powerful and universal idea towards addressing reward maximization.

It is also worth pointing out that the statements about policy iteration above \textit{did not necessarily} assume that policy evaluation algorithms run until convergence and converges to the correct values. This is because if the policy evaluation algorithm is guaranteed to converge, then policy improvement attempts based on the prematurely terminated value estimates could result in correct policy improvements as well, giving rise to Generalized Policy Iteration (GPI) \citep{sutton2018reinforcement}. Thus, the assumption is not necessary for the cases of dynamic programming and certain approximate RL cases, to be introduced soon.

We will introduce detailed policy evaluation and policy improvement methods in the following sections regarding dynamic programming and (approximate) RL, respectively. 

\section{Dynamic Programming (DP)}
\label{sec:dynamic_programming}

Dynamic Programming (DP) is an optimization algorithm that simplifies a complicated problem by breaking it down into simpler sub-problems in a recursive manner \citep{bellman1957dynamic}.

In the context of RL, DP mostly refers to a family of heavily-used algorithms that, given an environment MDP and access to all state-action combinations, computationally solve interesting quantities such as the values function given a certain policy, finding the optimal policies, \etc{}. Note that DP methods should be distinguished from the approximate RL methods, which are to be introduced later and are often call ``RL methods'' for short, as approximate RL methods do not assume access to the environment MDP and all state-action combinations. This should be intuitive, since DP methods do not employ the core RL strategy of learning from interactions. However, because of DP methods' provable convergence in solving the interesting quantities, they often serve as the bases of developments of approximate RL methods which can scale to real-world problem-solving, to be introduced later.

DP is widely used to analytically calculate the ground truth values of relatively small sized environments, which can be used to analyze the performance of approximate RL methods which does not have the access to the environment MDP. Take the contributions of this thesis for example, we solve the optimal policies in Chap.~\ref{cha:CP} to compare the planned actions against the ground truth optimals, as well as the ground truth relationships between pairs of states in Chap.~\ref{cha:skipper} to understand how well the estimators work. Despite that DP could be used to solve an optimal policy, its use is limited in practice because of the need of access to environment MDP, as well as the expensive computational cost.

\subsection{DP for Value Function}

First, we consider the method of computing the state-value function $v_{\pi}$ for an arbitrary policy $\pi$,\ie{}, policy evaluation via DP. This is enabled by the following:

\begin{coloredfact}{Existence \& Uniqueness of State-Value Function}{vfunc}
The existence and the uniqueness of state-value function $v_{\pi}$ are guaranteed as long as either $\gamma < 1$ or termination will be reached from any state following $\pi$.
\end{coloredfact}

From the Bellman equation (\ref{eq:bellman_v_general}), we have

$$v_{\pi}(s) = \sum_{a}{\pi(a|s) \sum_{s',r}{p(s',r|s,a)\left[ r + \gamma \cdot v_{\pi}(s')\right]}}$$

This means, when the environment dynamics ($4$-argument $p$) is known, the Bellman equation gives a system of $|\scriptS|$ linear equations with $|\scriptS|$ unknowns, unsurprisingly solvable. In a compact matrix form, this linear system can be presented as:

$$\bm{v}_{\pi} = \bm{r}_{\pi} + \gamma P_{\pi} \bm{v}_{\pi}$$

where $\bm{r}_{\pi}$ is a $|\scriptS{}| \times 1$ vector in which $\bm{r}_{\pi}[i] = \sum_{a}{\pi(a|s_i) \sum_{s_j, r}{r \cdot p(s_j, r | s_i, a)}}$, $P$ is a $|\scriptS{}| \times |\scriptS{}|$ matrix in which $P_{\pi}[i,j] = \sum_{a}{\pi(a|s_i) p(s_j, r | s_i, a)}$.

With the system in hand, the rest is just to solve it as $\bm{v}_{\pi} = (I - \gamma P_{\pi}) \backslash \bm{r}_{\pi}$ or $\bm{v}_{\pi} = (I - \gamma P_{\pi})^{-1} \bm{r}_{\pi}$. 

Note that the $\scriptO({|\scriptS|}^3)$ complexity is a nightmare for problems with large state spaces. Thus, it is desirable to change this method into an iterative method with lower computational complexity. 




Thus, we arrive at the \textit{iterative policy evaluation method}. It is simple and powerful, turning Bellman equation into an iterative formula achieves the convergence to the true values. One can also prove the convergence of iterative policy evaluation using Banach's fixed point theorem, by showing that the Bellman operator is a contraction.

\phantomsection
\label{sec:contraction}

\begin{coloreddefinition}{Bellman Operator}{bellmanopr}
Given an MDP with its dynamics $p$, a policy $\pi$ and discount function $\gamma$, the \textit{Bellman operator} $\scriptB_{\pi}: \doubleR^{|\scriptS|} \to \doubleR^{|\scriptS|}$ is defined by
\begin{equation}\label{eq:bellman_operator}
(\scriptB_{\pi}v)(s) \coloneqq \sum_{a}{\pi(a|s) \sum_{s', r}{p(s',r|s,a)[r + \gamma v(s')]}}
\end{equation}

Or equivalently in matrix form,
\begin{equation}\label{eq:bellman_operator_matrix}
\scriptB_{\pi} \bm{V} \coloneqq \bm{r}_{\pi} + \gamma P_{\pi} \bm{V}
\end{equation}
where $\bm{r}_{\pi}$ is a $|\scriptS{}| \times 1$ vector in which $\bm{r}_{\pi}[i] = \sum_{a}{\pi(a|s_i) \sum_{s_j, r}{r \cdot p(s_j, r | s_i, a)}}$, $P$ is a $|\scriptS{}| \times |\scriptS{}|$ matrix in which $P_{\pi}[i,j] = \sum_{a}{\pi(a|s_i) p(s_j, r | s_i, a)}$.
\end{coloreddefinition}

\begin{coloreddefinition}{Contraction}{contraction}
Let $\langle X, d \rangle$ be a complete metric space. Then a map $\scriptT: X \to X$ is called a \textbf{contraction mapping} on $X$ if there exists $q \in [0, 1)$ \st{}
$$\forall x, y \in X, d(\scriptT(x), \scriptT(y)) \leq q \cdot d(x, y)$$
\end{coloreddefinition}


We can prove that the Bellman operator, which is essentially turning the Bellman equation into an iterative formula, on the \textit{estimated value function} or simply \textit{value estimate} is a contraction. The unique fixed point must be the true value because that is when the Bellman equation holds. 

The state-action value function $q_{\pi}$ can be computed trivially by combining the computed $v_{\pi}$ with the tabular reward function:

\begin{equation}
\label{eq:vqconversion}
q(s,a) = r(s,a) + \gamma \cdot \sum_{s'} p_\pi(s,s') v(s'), \forall s \in \scriptS, \forall a \in \scriptA
\end{equation}

Now that we have an algorithm to reliably solve the true values of a policy, we can think about how to generalize this algorithm to compute other interesting quantities, including for more complex cases involving goal-conditioned policies. I present the following examples:

In Chap.~\ref{cha:skipper}, we solved the cumulative rewards the agent would get by following a certain (goal-conditioned) policy $\pi$ going from one state to another. This is done by augmenting $\scriptS^{+}$ with the target state as an additional terminal state. Similarly, we also solved the cumulative discount from one state to another by making $r$ all zeros except when reaching the target state, giving $1$ instead;

In Chap.~\ref{cha:delusions}, we solved the distance between a pair of states, under a certain policy $\pi$, \ie{}, how long it takes for policy $\pi$ to travel from one state to another, are computed by replacing the rewards $\bm{r}_{\pi}$ with a vector full of $-1$s and setting $\gamma$ to be $1$. The convergence was assisted by the fact that the used environments were abundant with terminal states and the agent's policies will almost surely send the agent to a terminal state within finite timesteps, per Fact.~\ref{fact:vfunc}.

\subsection{DP for Optimal Policy}
Relying additionally on the policy improvement theorem \citep{bellman1957dynamic}, DP are empowered to find better policies, which ultimately leads to the optimal policies.

\phantomsection
\label{sec:VI}

DP methods can naturally rely on \textit{policy iteration} to find better policies, essentially alternating policy evaluation and policy improvement. However, such alternation may not always be trivially explicit. For example, \textbf{Value Iteration (VI)} is a representative DP method for finding the optimal value function by applying the Bellman optimality equations iteratively. VI conducts search in the constrained space of deterministic greedy policies and combines policy evaluation and policy improvement in one unifying step. 


With the computed V-value function (for the optimal policy), we can recover the policy from the converted Q-value function via Eq.~\ref{eq:vqconversion}.

VI has other uses and can take in alternative forms. In Sec.~\ref{sec:q_learning} on Page.~\pageref{sec:q_learning}, we will introduce Q-learning, an approximate RL algorithm that draws heavy similarity to VI; In Chap.~\ref{cha:skipper} (Page.~\pageref{cha:skipper}), we use VI (over options) to solve an approximate plan given an estimated proxy problem, which essentially takes the same form as an MDP, to make plans, \st{} the agent can know which subgoal is to follow.

Now that we have finished introducing the DP algorithms relevant to this thesis (while skipping other less relevant use cases), we transition to the canonical RL methods.

\section{(Approximate) RL}
Approximate RL methods are used when there is no access to the MDP ground truth nor all combinations of state-action pairs. Thus, approximate RL represents the case most true to trial-and-error based sequential decision-making and is often abbreviated as just ``RL''. Learning from actual experience is striking because it requires no prior knowledge of the environment’s dynamics, yet could still attain optimal behavior.

Approximate RL methods build on the principles of DP. Many RL algorithms, such as Q-learning (Sec.~\ref{sec:q_learning}, Page.~\pageref{sec:q_learning}) and actor-critic methods (Sec.~\ref{sec:actor_critic}, Page.~\pageref{sec:actor_critic}), use DP ideas but adapt them for function approximation.

The similarity between approximate RL and DP is that it also relies on the two key steps of:

\begin{itemize}[leftmargin=*]
\item \textbf{Policy Evaluation}: Just like DP methods, approximate RL aims to estimate value functions, but it does so in a way that does not assume access to environment MDP, and can often rely on function approximations when the states need to be inferred.

\item \textbf{Policy Improvement}: Approximate RL and DP both use similar principles for policy improvement. In approximate RL, the learned value functions can guide policy updates, analogous to how DP updates policies based on value function estimates.
\end{itemize}
The difference is more pronounced in the aspects of:

\begin{itemize}[leftmargin=*]
\item \textbf{Convergence and Stability}: While DP methods provide guarantees of convergence to the optimal solutions, approximate RL methods can struggle with convergence due to function approximation, particularly when using non-linear approximators like neural networks.
\item \textbf{Exploration-Exploitation Tradeoff}: Because of the lack of access to all state-action pairs, approximate RL methods need to optimize for a fundamental tradeoff between exploration and exploitation, which correspond to trying diverse actions to find better previously unknown rewarding trajectories \vs{} committing to known rewarding trajectories for better returns. The tradeoff also affects an RL agent in other ways. For example, from a ``dataset''-label perspective, we could say that in RL problems, the data samples are dynamically collected by the agents' decisions. This also means the quality of the ``dataset'' is also determined by the quality of the past decisions, making learning quality dependent on behavior. 
\end{itemize}


\subsection{RL Agent Components}
\label{sec:RL_agent_components}

To implement the two core mechanisms, \ie{}, policy evaluation and policy improvement, an (approximate) RL agent is often equipped with the follow components:

\begin{itemize}[leftmargin=*]
    \item A \textbf{value estimator} that approximates the (true) value function to conduct policy evaluation. Depending on the learning algorithm and learning capacity, policy evaluation by a value estimator may be misaligned with the objective performance it has in the environments, \ie{}, in approximate RL, value estimates do not necessarily converge to the true values of the value function. Value estimators could also be used to estimate other interesting auxiliary ``values'', those beyond the expected on-policy future return, as in Chap.~\ref{cha:skipper}. 
    \item A \textbf{policy} that can be improved based on the value estimates, the output of the value estimator. The improvements can be made by a search or optimization algorithms. The policy component of an agent could also be implicit. For example, we can always extract the greedy policy from a Q-value estimator if convenient, as shown later.
    \item An optional \textbf{model} of the environment. A model is something that mimics the behavior of certain aspects of the environment, or more generally, that allows inferences to be made about how the environment will behave. For example, given a state and action, the model might predict the resultant next state and next reward. Models are used for planning, by which we are not limited to deciding on a course of action by considering possible futures before they are actually experienced. A model could assist policy evaluation, as to be discussed as background planning, or could be an active component of the agent's policy, as to be discussed as decision-time planning. A model could also take many forms, as some models mimic the environment MDP, predicting the next state and reward out of the current state and intended action, while others could even predict partial aspects of distant future states, \etc{}. Methods for solving RL problems that use models and planning are called \textbf{model-based methods}, as opposed to \textbf{model-free methods} that are explicitly trials-and-errors, viewed as almost the opposite of planning. A model-based method contains all components of a model-free method, and the latter can be seen as a foundation of the former. We are particularly interested in and primarily dealing with model-based methods in this thesis.
\end{itemize}



\subsection{Train \& Evaluate}
An RL agent is deployed into its training environments with a certain budget of agent-environment interactions. During its interactions, the agent is expected to \textit{autonomously} collect data, estimate the values of its policies and make improvements accordingly. 

To understand the learned capabilities, an agent's performance is evaluated in expectation over environment instantiations, where in the real-world, these evaluation environments can often be different from the training environments.

One important objective of RL is to achieve high (generalization) performance on evaluation tasks after learning from a limited number of training tasks, where the evaluation and training distributions may differ; for instance, a policy for a robot may need to be trained in a simulated environment for safety reasons, but would need to be deployed on a physical device, a setting called sim2real. Discrepancy between task distributions is often recognized as a major reason why RL agents are yet to be applied pervasively in the real world  \citep{igl2019generalization}.

In the chapters describing the contribution of this thesis, we employ on an experimental setting that emphasizes the evaluation of zero-shot generalization skills. Intuitively, these experiments evaluate if an agent could truly learn generalizable skills, instead of relying on memorization. 

Our experiment settings often involve training on limited number of environments and testing on a whole distribution of unseen environments with the same nature as the training tasks (skills needed to finish the task remain consistent). To generalize well, the agents need to build learned skills which capture the consistent knowledge across tasks. I will introduce the details in the next chapters. 

\subsection{Temporal Difference Learning}
\label{sec:TD_learning}

Temporal Difference (TD) learning is a fundamental concept and methodology in credit assignment within approximate RL. Credit assignment refers to the challenge of determining which actions lead to a particular outcome \citep{minsky1961steps}, and in the case of TD, it involves associating returns with states or state-action pairs.

TD learning combines elements of Monte Carlo (MC) simulation and Dynamic Programming (DP) to enable efficient policy evaluation. Like MC methods, TD can learn directly from raw experience without access to a model of the environment’s dynamics. Like DP, TD methods \textbf{bootstrap}, updating estimates based partly on other learned estimates, without waiting for the final outcome.

While MC methods must wait until the end of an episode to update the value of $V(S_t)$ (only after $G_t$ is known), TD methods can update their estimates at each timestep. At timestep $t+1$, they immediately make an update using the observed reward $R_{t+1}$ and the estimate $V(S_{t+1})$. The simplest TD method performs the following update:

\begin{equation}
\label{eq:TD_1}
V(S_t) = V(S_t) + \alpha \left[ R_{t+1} + \gamma V(S_{t+1}) - V(S_{t}) \right]
\end{equation}
immediately on transition to $S_{t+1}$ and receiving $R_{t+1}$. The update rule \ref{eq:TD_1} is called the \textbf{$1$-step TD update}, where we recognize $R_{t+1} + \gamma V(S_{t+1}) - V(S_{t})$ as the \textbf{($1$-step) TD error} and $R_{t+1} + \gamma V(S_{t+1})$ as the \textbf{(TD-)update target}. Every $1$-step TD update can be understood as: walk towards the update target $R_{t+1} + \gamma V(S_{t+1})$ from the current (estimated) value $V(S_{t})$ with a step length of $\alpha \left[ R_{t+1} + \gamma V(S_{t+1}) - V(S_{t}) \right]$ (decreasing the distance by ratio $\alpha$). Note that the update target $R_{t+1} + \gamma V(S_{t+1})$ is also a random variable. 



\begin{coloredfact}{Convergence of $1$-step TD}{convtdzero}
Under the episodic setting, given an MDP and a policy $\pi$, either discounted or not, $1$-step TD achieves convergence to $v_\pi$ asymptotically (in tabular setting).
\end{coloredfact}

The flexibility of TD learning, as well as its convergence guarantees in tabular and linear cases, set itself as a foundation for most RL methods, keeping generalized policy iteration valid even without direct access to MDPs.

TD has also received attention in neuroscience because of its connections to the reward-prediction-error hypothesis. It was discovered that the firing rate of dopamine neurons in the ventral tegmental area and substantia nigra appear to mimic the error function in the algorithm \citep{schultz1997neural}.

Note that TD updates can also use targets constructed from value estimates over multiple steps, known as \textbf{multi-step TD}. However, we will omit a detailed discussion of these methods due to their limited relevance to this thesis.

\subsection{Off-Policy Evaluation}
\label{sec:off_policy_evaluation}

In approximate RL, possibly because of the limited agent-environment interaction budgets or others, an agent is often required to estimate the values of one policy when acting upon another. Let us call the policy to learn about the \textit{target policy}, and the policy used to generate behavior the \textit{behavior policy}. In this case, we say that learning is from data ``off'' the \textit{target policy}, and the overall process is termed off-policy learning, in contrast with on-policy learning, where the agents act accordingly to the values of the policy it estimates.

In \citet{watkins1992q}, the author proved that learning a Q-value estimator with $1$-step TD yields convergent value estimation results in tabular cases. And this even applies to off-policy learning if the update targets are constructed carefully using the target policy. This analysis, in theory, granted RL agents freedom to learn a Q-value estimator from any experience about the target policies. However, as later years has shown, when combined with function approximation, this approach still faces lots of challenges in practice, especially when the agent is asked to learn on experiences acquired by other agents, without access to their policies \citep{fujimoto2021minimalist}.

Off-policy learning can also be conducted with V-value estimators using importance sampling. However, I will skip further discussions, because it is mostly irrelevant to the contributions of this thesis.

\subsection{Q-Learning}
\label{sec:q_learning}

We now turn to Q-learning, an approximate RL method of learning optimal policy, through TD learning of state-action values, commonly referred to as \textbf{Q-values}. Q-learning is off-policy compatible, meaning it can learn from experiences generated by a policy different from the one being optimized. This makes Q-learning particularly flexible and widely applicable. Q-learning is an approximate RL algorithm that draws heavy similarity to value iteration used in DP, when access to the environment MDP is provided (Sec.~\ref{sec:VI}, Page.~\pageref{sec:VI}).

The Q-learning update rule is defined as follows:

\begin{equation}
Q(s_t, a_t) \leftarrow Q(s_t, a_t) + \alpha \left( r_{t+1} + \gamma \max_{a'} Q(s_{t+1}, a') - Q(s_t, a_t) \right)
\end{equation}

where $Q(s_t, a_t)$ is the estimated value following taking action $a_t$ in state $s_t$, $r_{t+1}$ is the reward received during the transition, $\alpha$ is the learning rate, equivalent to a step size and $\max_{a'} Q(s_{t+1}, a')$ is the maximum estimated Q-value for the next state, reflecting the maximum value following taking the best action in the next state.

For a Q-learning update, the combined term $r_{t+1} + \gamma \max_{a'} Q(s_{t+1}, a')$ is recognized as the  \textbf{update target} of Q-learning. In the variants of Q-learning, such as double Q-learning \citep{hasselt2015double}, the update target can be substituted with other terms. We will discuss this further in Sec.\ref{sec:DDQN} on Page.~\pageref{sec:DDQN}.

Q-learning converges to the optimal Q-values under certain conditions (\eg{}, sufficient exploration of the state space, decaying learning rate) and ultimately leads to an optimal policy in the tabular case \citep{watkins1992q}.

One of the most important modern deep RL methods, \textbf{Deep Q-Learning (\DQN{})}, combines Q-learning with \textbf{function approximation} using deep neural networks. We will introduce \DQN{} in detail in Sec.~\ref{sec:DQN} on Page.~\pageref{sec:DQN}.

\subsection{Function Approximation}
In classic MDPs and DP methods, states are typically enumerated in a tabular form, where all states are indexed and open-for-access explicitly. In this tabular setting, the agent can only be in one state at a time, and state estimates do not influence each other. While tabular approaches are useful for theoretical analysis, they become impractical in real-world scenarios, especially when the state space is large or continuous and cannot be easily discretized.

In the more real-world setting - the \textbf{function approximation} case, an agent may be instead given observations that are representations of the states, or may have to use a function approximator to infer the states from the history of interactions. In either case, it has to employ a value estimator to map the state representations into estimated values. 

In RL, function approximation primarily refers to using function approximators for value estimation, \ie{}, policy evaluation. While, policies themselves, could have function approximators of their own.

When using observational inputs for value estimation, function approximators can help construct state representations - transformed or abstracted versions of the environment's true state, which is often abbreviated as simply the \textbf{state}. A state refers to the full configuration of the environment at a given time, containing all relevant information for decision-making, \ie{}, the sufficient statistics; In contrast, a state representation is a processed version of this state. When a state representation is formed by the agent itself, it typically involves feature extraction, dimensionality reduction, or other abstractions to make it easier for the agent to learn and generalize. With deep learning, these representations are often lower-dimensional embeddings that capture essential features while discarding irrelevant information. A state representation can take the shape of a vector, real-valued or even binary, or even a set of unordered objects, as shown in Chap.~\ref{cha:CP} (Sec.~\ref{sec:set_encoder}, Page.~\pageref{sec:set_encoder}).

With function approximation, the two key problems of RL can be re-aligned towards representation (learning) and exploration, as suggested in \citet{amortila2024reinforcement}.

In the tabular setting, methods like Temporal Difference (TD) learning can converge exactly to the value function because updates to each state are independent. However, with function approximation, updates to one state’s value may affect the estimates of many other states, making exact convergence difficult. This introduces a challenge: improving the estimate of one state might degrade the estimates of others, especially when the number of states exceeds the learnable parameters. This poses a \textbf{generalization dilemma}: while interference between state estimates is problematic, it can also accelerate learning by improving estimates for similar states.

With optimization methods like gradient descent, surrogate losses (\eg{}, $L_1$ or $L_2$ distances) are often used to guide learning, typically in conjunction. This means, a loss that encourage the value estimates to be closer to their update targets will be established, and an optimizer will be applied to minimize such loss to achieve the convergence to the update targets. In contrast, vanilla TD learning does not seek to optimize a loss but provably converges in the respective cases.

The learnable parts of the approximator are often abstracted and compiled into a collection of learnable \textbf{parameters}, which are also sometimes called \textbf{weights}.

\subsection{Semi-Gradient Methods for Learning Function Approximators}

Popularly, we use differentiable value estimators $V(s; \bm{w})$ parameterized by a weight vector $\bm{w}$ to enable stochastic gradient-descent methods for approaching the update targets.


$\bm{w}$ can be updated at each of a series of discrete timesteps as before, $t \in \{1, 2, \dots\}$, trying to minimize the losses based on the value errors. Stochastic gradient-descent (SGD) methods do this by adjusting the weight vector based on each example by a small amount in the direction that would most reduce the error on that example:

\begin{align}
\label{eq:fa_true_gradient}
\begin{split}
\bm{w}_{t+1} \leftarrow  \bm{w}_{t} - \frac{1}{2} \alpha \nabla \left[v_{\pi}(S_t) - V_{\pi}(S_t, \bm{w}_t)\right]^2 = \bm{w}_{t} + \alpha \left[v_{\pi}(S_t) - V_{\pi}(S_t, \bm{w}_t)\right] \nabla V_{\pi}(S_t, \bm{w}_t)
\end{split}
\end{align}
where $\alpha$ is the learning rate, a positive step-size hyperparameter or \textbf{learning rate} for short.

Gradient descent methods are called ``stochastic'' when the update is done on only one or a few examples, selected stochastically. Over many steps, the overall effect is to minimize an average performance measure, such as the overall value error across all states.

Obviously, we cannot use (Eq.~\ref{eq:fa_true_gradient}) to do update because the true value $v_{\pi}(S_t)$ is unknown. Thus, we must replace the update target $v_{\pi}(S_t)$ with an estimate $U_{t}$. If $U_{t}$ is unbiased, \ie{}, $\doubleE[U_t | S_t = s] = v_{\pi}(S_t), \forall t$, then $\bm{w}_t$ is guaranteed to converge to a local optimum under the usual SGD conditions with decreasing $\alpha$. One simplest instance of this kind of method is to use the MC returns as the update targets, which leads to the gradient-MC method.

Having discussed the inefficiencies of using MC returns as the value estimation update targets, we naturally turn to the possibility of combining TD methods with function approximation. Unfortunately, despite TD's convergence in the tabular case, TD always employs biased targets, because TD update targets are constructed by  bootstrapping existing value estimates. Combined with the requirements from gradient descent, this implies that TD methods will not produce a true gradient method. Formally, it can be shown that bootstrapping methods are not in fact instances of true gradient descent \citep{barnard1993temporal}, as they take into account the effect of changing the weight $\bm{w}_t$ on the estimate ($\nabla V_{\pi}(S_t, \bm{w}_t)$ in Eq.~\ref{eq:fa_true_gradient}) but ignore its effect on the target ($\left[v_{\pi}(S_t) - V_{\pi}(S_t, \bm{w}_t)\right]$ in Eq.~\ref{eq:fa_true_gradient}).

The methods that combine TD-learning and gradient-based optimization are recognized as \textit{semi-gradient} methods, because they only take into consideration a part of the gradient. Although semi-gradient bootstrapping methods do not converge as robustly as gradient methods, they have shown generally good performance in application, especially the deep RL methods using neural networks as function approximators. 

\subsection{Off-policy Methods with Function Approximation}

When learning off-policy with function approximation, new troubles emerge for semi-gradient methods. First, the update targets may need to be fixed with importance sampling ratios, depending on if a Q-value estimator is used; Second and most importantly, the state distribution will no longer match the target policy. 

Let us first look into why off-policy learning is more difficult with function approximation compared to the tabular case. With function approximation, value estimation becomes dependent on state representations. Thus, the updates for one state could affect multiple states with similar representations, whereas in the tabular case, the updates for one state have no influence on others. This means that, in the off-policy case, the tabular updates do not have to care about the state-frequencies when doing updates as long as the update targets are fixed using the importance sampling ratios. The blessing that the tabular case updates do not rely on any special distribution for stability has not been passed to the function approximation cases. In the function approximation case, the semi-gradient methods that we have introduced before rely on the state-frequencies for updates. This means we either have to ``reweight'' the updates, \ie{}, to warp the update distribution back to the on-policy distribution using importance sampling methods, or we have to develop true gradient methods that do not rely on any special distribution for stability, which are not yet available for non-linear function approximators such as neural networks \citep{zhao2020meta}. In fact, the problem of the coexistence of bootstrapping, off-policy learning and function approximation is so troublesome that it is considered as ``the deadly triad'', which is known for the resulting divergent value estimations \citep{sutton2018reinforcement}.

\section{Temporal Abstraction}
As discussed, the RL formulation of sequential decision-making relies on a one-step state transition model (decisions are made at the most atomic timesteps), where the action taken at timestep $t$ impacts the state and the immediate reward at $t + 1$. This prompts RL methods to overwhelmingly learn to directly work on these finest-grain courses of action. 

However, to be able to efficiently and effectively reason longer-term in the face of novelty, an agent must have the ability to utilize and behave, according to different appropriate timescales \citep{sutton2018reinforcement}. This is because an agent can learn more effectively if uninteresting details are abstracted away, and focus its computations on important decision timings. This is coupled with the fact that some predictions about the environment are counterintuitively more difficult in the finer timescales than the more coarse, where in the latter case, unnecessary details may be in a way effectively marginalized. This, together with the fact that the accumulation of errors by imperfect models during multistep planning, motivate the learning for temporal abstractions in the framework of RL, \eg{} options.

Learning temporal abstractions is a longstanding problem and existing methods cover a variety of approaches, \eg{} discovering partially defined policies skills \citep{thrun1994finding}, learning more and more complex behaviors using temporal-transition hierarchies \citep{ring1997child}, a feudal approach where high level managers learn to allocate tasks to their sub-managers, which in turn learn how to satisfy them \citep{dayan1992feudal}, a framework to consider the augmented MDP but also the underlying MDP in a seamless fashion \citep{sutton1999between}, \etc{}. 

We lay the foundations of temporal abstractions including options and options models, discuss the discovery problem, then review both model-based (planning with option models) and model-free (option discovery) algorithms relevant to the contributions of this thesis.

\subsection{Semi-Markov Decision Processes (SMDPs)}
\label{sec:SMDP}

A Semi-Markov Decision Process (SMDP) provides a foundation for temporal abstractions, where, compared to MDPs, the amount of time between two decision points can vary \citep{puterman2014markov}. In SMDPs, the transition functions $\doubleP\{S^\tau = s' | S^0 = s, \pi_i\}$ are defined additionally over a \textit{transition time} $\tau$, for an agent to enter a next interesting state $s'$ from $s$. 

\phantomsection
\label{sec:macro_actions}

Building upon SMDPs, we can replace the action spaces with ``macro'' actions induced by different policies, abstracting the decision-making process, which leads to the options framework. These macro actions can also be nested, \ie{}, consisted of a sequence of other macro actions, naturally resulting in a potential of hierarchies corresponding to different timescales. The aim of hierarchical RL is to find closed-loop policies at several levels of abstraction, also known as temporally-extended actions. Discovering useful and reusable temporally extended actions are core to temporal abstraction.

Instead of primitive actions, let us consider an agent with a set of policies to choose from at each decision-time, \ie{}, the action space is replaced with a set of policies. Let the cumulative discounted reward under $\pi_i$ be $R_{s}^{\pi_i}$. Considering the case, where the decision time for actions are only at discrete events, we have:

\begin{coloredfact}{Bellman Optimality Equations for SMDP}{smdpbellman}
\begin{equation}\label{eq:bellman_operator_SMDP_V}
V^*(s) = \max_{\pi_i}{\left[ R_{s}^{\pi_i} + \sum_{\tau=1}^{\infty}{\gamma^{\tau - 1}\cdot \sum_{s'}{\doubleP\{S^\tau = s' | S^0 = s, \pi_i\} V^*(s')}}   \right]}
\end{equation}

\begin{equation}\label{eq:bellman_operator_SMDP_Q}
Q^*(s,\pi_i) = R_{s}^{\pi_i} + \sum_{\tau=1}^{\infty}{\gamma^{\tau - 1}\cdot \sum_{s'}{\doubleP\{S^\tau = s' | S^0 = s, \pi_i\} \max_{\pi_{i}^{'}}{Q^*(s',\pi_{i}^{'})}}}
\end{equation}
\end{coloredfact}

The equations above form sequences of actions defined over macro-actions. A sequence of actions forming a ``macro'' is one of the simplest kinds of abstraction. A macro can also be obtained as a sequence of other macros, which naturally results in a hierarchy in architecture. The equations indicate the potential of using macro-actions in sequential decision-making problems, giving rise to option-based frameworks. 

\subsection{Options: SMDPs with Policies as Macro Actions}
\label{sec:options}
Temporally extended actions are usually defined over a subset of the state space, with the primary aim to reduce the number of steps needed for the agent to solve a task. It serves as motivation to learn abstractions, which are partial solutions to a task that could be reused for other tasks.

Options are a way to achieve temporal abstraction in RL, \ie{}, finding useful action sequences that span over multiple decision intervals \citep{sutton1999between}. The usefulness of these options can be evaluated by their robustness, re-usability, \etc{}. The appeal of options is that they are in some ways interchangeable with actions. Temporal abstraction allows agents to use sub-policies, and to model the environment over extended time scales, to achieve both better generalization and the divide and conquer of larger problems.

Each option comprises a way of behaving (a policy) and a way of stopping. Formally, for any set of options defined on any MDP, the decision process that selects only among those options, executing each to termination, is a SMDP \citep{puterman2014markov}. An SMDP consists of 1) a set of states $\scriptS$, 2) a set of options $\mathcal{O}$, 3) for each pair of state and option, an expected cumulative discounted reward, and (4) a well-defined joint distribution of the next state and transit time \citep{bacon2017option}. Naturally, options give rise to option-associated dynamic functions, value functions, Bellman optimality equations, \etc{}

\citet{sutton1999between} demonstrated the empirical potential to plan and learn at multiple time scales in the options framework, indicating options' effectiveness for speeding up learning, improving robustness and generalization abilities, \etc{}. However, the original experiments in \citet{sutton1999between} required the designer to use prior knowledge about the task to add pre-defined options, either providing the option-specific reward functions, or providing complete option policies.

This raises the critical question of where the options should come from, giving rise to the \textbf{option discovery} problem. Besides principled approaches such as option-critic \citep{bacon2017option}, common approaches to option discovery involve posing subsidiary tasks such as reaching a bottleneck state or maximizing the cumulative sum of a sensory signal other than reward. Given such subtasks, the agent can develop temporally abstract structure for its cognition by following a standard progression in which each subtask is solved to produce an option, the option’s consequences are learned to produce a model, and the model is used in planning. Interestingly, option discovery problems can be bypassed to a certain degree, as what my collaborators and I have done for Chap.~\ref{cha:skipper} with goal-conditioned planning.

Just as we can model the consequences of primitive actions and plan accordingly, so we can learn and plan with models of options’ effects. However, this will not be as straight forward as operating on primitive actions. I will expand on this point in the next chapter.

\subsection{Goal Conditioned RL \& Options}

Notably, goal-conditioned RL seeks to train agents to achieve a goal or a sequence of goals, and hence can be viewed as a formulation of instantiating option learning whose policies are shaped towards achieving certain outcomes \citep{sutton2023reward}.

In some sequential decision-making problems, what matters is the achievement of certain goals instead of the maximization of returns. Researchers have thus tried to design reward functions that would align with the objectives of the problems. However, this remains an open-problem since the alignment can be non-trivial to establish, \ie{}, maximizing the accumulation of the designed rewards often do not lead to the achievement of the important goals.

In Chap.~\ref{cha:skipper}, we will focus on goal-conditioned options, where the initiation set covers the whole state space $\scriptS$. Each such option is a tuple $o = \langle \pi, \beta \rangle$, where $\pi: \scriptS \to \text{Dist}(\scriptA)$ is the (intra-)option policy and $\beta: \scriptS \to \{0, 1\}$ indicates when a goal state is reached.

%% file: chapter_3_preliminary.tex
\chapter{Literature Review: Model-Based Deep RL}
\label{cha:prelim}
\textit{\small This chapter presents the discussions of related works to the contributions of this thesis, as well as some more advanced preliminary background knowledge for understanding the methods used in later chapters.}

\minitoc

\section{Deep RL: Neural Network based Methods}
When function approximators in approximate RL methods are implemented with Neural Networks (neural nets, NNs)s, the resulting agents are often recognized as \textbf{Deep RL} agents, or \textbf{DRL} agents for short. This means, at least the value estimator will be parameterized as a neural network and will be optimized using a surrogate loss with certain neural network optimizers. The neural network optimizers are mostly based on gradient descent, \ie{}, evolving the parameters roughly towards the direction of lower losses following the suggestion of the gradient directions. Gradient descent in neural nets is efficiently implemented with the Back-Propagation (BP) algorithm, which takes advantage of the nature of neural nets being composite functions and the chain rule of gradients \citep{rumelhart1986learning}.

Neural nets are a popular choice for function approximator in RL, because of their abilities to learn complex mappings from observations or state representations to actions or values. While tabular RL algorithms struggle with high-dimensional observation / state / action spaces, neural networks allow for efficient generalization in large or continuous spaces by learning compact, useful representations, enabling RL agents to solve previously intractable problems. Notably, neural nets can approximate value functions (\eg{}, Q-values or V-values) or policy functions. This ability paved the way for the renaissance of modern DRL, including algorithms like Deep Q-Networks (\DQN{}, to be introduced in detail soon) and policy gradient methods\footnote{We will skip the discussions of policy gradient methods in this thesis, since they are not too relevant to the contributions in the following chapters.}.

Despite being subjected to the dangers of the deadly triad, we know from practice that DRL's value estimations, once meticulously tuned, can acquire convergence. Indeed, when we discuss DRL methods, we are most likely in a territory without guarantees.

To prepare the readers for a clear understanding of the methods used in the following chapters, we introduce one of the most impactful RL method - \DQN{} \citep{mnih2015human}, that is fundamental to DRL methods dealing with discrete action spaces.

\subsection{\texorpdfstring{\DQN{}}{DQN}}
\label{sec:DQN}

Based on Q-Learning (introduced in Sec.~\ref{sec:q_learning}, Page.~\pageref{sec:q_learning}, originally proposed in \citet{watkins1992q}), Deep Q-Network (\DQN{}) is a DRL method operating in discrete action spaces, based on off-policy Q-value estimation without an explicitly parameterized policy \citep{mnih2015human}. Before the emergence of \DQN{}, DRL methods suffered greatly from training instabilities and sensitivity to hyperparameters. As a truly groundbreaking contribution, \DQN{} achieved generally human-level performance on Atari games, and kickstarted the consequent progress in DRL. \DQN{} set itself apart with the following features:

\textbf{State Encoder \& Value Estimator as Neural Networks}: both the state representation encoder and the value estimator on top are implemented with neural networks and optimized with neural network parameter optimizers. \DQN{} is fully parameterized, with learned state representations extracted from high-dimensional pixel-based observations.

\textbf{Target Network for Training Stability}: the agent maintains a time-delayed clone of the ``policy network'', \ie{}, the bundle containing the state encoder and the value estimator. The clone is named the ``target network'' and is responsible for providing TD update targets for value estimator training (of the policy network), based on the experimental observation that this would produce more stable update targets for value estimator learning. Periodically, the parameters of the target network will be synchronized with the latest parameters in the policy network.

\textbf{Q-learning with Training Loss}: In \DQN{}, the value estimator is updated by the gradient descent-based optimizer to minimize a loss function. \DQN{} uses surrogate loss functions, such as $L_2$ or Huber loss, to pull the estimated values of the current state towards an update target $y$ constructed in the Q-learning fashion (introduced in Sec.~\ref{sec:q_learning}, Page.~\pageref{sec:q_learning}). For example, over a sampled transition $\langle s, a, r, s', \omega'\rangle$, where $\omega'$ is a binary indicator of if the next state $s'$ is terminal, the simple $L_2$ surrogate loss takes the following form:

$$\scriptL_{\text{\DQN{}}} \coloneqq (\hat{Q}_\theta(s,a) - y)^2$$

where $y$, in \citet{mnih2015human} (the original \DQN{} paper), is an update target constructed with the help of the target network $\theta'$:

\begin{singlespace}
\begin{equation}
\label{eq:DQN_target}
y_{\text{\DQN{}}} \coloneqq \begin{cases}
r &\text{if } \omega \text{ is true}\\
r + \gamma \max_{a'}{\hat{Q}_{\theta'}(s',a')} &\text{otherwise}\\
\end{cases}
\end{equation}
\end{singlespace}

\phantomsection
\label{sec:DDQN}

An alternative method of constructing update targets, named double \DQN{} or \DDQN{} for short, has shown more promising performance against the overestimation problem of Q-learning induced by the $\max$ operator \citep{hasselt2015double}. With \DDQN{}, the update target is constructed jointly by the policy network $\theta$ and the target network $\theta'$:

\begin{singlespace}
\begin{equation}
\label{eq:DDQN_target}
y_{\text{\DDQN{}}} \coloneqq \begin{cases}
r &\text{if } \omega \text{ is true}\\
r + \gamma \hat{Q}_{\theta}(s',\argmax_{a'}{\hat{Q}_{\theta'}(s',a')}) &\text{otherwise}\\
\end{cases}
\end{equation}
\end{singlespace}

Note that notations are abused here for simplicity: since both the policy and the target networks have their own paired state encoder, thus the $s'$ input of $\hat{Q}_{\theta}$ and $\hat{Q}_{\theta'}$ are different, \ie{}, they are produced by their respective state encoders. 

\begin{figure}[htbp]
\centering
\captionsetup{justification = centering}
\includegraphics[width=0.8\textwidth]{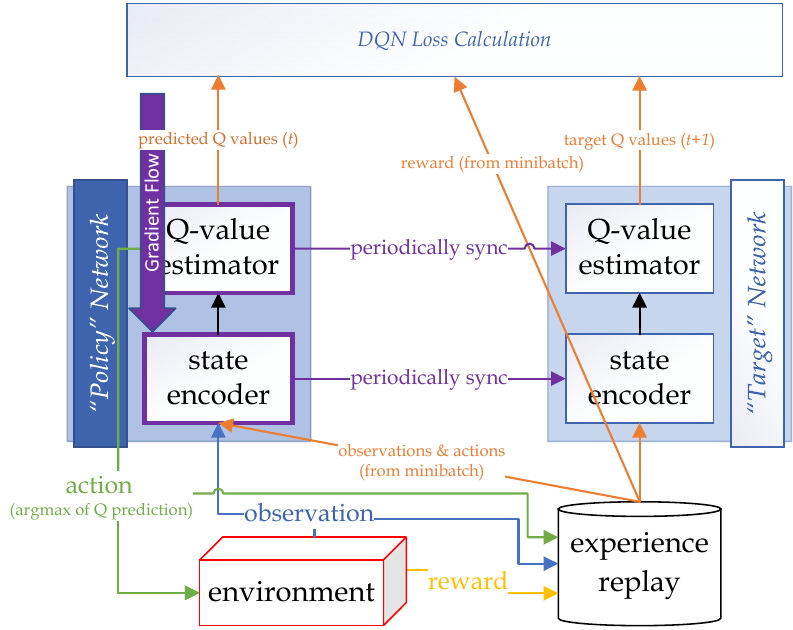}
\caption[Dissection of \texorpdfstring{\DQN{}}{DQN}]{\textbf{Dissection of \DQN{}}: The policy network and the target network architectures are identical, both consists of a state encoder and a Q-value estimator. The target network is not trained, but updated periodically by synchronizing the parameters from the policy network. Only the policy network is trained through gradient descent-based optimization, and it can be extracted as an inference-only agent, \ie{}, all components except the policy network are for training purposes only. While in the literature, we often see the policy network as one whole unit, for more unified discussions, we decompose the policy network into the two components of the state encoder and the Q-value estimator.}
\label{fig:DQN}
\end{figure}

\textbf{Stores and Trains on Transitions with Experience Replay}: \DQN{} stores the interaction history as transitions in the experience replay, and samples the transitions in minibatches to conduct optimization based on batched stochastic gradient descent. An \textbf{experience replay} is a buffer for the interaction history between the agent and the environment. The data buffered in the experience replay can be later used for learning purposes, and can be organized in different ways, such as transitions (in \DQN{}) or trajectories (in other methods). Agents that do not require the assistance of experience replay are often distinctively recognized as streaming methods \citep{elsayed2024streaming}.

\textbf{$\epsilon$-greedy Exploration}: \DQN{} employs $\epsilon$-greedy exploration policy as the behavior policy. Let the target policy given the current value estimate be $\pi$, the $\epsilon$-greedy is defined as:

\begin{singlespace}
\begin{equation*}
\pi_{\epsilon\text{-greedy}} = \begin{cases}
\text{uniform random policy} &\text{\textit{w.p.} } \epsilon\\
\pi(s) &\text{otherwise}\\
\end{cases}
\end{equation*}
\end{singlespace}

Note that because of \DQN{}'s compatibility with off-policy learning, its implementations often take advantage of a controlled annealing from $\epsilon=1.0$ to a very small value to control the exploration-exploitation tradeoff within a limited agent-environment interaction budget. Detailed implementations can be flexible.

An overall illustration of \DQN{}'s training and behaviors is presented in Fig.~\ref{fig:DQN}.

\subsection{Distributional RL}
\label{sec:distoutputs}

Distributional RL is a useful DRL technique that enables the estimation of the \textit{distribution} of returns, instead of only the scalar expectation of the return. Distributional RL can be similarly analyzed through the lens of Bellman operators and contractions (as suggested in Sec.~\ref{sec:contraction}, Page.~\pageref{sec:contraction}), which will show that the convergence guarantees of scalar RL updates are preserved \citep{bellemare2017distributional}.

\begin{figure}[htbp]
\centering
\captionsetup{justification = centering}
\includegraphics[width=0.5\textwidth]{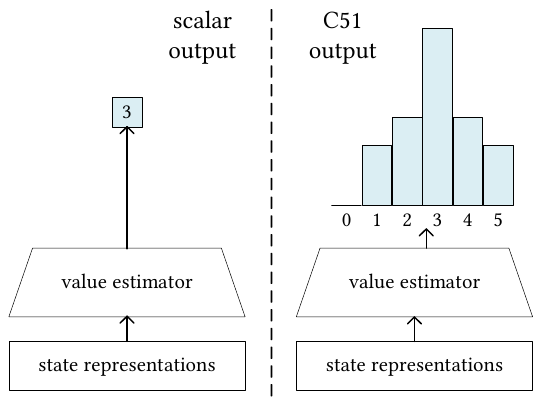}
\caption[C51 Distributional Variant of \texorpdfstring{\DQN{}}{DQN} Value Estimators]{\textbf{C51 Distributional Variant of \DQN{} Value Estimators}: Q-value estimators output a prediction of values based on the input pair $\langle s, a \rangle$. For vanilla \DQN{}, the output is a scalar; For C51 \citep{bellemare2017distributional}, the output is similar to histograms where the bins within a certain range of support are preset. For the right side (C51), the support is set to be $\{0, 1, 2, 3, 4, 5\}$, and the outputs are \softmax{}-ed logits of the $6$-dimensional outputs from the value estimator.}
\label{fig:distributionals}
\end{figure}

In short, distributional RL changes the architecture of the value estimator's function approximator from having a scalar output to having a vectorized output, where the vectorized output is a representation of certain estimated distributions, as shown in Fig.~\ref{fig:distributionals}. Naturally, the old surrogate losses used for scalar prediction must be also replaced, so they can help the predicted distributions converge to the update target distributions with the help of the gradient descent-based optimizers, depending on how the distributions are represented. 

There are several popular choices for the representation of the estimated distributions. For the contents of this thesis, we are particularly interested in the so-called C51-style distributional outputs \citep{bellemare2017distributional}. 

C51 outputs a histogram-like representation of a discretized distribution over a preset support over certain value ranges. A histogram over the preset support will be the output, where the bins of the histograms do not necessarily have to be uniform. The expectation can be recovered by a weighted sum over the preset support. The surrogate losses for convergence take the forms of Kullback-Leibler divergence (KL-divergence) and the update targets are the histograms converted from the scalar update targets bootstrapping the weighted-sum expectation estimates (Eq.~\ref{eq:DQN_target} \& Eq.~\ref{eq:DDQN_target}).


\phantomsection
\label{sec:auxiliary_learners}

C51 greatly alleviates the problem introduced by the difference in the magnitude of outputs, and thus agents with auxiliary learners (that learn to estimate not only the values) are less subjected to conflicts of training losses due to imbalanced magnitudes. \textbf{Auxiliary learners} are additional estimators introduced during the training, typically used to improve performance in a specific way or help the model learn more robust features. Auxiliary learners and their training signals - auxiliary losses, may not be the primary objective of the model, but are designed to assist the primary learner (the primary loss function) by providing supplementary guidance. Sometimes in RL, only the value estimator is recognized as the non-auxiliary learner, among all learners in the RL system.

In this thesis, we by default used a variant of C51 distributional TD learning used in \citet{schrittwieser2019mastering} on all \DQN{}-based baseline and agents. That is, the estimators output histograms instead of scalars. We regress the histogram towards the targets, where these targets are skewed histograms of scalar values, towards which KL-divergence is used to train. Note that this variant of C51 does not use the full distributional estimates of the subsequent states to construct update targets that span the full support, rather, only converting the scalar update targets into two-hot histograms \citep{schrittwieser2019mastering}. The variant was preferred for its flexibility of not having to know the full distributions of the update targets, useful in planning situations.

Also, in Chap.~\ref{cha:skipper}, we proposed a technique upon C51 distributional outputs, to interchangeably learn the distributions of cumulative discounts and cumulative distances simultaneously.

\subsection{Goal-Conditioned RL: Source-Target Pairs \& Hindsight Relabeling}
\label{sec:source_target_pair_hindsight_relabeling}
Goal-conditioned RL is an extension of traditional RL where an agent's objective is to achieve specific goals rather than merely maximizing cumulative rewards. In MDPs, a goal is defined to be a set of states that match certain criteria, where such state set can be empty, a singleton or even infinite \citep{ghosh2018learning}.

In goal-conditioned RL, the agent conditions its behaviors additionally on a goal (\eg{}, a target state or a desired outcome) at each timestep. This is to say that goal-conditioned agents must learn a \textbf{goal-conditioned policy} that, given both the current state and the goal, takes actions to achieve the goal effectively. The goals could come from the agent itself, or an external instructor that is a part of the environment, \etc{}. 

Goal-conditioned policies enable more flexible and adaptable learning, as the agent can generalize to different tasks by conditioning its behavior on various goals, rather than learning a separate policy for each task. From the viewpoint of options (introduced in Sec.~\ref{sec:options}, Page.~\pageref{sec:options}), this also mean that one single goal-conditioned policy may be used to substitute a whole set of options. Also, evidently, a goal-conditioned policy can also be viewed as a special case of constrained option, whose initiation set is the whole state space $\scriptS$ and the termination condition is the goal criterion.

Some goal-conditioned agents can be viewed from the perspectives of the feudal RL framework, a hierarchical RL framework in which a high-level \textit{manager} provides low-level goals to its \textit{workers} \citep{dayan1992feudal,parr1997hierarchies,vezhnevets2017feudal}. Feudal RL framework is highly similar to the target-assisted planning framework, which my collaborators and I have abstracted for existing planning agents that utilizes generative models to produce targets to achieve (in Chap.~\ref{cha:delusions}, Page.~\pageref{cha:delusions}), whose emphases are rather on the generator-estimator duo, inspired by the belief formation and belief evaluation systems that underpin the delusion mechanisms in the human brain.

Goal-conditioned agents often rely on ``source-target pairs'' for training, where ``source'' often represent a current state, \ie{}, the state that the agent should make decisions from, and the ``target'' correspond to one of the states that match the goal (a state that belongs to the set of states that match the goal criterion). Source-target pairs are organized training data that could be used to let the agents learn about the relationship between one state and another. Note that pairs do not need to be explicitly stored: for example, two ordinarily sampled transitions can be trivially assembled during training to form a source-target pair, as follows

$$\langle s_1,a_1,r_1,s_1',w_1'\rangle \bigoplus \langle s_2,a_2,r_2,s_2',w_2'\rangle \rightarrow \langle s_1,a_1,r_1,s_1',w_1', s_2\rangle$$

where $s_1$ acts as the source state and $s_2$ acts as the target state.

There are many ways to construct a source-target pair, but undoubtedly, \textbf{Hindsight Experience Replay (HER)} is among the most popular \citep{andrychowicz2017hindsight}. HER's core mechanism is \textbf{hindsight relabeling}, where agent-environment interaction histories buffered in the experience replay while following their respective contemporary goals are relabeled with other goals (the ones that the agent was not following). Intuitively, this makes the agent pretend that certain experience was gathered when following the relabeled goals. 

Training target-assisted planning agents with only contemporary targets (the ones being followed) can lead to poor performance \citep{dai2021diversity}, since contemporary targets may be low-quality, hard to achieve, or lack in diversity \citep{moro2022goaldirected,davchev2021wish}. The fact that hindsight relabeling creates much more diverse source-target pairs (compared to those with only the contemporary goals as targets) for the goal-conditioned estimators to generalize and gain deeper understanding of the relationship inside the source-target pairs, which in turn contributed to its massive success in training goal-conditioned policies.

From another perspective, HER is crucial for enhancing \emph{sample efficiency} in goal-directed RL \citep{andrychowicz2017hindsight}, as it enables learning from failed experiences \citep{dai2021diversity}. We will discuss HER more in Sec.~\ref{sec:related_work_HER} (Page.~\pageref{sec:related_work_HER}), about its connections with our contributions and its existing limitations.

Formally, HER augments a transition $\langle x_t, a_t, r_{t+1}, x_{t+1} \rangle$ with an additional observation $x^{\odot}$ (or its encoding), the relabeled target, creating a source-target pair, with one decision point for the current step and another for the relabeled target. \textbf{Relabeling strategies}, which correspond to how $x^{\odot}$ is selected, are critical for the performance of HER-trained agents \citep{shams2022addressing}. Most popular choices are \textit{trajectory-level}, meaning $x^{\odot}$ comes from the same trajectory as $x_t$. These include \futurestr{}, where $x^{\odot} = x_{t'}$ with $t' > t$, and \episodestr{}, with $0 \leq t' \leq T_\perp$.

The introduction of HER greatly enhanced the sample efficiency of learning about experienced targets. Meanwhile, the incompleteness of the accompanying relabeling strategies planted a hidden risk of delusions towards hallucinated targets, which will be discussed later in Chap.~\ref{cha:delusions} (Page.~\pageref{cha:delusions}), where we will also discuss relabeling strategies in detail.

\section{Deep Learning Preliminaries}
\subsection{Generative Models: VAEs \& Conditional VAEs}
\label{sec:VAE}

\begin{figure}[htbp]
\centering
\captionsetup{justification = centering}
\includegraphics[width=0.9\textwidth]{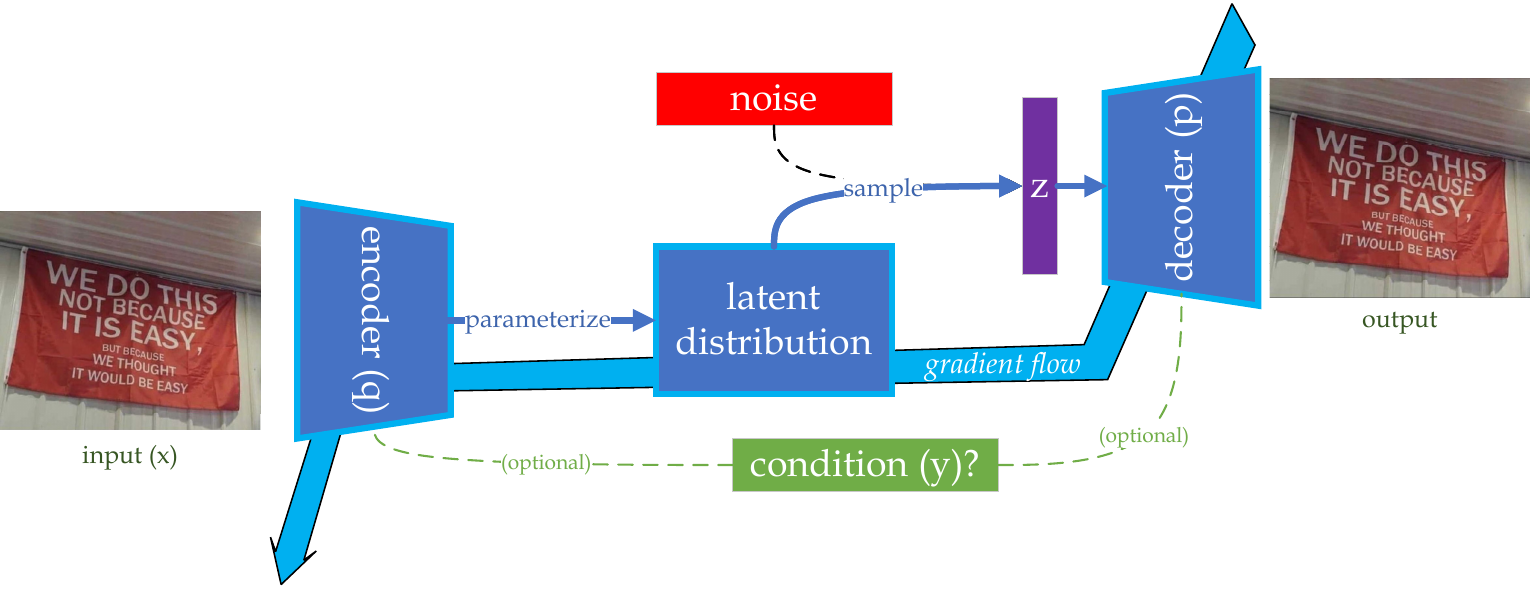}
\caption[VAE Mechanisms]{\textbf{VAE Mechanisms}: the encoder $q$ extracts parameters that define the distribution of the latent variable $z$, which is sampled and then used with the decoder $z$ to parameterize the likelihood of the reconstructed output $\hat{x}$. The gradient flows back from the output, to the decoder and the encoder. The gradient flow through the sampled $z$ is enabled by reparameterizing $z$ with the help of an external noise \citep{kingma2013auto}.
}
\label{fig:VAE_explanation}
\end{figure}

Aiming to maximize the likelihood of the observed data through a latent variable model, Variational Autoencoders (VAEs) are a type of generative model that learn to approximate the distribution of training data. Given an observed dataset $\{ \bm{x} \}$, a VAE models the joint distribution $p(\bm{x}, \mathbf{z})$, where $\mathbf{z}$ represents its discovered latent variables \citep{kingma2013auto}. We illustrate the mechanism of VAEs in Fig.~\ref{fig:VAE_explanation}.

Despite the effort to model the joint $p(\bm{x}, \mathbf{z})$, the ultimate objective of a VAE is to maximize the marginal likelihood \( p(\bm{x}) \), which is given by:

$$p(\bm{x}) = \int{p(\bm{x} | \mathbf{z}) p(\mathbf{z})} \, d\mathbf{z}$$

However, directly computing this integral is intractable. To address this, VAEs use variational inference to approximate the true posterior $p(\bm{z} | \bm{x})$ with a simpler, tractable surrogate $q(\bm{z} | \bm{x})$.


Such variational approach involves maximizing the Evidence Lower Bound (ELBO), which is a lower bound on the log-likelihood of the data, which a VAE seeks to learn. We can derive the ELBO from the log of the marginal likelihood $p(\bm{x})$:

\begin{align}
\begin{split}
\log p(\bm{x}) & = \log \int p(\bm{x} | \mathbf{z}) p(\mathbf{z}) \, d\bm{z} = \log \int \frac{q(\bm{z} | \bm{x})}{q(\bm{z} | \bm{x})} p(\bm{x} | \mathbf{z}) p(\mathbf{z}) \, d\bm{z} \\
& = \mathbb{E}_{q(\bm{z} | \bm{x})} \left[ \log \frac{p(\bm{x}, \mathbf{z})}{q(\bm{z} | \bm{x})} \right] + \text{KL}(q(\bm{z} | \bm{x}) \| p(\bm{z} | \bm{x}))
\end{split}
\end{align}





The first term is the ELBO, and the second term is the Kullback-Leibler (KL) divergence between the variational distribution and the prior over the latent variables, which will be discarded due to intractability. The ELBO can be expanded into two terms:

\[
\text{ELBO}(\theta, \phi) = \mathbb{E}_{q(\bm{z} | \bm{x})} \left[ \log p(\bm{x} | \mathbf{z}) \right] - \text{KL}(q(\bm{z} | \bm{x}) \| p(\mathbf{z}))
\]

In practice, the VAE learns the parameters \( \theta \) and \( \phi \) of the neural networks that parameterize \( p(\bm{x} | \mathbf{z}) \) and \( q(\bm{z} | \bm{x}) \), respectively, by maximizing the ELBO. 


Maximizing the ELBO encourages the model to generate accurate reconstructions of the data while keeping the variational posterior close to the prior distribution. Depending on how the latent space is defined, the ELBO can be instantiated in different forms, some even with closed form solutions, such as in \citet{kingma2013auto}, where an encoder extracts from input the parameters for defining a normal distribution. In Chap.~\ref{cha:skipper}, we used the encoder to instead parameterize a discrete bottleneck consists of a bundle of categorical distributions, from which a joint sample is treated as $z$, used as a partial description of a state (see Sec.~\ref{sec:skipper_discrete_partial_desc}).

A hyperparameter $\beta$ can be introduced to weight the two terms to control the learning, giving rise to $\beta$-VAE \citep{higgins2017beta}:

$$ \mathcal{L}(\theta, \phi, \beta) = -\mathbb{E}_{q(\bm{z} | \bm{x})} \left[ \log p(\bm{x} | \mathbf{z}) \right] + \beta \cdot \text{KL}(q(\bm{z} | \bm{x}) \| p(\mathbf{z})) $$

Here, $\beta$ is a scalar that controls the trade-off between the reconstruction term and the KL divergence term. When \( \beta = 1 \), this reduces to the standard VAE loss. 



The \textbf{Conditional VAE (CVAE)} is an extension of standard VAE, where both the encoder and the decoder are conditioned on additional information, such as class labels or other external attributes. This allows the model to generate data conditioned on specific variables, making it suitable for tasks like conditional goal generation, where the generated goals depend on information collected up to the current state.

Formally, in a CVAE, the objective is to learn the conditional distribution $p(\bm{x} | \bm{y})$. This is done by learning the joint distribution $p(\bm{x}, \bm{y})$, where $\bm{y}$ is the condition, by marginalizing over latent $z$ in the joint $p(\bm{x}, \bm{y}, \mathbf{z})$. The encoder and decoder are modified as follows:

\textbf{Encoder}: The distribution $q(\bm{z} | \bm{x}, \bm{y})$ is conditioned on both the data $\bm{x}$ and the condition $\bm{y}$.
   
\textbf{Decoder}: The likelihood $ p(\bm{x} | \mathbf{z}, \bm{y})$ is conditioned on both the latent $\mathbf{z}$ and the condition $\bm{y}$.

The ELBO inequality for CVAE is:

$$
\log p(\bm{x} | \bm{y}) \geq \mathbb{E}_{q(\bm{z} | \bm{x}, \bm{y})} \left[ \log \frac{p(\bm{x}, \bm{z} | \bm{y})}{q(\bm{z} | \bm{x}, \bm{y})} \right]
$$

With the same algebra, this results in the following objective function (for minimization):

$$
\mathcal{L}_{\text{CVAE}}(\theta, \phi) = -\mathbb{E}_{q(\bm{z} | \bm{x}, \bm{y})} \left[ \log p(\bm{x} | \mathbf{z}, \bm{y}) \right] + \text{KL}(q(\bm{z} | \bm{x}, \bm{y}) \| p(\mathbf{z}))
$$

In Chap.\ref{cha:skipper} and Chap.\ref{cha:delusions}, we use CVAE in conjunction with hindsight experience replay to train generative models that can imagine distance future states, conditioning on the current state.

\subsection{Attention}
\label{sec:attention}

Attention mechanisms, first popularized by \citet{bahdanau2014neural} in the context of machine translation, have evolved to become one of the most powerful and widely used tools in machine learning. Allowing models to focus on different parts of the input sequence or data, the ability of attention has made it central to language models powered by the Transformer architecture \citep{vaswani2017attention}. The ability to query sets of objects with varying degrees of attention has led to improvements in tasks such as question answering \citep{devlin2018bert}, image captioning \citep{xu2015show}, and even RL \citep{chen2021decision}. Through the introduction of mechanisms like multi-head attention and semi-hard attention, these models can capture more complex patterns and relationships within data, enabling a wide range of applications. One of the core contributions of this thesis, the consciousness-inspired - ``spatial abstraction'' mechanism, is implemented as a top-down semi-hard attention bottleneck, in Sec.~\ref{sec:CP_bottleneck} (Page.~\pageref{sec:CP_bottleneck}) and improved to work in synergy with temporal abstraction in Sec.~\ref{sec:skipper_spatial_abstraction} (Page.~\pageref{sec:skipper_spatial_abstraction}).

\begin{figure}[htbp]
\centering
\captionsetup{justification = centering}
\includegraphics[width=0.9\textwidth]{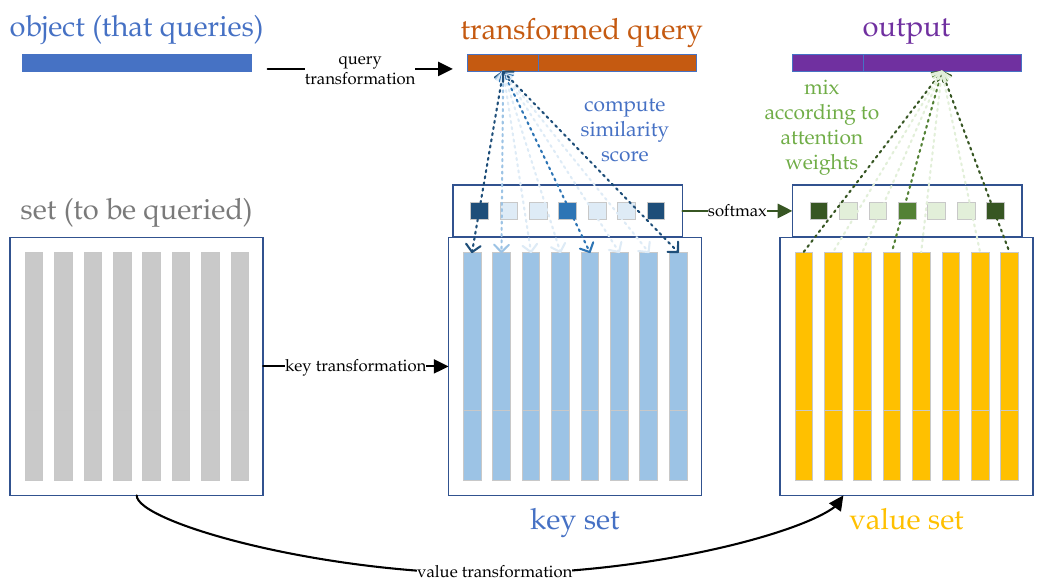}
\caption[Querying A Set with an Object]{\textbf{Querying A Set with an Object}: the \textit{object} is first transformed into a \red{query vector}. Then, \darkgreen{attention weights} are computed based on the query's \blue{similarity} to the transformed keys (from the set of objects being queried). Finally, the \orange{transformed values} are fused to form the \purple{output} according to the \darkgreen{attention weights}). Top-$k$ attention (Sec.~\ref{sec:semihard_attention} on Page.~\pageref{sec:semihard_attention}) can be implemented by setting the non-top-$k$ \darkgreen{attention weights} to $0$, or equivalently the non-top-$k$ scores to $-\infty$, due to the choice of \softmax{}.
}
\label{fig:attention_query}
\end{figure}

To understand formally how attention works, we now revisit a generic set query procedure (Routine \ref{rtn:objectqueryset}), which is also illustrated in Fig.~\ref{fig:attention_query}.

\begin{coloredroutine}{Querying a Set of objects $\{ \bm{x}_i \}$ with an Object $\bm{x}$}{objectqueryset}
\begin{singlespace}
\begin{enumerate}[leftmargin=*]
    \item The object $\bm{x}$ is transformed into a \textbf{query vector}. This is generally done via a linear transformation, typically denoted as:
    
    $$\bm{q} = W_q \bm{x}$$
    
    where \( W_q \) is a learnable weight matrix.
    
    \item Each object in the set being queried $\{ \bm{x}_i \}$ is independently transformed into two other vectors, forming two sets of the same cardinality, named the \textbf{key set} $\{ \bm{k}_i \}$ and the \textbf{value set} $\{ \bm{v}_i \}$, respectively. With linear transformations, these are computed as:
    $$
    \bm{k}_i = W_k \bm{x}_i, \quad \bm{v}_i = W_v \bm{x}_i
    $$
    where $\bm{k}_i$ and $\bm{v}_i$ are the transformed key and value vectors for $\bm{x}_i$ in set $\{ \bm{x}_i \}$, and \( W_k \), \( W_v \) are the learnable weight matrices for the key and value transformations.

    \item Each key vector $\bm{k}_i$ in the key set is scored by similarity to the query vector $\bm{q}$, according to some similarity metric, typically the scaled dot-product similarity:
    $$
    \text{score}(\bm{q}, \bm{k}_i) = \frac{\bm{q}^T \bm{k}_i}{\sqrt{d_k}}
    $$
    where $d_k$ is the dimensionality of the key vectors, and the similarity score is often scaled by $\sqrt{d_k}$ to prevent large values during training, which could lead to vanishing gradients. 

    \item The \textbf{attention weights} are then computed by applying a \softmax{} to the scores:
    $$
    \alpha_i = \frac{\exp(\text{score}(\bm{q}, \bm{k}_i))}{\sum_j \exp(\text{score}(\bm{q}, \bm{k}_j))}
    $$
    where $\alpha_i$ represents the normalized attention weight for the \( i \)-th key, and the denominator sums over all key vectors to ensure the attention weights sum to $1$, to form a probability distribution.

    \item Finally, the value vectors are weighted by the normalized attention weight vector, and thus combined to yield the output vector:
    $$
    \bm{y} = \sum_i \alpha_i \bm{v}_i
    $$
    The output $\bm{y}$ represents a weighted combination of the values $\{\bm{v}_i\}$, where the weights correspond to the degree of ``attention'' the model pays to each key-value pair. Intuitively, this is the ``answer'' to the query of $\bm{x}$ given by the set $\{ \bm{x}_i \}$.

\end{enumerate}
\end{singlespace}
\end{coloredroutine}

Querying a set with \textit{another set of vectors} is no different from independently applying the described procedure multiple times, for each object in the set. The number of outputs always matches the size of the query set, as each query element $\bm{q}_i$ generates an output vector.

Now let us look into some more advanced use cases of attention:

\subsubsection{Self-Attention}

When a set queries itself, \ie{}, when the one set is queried by a set identical to itself, such procedure is referred to as \textit{self-attention}. In self-attention, the query, key, and value vectors are all derived from the same set of objects, and the model computes attention scores between elements within the same set. This allows each element in the set to attend to every other element, facilitating the extraction of long-range dependencies in sequences. The self-attention mechanism is key to models like the Transformer, where it enables the model to capture complex relationships within an input sequence \citep{vaswani2017attention}.

\subsubsection{Multi-Head Attention}

\textbf{Multi-head attention} involves performing attention computations multiple times independently, each using a distinct set of transformations, such as different learned linear weights. The outputs from each transformation set (or ``head'') are then concatenated and projected back into the desired dimensionality. This approach enables the model to learn attention from multiple subspaces simultaneously, capturing various aspects of the relationships between objects.

In the linear case, multi-head attention can be formulated as:
$$
\text{MultiHead}(Q, K, V) = \text{Concat}(\bm{y}_1, \bm{y}_2, \dots) W_O
$$
where $Q$, $K$ and $V$ are the concatenated matrix representations of the query set, the key set and the value set, respectively. Each head $\bm{y}_i$ is defined as:
$$
\bm{y}_i = \text{Attention}(Q W_Q^i, K W_K^i, V W_V^i)
$$
where $W_Q^i$, $W_K^i$ and $W_V^i$ are the learnable linear transformation weights for the query, key, value of head $i$, respectively, with $W_O$ being a final output transformation matrix. This allows the model to aggregate different attention patterns from different subspaces.

Multi-head attention was used in the set-based bottlenecked dynamic models introduced in Chap.~\ref{cha:CP} (Sec.~\ref{sec:CP_bottleneck}, Page.~\pageref{cha:CP}).

\subsubsection{Semi-Hard Attention}
\label{sec:semihard_attention}

When computing attention weights, it can be beneficial to ignore the unrelated values and focus more strongly on the more matching elements. A modification of attention called \textbf{semi-hard attention} involves keeping only the top-$k$ attention weights, where the output of a vector-set query is only a weighted sum over the top-$k$ most matching values. Note that the output vector will still be a \textit{soft} mixture of the top-$k$ values, while the non-top-$k$ values will be \textit{hard}-discarded \citep{gupta2021memory}. Here, the classical terminology about soft and hard attention weights is used: when the weights are binary values, a selection is essentially conducted, which is recognized as ``hard''; While, if the attention weights are continuous values, the attention is recognized as ``soft''. Top-$k$ attention weights are soft on the top-$k$ indices of the attention weight vector, while being hard $0$s on others, separating the influence of the non top-$k$ ones.


Formally, semi-hard weights are obtained by selecting the top-$k$ scores from the full attention weights, setting the remaining weights to zero, and then renormalizing the selected weights:
\begin{singlespace}
\begin{equation}
\alpha'_i = \begin{cases} 
\frac{\alpha_i}{\sum_{j\in \text{top-$k$}}{\alpha_j}} & \text{if } \alpha_i \text{ is one of the top-k scores}, \\
0 & \text{otherwise}.
\end{cases}
\end{equation}
\end{singlespace}

The formula above shows that a linear renormalization is conducted such that the selected top-$k$ attention weights sum to $1$. Then, as usual, the renormalized weights are applied to the value vectors. The renormalization can alternatively be implemented more efficiently by setting the non-top-$k$ scores to be $-\infty$ before the \softmax{} operation (Step 3 \& 4 in Routine.~\ref{rtn:objectqueryset} on Page.~\pageref{rtn:objectqueryset}).

When querying a set $\{ \bm{x}_1^K, \dots, \bm{x}_n^K \}$ containing $n > k$ objects with another set $\{ \bm{x}_1^Q, \dots, \bm{x}_k^Q \}$ of exactly $k$ objects, the score matrix will be $k \times n$ (see Step 3 in Routine.~\ref{rtn:objectqueryset} on Page.~\pageref{rtn:objectqueryset}). With ordinary semi-hard top-$k$ attention, this query would result in an output set of $k$ objects, each a transformation of its own top-$k$ most matching object from $\{ \bm{x}_i^K \}$. This means, the output set will be a set with $k$ to $k^2$ objects, where each object is a transformation of $k$ best-matching objects from $\{ \bm{x}_i^K \}$, essentially creating a transformed subset of $\{ \bm{x}_i^K \}$. This is how we used a semi-hard top-$k$ attention as a mechanism to soft-select a subset of objects out of a larger set, with it created the backbone mechanism of spatial abstraction proposed in Chap.~\ref{cha:CP}. In Chap.~\ref{cha:skipper}, spatial abstraction was made more generic to non-object-oriented state representations, \eg{}, to work with feature maps of local patches of observations extracted by Convolutional Neural Networks (CNNs, \citet{lecun1989backpropagation}). 

\subsubsection{Bottom-Up \vs{} Top-Down Attention}
In both machine learning and human cognitive science, attention mechanisms can be classified into two general types based on the source of the attentional signal, \ie{}, the signal that conditions the attention: \textbf{bottom-up attention} and \textbf{top-down attention}. While both types use learned attention weights to focus on different parts of the input data, the sources of conditioning signals in each approach differ significantly.

\textbf{Bottom-Up} attention is driven by the intrinsic properties of the input data. In this approach, the model attends to salient or noteworthy features in the input data based on its sensory characteristics, without any explicit guidance or prior knowledge. For example, in computer vision, regions of an image that contain high contrast, bright colors, or movement might naturally attract more attention \citep{mnih2014recurrent}. Bottom-Up attention is data-driven and reactive, often used in tasks where the model must extract relevant information from raw data with minimal preconceptions.

\textbf{Top-Down} attention, instead, is guided by higher-level goals, intentions, or prior knowledge. In this approach, the attention mechanism is influenced by the intention of the model, allowing the model to focus on certain aspects of the data that are most relevant to achieving the desired outcome. For example, in a visual question answering (VQA) task, the model may direct its attention toward specific regions of an image that are relevant to answering the given question, based on context and prior knowledge, rather than simply attending to all potentially salient features \citep{antol2015vqa}.

We used top-down attention mechanisms to implement spatial abstractions: 1) in Chap.~\ref{cha:CP} to make the model pay attention to the aspects of the state representations most relevant to the intended actions during planning (Sec.~\ref{sec:CP_bottleneck}, Page.~\pageref{sec:CP_bottleneck}), and 2) in Chap.~\ref{cha:skipper} to make the estimators to focus on parts of the state representations most relevant to transitioning from one state to another target state (Sec.~\ref{sec:skipper_spatial_abstraction}, Page.~\pageref{sec:skipper_spatial_abstraction}).

\section{Model-based RL}
\label{sec:model_basedL_drl}

Despite significant progress in applying RL to a variety of problems, its real-world application remains hindered by challenges such as poor sample efficiency, unsatisfactory generalization, and other issues \citep{lillicrap2015continuous}. In a standard RL system, the primary focus is on estimating a single quantity: the returns. While value estimation alone can suffice for optimal decision-making \citep{silver2021reward}, relying solely on returns often proves inefficient or even intractable in many real-world scenarios.

\textbf{Model-Based RL} (\textbf{MBRL}) offers an alternative approach by incorporating predictive or generative models, which provide estimations of additional quantities beyond just returns. This inclusion of models opens up new avenues for improving efficiency and performance in RL.

The versatility of MBRL lies in its creative applications of models. For instance, models can be derived from domain expertise, learned in advance, or developed during the RL agent’s training process. Furthermore, models in MBRL can be utilized at different stages of decision-making. \textbf{Decision-time} planning explicitly uses a model to help make decisions at decision-time \citep{alver2022understanding}. For example, \MuZero{} performs a tree search at each timestep, querying a learned model to simulate the most desirable future actions \citep{schrittwieser2019mastering}. In contrast, \textbf{background} planning methods typically function as model-free during decision time, but leverage the model to enhance learning during training. Prominent examples of such methods include \Dyna{} \citep{sutton1991dyna} and \Dreamer{} \citep{hafner2023mastering}, among others.

\subsection{Sources of Models: Where They Come From}
We begin by discussing the origins of models in MBRL agents, specifically distinguishing between \textbf{acquired models}, which do not require learning, and \textbf{learned models}, which are acquired during RL training.

\subsubsection{Known Perfect Models}
Computer Go has been successfully conquered by the MBRL agent \AlphaGo{} and its variants \citep{silver2016mastering}. In this case, the model used for decision-time planning is the Go simulator itself, which provides perfectly accurate dynamics. With a perfect model, planning can be highly effective when combined with classical AI search methods, such as those with guaranteed performance \citep{kocsis2006bandit}. However, task-specific models are not always available for all tasks. A more general and practical approach involves learning a model of the environment.

\subsubsection{Models that are Learned}
Many methods involve learning the models they utilize. Notable examples include the use of learned models in planning algorithms, such as the World Models framework \citep{ha2018world,zhou2024dino} and Model-Based Policy Optimization \citep{levine2016end}. These approaches have shown success in both simulated and real-world environments.

We now briefly review model learning strategies, distinguishing between approaches where the model and the RL agent are trained together as a unified system or separately.


\begin{itemize}[leftmargin=*]
\item
\textbf{Pre-Trained Models}: Many methods employ separate training stages for the model and the core RL components. For example, consider the unsupervised training methodology.

Approaches like World Models \citep{ha2018world,zhou2024dino} benefit from a separated phase for model learning, known as the \textbf{exploration} or \textbf{pretraining} phase, where the RL agent is not involved. By using a fixed exploration policy, this strategy transforms the non-stationary nature of model learning in an RL setting into a more stable, stationary learning problem, thereby sidestepping certain challenges.

Compared to end-to-end training methods, to be introduced shortly, a key disadvantage is that these approaches allocate a significant portion of the agent-environment interaction budget to a potentially long exploration phase (assuming the ``unsupervised'' training phase is not freely provided). Another issue is that the consequent models are trained only on the trajectories close to the initial states, as uniformly random policies usually do not get the agents far. Thus, these methods may struggle in environments with distinct local features that require the model to continually learn.

\item 
\textbf{End-to-End Trained Models}: In supervised learning, ``end-to-end'' training refers to optimizing a model directly from inputs to outputs. While in RL, end-to-end indicates that both the model and the RL agent are trained together, where the model can aid in decision-making and value estimation. This means both the model and other system components learn from scratch.

However, end-to-end training does not necessarily imply that the model and RL components are tightly coupled. For example, in \Dyna{}, although the model and value estimator are learned concurrently, their learning processes are largely independent.

End-to-end training is attractive due to its procedural simplicity, especially in deep learning scenarios. However, it is not always suitable for every method. Since end-to-end models are learned from scratch alongside the RL system, the initial stages of RL learning can be difficult without an effective model. The model must rapidly learn to assist the RL system in a meaningful way.

Moreover, training a model end-to-end in RL control tasks is technically challenging, particularly due to non-stationarity. The data distribution evolves as the agent's policy changes. Additionally, when the model is coupled with RL components (\eg{}, sharing the same state representation encoder), training signals often compete with each other. Different optimization objectives may not align, and an inadequately tuned bottleneck for all signals can lead to undesirable parameter landscapes. Although scalar trade-offs and alternative optimization objectives are commonly used to mitigate these issues, they are not always effective.
\end{itemize}

\subsection{Timing of Planning: Decision-Time \vs{} Background}
\label{sec:planning}

Planning is a term referring to the use of more computation to improve predictions and behavior without additional agent-environment interactions, \ie{}, essentially trading computation for sample efficiency \citep{hasselt2019when}.

For clear initiation of the discussions and the coherence with terminologies proposed in \citet{alver2022understanding}, based on the timing of the model usage, we roughly categorize some existing planning methods into two categories: \textbf{decision-time} planning methods and \textbf{background} planning methods. While a decision-time planning agent \textit{utilizes the model to directly enhance its behavior} when interacting with the environment, a background planning agent often seeks to \textit{improve itself with a model when not interacting with the real environment}.

\subsubsection{Decision-Time Planning}
Decision-time planning is a computational embodiment of proactive decision-making. Intuitively, a decision-time planning agent evaluates potential future outcomes based on its understanding of the effects of actions and current value estimates, then selects the most promising action or sequence of actions to pursue. This knowledge is derived, at least in part, from the model.

A key advantage of decision-time planning is that, with an accurate model, behavior policies can quickly adapt toward optimality. This proactive approach also enhances adaptability in unexpected situations. Notable decision-time planning agents include \MuZero{} \citep{schrittwieser2019mastering} and \PlaNet{} \citep{hafner2019learning}.

Decision-time planning can also be framed as a search problem. We will now explore a representative methodology: Model Predictive Control (MPC).

\subsubsection{Model Predictive Control (MPC) \& Tree Search}
\label{sec:MPC}

Model Predictive Control (MPC) is a decision-time planning methodology that involves optimizing control inputs over a finite prediction horizon \citep{hansen2023td}. At each time step, MPC solves an optimization problem using a model to predict future states and determine the optimal control sequence. The optimization typically minimizes a cost function that balances objectives like tracking a desired trajectory, minimizing energy consumption, or avoiding constraints. After solving the optimization, only the first action is applied, and the process can be repeated at the next timestep, updating the predictions with new observations. Beyond RL, MPC is also widely used in systems where constraints (\eg{}, physical limits or safety requirements) must be explicitly respected. MPC serves as the foundation of many modern successful applications, \eg{}, Monte-Carlo Tree Search (MCTS) which powers state-of-the-art game RL game-playing agents \citep{schrittwieser2019mastering}.

We can implement MPC with tree-search. The search tree structure is used to represent and evaluate possible future states of the system. Below are key aspects of how MPC utilizes the search tree:

\begin{enumerate}[leftmargin=*]
    \item \textbf{Nodes in the Search Tree}: In the context of MPC, each node in the search tree represents a possible state at a specific time step. MPC expands the search tree by appending new nodes corresponding to the predicted future states, which are generated by applying potential actions to the current state using the model;

    \item \textbf{Budget-Constrained Search with Priority Queue}: The number of simulations (or steps of search) is typically constrained by a budget, meaning the agent performs a limited number of tree expansions. Once the budget is exhausted, the agent selects the immediate action based on the most promising path found within the search tree. For this, MPC tree search algorithm uses a priority search queue to manage which branches of the search tree are explored first. For example, some MPC's search priorities of the nodes are set according to a best-first search heuristic, which is designed to prioritize nodes that are more likely to lead to a favorable outcome. The priority heuristic in the MPC search process influences the order in which different branches of the search tree are explored, though it does not affect the finalization of the decision regarding the optimal action. The optimal action is always the one that leads to the most promising trajectory, as defined by the optimization objective. Only the first action of the optimal sequence is applied, and the search tree can be reconstructed at the next timestep if necessary, taking into account the new observations.

    \item \textbf{Action Selection and Node Evaluation}: At each planning step, the MPC evaluates the nodes by calculating the expected future return (or other surrogate values) associated with each branch. The optimization process selects the action corresponding to the most promising node;
\end{enumerate}

We now introduce a simple instance of the tree search algorithm used in Chap.~\ref{cha:CP}, characterized by a priority search queue determined by certain priority heuristic.

\begin{algorithm*}[htbp]
\caption{Prioritized Tree-Search MPC}
\label{alg:bfsearch}
\KwIn{$s_0$ (current state), $\scriptA$ (action set), $\scriptM$ (model), $\scriptQ$ (value estimator), $\gamma$ (discount)}
\KwOut{$a^*$ (action to be taken)}

$q = \text{queue}()$; $q_T = \text{queue}()$ \textcolor{darkgreen}{//$q_T$ for terminal nodes}\\

$n_u = \text{NODE}(s_0, \text{root}=\text{True})$ \textcolor{darkgreen}{//$n_u$ denotes a node with branches unprocessed nor in $q$}\\

\While{True}{
    \If{$n_u.\omega$}{
        $q_T.\text{add}(\langle n_u, n_u.\sigma \rangle)$ \textcolor{darkgreen}{//identified as a terminal state. $n_u$ is added to $q_T$ using bisection, together with the discounted sum of the simulated rewards along the way $n_u.\sigma$}
    }
    \Else{
        \lFor{$a \in \scriptA$}{
            $q.\text{add}(\langle n_u, a, n_u.\sigma +  \gamma ^{n_u.\text{depth}} \cdot Q(n_u.s,a) \rangle)$ \textcolor{darkgreen}{//bisect \wrt{} priority}
        }
    }
    \If{$\text{isempty}(q)$}{\textbf{break} \textcolor{darkgreen}{//tree depleted}}
    $n_c,a_c,v_e=q.\text{pop}()$ \textcolor{darkgreen}{//get branch with highest priority; for in-distribution setting, priority is the estimated value of the leaf trajectory}\\
    \If{budget depleted}{\textbf{break} \textcolor{darkgreen}{//termination criterion met}}
    $\hat{s}, \hat{r}, \hat{\omega} = \scriptM(n_c.s, a_c)$ \textcolor{darkgreen}{//simulate the chosen branch}\\
    $n_u = \text{NODE}(\hat{s}, \text{parent}=n_c)$\\
    \If{$n_c.\text{depth}>0$}{$n_u.a_b=n_c.a_b$}
    \Else{$n_u.a_b=a_c$ \textcolor{darkgreen}{//descendants trace root action}}
    $n_u.\omega=\hat{\omega}$; $n_u.\sigma=n_c.\sigma + \gamma^{n_c.\text{depth}} \cdot \hat{t}$
}

$n_c,a_c,v_e=q.\text{pop}(\text{`highest value'})$ \textcolor{darkgreen}{//get branch with highest \textbf{value} within the expandables}\\


$n^* = n_c$;\\

\If{$\neg\text{isempty}(q_T)$}{

$n_T = q_T.\text{pop}(\text{`highest value'})$ \textcolor{darkgreen}{//get node with highest \textbf{value} within simulated terminal states}\\

\If{$n_T.\text{value} \geq v_e \vee \text{isempty}(q)$}{$n^* = n_T$}
} 

\If{$\text{isroot}(n^*)$}{$a^*=a_c$}
\Else{$a^*=n^*.a_b$}
\end{algorithm*}

\begin{figure*}[htbp]
\centering
\captionsetup{justification = centering}
\includegraphics[width=0.90\textwidth]{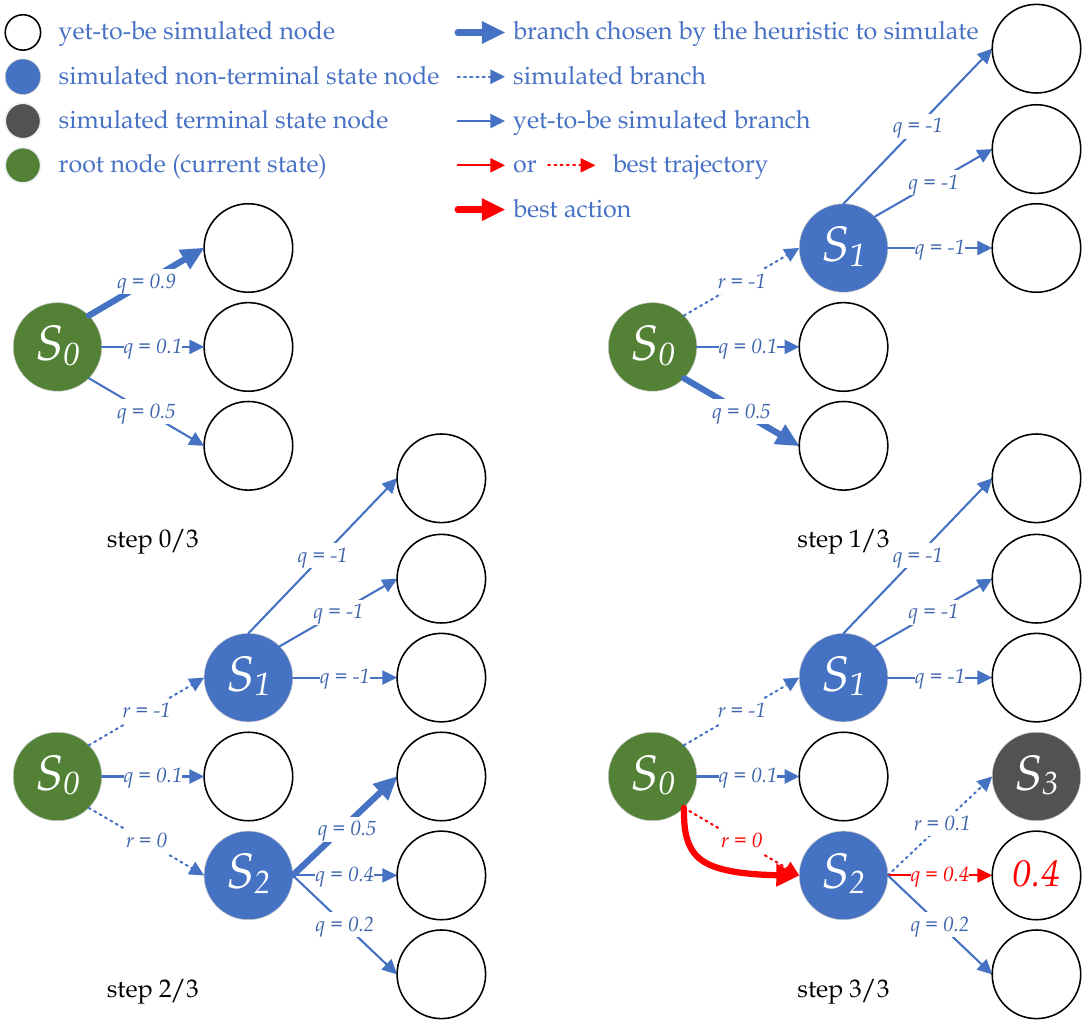}

\caption[Best-First Tree Search for MPC]{\textbf{Example of the Best-First Heuristic-based Tree Search (for MPC)}: Step 0 / 3) Start of planning, with the root node and three branches. The branch $\langle s_0, a_0 \rangle$ is chosen due to the best-first heuristic. If we employ the random search heuristic, like what we do in OOD evaluation in Sec.~\ref{sec:CP_experiments} (Page.~\pageref{cha:CP}), a random branch would be chosen; Step 1 / 3) We expand the chosen branch, popped out of the priority queue. A new node is constructed, together with its out-reaching branches, which are added to the queue. Now the queue has $5$ branches in it. The heuristic marks $\langle s_0, a_2 \rangle$ to be the next simulated branch; Step 2 / 3) Simulation of $\langle s_0, a_2 \rangle$ is finished and $\langle s_2, a_0 \rangle$ is marked; Step 3 / 3) Node S3 is imagined via $\langle s_2, a_0 \rangle$ but it is estimated to be a terminal state. Now, the tree search budget is depleted. We locate the root node branch $\langle s_0, a_2 \rangle$ which leads to the trajectory with the most promising return $0.4$.}
\label{fig:bfsearch}
\end{figure*}

For more intuitive understanding, we provide in Fig.~\ref{fig:bfsearch} an example showing how the algorithm works with a best-first heuristic with $\gamma=1$ and $|\scriptA{}| = 3$ and maximum planning steps of $3$.

In the case of Fig.~\ref{fig:bfsearch}, which depicts the tree-search algorithm used in decision-time planning in Chap.~\ref{cha:CP}, the agent can choose to (re-)plan at every timestep using the learned model to maximize adaptability and to correct deviations in the outdated planning results.

Equivalence could be drawn from this planning approach to Monte-Carlo Tree Search (MCTS) \citep{silver2016mastering,silver2017mastering}. While this method is more simplistic and assumes deterministic models to operate with.

\subsubsection{Background Planning}
\label{sec:background_planning}

Background Planning often involves more reactive model usages and does not directly optimize for the action to take at decision-time. We present one representative framework - \Dyna{}, to showcase the potentials of this approach\footnote{In some sense, classical methodologies such as \Dyna{}mic Programming (DP) can also arguably fit in the category of background planning, as they seek to estimate the value function and optimal policy \citep{howard1960dynamic}} \citep{sutton1991dyna}.

\Dyna{} was first introduced in \citet{sutton1991dyna} as an early RL planning framework designed to reinforce an agent's policy in a simulated environment, while it interacts with the real world\footnote{Though it may seem counterintuitive to describe \Dyna{}'s behavior as planning, the authors argued that conducting RL in a simulated environment enhances the agent's behavior and can therefore be interpreted as a form of planning.}.

\Dyna{} trains its model based on the experiences from agent-environment interactions, enabling the model to generate simulated transitions. It then uses these simulated transitions to augment the training data for the value estimator. Importantly, there is no fundamental change to the learning process of the value estimator; the only difference is that the training data is enriched with simulations from the model. For more discussions regarding \Dyna{}, please check Sec.~\ref{sec:exprepgen} (Page.~\pageref{sec:exprepgen}).

Background planning frameworks like \Dyna{} offer sample efficiency benefits and are computationally less demanding during agent-environment interactions, as their additional computations occur offline. However, they cannot provide immediate behavior adaptations in the states the agent encounters. Furthermore, \Dyna{} typically only learns to generate transitions from states visited in the past. As a result, \Dyna{} is focused on the data distribution during training time, which can limit its ability to generalize OOD. 

There are some notable works that followed the idea of \Dyna{}, \eg{}, \citet{hafner2020mastering,kaiser2019model}.

\subsection{Usage of Models: What They Estimate \& How They Help}
As discussed, the versatility of MBRL methods makes them promising for improving RL performance across various dimensions. A model does not have to estimate the transition functions defined in an MDP; instead, it can capture other interesting and potentially partial aspects of the environment. Below, we explore several representative use cases of models:

\subsubsection{Search for Best Actions with Dynamics Predictors}
\label{sec:dynamics_predictors}

The most straightforward use of a model is to estimate the transition dynamics of the environment, \ie{}, to predict how the environment changes given the actions that the agent may take.

In tabular settings (finite state and action spaces), learning transition dynamics is relatively simple. However, in non-tabular tasks, dynamics predictors may have to reconstruct future observations or given state representations, which can be noisy or contain irrelevant details. Despite progress, such as in \citet{kaiser2019model}, observation-level reconstruction often suffers from inefficiencies that complicate decision-making. Moreover, model architecture may have to be highly task-specific to handle these detailed reconstructions. For instance, object-oriented dynamics models work well for grid-like environments, while models suited for differential equations are better for continuous control tasks.

An alternative to detailed reconstruction is to build models in learned state representation spaces, where meaningful features are extracted and irrelevant details to decision-making are neglected. However, state representation learning introduces potential degeneracy: state transitions may be accurate yet meaningless without proper training signals. To address this, methods like \citet{kaiser2019model} use reconstruction to seek establishing bijections between observations and states, aiming for lossless compression of observations. However, these methods, to a degree, still share the challenges of observation-level reconstruction \citep{silver2016predictron, schrittwieser2019mastering}. To mitigate these issues, additional signals, such as value estimation, are needed to constrain state representations and ensure they are meaningful for prediction.

The Predictron \citep{silver2016predictron} is a model-based approach that learns state representations specialized for accurate value estimation and a model that operates within. By matching the Monte Carlo returns from the real environment, the model ensures consistency between generated and real environment returns. This approach enhances TD-based value estimation and has been extended to \MuZero{} \citep{schrittwieser2019mastering}, which uses decision-time planning with MCTS. \MuZero{} achieves state-of-the-art performance in board games and Atari by employing Predictron-based models for planning. However, training solely with task-specific reward signals can lead to overfitting and overreliance on memorization, limiting generalization across tasks. We will discuss this more in Chap.~\ref{cha:CP}.

These dynamics predictors can be used effectively in tree-search algorithms during decision-time planning. A model that can estimate state transitions can also serve as environment simulators for background planning, which is to be introduced later separately.

Dynamics predictors can also be used to predict the outcomes of macro-action, giving rise to ideas such as \textbf{option models}. However, these models are often haunted by the uncertainty accumulated over the time horizon of the macro actions and can be highly unreliable, if sources of stochasticity are involved.

\subsubsection{Propose Beneficial Goals with Temporarily-Abstract Target Generators}
Models can also be used to directly generate goals that the agent may seek to achieve. This requires that the model are modelled in a temporally-abstract way, since the possible states / observations at the very next timestep typically do not serve as meaningful goals. For instance, \LEAP{} used variational autoencoders (VAEs) to generate observations that may be used as waypoints to guide the overall navigation; While, in Chap.~\ref{cha:skipper}, \Skipper{} used conditional VAE (CVAEs) to generate possible states given the environmental contexts, enabling the OOD generation of targets that help decompose the overall tasks into smaller and more manageable steps.

\subsubsection{Learn for Free with Experience Generators \& Experience Replays}
\label{sec:exprepgen}
\Dyna{} and recent methods like \citet{ha2018world, kaiser2019model} use models to generate simulated transitions, building on generative modeling approaches. Recent work focuses on stochastic experience generation to improve MBRL in stochastic environments \citep{kaiser2019model,du2019energy}.

Experience Replay, popularized by \DQN{}, stores transitions or trajectories experienced by the agent \citep{lin1992self,mnih2015human}. During background planning, the agent samples from the replay buffer to improve sample efficiency by reusing experiences, trading off large storage space requirements \citep{sutton1991dyna}.

Although \DQN{} popularized experience replay, it is not always considered an MBRL method. In DQN, experience replay is used in a model-free, off-policy Q-learning fashion (See Sec.~\ref{sec:q_learning}, Page.~\pageref{sec:q_learning} for more details). \citet{hasselt2019when} unifies \DQN{} with \Dyna{}, where experience replay serves as a model that "generates" on-policy transitions. This makes \DQN{} a special case of \Dyna{}, with a model trained by storing transitions.

From this unification, an experience generator can be seen as a learnable version of experience replay \citep{seijen2015replay,hasselt2019when}. Parametric experience generators, trained to reduce storage requirements, can generate samples that may improve generalization if the model parameterization supports it. However, in deep learning, these generators often struggle to produce high-quality samples due to model inaccuracies. As \citet{hasselt2019when} notes, inaccurate transitions can introduce uncommon or impossible states, damaging the value estimator by promoting undesirable generalization. Observation-level generators are especially prone to this issue \citep{silver2016predictron,schrittwieser2019mastering}. In contrast, experience replay avoids these risks but is limited to samples from seen trajectories.

\subsubsection{Enhance State Representations with Extra Predictions}
It has been shown that incorporating additional predictive models that can back-propagate into the state encoder can improve the quality of state representations. These extra predictions may be used during decision-time planning, or they may never be called upon. In the latter case, the predictive models exist solely to shape the state representation through training, effectively acting as an implicit regularization signal.

In the former case, the agent proposed in Chap.~\ref{cha:CP} enhances its object-oriented state representation encoder with predictive losses, including state dynamics, reward and termination predictions, as well as value estimation, with all predicted quantities being explicitly utilized during decision-time planning. In contrast, for the latter case, \UNREAL{} uses reward and termination signal predictions to prevent the state representation from overfitting to value estimation, making it more aware of the environment's dynamics \citep{jaderberg2016unreal}.

\subsubsection{Optimize Policy Directly with Differentiable Models}
Differentiable models can play an important role in enabling the direct optimization of policies by providing a smooth, learnable approximation of the environment's dynamics. These models allow for the end-to-end optimization of both the policy and the model in a unified framework. By using differentiable dynamics models, gradient-based optimization techniques can be used to directly update the policy based on planning predictions. Differentiable models enable the use of techniques such as MPC within a differentiable framework, allowing for direct gradient updates to both the model and the policy, improving overall performance \citep{heess2015learning}. 


\subsection{Inspiring Developments in Model-Based RL}
\label{sec:newideas}

We turn to more recent literature and survey the interesting progress for MBRL research related to this thesis, focusing on perspective methods seeking to address the innate difficulties of existing approaches from different perspectives.

\subsubsection{Partial Planning}
\label{sec:partial_planning}

If the state space or feature space is large, then the expected next state or distribution over it can be difficult for models to estimate, as has been repeatedly shown \citep{alver2024partial}.

In MBRL, \textbf{partial models} refer to models that predict partial aspects of the environment rather than capturing the full dynamics, which may be more complex or less relevant to the agent's decision-making process. For instance, partial models may predict only the immediate rewards associated with state-action pairs, or generate high-level abstractions such as goals or waypoints, as seen in frameworks like \LEAP{} and \Skipper{}. By simplifying the environment representation, partial models can help reduce the complexity of learning and planning, focusing the agent’s attention on the most important aspects of the environment for its current objective. This approach can improve computational efficiency and allow for better generalization \citep{bengio2017consciousness}.


Partial models resemble the idea of spatial abstraction we mentioned in Chap.~\ref{cha:CP} and Chap.~\ref{cha:skipper}. 

Similarly, partial models can also reflect abstraction in the temporal dimension \citep{alver2024partial}. For example, the model in \citet{talvitie2008local} makes certain predictions based on certain few points in history (past observations) when triggered at a certain decision point (timestep). This idea predates the sequence modeling methods based on modern deep learning, which take advantage of the recent advances of deep learning to facilitate attention to history back-in-time.




\subsubsection{Planning with Temporal Abstraction}
When planning a long trajectory, searching at minuscule timescales is computationally demanding and in nature difficult, due to the exponentially growing search outcomes and imperfectness of the learned model.

An \textit{option model} seeks to capture the dynamics given by a set of options inside an MDP. As pointed out in \citet{sutton1999between}, planning with an option-model is preferred by its sample efficiency and the less exposure to the error accumulation problem of atomic timestep planning using an imperfect model. Though option models in real practice face the challenges brought by the discovery of options, the changing of option, \etc{}, I have proposed a solution as the \Skipper{} framework in Chap. \ref{cha:skipper}.

\subsubsection{Directions of Planning}
Our previous discussions on model behaviors focused on the forward simulation from certain states, whether from present, to the relative future. Somewhat counter-intuitively, \citet{hasselt2019when} investigates benefits of using a \textit{backward model}, \ie{}, a \Dyna{} transition generator which simulates a transition from the current state and the last action to the last state. When the backward model is still learning much, \ie{}, model is still inaccurate, temporal credit assignment, \eg{}, temporal-difference updates, will be conducted on fictional starting state with the target constructed from a true state, therefore lowering the chance of catastrophic delusional updates \citep{hasselt2019when}. When the model is accurate enough, this would perform in the same way as we would expect from a good forward \Dyna{} \citep{hasselt2019when}. The concept of delusions in RL is heavily discussed in Chap.~\ref{cha:delusions}.

\section{Other Related Background Knowledge}

\subsection{Consciousness in the First Sense \& Conscious Planning}
In this thesis, my collaborators and I frequently draw inspiration from human conscious planning when designing agent behaviors targeting the challenges of generalization. Note that this thesis does not claim to understand the exact mechanisms behind consciousness in the human brain. Instead, the methods presented in this thesis aim to endow agents with capabilities akin to conscious planning. In other words, we seek to design agents that behave similarly to conscious planning agents, leveraging consciousness-driven behaviors such as Out-Of-Distribution (OOD) generalization \citep{bengio2017consciousness}.

In \citet{dehane2017consciousness}, the authors outline conscious behaviors along two orthogonal axes. The first axis, corresponding to consciousness in the first sense, often abbreviated as C1, describes how information processing is selectively focused. When humans reason for decision-making, their attention is directed to the most relevant aspects of space and time, facilitating efficient generalization. This form of consciousness is often modeled by the global workspace hypothesis \citep{baars1997theater}. The second axis, representing consciousness in the second sense, pertains to an agent's awareness of its own internal state \citep{vangulick2004consciousness}. While this axis is intriguing and often receives significant attention, it is essential not to overlook the more fundamental axis of conscious behaviors (the first sense).

The top-down attention-based spatial abstraction mechanism proposed in Chap.~\ref{cha:CP} and used in Chap.~\ref{cha:skipper} draws heavy inspirations from consciousness in the first sense.

More discussions regarding consciousness and its impact on decision-making can be found in Sec.~\ref{sec:CP_intro} on Page.~\pageref{sec:CP_intro}.

\subsection{Generalization-Focused Environments \& Experimental Settings}
In existing works, the failure modes of RL agents are often overlooked, possibly due to a lack of access to environmental ground truth in benchmark environments. To address this, my collaborators and I developed several sets of environments aimed at evaluating the generalization capabilities of RL agents, specifically in the face of distributional shifts, compositional challenges, and OOD systematic generalization scenarios \citep{quinonero2022dataset}.

\subsubsection{Generalization-Focused Environments}

The two most representative sets of these environments are \RDSfull{} and \SSMfull{}, or simply \RDS{} and \SSM{}, which are the primary focus of later introductions. These environments were designed with the goal of testing how agents adapt to the challenges of generalization. Unlike traditional environments, however, \RDS{} and \SSM{} are environment factories that can generate specific instances of tasks based on our design specifications.

Both \RDS{} and \SSM{} are built upon the {\fontfamily{qcr}\selectfont DistShift}\footnote{\url{https://minigrid.farama.org/environments/minigrid/DistShiftEnv/}} environment from the MiniGrid repository \citep{chevalierboisvert2018minigrid,chevalierboisvert2023minigrid}, which itself is based on the work by \citet{leike2017ai}.

A significant advantage of using gridworlds as the foundation for these environments is that all instances are solvable by dynamic programming (DP), due to the simplicity of enumerating state-action pairs and constructing a transition table.

It is important to note that the challenges presented by these tasks are more complex than they may initially appear. For instance, the random placement of episode-terminating lava traps introduces compositionality challenges, while the abundance of terminal states complicates path planning. Certain locations in some environment instances become inaccessible to the agent, leading to problematic targets, as discussed in Chap.~\ref{cha:delusions}.

We deliberately made both \RDS{} and \SSM{} fully observable by modifying the original agent-centric views (which render the observation) to a top-down bird's-eye view of the entire gridworld. These fully observable tasks prioritize reasoning over causal mechanisms rather than learning representations from complicated observations.

\subsubsection{\RDS{}}
\label{sec:RDS}

\RDS{} is a navigation task where the agent must navigate to the goal without stepping into the episode-terminating lava grids. A terminal sparse reward of $+1$ is given when the agent successfully gets to the goal grid.

\begin{figure}[htbp]
\centering
\subfloat[left-right, $\delta = 0.25$]{
\captionsetup{justification = centering}
\includegraphics[height=0.24\textwidth]{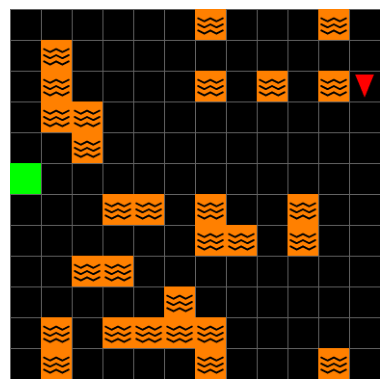}}
\hfill
\subfloat[left-right, $\delta = 0.35$]{
\captionsetup{justification = centering}
\includegraphics[height=0.24\textwidth]{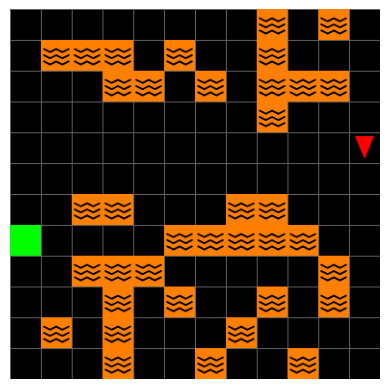}}
\hfill
\subfloat[left-right, $\delta = 0.45$]{
\captionsetup{justification = centering}
\includegraphics[height=0.24\textwidth]{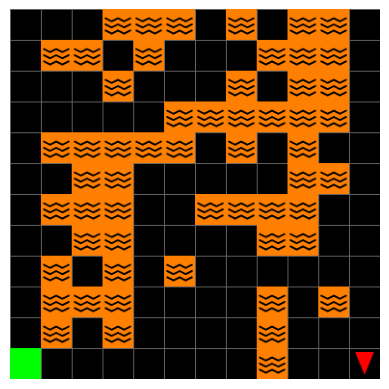}}
\hfill
\subfloat[left-right, $\delta = 0.55$]{
\captionsetup{justification = centering}
\includegraphics[height=0.24\textwidth]{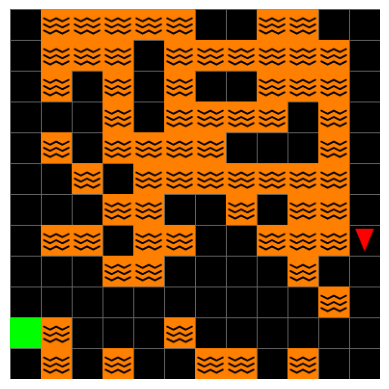}}

\subfloat[top-bottom, $\delta = 0.25$]{
\captionsetup{justification = centering}
\includegraphics[height=0.24\textwidth]{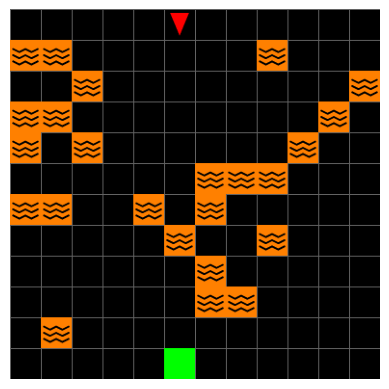}}
\hfill
\subfloat[top-bottom, $\delta = 0.35$]{
\captionsetup{justification = centering}
\includegraphics[height=0.24\textwidth]{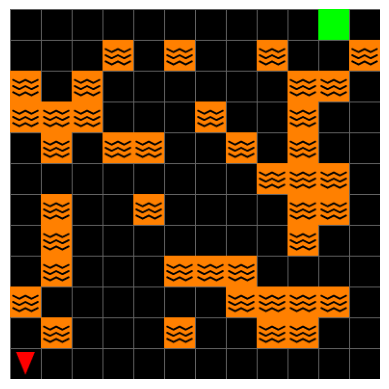}}
\hfill
\subfloat[top-bottom, $\delta = 0.45$]{
\captionsetup{justification = centering}
\includegraphics[height=0.24\textwidth]{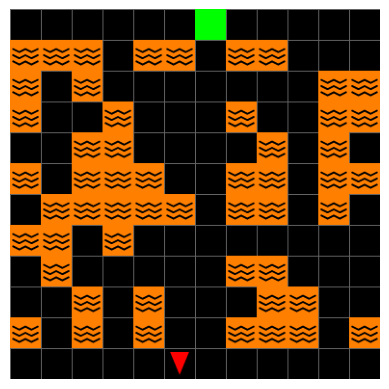}}
\hfill
\subfloat[top-bottom, $\delta = 0.55$]{
\captionsetup{justification = centering}
\includegraphics[height=0.24\textwidth]{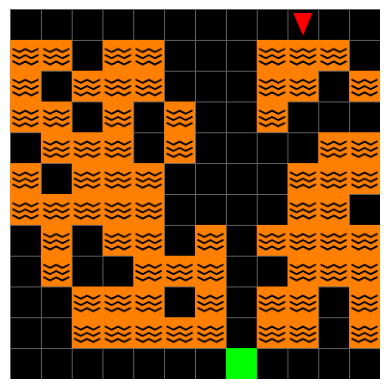}}

\caption[Rendering \RDS{} Instances]{\textbf{Left-Right \& Top-Down \RDS{} Instances}: We render observations of \RDS{} instances of different orientation (left-right or top-down) and different difficulties $\delta \in \{ 0.25, 0.35, 0.45, 0.55 \}$. The agent location and facing direction are rendered with a red triangle. The lava grids are rendered with orange boxes with waves. The goal locations are rendered as green squares. In Chap.~\ref{cha:CP}, the agents are trained on left-right instances and evaluated on only top-down instances, which are outside the training distributions. Such experimental setting is how we evaluated the zero-shot OOD generalization abilities of the agent in Chap.~\ref{cha:CP}.
}
\label{fig:render_RDS}
\end{figure}

In each \RDS{} instance, the agent and the goal spawn at opposite edges of the field. Between these edges, lava is randomly placed according to a difficulty parameter, $\delta$, which ensures that a viable path to the goal exists. The agent must navigate through this lava field to reach the goal and receive the terminal reward. When we constrain that the agent only spawns at the left or the right edge of the grid world, the corresponding \RDS{} instances are out of the distribution of those instances generated by constraining the spawning at the top or bottom edges. We illustrate these in Fig.~\ref{fig:render_RDS}.

The evaluative task instances propose a systematic generalization challenge to the agents \citep{frank2009connectionist}. Systematic generalization refers to the ability of a cognitive system (a human brain or a computational agent) to apply learned knowledge or rules to new, previously unseen situations in a consistent and structured manner.

One key feature of \RDS{} instances is that all non-terminal states form a connected component. This means that, starting from any non-terminal state, the agent can always reach another non-terminal state within a limited number of steps.

Each \RDS{} instance can be paired with three different action spaces, each defining a distinct transition kernel:

\begin{itemize}[leftmargin=*]
\item \textbf{``Turn-or-Forward'' dynamics}: Depending on the agent's facing direction, it can either turn clockwise, turn counterclockwise, or move forward to the grid in front.

\item \textbf{``Absolute-Direction'' dynamics}: The agent chooses to move to one of the four adjacent grids based on absolute directions, regardless of its facing direction.

\item \textbf{``Turn-and-Forward'' dynamics}: The agent chooses to turn clockwise, counterclockwise, or $180^\circ$ and move forward, or simply step forward without turning.
\end{itemize}

While \RDS{} instances are generally deterministic, randomness is occasionally introduced to test the agent's ability to handle stochasticity.

\SSM{} was built on top of the features of \RDS{}.

\subsubsection{\SSM{}}

In \SSM{}, the agent moves one step at a time in four absolute directions to navigate the field, implementing only the ``Absolute Direction'' dynamics. The agent must collect both a sword and a shield, which are randomly placed, before reaching the ``monster'' guarding the treasure. If the agent reaches the monster prematurely, the episode ends without reward. The paths to the sword, shield, and monster are guaranteed to exist in all environmental instances. The sword and shield are collected by moving to their respective grid cells, and once acquired, they cannot be dropped.

This design introduces temporary unreachability in the state structure, as non-terminal states are not fully traversable from one to another. Semantically, this segments the states of \SSM{} into four situations, \ie{}, equivalence classes, determined by two binary indicators: the agent's possession of the sword and shield. For example, $\langle 0, 1 \rangle$ denotes that the sword has not been acquired, but the shield has been acquired.

\begin{figure}[htbp]
\centering
\subfloat[$\delta = 0.25$]{
\captionsetup{justification = centering}
\includegraphics[height=0.24\textwidth]{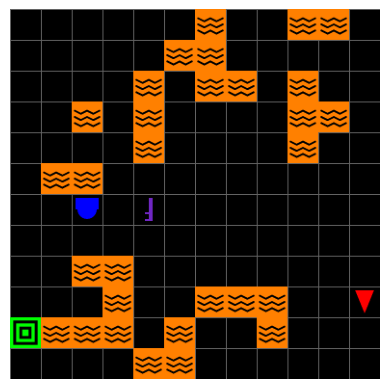}}
\hfill
\subfloat[$\delta = 0.35$]{
\captionsetup{justification = centering}
\includegraphics[height=0.24\textwidth]{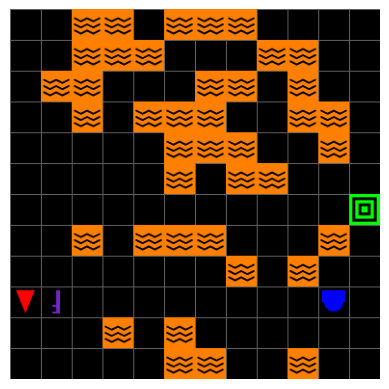}}
\hfill
\subfloat[$\delta = 0.45$]{
\captionsetup{justification = centering}
\includegraphics[height=0.24\textwidth]{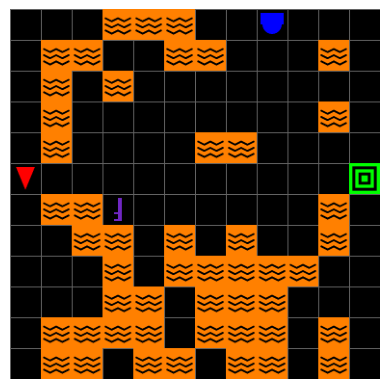}}
\hfill
\subfloat[$\delta = 0.55$]{
\captionsetup{justification = centering}
\includegraphics[height=0.24\textwidth]{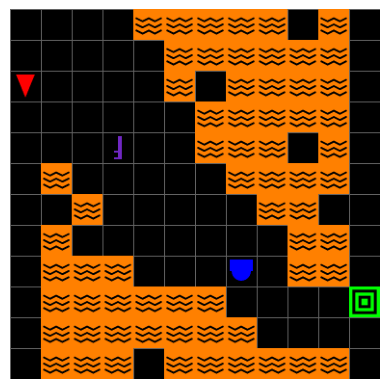}}

\caption[Rendering \SSM{} Instances]{\textbf{\SSM{} Instances}: We render observations of \SSM{} instances of difficulties $\delta \in \{ 0.25, 0.35, 0.45, 0.55 \}$. The renderings of the agent location and the lava grids are the same as in \RDS{}. While, the locations of the newly introduced sword and shield are rendered as a purple rapier shape and a blue heater shield shape, respectively. The monster location is marked with green-black loops. \SSM{} was introduced in Chap.~\ref{cha:delusions} (Page.~\pageref{cha:delusions}) to expose target-assisted planning agents' failure modes regarding problematic targets. In Sec.~\ref{sec:delusions_exp} (Page.~\pageref{sec:delusions_exp}), the agents are trained only on instances with $\delta = 0.4$ and then evaluated on a range of OOD difficulties $\delta \in \{ 0.25, 0.35, 0.45, 0.55 \}$. Such experimental setting is how we evaluated the zero-shot OOD generalization abilities of the agent in both Chap.~\ref{cha:skipper} \& Chap.~\ref{cha:delusions}. In Chap.~\ref{cha:skipper}, only \RDS{} is used. While in Chap.~\ref{cha:delusions}, both \RDS{} and \SSM{} are used.
}
\label{fig:render_SSM}
\end{figure}

\begin{figure}[htbp]
\centering
\includegraphics[width=0.75\textwidth]{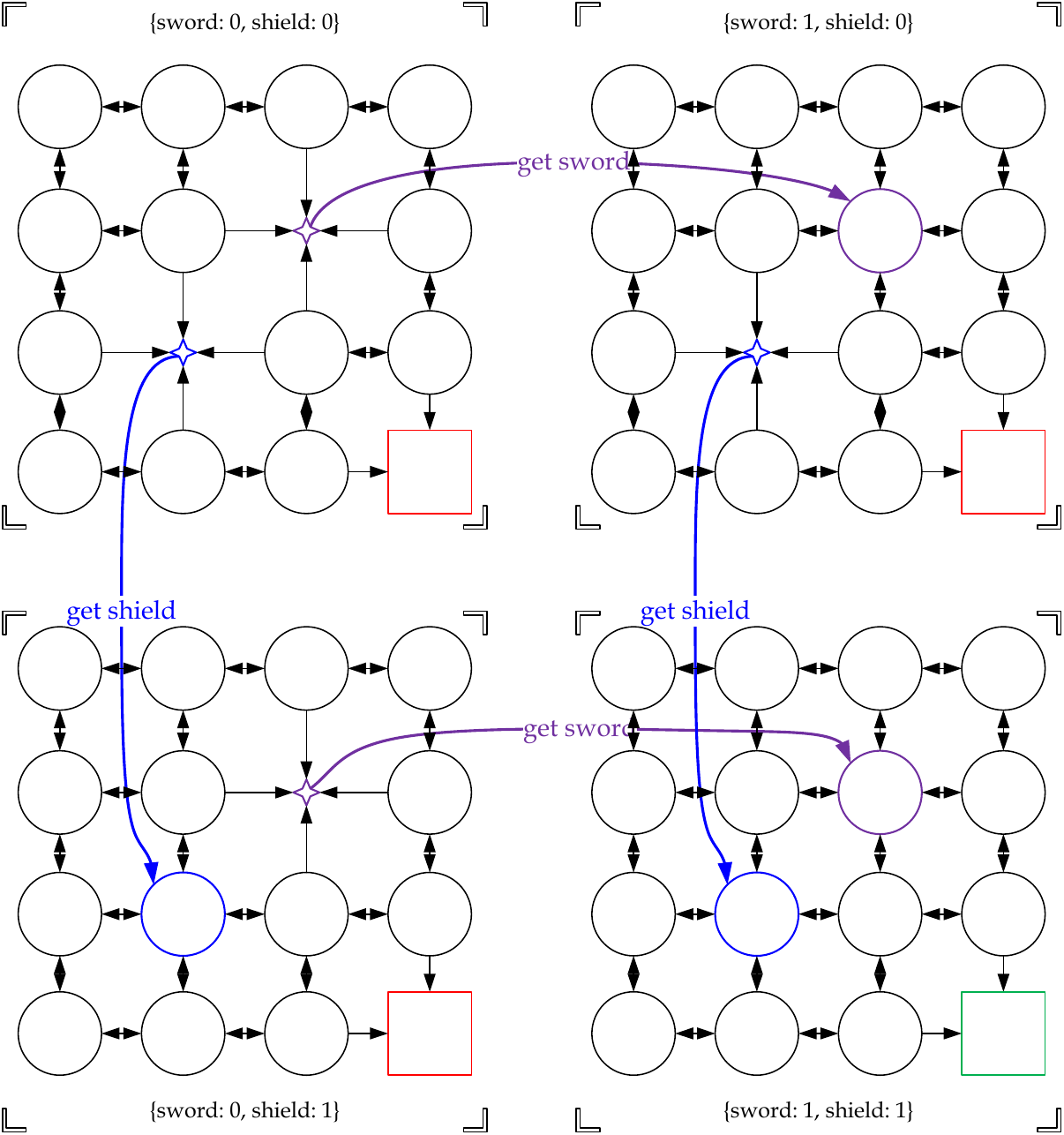}
\caption[\SSM{} State Structure]{\textbf{Visualization of \SSM{} State Structure}: we analyze an instance of $4\times4$ \SSM{} task instance with difficulty $\delta=0$, \ie{}, no lava traps generated, approximately $4$ times as large as a corresponding \RDS{} instance. It can be seen that intra-situation transitions are only facilitated with pivotal transitions that involve the acquisition of the sword or the shield. The only possible situation transitions are $\langle 0, 0 \rangle$ to $\langle 0, 1 \rangle$, $\langle 0, 0 \rangle$ to $\langle 1, 0 \rangle$, $\langle 1, 0 \rangle$ to $\langle 1, 1 \rangle$, and $\langle 0, 1 \rangle$ to $\langle 1, 1 \rangle$. To effectively solve an unseen \SSM{} instance, an agent needs to understand the environment, make plans based on its location and the world layout, deciding if it is the sword or the shield it should be getting first, before success.}
\label{fig:SSM_states}
\end{figure}

Visualizations of \SSM{} instances are in Fig.~\ref{fig:render_SSM}. The state structure and the situations in \SSM{} are visualized in Fig.~\ref{fig:SSM_states}. \SSM{} was designed with the expectation to handle future work directions regarding abstract planning, as explained in Sec.~\ref{sec:future_work} on Page.~\pageref{sec:future_work}.

\subsubsection{Generalization-Focused Experimental Settings}
\label{sec:OOD_settings}

We designed our customized MiniGrid-based environments with the intention to use them in a generalization-focused experimental setting. In this setting, agents are trained in a gridworld with one or a limited number of layouts and then tested on unseen layouts. This setup, referred to as {\fontfamily{qcr}\selectfont DistShift}, is designed to evaluate how well agents generalize to different tasks \citep{mendonca2020meta}. This setting not only reflects how RL may be applied in real-world scenarios, but also tests the agents' abilities to generalize via \textit{understanding} instead of \textit{memorization}. I provide a brief summary of the experimental settings used in the main thesis chapters in Tab.~\ref{tab:exp_settings}.

\begin{table}[ht]
\centering
\begin{tabular}{|c|p{0.15\textwidth}|p{0.1\textwidth}|p{0.25\textwidth}|p{0.25\textwidth}|}
\hline
\textbf{Setting} & \textbf{\RDS{}} & \textbf{\SSM{}} & \textbf{Training} & \textbf{Evaluation} \\ \hline
Chap.~\ref{cha:CP}   & Used \newline (sizes from $6 \times 6$ to $10 \times 10$, $3$ action spaces \& variants) & Not used & Train with $\infty$ environment instances with $\delta = 0.35$ ($2.5\times 10^{6}$ interactions) & Evaluate with \textit{transposed} instances with $\delta$ from \newline $\{ 0.25, 0.35, 0.45, 0.55 \}$ \\ \hline
Chap.~\ref{cha:skipper}              & Used \newline (size $12 \times 12$) & Not used & Train with $\{1, 5, 25, 50, 100, \infty \}$ environment instances with $\delta = 0.4$, random initialization ($1.5\times 10^{6}$ interactions) & Evaluate with $\delta$ from \newline $\{ 0.25, 0.35, 0.45, 0.55 \}$ \\ \hline
Chap.~\ref{cha:delusions}           & Used \newline (size $12 \times 12$) & \textit{Used} (sizes $8 \times 8$ \& $12 \times 12$) & Train with 50 environment instances with $\delta=0.4$, random initialization ($1.5\times 10^{6}$ interactions) & Evaluate with $\delta$ from \newline $\{ 0.25, 0.35, 0.45, 0.55 \}$ \\ \hline
\end{tabular}
\caption[OOD-Focused Experiment Settings for Different Chapters]{\textbf{OOD-Focused Experiment Settings for Different Chapters}:
The experimental settings have evolved throughout the course of my PhD, I will introduce them in more detail in later chapters where they are used. In Chap.~\ref{cha:CP}, we explored the most diverse set of experiments, ranging from transposed instances to varying world sizes. In Chap.~\ref{cha:skipper}, we adjusted the training difficulty to $\delta = 0.4$ to be OOD from all evaluation, and removed the transposed instances to better capture the generalization gap and align with the notion of distribution shifts \citep{soulos2024compositional}. We accelerated training by granting the agents a uniformly random initialization (in terms of their initial locations in the gridworlds). Furthermore, we limited the action space to the ``absolute-direction'' dynamics. We also mostly limited the number of training environments to reflect limited training scenarios in the real world. We tested \Skipper{}'s performance scalability in terms of the number of training environments. In Chap.~\ref{cha:delusions}, we converged at an OOD-focused setting, where the generalization gap can be intuitively visualized, while introducing the new \SSM{} environments.}
\label{tab:exp_settings}
\end{table}

It is similar to the ``train-on-few then test-on-unseen'' approach used in the non-curriculum experiments in ProcGen \citep{cobbe2019procgen}, which focuses on the challenges of learning representations from image-based observations. However, unlike ProcGen, \RDS{} and \SSM{} utilize simple observations to minimize the challenges of representation learning, instead focusing on the inherent difficulty of the tasks themselves. This deliberate design choice isolates the complexities of decision-making and generalization, which are the central focus of this thesis, from the challenges posed by representation learning.

\phantomsection
\label{sec:transient_continual}

Note that, despite the focus on generalization, our experimental setting is considered a \textbf{transient} learning setting with its limitations, instead of a \textit{continual} learning setting that would be necessary to achieve more general intelligence.


%% file: chapter_4_CP.tex
\chapter{Conscious Planning: Spatially-Abstract Decision-Time Reasoning}
\label{cha:CP}


\minitoc

\section{Overview of This Thesis Milestone}
\label{sec:CP_intro}

\textit{We present an end-to-end, model-based deep RL agent that dynamically attends to relevant parts of the environment during its decision-time planning. The agent uses a top-down attention based bottleneck mechanism over a set-based (object-oriented) state representation to limit the number of entities considered during planning. In experiments, we investigate the bottleneck-equipped agent with several sets of customized environments featuring different challenges. We consistently observe that the design allows the planning agents to better generalize their learned task-solving abilities in compatible unseen environments by attending to the relevant objects, leading to better out-of-distribution generalization performance.}

Whether when humans plan our paths home from the offices or from a hotel to an airport, we typically focus on a small subset of relevant variables, \eg{}, the change in position or the presence of traffic. An interesting hypothesis of how this path planning skill generalizes across scenarios is that it is based on computation associated with the conscious processing of information \citep{baars1993cognitive}. Conscious attention focuses on a few necessary environment elements, with the help of an internal abstract representation of the world \citep{vangulick2004consciousness}. This behavioral pattern, also known as consciousness in the \nth{1} sense (C1) \citep{dehane2017consciousness}, was theorized to be the foundation of humans' exceptional adaptability and learning efficiency \citep{baars2002conscious}. A central characterization of conscious processing is that it involves a computational \textbf{bottleneck}, which forces the human brain to handle dependencies among only limited few environmental characteristics at a time. Though focusing on a subset of the available information may seem limited, such limitation counterintuitively facilitates generalization abilities to situations where the ignored variables are different and irrelevant \citep{bengio2017consciousness,goyal2022inductive}.

In this project, we (the collaborators and I) introduce these ideas into RL, as most of the big successes of RL have been obtained by deep, model-free agents \citep{mnih2015human,silver2016mastering,silver2017mastering}. While model-based RL has generated significant research due to the potentials of using an extra model \citep{moerland2020model}, its empirical performance has typically lagged behind, with some recent notable exceptions \citep{schrittwieser2019mastering,kaiser2019model,hafner2020mastering}.

We take inspiration from human conscious planning behaviors to build set-based (object-oriented) neural network architectures which learn a useful state representation and in which attention can be focused on a small set of variables at decision-time, where the aspect of ``partial planning'' (see Sec.~\ref{sec:partial_planning} on Page.~\pageref{sec:partial_planning}) is enabled by modern deep RL techniques \citep{talvitie2008local,khetarpal2021temporally}. Specifically, we propose an end-to-end latent-space model-based RL agent which does not require reconstructing the observations, as in most existing works, and uses Model Predictive Control (MPC) for decision-time planning \citep{richards2005robust,rao2009survey}, which we introduced in Sec.~\ref{sec:MPC} on Page.~\pageref{sec:MPC}. From an observation, the agent encodes a set of objects as a state, with an attention bottleneck mechanism to \textbf{reason} over selected subsets of the state. Our experiments show that the introduced inductive biases improve a systematic OOD generalization, where consistent environmental dynamics are preserved across seemingly different tasks.

\section{Discussions of Methodologies}

Much of the background and context for this project have already been covered in Chap.~\ref{cha:basics} and Chap.~\ref{cha:prelim}. We now focus on aspects specific to this project that will help explain the rationale behind our design choices.

\subsection{About Reconstruction, Observation-Level \& State-Level Planning}
Many model-based RL methods utilize models that operate in the observation space or over state representations obtained with reconstruction-based losses  \citep{wang2018leap,kaiser2019model}. Appropriate as these models might be for tasks with few sensory inputs, \eg{}, continuous control of robots with joint states, these approaches encounter difficulties with high-dimensional inputs such as images, because they must focus on predictable yet potentially useless aspects of the environmental observations \citep{moerland2020model}. Besides the need to reconstruct irrelevant parts of the environment, it is unclear if representations built by a reconstruction loss (\eg{}, $L_2$ in the observation space) are effective for an model-based RL agent to plan or produce desired behaviors at all \citep{silver2016predictron,hafner2020mastering,hamrick2020role}, \eg{}, values, rewards, \etc{}. Taking a similar approach to those in \citet{silver2016predictron,schrittwieser2019mastering}, in this chapter, we build a latent space representation jointly shaped by all the training signals without using reconstruction (to serve value estimation and planning). We briefly discussed this topic previously in Sec.~\ref{sec:dynamics_predictors} on Page.~\pageref{sec:dynamics_predictors}.

\subsection{About Staged Training with World Models}
\label{sec:world_models}

\textbf{World Models} are a popular methodology for model-based RL agents  \citep{ha2018world,kaiser2019model,moerland2020model}. These models require two explicit stages of training: 1) an representation of the environment is trained using exploration, sometimes in an unsupervised fashion; 2) the representation is fixed and used for planning and model-based RL. World models are favored for the lack of distributional shift and their adjacency to supervised deep learning methods. However, despite these advantages, the world model methodology relies on the implicit assumption that the  environments are highly homogeneous, \ie{}, the trained model that could handle the experiences from the exploration stage can handle the more rewarding trajectories as well. This is obviously not true in many environments, and the applicability of such method deteriorates rapidly with the increment of exploration difficulties. Such implicit assumption also demonstrates the limited OOD generalization abilities of such approach and indicate that world models may not be robust enough to generalize in novel situations. This can be viewed from a transient learning \vs{} continual learning perspective, discussed in Sec.~\ref{sec:transient_continual} on Page.~\pageref{sec:transient_continual}. In certain existing literature, world models are also understood to be the methodology of building predictive models based on the actions taken by the agents. We do not use such aspects when we discuss the world models.

Furthermore, despite that the learned representations of a world model are expected to be effective in predicting the evolution of environmental states, they should not be expected to be effective for value estimation at all. This is because of the properties of neural networks: without a value estimation training signal, the learned representation would typically not be effective for value estimation; End-to-end model-based RL agents, \eg{}, \citet{silver2016predictron,schrittwieser2019mastering}, are able to learn the representation of the world simultaneously with the value estimator, hence have the potential to adapt better to non-stationarity in the control tasks.

\subsection{About Vectorized \vs{} Set-based State Representations}
\label{sec:set_encoder}

Most existing deep RL methods employ vectorized state representations, where the agents' observation is encoded into a vector of fixed dimensionality of features \citep{mnih2015human,hessel2017rainbow}. while, set-based encoders, \aka{} object-oriented architectures, are designed to extract a set of unordered vectors (often named \textbf{objects} in such context), from which the desired signals can be predicted via permutation-invariant computations, which is extensively investigated in  \citet{zaheer2017sets,lee2019set}. We illustrate the difference between the two approaches in Fig.~\ref{fig:CP_encoder}. In recent years, the potential of set-based representations are discovered for RL, in terms of effectively building environmental state representations, in terms of compositional generalization \citep{wang2018nervenet,vinyals2019grandmaster, davidson2020investigating,mu2020refactoring,lowe2020contrasting}. In this chapter, we exploit the compositionality of these set-based state representations to enable the discovery of sparse interactions patterns among the encoded objects, as well as to facilitate the core bottleneck mechanism, analogous to the dynamic selection empowered by consciousness in the \nth{1} sense (C1). The combination of the set-based representation and the bottleneck provides a set of inductive biases consistent with dynamically selecting only the relevant aspects of the environmental state through the proposed attention mechanism. The sparsity of the dependencies captured by the learned dynamics model is enforced by the small size of the working memory bottleneck: each transition can only relate a few objects together, no more than the size of the bottleneck \citep{bengio2017consciousness}.

\section{Methodology: Model-based RL with Set Representations}
\label{sec:UP}

We present a baseline end-to-end agent, which uses a set-based representation and carries out latent space decision-time planning, but \textbf{without} a consciousness-inspired small bottleneck \citep{alver2022understanding}. This agent serves as a baseline to investigate the OOD generalization capabilities brought by the bottleneck, which is to be introduced in Sec.~\ref{sec:CP_bottleneck} on Page.~\pageref{sec:CP_bottleneck}.

As discussed previously, a model-free agent (or a model-free part of a model-based agent) relies on the mapping from observations to values, which is in turn a combination of an \textbf{state representation encoder} (encoder for short) and a \textbf{value estimator}. Aiming at a set-based representation, we designed the encoder to map an observation vector to a set of objects. Also, the value estimator is designed to be a permutation-invariant set-to-vector neural network, mapping the learned state representation to a value estimate. With our design, to maximize the gain from end-to-end learning, the same state representation is shared for all the agents' predictions, including the dynamics model's prediction of future states, rewards, \etc{}, to be discussed soon.

\subsection{State Representation Encoder}
\begin{figure}[htbp]
\centering
\captionsetup{justification = centering}
\includegraphics[width=0.8\textwidth]{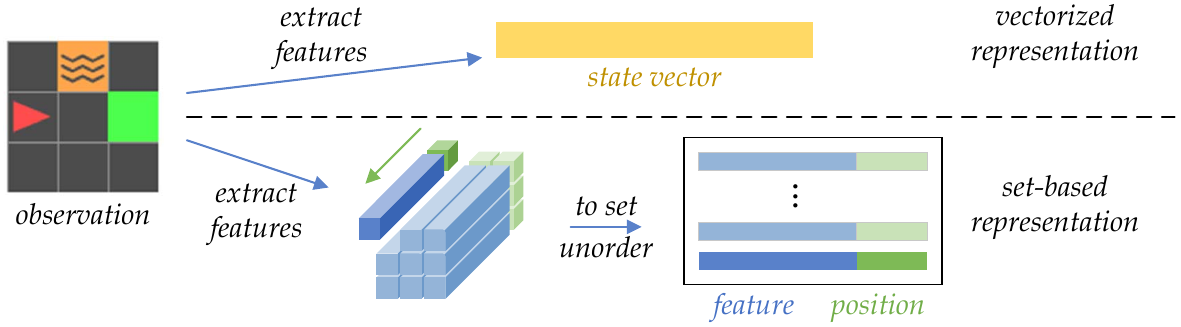}
\caption[Set-based State Representation Encoder]{\textbf{Set-based State Representation Encoder}: Compared to vectorized encoders, the feature map extracted by some feature extractor, \eg{}, a CNN, is ``sliced'' into individual feature vectors and concatenated with positional information. Note that such concatenation is a design choice that serves the dynamics model training purposes. Post concatenation, all resulting vectors are treated as \textit{objects}, indicating all entities an agent could perceive given the current observation.}
\label{fig:CP_encoder}
\end{figure}

We propose a neural network architecture targeting image-based observations. For this, we ``slice'' the CNN-output feature maps each position to characterize the feature of an object, similar to the approach in \citet{carion2020end}, as shown in Fig.~\ref{fig:CP_encoder}. 

When a 3-dimensional CNN feature tensor is sliced into an unordered set of 1-dimensional feature vectors, the positional information of each feature will be lost. To recover such lost information, we concatenate each obtained feature vector with a positional embedding. This method is different from the common practice of mixing positional information by addition, as often used in transformer architectures \citep{vaswani2017attention}. This particular design aims to address a challenge in training a dynamics model with set-based representations, which is to be discussed in Sec.~\ref{sec:CP_model_alignment} on Page.~\pageref{sec:CP_model_alignment}.

\subsection{(State-Action) Q-Value Estimator} 
\begin{figure}[htbp]
\centering
\captionsetup{justification = centering}
\includegraphics[width=1.0\textwidth]{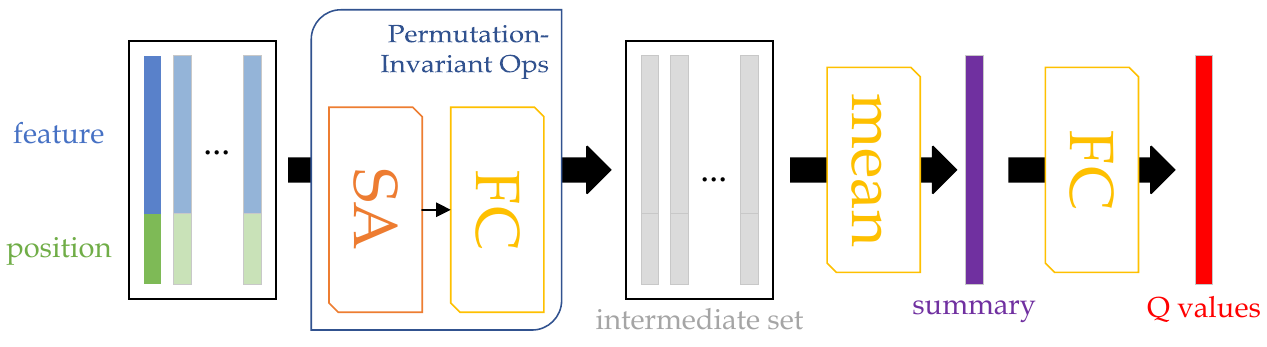}
\caption[Value Estimator \& Set-to-Vec Architecture]{\textbf{Value Estimator} $Q$ \& \textbf{Generic Set-to-Vec Architecture}: We improved upon the DeepSets and Set Transformer architectures \citep{zaheer2017sets,lee2019set} by swapping the Fully Connected (FC) Multi-Layer Perceptron (MLP) before pooling with transformer layers (Self-Attention (SA) + object-wise Fully Connected (FC)) \citep{vaswani2017attention}. After passing through the transformer layers, the intermediate set (gray) will entangle features and positions, \ie{}, the feature and the positional information of each object no longer have their own separate dimensions.}
\label{fig:CP_value_estimator}
\end{figure}

This Q-value estimator is implemented in the form $Q: \scriptS \to \doubleR^{|\scriptA|}$, where $\scriptS$ is the \textit{learned} state representation space by the previously introduced set-based encoder and $\scriptA$ is a discrete action set. The architecture we propose for this component uses an improved architecture upon DeepSets \citep{zaheer2017sets} and Set Transformer \citep{lee2019set}, as shown in Fig.~\ref{fig:CP_value_estimator}. This architecture performs reasoning based on the relationship among the encoded objects of the input set, resembling token-based methods in natural language processing \citep{porada2021modeling}.

\subsection{Transition Model}
We design a set-to-set transition model that maps from $s_t, a_t$ to $\hat{s}_{t+1}$, $\hat{r}_t$ and $\hat{\omega}_{t+1}$. The model is a sample model that predicts the outcome of a transition.

For clarity, we separate the computational flow of the transition model into: 1) the \textbf{dynamics model}, responsible for simulating how the state would evolve with an intended action $a_t$ and 2) the \textbf{reward-termination estimator} which maps $s_t, a_t$ to $\hat{r}_t$ and $\hat{\omega}_{t+1}$.

\begin{figure}[htbp]
\centering
\captionsetup{justification = centering}
\includegraphics[width=0.95\textwidth]{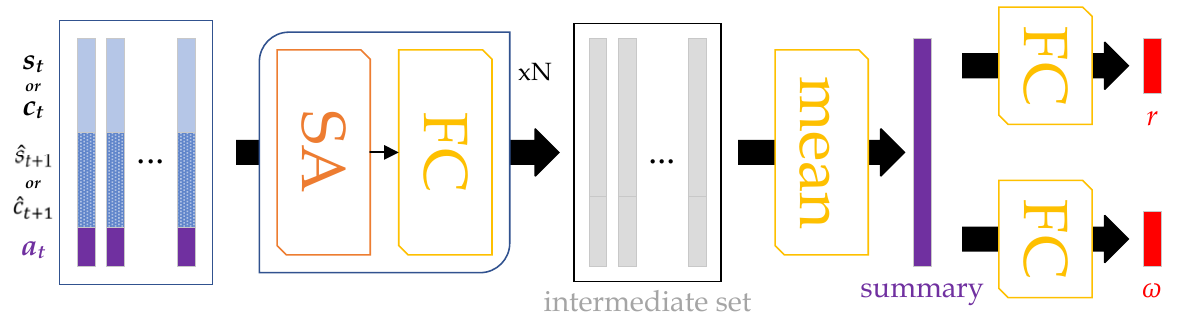}
\caption[Reward-Termination Estimator based on Set Transitions]{\textbf{Reward-Termination Estimator based on Set Transitions}: The computational flow requires $3$ inputs. The \nth{1} input is either the full set $s_t$ for agents without the bottleneck, or bottleneck set $c_t$, for the bottlenecked agents (from the selection). The \nth{2} input is the respective output of the dynamics model for the next timestep, \ie{}, the predicted full set $\hat{s}_{t+1}$ or $\hat{c}_{t+1}$. The \nth{3} input is the embedding of the action that governs the simulated transition, \ie{}, the intention of the agent in the search step. The first two inputs are aligned and concatenated with the tiled action embeddings to form an augmented set, as shown to the left of the figure. The augmented set is then passed through the transformer layers to form an intermediate set, which is then mean-pooled and projected to the two predictions. With deterministic environments, it will be sufficient to predict the reward and termination with only $s_t$ and $a_t$.}
\label{fig:rt_estimator}
\end{figure}

\phantomsection
\label{sec:CP_model_alignment}

\begin{figure}[htbp]
\centering
\captionsetup{justification = centering}
\includegraphics[width=1.0\textwidth]{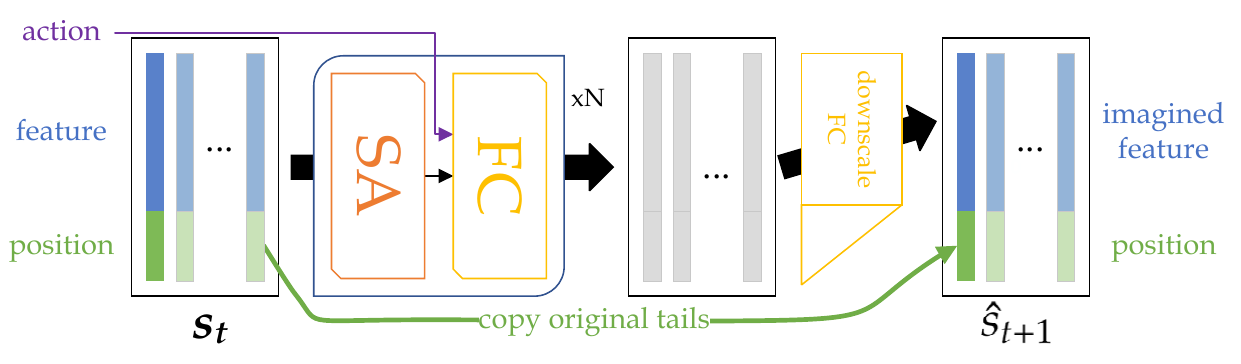}
\caption[Set-Based Dynamics Model]{\textbf{Set-Based Dynamics Model}: This dynamics model architecture takes state set $s_t$ and action embedding $a_t$ as input and outputs the imagined next state $\hat{s}_{t+1}$. For the FC sub-layers of the transformer layers, we inject an action embedding \st{} the transformer computations become action-conditioned. This process is illustrated in detail in Fig.~\ref{fig:layer_transformer_conditioned}. After getting the intermediate set, we use linear projections to downscale each object to the correct dimensionality, while forcing the positions untouched and directly copied from the input $s_t$. Note that though the objects in the sets (input-intermediate-output) are inter-aligned, within each set they are still unordered, \ie{}, permutation-invariant.}
\label{fig:CP_dynamics_UP}
\end{figure}

While designing a reward-termination estimator should be  straightforward (a 2-headed augmented architecture similar to the value estimator, as shown in Fig.~\ref{fig:rt_estimator}), the dynamics model requires predicting changes on sets of \textit{unordered} objects (set-to-set). However, since the position and the feature of each object can change at the same time, the predicted set of objects may not align with their counterparts from the input set trivially. These additional degrees of freedom raise the computational difficulties of set-to-set predictions, often rendering the learning process ineffective. To address such challenge, in existing deep learning literature, a common approach is to use alignment methods, \eg{}, aligned loss functions such as Hausdorff distance or Chamfer matching. Nevertheless, these approaches are computationally inefficient and subject to local optima \citep{barrow1977parametric,borgefors1988hierarchical,kosiorek2020conditional}. With these drawbacks in mind, we deliberately separated the dimensions of feature and position in our objects: this design not only make the permutation-invariant computations position-aware, but also allow surprisingly simple end-to-end training losses over the dynamics, similar to those used for agents based on vectorized state representations. This is made possible by forcing the dimensions of positional information, \ie{}, the ``position tails'', to stay unchanged during the dynamics prediction. Such anchor addresses the difficulty of  alignment by design: the degree of freedom of changing positional tails is no longer allowed. Objects ``labeled'' with the same positional information in the output of the dynamics model $\hat{s}_{t+1}$ and the training sample input  $s_{t+1}$ must be aligned, forming pairs of objects with changes solely in the feature dimensions. This is further explained in Fig.~\ref{fig:CP_dynamics_UP}.

\begin{figure}[htbp]
\centering
\captionsetup{justification = centering}
\includegraphics[width=0.95\textwidth]{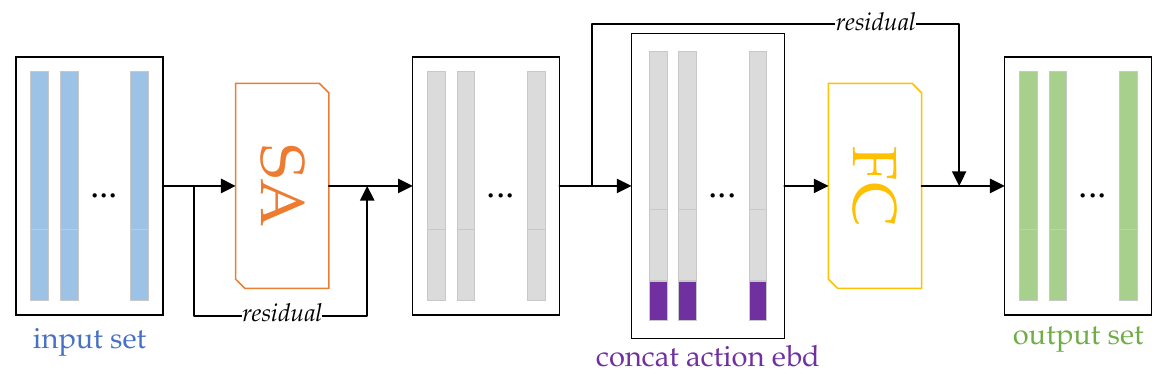}
\caption[Action-Conditioned Transformer Layer]{\textbf{Action-Conditioned Transformer Layer}: compared to the vanilla transformer layers, we concatenate additionally the action embedding to every intermediate object embeddings in the FC sub-layers of transformer layers. The FC sub-layer implements $X' = X + f(cat[X,a])$, where $X$ is the input set and  $cat([X,a]])$ is the concatenation of action embedding $\bm{a}$ to every object embedding in $X$, with $X'$ being the output set. $f$ needs to correctly downscale the dimensionality of its input to match that of $X$. Layer normalization layers are omitted for simplicity.}
\label{fig:layer_transformer_conditioned}
\end{figure}

The action-conditioned transformer layer serves as the backbone of set-to-set learning in this chapter. A vanilla transformer layer consists of two consecutive sub-layers, the multi-head SA and the fully connected, each containing a residual pass. To make a vanilla transformer layer action-conditioned, we first embed the discrete actions into a vector and then concatenate it to every intermediate object output by the SA sub-layer. A detailed illustration of the action-conditioned transformer layer is provided in Fig.~\ref{fig:layer_transformer_conditioned}.

\subsection{Training}
\label{sec:CP_training}
The baseline model-based agent is trained the following losses (over sampled transitions):

\begin{itemize}[leftmargin=*]
\item 
Value Estimation $\scriptL_{\text{TD}}$: push the current value estimates towards the update targets. These targets can be acquired with techniques such as Double DQN (\DDQN{}) \citep{hasselt2015double}. In experiments, C51-style distributional architectures as used for all scalar predictions including values and rewards, making $\scriptL_{\text{TD}}$ a KL-divergence \citep{bellemare2017distributional}.
\item
Dynamics Prediction $\scriptL_{\text{dyn}}$: A $L_2$ penalty established between the aligned $\hat{s}_{t+1}$ and $s_{t+1}$, where $sg(\hat{s}_{t+1})$ is the imagined next state given $o_t, a_t$ and $s_{t+1}$ is the true next state encoded from $o_{t+1}$. The stop-gradient function $sg$ indicates that the gradient updates will treat $\hat{s}_{t+1})$ as a constant tensor and do not compute the associated partial derivatives. This loss is made possible by our separate-dimension design of feature-position object representations.
\item
Reward Prediction $\scriptL_{r}$: given the C51 architecture, the KL-divergence between the imagined reward $\hat{r}_{t+1}$ predicted by the model and the true reward $r_{t+1}$ of the observed transition.
\item
Termination Prediction $\scriptL_{\omega}$: this is an optional binary cross-entropy loss from the imagined termination $\hat{\omega}_{t+1}$ to the ground truth $\omega_{t+1}$, obtained from environment feedback.
\end{itemize}

The overall loss for end-to-end training of this set-based model-based baseline agent is thus:
\[
    \mathcal{L} = \scriptL_{\text{TD}} + \scriptL_{\text{dyn}} + \scriptL_{r} + \scriptL_{\omega}
\]
As discussed earlier, jointly shaping the state representations avoids the collapse to trivial solutions and makes the representations useful for predicting all signals  of interest.

In our implementation, no re-weighting is used for each loss term. This simplicity is possible because most individual losses are entropy-like and thus similar in magnitudes.

\section{Methodology: Consciousness-Inspired Bottleneck}
\label{sec:CP_bottleneck}

Planning should focus on the relevant parts of the environment that matter the most for the intended plan. In this section, we introduce an inductive bias which facilitates C1-capable planning, which can be applied to the baseline agent introduced above. Such inductive bias is built on the observation that, with each action taken, only a limited few aspects of the environment should change, as in the real world. 

\begin{figure}[htbp]
\centering
\captionsetup{justification = centering}
\includegraphics[width=0.85\textwidth]{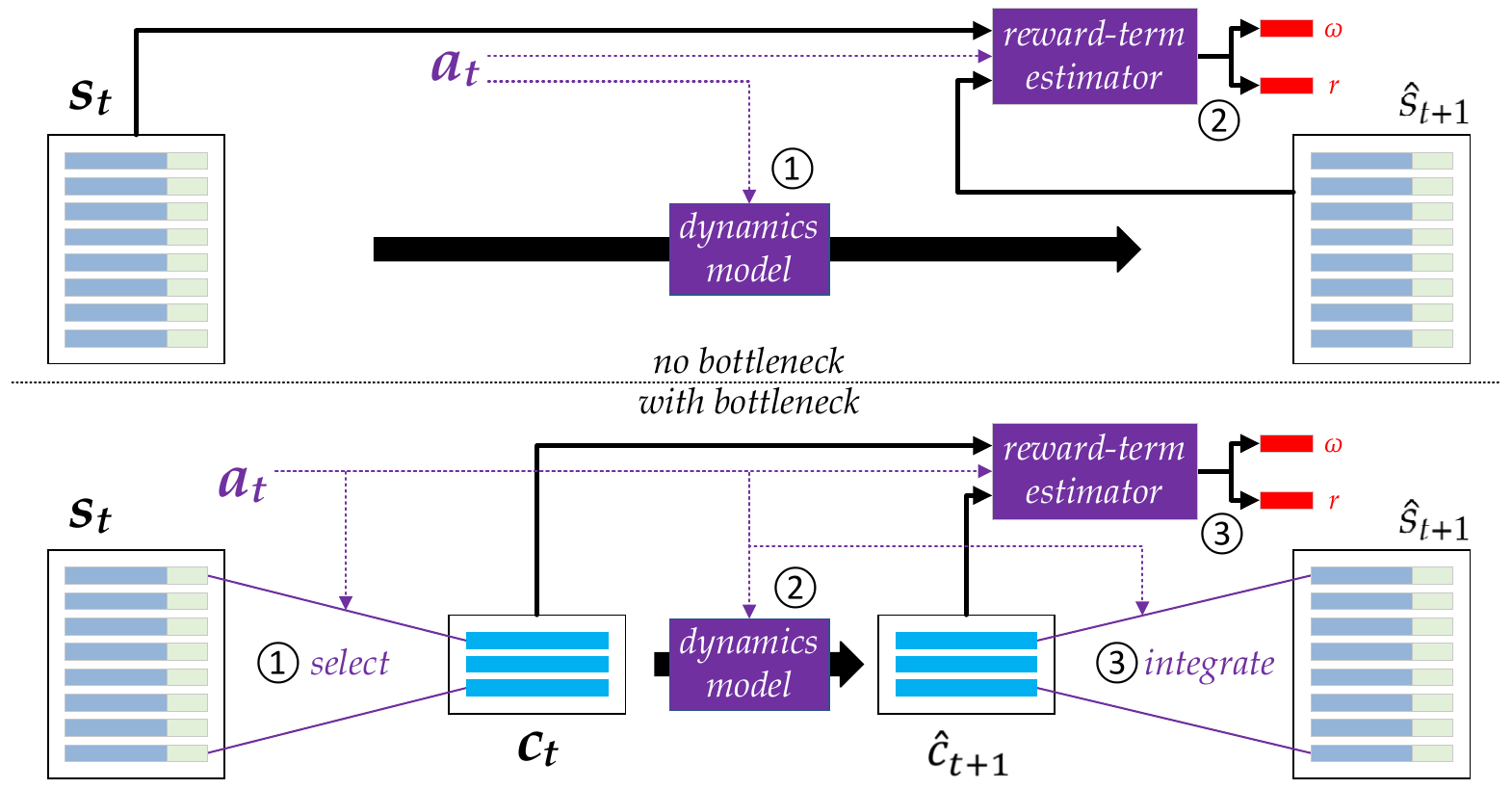}
\caption[Bottleneck-Enabled Set-Based Dynamics Model]{\textbf{Stages of Bottleneck-Enabled Set-Based Dynamics Model}: Similar to the baseline model illustrated in Fig.~\ref{fig:CP_dynamics_UP}, this component also takes the state set $s_t$ and the action embedding $a_t$ as input, and outputs the predicted next state $\hat{s}_{t+1}$, together with the transitional reward $r$ and the termination signal $\omega$. However, it produces some very different intermediate predictions / representations. 1) a bottleneck set $c_t$ is soft-selected from the whole state set $s_t$ via semi-hard top-down attention, conditioned on the intended action. Details of this operation are illustrated in Fig.~\ref{fig:compressor}; 2) dynamics are applied to the bottleneck set $c_t$ to form $\hat{c}_{t+1}$. This part is implemented with the same architecture presented in Fig.~\ref{fig:layer_transformer_conditioned}; 3) similarly, the reward and termination signals are predicted out of $c_t$, $\hat{c}_{t+1}$ and $a_t$, as depicted in Fig.~\ref{fig:rt_estimator}. At the same time, the changes introduced in $\hat{c}_{t+1}$ are \lowercase{\decompress{}}d with $s_t$ to obtain $\hat{s}_{t+1}$, the imagined next state, with an attention-like operation, as shown in Fig.~\ref{fig:decompressor}. The two computational flows in stage $3$ are naturally parallelizable.}
\label{fig:CP_bottleneck}
\end{figure}

From the input of a full state set $s_t$, we aim to capture all the relevant aspects of the state in a \textit{small} bottleneck set, which is expected to \textit{contain all the important transition-related information}. With such bottleneck set $c_t$, we can perform predictions about the change in the environmental state, the immediate reward, \etc{}. To be more precise, as illustrated in Fig.~\ref{fig:CP_bottleneck}, the model performs 1) selection of the bottleneck set from the full state-set (to acquire $c_t$), 2) dynamics simulation on the bottleneck set (from $c_t$ to $\hat{c}_{t+1}$) and 3) integration of predicted bottleneck set to form the predicted next state (to acquire $\hat{s}_{t+1}$).

\subsection{Conditional State Selection}
\begin{figure}
\centering
\captionsetup{justification = centering}
\includegraphics[width=0.8\textwidth]{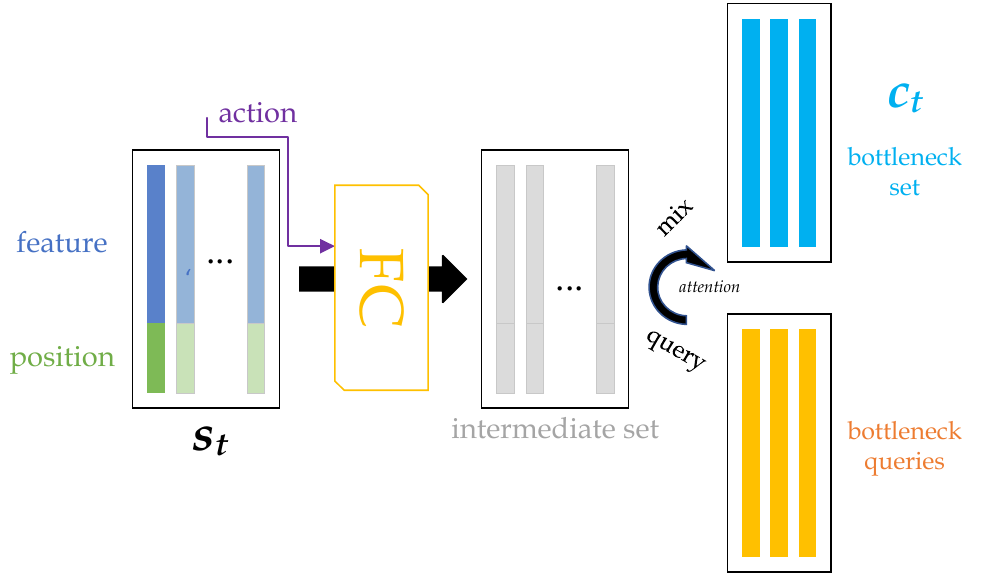}
\caption[Design of the Bottleneck \compressor{}]{\textbf{Design of the Bottleneck \compressor{}}: the bottleneck set $c_t$ is obtained by querying the (transformed) state set $s_t$, denoted as the \textit{intermediate set}, with a learned query set of size $k$, denoted as the \textit{bottleneck queries}, using a custom semi-hard attention \citep{gupta2021memory}. The selection is conditioned on the chosen action. Note that, for this component, self-attention should NOT be used to transform $s_t$ to create the value set (in terms of key-query-value) for the output, otherwise a mixture of all objects would be created, defeating the purpose of the bottleneck.
}
\label{fig:compressor}
\end{figure}

We soft-select a bottleneck set $c_t$ of $n$ objects from the potentially large input state set $s_t$ of $m \gg n$ objects. Then we only model the transition for the selected objects in $c_t$. To implement this, we creatively implement an action-conditioned key-query-value attention mechanism, where the source of keys and values come from $s_t$ and $a_t$, and the queries come from some learned dedicated set of vectors and of the action considered, as shown in Fig.~\ref{fig:compressor}.

To dodge the difficulties of a hard attention bottleneck, we use a semi-hard top-$k$ attention mechanism to facilitate the selection of the bottleneck set. This semi-hard attention technique limits the influence of the irrelevant objects on the bottleneck set $c_t$ while allowing for a gradient to propagate on the assignment of relative weight to different objects. With purely soft attention, weights for irrelevant objects are never $0$ and learning to disentangle objects may be more difficult. Please refer to Sec.~\ref{sec:semihard_attention} on Page.~\pageref{sec:semihard_attention} for more details regarding how we used semi-hard top-$k$ attention to make sure that the output set $c_t$ is a transformed subset of the state set $s_t$ (the set being queried).

\subsection{Dynamics / Reward-Termination Prediction on Bottleneck Sets}

We use the same architecture as described in Sec.~\ref{sec:UP} (in Fig.~\ref{fig:rt_estimator}), but taking the bottleneck objects as input rather than the full state set.

\subsection{Change Integration}

\begin{figure}
\centering
\captionsetup{justification = centering}
\includegraphics[width=0.95\textwidth]{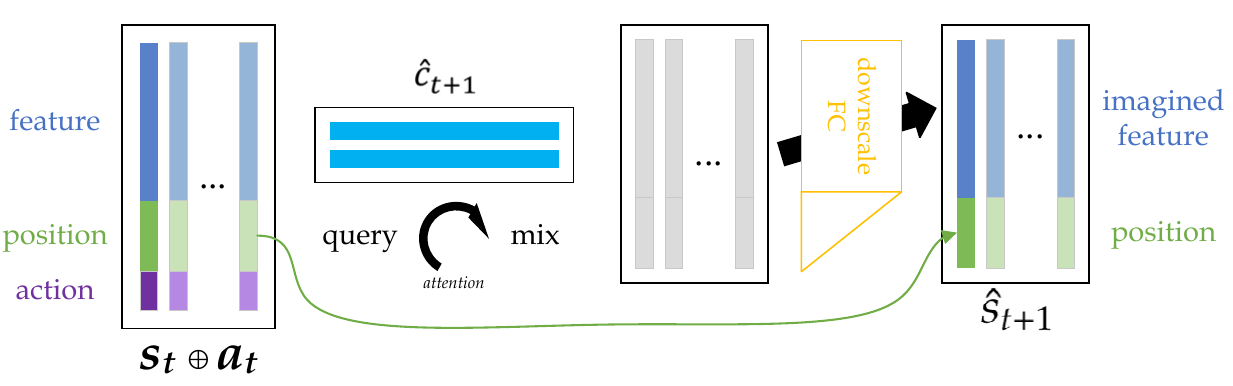}
\caption[Design of the Bottleneck \decompressor{}]{\textbf{Design of the Bottleneck \decompressor{}}: This component takes the $3$ inputs, the state set $s_t$, the bottleneck set $\hat{c}_{t+1}$ and the action embedding $a_t$. The output $\hat{s}_{t+1}$ is generated by using the action-augmented $s_t$ to query the imagined bottleneck set $\hat{c}_{t+1}$ with an attention-like operation without the \softmax{} on attention weights. Note that there is the similar operation of downscaling objects to features and copying the positional tails. For more details of the query operation, please refer to Sec.~\ref{sec:attention} on Page.~\pageref{sec:attention}. Note that because there is no SA used to transform the full state set $s_t$, meaningful changes during $s_t \to s_{t+1}$ must be modeled by the query of the bottleneck set $\hat{c}_{t+1}$. Thus, the integrator cannot be used to bypass the bottleneck.
}
\label{fig:decompressor}
\end{figure}

The evolution on the bottleneck set should cover the change of the full state set. Thus, we implement an integration operation to `soft paste-back' the changes of the bottleneck state onto the state set $s_t$, yielding the imagined next state set $\hat{s}_{t+1}$. Intuitively, this integration is somewhat similar to an inverse operation of selection process. This `soft paste-back' is implemented with attention-like operations, more specifically querying $\hat{c}_{t+1}$ with $s_t$, conditioned on the action $a_t$. We emphasize attention-``like'' here because, since certain objects in $s_t$ should not interact with anything from the bottleneck set at all, we allow the attention weights to be $0$ by forgoing the \softmax{} operation when computing the attention weights. Details of the architecture regarding change integration are in Fig.~\ref{fig:decompressor}.

\subsection{Discussion}
The proposed bottleneck is a natural complement to the baseline agent introduced previously. In particular, planning and training are carried out the same way as discussed in Sec.~\ref{sec:UP}.

One may wonder if it is appropriate to have the value estimator share the bottleneck with the transition model. To estimate a value, the estimator needs to consider the future object interactions, which likely include the objects that do not matter for the current transition; Additionally, it is not necessary to implement an explicit bottleneck mechanism on the value estimator, since the mean-pooling operation functions as the selection if some objects are learned to be ignored.

We call the baseline agent equipped with the bottlenecked inductive biases in this section the Conscious Planning (CP) agent. We expect the CP agent to demonstrate the following advantages:
\begin{itemize}[leftmargin=*]
\item
More Effective Generalization: only relevant objects participate in each planning step, thus each planning step is handling a smaller scale, more manageable problem. Thus, generalization should be improved even in OOD scenarios, because the transitions do not depend on the parts of the state ignored by the bottleneck.
\item
Lower Computational Cost: directly employing transformers to simulate the full state dynamics results in a complexity of $\scriptO(|s_t|^2 d)$, where $d$ is the dimensionality of the objects, due to the use of self-attention, while the bottlenecked variants lowers the related parts to $\scriptO(|s_t||c_t| d)$.
\end{itemize}

The partial predictions enabled by the bottleneck can be viewed from the perspectives of partial / local models. While, previous works have more emphasis on the selective attention towards points in history (temporal focus) \citep{talvitie2008local}, and our bottleneck-equipped model has the attention towards aspects within state and state transitions.

\subsection{Birdseye View of Overall Design}
\begin{figure}[htbp]
\centering
\captionsetup{justification = centering}
\includegraphics[width=0.85\textwidth]{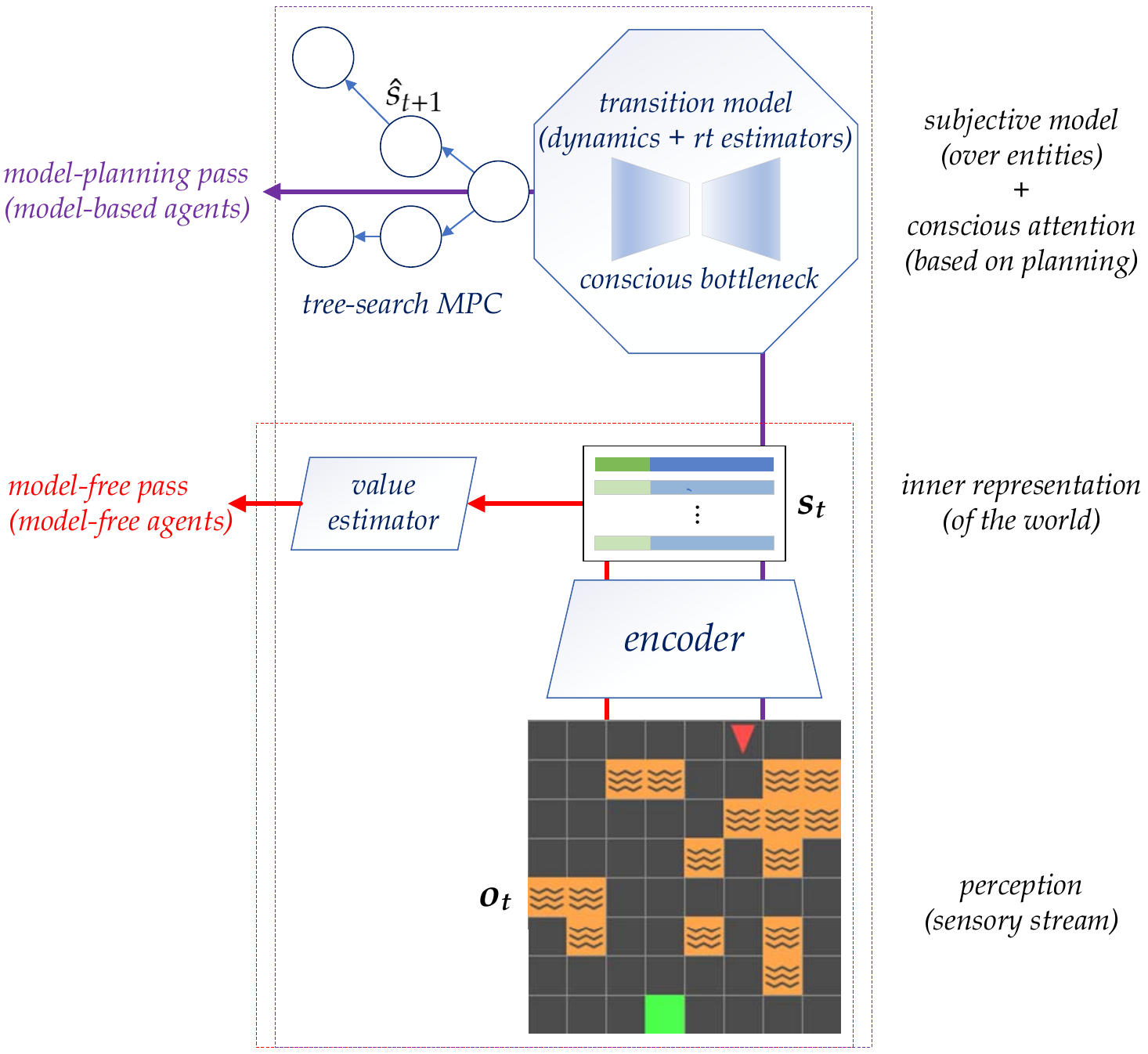}
\caption[Overall Organization of Proposed Components for CP Agent]{\textbf{Overall Organization of Proposed Components for CP Agent}: Both the model-free baseline and the model-based baseline without the bottleneck can be extracted from the figure as well.}
\label{fig:birdseye}
\end{figure}

We assemble the components and present the organization of the proposed CP agent in Fig.~\ref{fig:birdseye}. 

\section{Research Findings: Experiments}
\label{sec:CP_experiments}

We wish to empirically validate the advantages of generalization brought by the introduced consciousness-inspired bottleneck. This is to be done with rigorously  controlled experiments and ablation studies.

\subsection{Environment / Task Description}

We use \RDS{} as the backbone environment in the experiments of this chapter. Crucial for the experimental insights and isolating the unwanted technical difficulties such as learning from complicated visual observations, \RDS{} tasks provide clear object definitions, with clear and intuitive dynamics based on object interactions. For more details regarding the environment, please check Sec.~\ref{sec:RDS} on Page.~\pageref{sec:RDS}.

For demonstration purposes, the main set of experiments we will be using in this chapter are conducted on $8 \times 8$-sized \RDS{} instances paired with ``turn-or-forward'' dynamics. In the later parts, we will provide additional test settings with different dynamics and different world sizes (Sec.~\ref{sec:CP_more_exp}, Page.~\pageref{sec:CP_more_exp}).

For better generalization, the agent needs to understand how to avoid lava in general (and not at specific locations, since their placement changes) and to reach the target locations as quickly as possible. For the agent to be able to \textit{understand} the environment dynamics instead of \textit{memorizing} specific task layouts, we generate a new environment for each episode (training or evaluation), facilitating essentially a multi-task setting. In each training episode, the agent starts at a random position on the leftmost or rightmost edge, and the goal is placed randomly somewhere on the opposite edge. Consistent with all \RDS{} instances, the difficulty parameter $\delta$ controls partially how seemingly different the OOD evaluation tasks are to the in-distribution training tasks, though we know the underlying dynamics of all these tasks are consistent. For training episodes, the difficulty is fixed to $\delta=0.35$\footnote{Based on \RDS{}, this was the \nth{1} OOD-focused experimental setting used in this thesis, The setting was later refined in Chap.~\ref{cha:skipper} and Chap.~\ref{cha:delusions}.}. 

To be specific, the OOD generalization we refer to in this chapter is \textit{the agents' ability to generalize its learned task skills across seemingly different tasks with common underlying dynamics}, \ie{}, systematic generalization \citep{frank2009connectionist}. For OOD evaluation, the agent is expected to adapt to new tasks with consistent underlying dynamics in a $0$-shot fashion, \ie{}, with the agent's parameters fixed \citep{sylvain2019locality}. In other words, the agent will be challenged with distribution shifts \citep{mendonca2020meta,quinonero2022dataset}. 

\begin{figure}
\centering
\subfloat[In-Dist, $\delta = 0.35$]{
\captionsetup{justification = centering}
\includegraphics[width=0.185\textwidth]{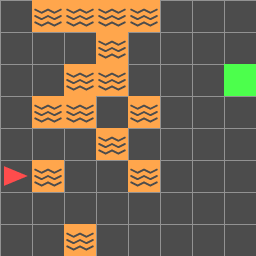}}
\hfill
\subfloat[OOD, $\delta = 0.25$]{
\captionsetup{justification = centering}
\includegraphics[width=0.185\textwidth]{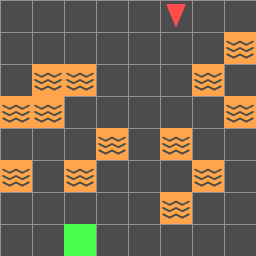}}
\hfill
\subfloat[OOD, $\delta = 0.35$]{
\captionsetup{justification = centering}
\includegraphics[width=0.185\textwidth]{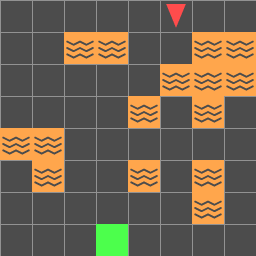}}
\hfill
\subfloat[OOD, $\delta = 0.45$]{
\captionsetup{justification = centering}
\includegraphics[width=0.185\textwidth]{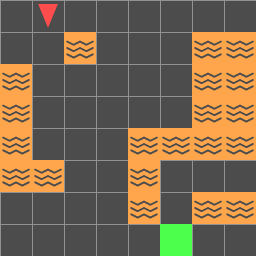}}
\hfill
\subfloat[OOD, $\delta = 0.55$]{
\captionsetup{justification = centering}
\includegraphics[width=0.185\textwidth]{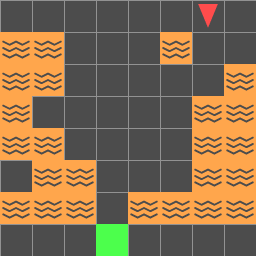}}
\caption[Multitask RL Setting, with In-Distribution and OOD Tasks on \RDS{}]{\textbf{Multitask RL Setting, with In-Distribution and OOD Tasks on \RDS{}}: \textbf{a)} example of training environments; \textbf{b) - e)} examples of OOD evaluation environments (transposed with a gradient of OOD difficulties $\delta$). For each episode (training or OOD), we randomly generate a new environmental instance, which we call a ``task'', from a sampling distribution controlled by certain $\delta$. Note that the training environments and the OOD testing environments have almost no intersections.}
\label{fig:distshift}
\end{figure}

The OOD evaluation tasks are designed to challenge the trained agents. These tasks include changes both in the support (orientation) and in the distribution (difficulty): the agents are deployed in \textit{transposed} layouts that they have never seen before with varying levels of difficulties ($\{0.25, \textbf{0.35}, 0.45, 0.55\}$) that they were not trained on. The differences of in-distribution (training) and OOD (evaluation) environments are illustrated in Fig.~\ref{fig:distshift}. To be more specific, in transposed tasks, an agent starts at the top or bottom edge and the goal grid is on the farthest edge (bottom or top), whereas a training environment has the agent and goal on the left or right edges.

\subsection{Compared Methods}
We based all compared agents on a common model-free baseline: a set-based variant of \DDQN{} (introduced in Sec.~\ref{sec:DDQN} on Page.~\pageref{sec:DDQN}, introduced in \citet{hasselt2015double}) with prioritized replay and distributional outputs. We make all compared methods share architectures as much as possible to ensure fair comparisons. Details of the compared methods, their adaptation, and hyperparameters are provided in the Appendix (Sec.~\ref{sec:CP_aux}, Page.~\pageref{sec:CP_aux}).

\subsection{Agent Variants}
We compare the proposed approach, labeled CP (for Conscious Planning) in the figure legends, against the following methods (variants):
\begin{itemize}[leftmargin=*]
\item
\textbf{UP} (for ``Unconscious Planning''): the agent proposed in Sec.~\ref{sec:UP}, lacking the bottleneck.
\item
\textbf{model-free}: this model-free set-based agent is the common backbone for all the compared set-based model-based agents. It consists of only the encoder and the value estimator, sharing their architectures with CP and UP, as shown in the red box in Fig.~\ref{fig:birdseye}.
\item
\textbf{\Dyna{}}: a set-based model-based RL agent which includes a model-free agent and an observation-level transition model, \ie{}, a transition generator, which shares the same architecture as the CP transition model (with the same hyperparameters as the best performing CP agent), yet applied on the observation-level without an encoder. If performing perfectly, the agent could essentially double the batch size of the model-free baseline by augmenting training batches with an equal number of generated transitions.
\item
\textbf{\Dyna{}*}: A \Dyna{} variant using the ground truth environment model for transition generation. This is expected to demonstrate \Dyna{}'s best-possible performance.
\item
\textbf{WM-CP}: A world model CP variant that is different for following a $2$-stage training procedure \citep{ha2018world}. First, the model (together with the state representation encoder) is trained with $10^{6}$ random transitions. After this, the encoder and the model are fixed and RL (value estimation) begins.
\item
\textbf{\NOSET{}}: A UP-counterpart with vectorized representations and no bottleneck mechanism.
\end{itemize}

\subsection{Performance Evaluation (\RDS{} with Turn-Or-Forward Dynamics)}

\subsubsection{In-Distribution}
\begin{SCfigure}[][htbp]
\includegraphics[width=0.4\textwidth]{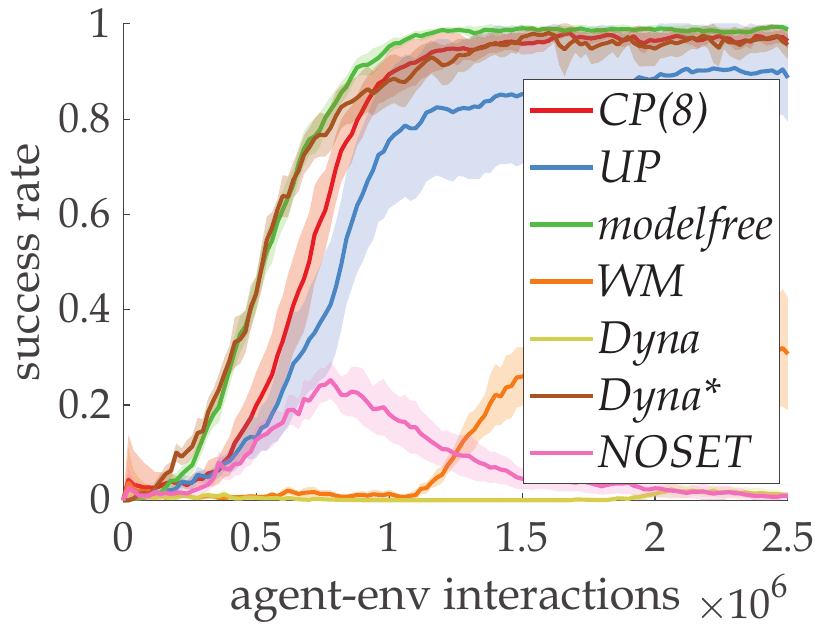}
\caption[In-Distribution Task Performance]{\textbf{In-Distribution Task Performance}: the $x$-axis shows the training progress ($2.5\times10^{6}$ agent-environment interactions). The $y$-axis values are generated by agent snapshots at times corresponding to the $x$-axis values. The bands denote 95\%-CI. CP, UP, model-free and \Dyna{}* agents all learn to solve the in-distribution tasks quickly. All error bars are 95\% confidence intervals obtained from $20$ independent seed runs.}
\label{fig:in_dist}
\end{SCfigure}

In Fig.~\ref{fig:in_dist}, we present the in-distribution training performance for the compared agents, \ie{}, the agents are evaluated with tasks sampled from the same distribution of tasks as that of training.

For \textbf{UP}, \textbf{CP} and the corresponding \textbf{modelfree} baselines, the performance curves indicate convergence to optimal success rates, with confidence intervals overlapping, showing no statistically significant difference in performance. These demonstrate that the $3$ agents are effective in learning to solve the in-distribution tasks. 

During \textbf{WM}'s ``warm-up'' period, the model learns a representation that captures the underlying dynamics. After the warm-up, the encoder and the model parameters are fixed, and only the value estimator learns to predict the state-action values based on the learned representation. From Fig.~\ref{fig:in_dist}, we can observe that the increase in performance is not only delayed due to the warm-up phase (during which rewards are not taken into account), but also visibly harmed, presumably because the value estimator has no ability to shape the representation to better suit its needs.

For in-distribution evaluation, \Dyna{} performs badly while \Dyna{}* perform relatively well. We suspect that this is due to the delusional transitions generated at the early stages of training, from which the value estimator never recovers\footnote{This was validated and investigated by concurrent work \citet{jafferjee2020hallucinating} and importantly in later work \citet{lo2024goal}, which is not only heavily connected to Chap.~\ref{cha:skipper}, but also draws great similarity to approaches used in Chap.~\ref{cha:delusions}.}. However, despite that \Dyna{}* is by design free of this trouble and exhibits satisfactory training performance, we can see that it does not achieve satisfactory OOD performance (later in Fig.~\ref{fig:comparison_OOD}), due to the limited OOD generalization abilities of background planning methodology \citep{alver2022understanding}.

\NOSET{} performs very badly even in-distribution, per Fig.~\ref{fig:in_dist}. In later experiments, we show that \NOSET{} seems only able to perform well in a more classical, monotask RL setting, suggesting its reliance on memorization. There, we provide more results regarding the model accuracy.

\subsubsection{OOD Evaluation Performance}

\begin{figure}[htbp]
\centering
\subfloat[OOD, $\delta = 0.25$]{
\captionsetup{justification = centering}
\includegraphics[width=0.242\textwidth]{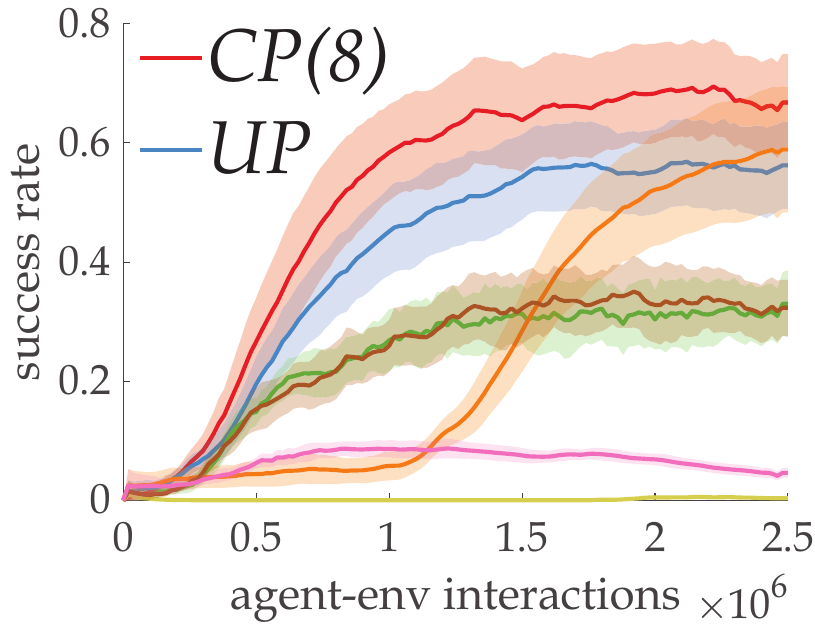}}
\hfill
\subfloat[OOD, $\delta = 0.35$]{
\captionsetup{justification = centering}
\includegraphics[width=0.242\textwidth]{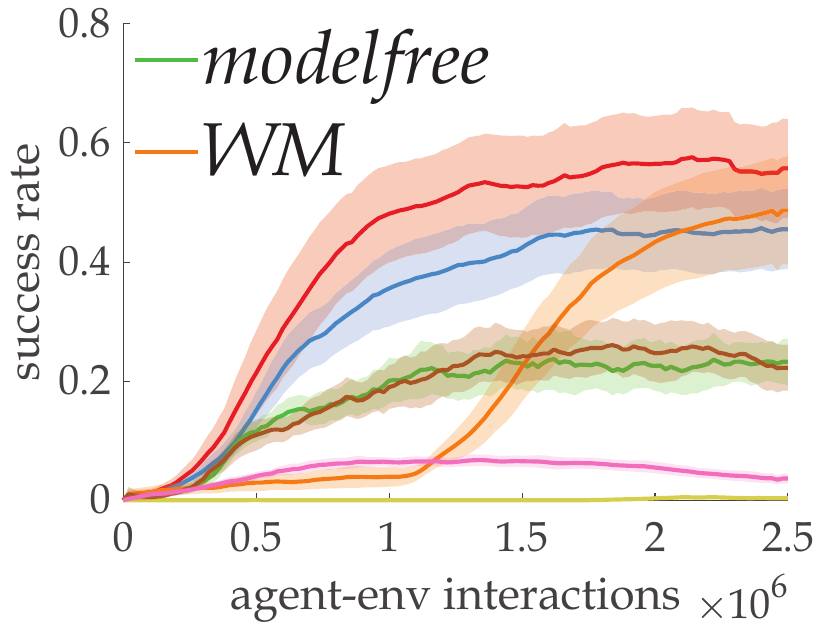}}
\hfill
\subfloat[OOD, $\delta = 0.45$]{
\captionsetup{justification = centering}
\includegraphics[width=0.242\textwidth]{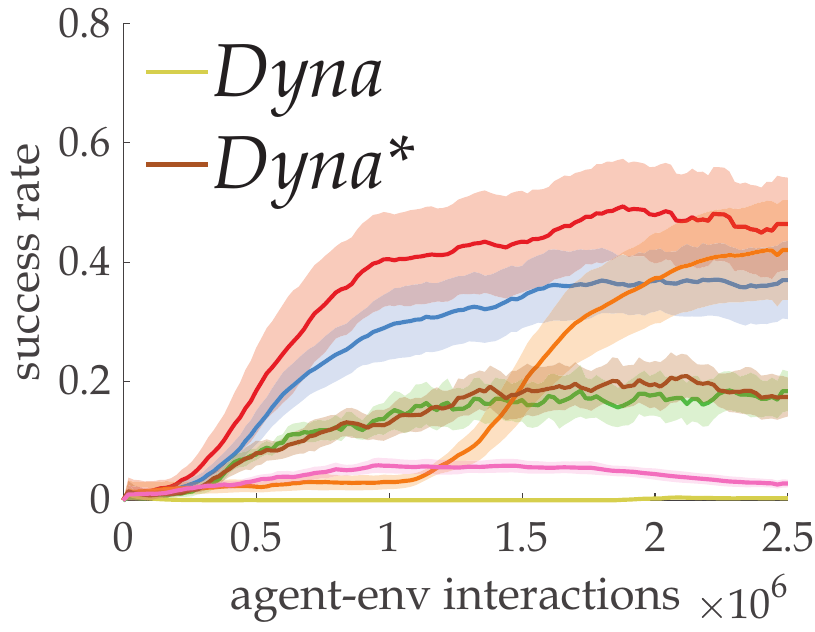}}
\hfill
\subfloat[OOD, $\delta = 0.55$]{
\captionsetup{justification = centering}
\includegraphics[width=0.242\textwidth]{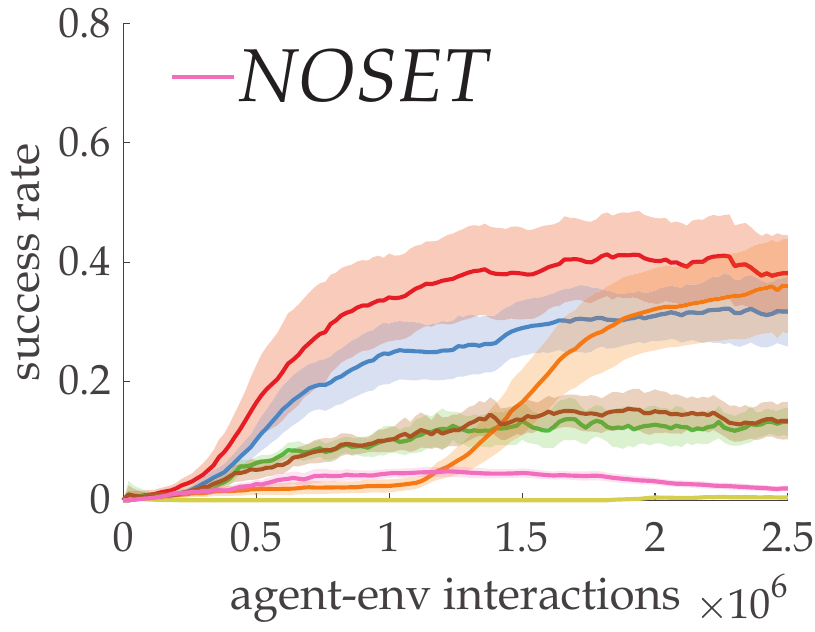}}

\caption[OOD performance of Compared Agents under a Gradient of Difficulties]{\textbf{OOD performance of Compared Agents under a Gradient of Difficulties}: The figures show a consistent pattern: the MPC-based end-to-end agent equipped with a bottleneck (\textbf{CP}) performs the best among all compared methods. All error bars are 95\%-CI obtained from $20$ independent seed runs. All sub-figures share the same $y$-range / scale.}
\label{fig:comparison_OOD}
\end{figure}

The zero-shot evaluations focus on testing agents' performance facing a gradient of OOD difficulties. The results are presented in Fig.~\ref{fig:comparison_OOD}. \textbf{CP(8)}, \textbf{CP} with $k=8$ for top-$k$ semihard attention, shows a clear performance advantage over UP, validating the OOD generalization capability. The \textbf{\Dyna{}*} baseline, essentially the performance upper bound of \textbf{\Dyna{}}-based planning methods, shows no significant performance gain in OOD tests compared to model-free methods. \textbf{WM} may have the potential to reach similar performance as CP, yet it needs to warm up the encoder with a significant portion of the agent-environment interaction budget. We investigate further into the potentials of the WM baseline in Sec.~\ref{sec:potential_WM} on Page.~\pageref{sec:potential_WM}.

\begin{SCfigure}[][htbp]
\includegraphics[width=0.4\textwidth]{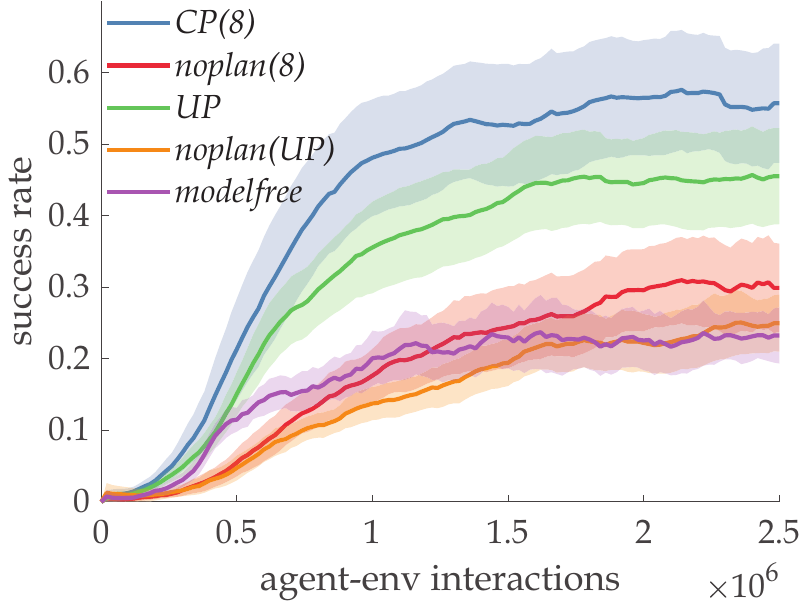}
\caption[Reasoning Addresses Generalization Gap, Bottleneck benefits OOD capability]{\textbf{Reasoning Addresses Generalization Gap, Bottleneck benefits OOD capability}: the $x$-axis shows the training progress ($2.5\times10^{6}$ agent-environment interactions). The $y$-axis values are generated by agent snapshots at times corresponding to the $x$-axis values. noplan(8) and noplan(UP) correspond to the CP(8) and UP variants with planning disabled during OOD tests. Comparing \textit{noplan} against \textit{modelfree}, we see that planning during training is beneficial for both value estimation and representation learning. All error bars are 95\% confidence intervals obtained from $20$ independent seed runs.}
\label{fig:CP_inductive}
\end{SCfigure}

We want to also verify how much decision-time planning in novel situations could address the generalization gap. For this, we compare the planning agents' performance by enabling / disabling planning (relying on the model-free pass only) in OOD evaluation scenarios. The results are shown in Fig.~\ref{fig:CP_inductive}.

\subsubsection{Verification of Selection}
\begin{figure}[htbp]

\subfloat[Turn Left]{
\captionsetup{justification = centering}
\includegraphics[width=0.25\textwidth]{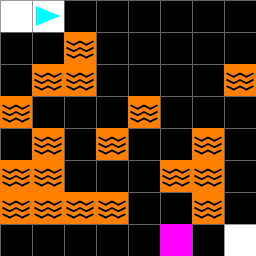}}
\hfill
\subfloat[Turn Right]{
\captionsetup{justification = centering}
\includegraphics[width=0.25\textwidth]{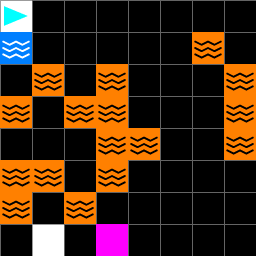}}
\hfill
\subfloat[Step Forward]{
\captionsetup{justification = centering}
\includegraphics[width=0.25\textwidth]{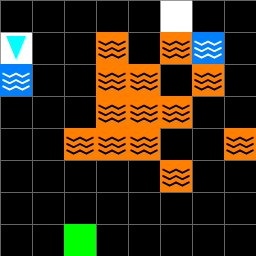}}

\caption[Visualization of Attention Selection]{\textbf{Visualization of Bottleneck Selection given Observation \& Specific Actions}: These figures are extracted from a fully trained CP agent under OOD evaluation. The bottleneck is set to very small for clearer visualization purposes. We invert the color of the selected grid by the selector's output semihard attention weights.}
\label{fig:visualize_att}
\end{figure}

We present some examples to visually verify  the object selection during the planning steps in Fig.~\ref{fig:visualize_att}. Note that though these visualizations provide an intuitive understanding of the agents' behavior, they do not serve statistical purposes. In \textbf{a)} and \textbf{b)}, the agent only turns in its current grid without changing any other grids, thus it exhibits quite random attention towards the other grids. While in c), we can see that the agent takes consideration into the grid (the blue lava grid, color-inverted from orange) that it is facing before taking a step-forward action, since it predicts that such grid will be changed given the intended actions.

Additionally, we collected the coverage ratio of all the relevant objects by the selection phase in all the in-distribution and OOD evaluation cases. The collected data on bottleneck sizes $k=4$, $k=8$ and $k=16$ indicate that the coverage is almost perfect very early on during training. We do not provide these curves because the convergence to $100\%$ is so fast that the curves would all coincide with the line $y=1$, with some minor fluctuations of the confidence intervals.

\subsection{Ablation Studies}
The ablation studies focus on validating the individual design choices by comparing the proposed agents to variants with certain designs altered.

\subsubsection{Bottleneck \& Planning}
\begin{figure}[htbp]
\subfloat[\textbf{Value-guided tree search does not generalize well in OOD evaluation}: random heuristic significantly outperforms best-first heuristic]{
\captionsetup{justification = centering}
\includegraphics[width=0.48\textwidth]{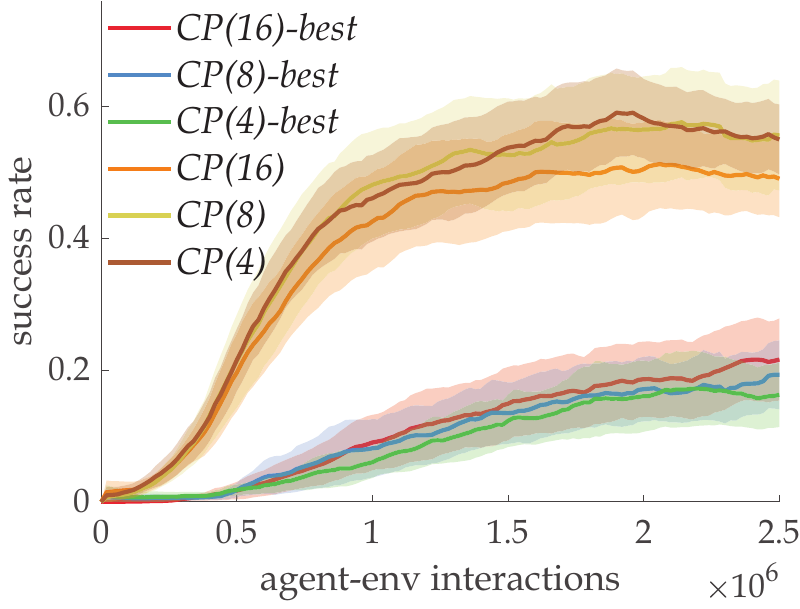}}
\hfill
\subfloat[\textbf{Attention Type}: when used in bottleneck selection, semi-hard attention outperforms soft attention]{
\captionsetup{justification = centering}
\includegraphics[width=0.48\textwidth]{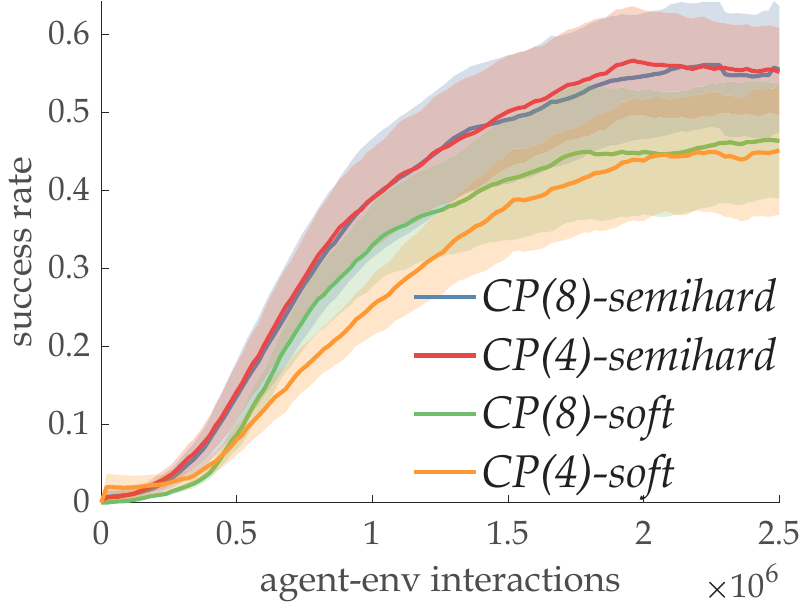}}

\subfloat[\textbf{Action Optimality}: for in-distribution evaluation, the methods both perform well. Interestingly, the model-free agent performs superior possibly due to its simple value-based greedy policy. However, in OOD evaluation, only the CP agent with the random heuristic shows neither significant deterioration nor signs of overfitting]{
\captionsetup{justification = centering}
\includegraphics[width=0.48\textwidth]{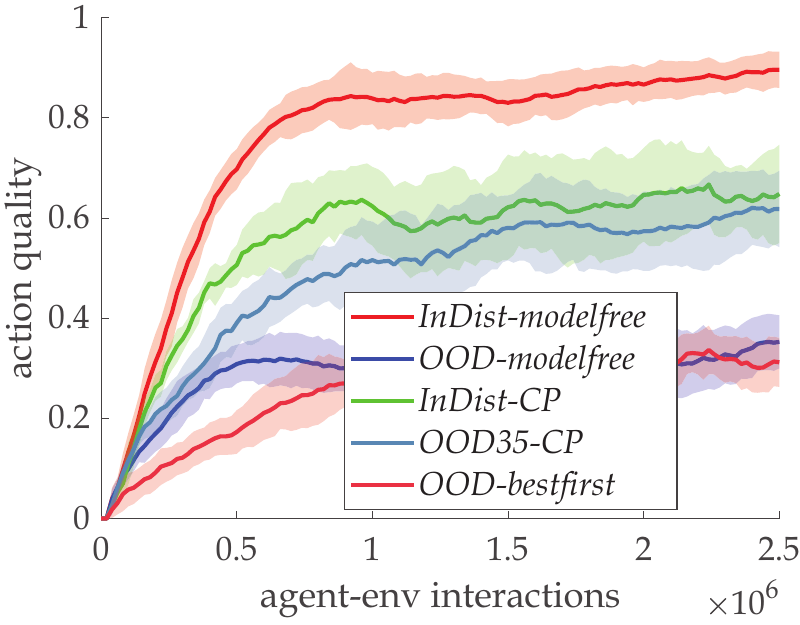}}
\hfill
\subfloat[\textbf{Tree Search Dynamics Accuracy}: we average over the cumulative $L_1$ error in the simulated state of the chosen trajectories during tree search. The curves show no signs of overfitting, as the cumulative trajectorial dynamics errors for OOD evaluation are decreasing continuously.]{
\captionsetup{justification = centering}
\includegraphics[width=0.48\textwidth]{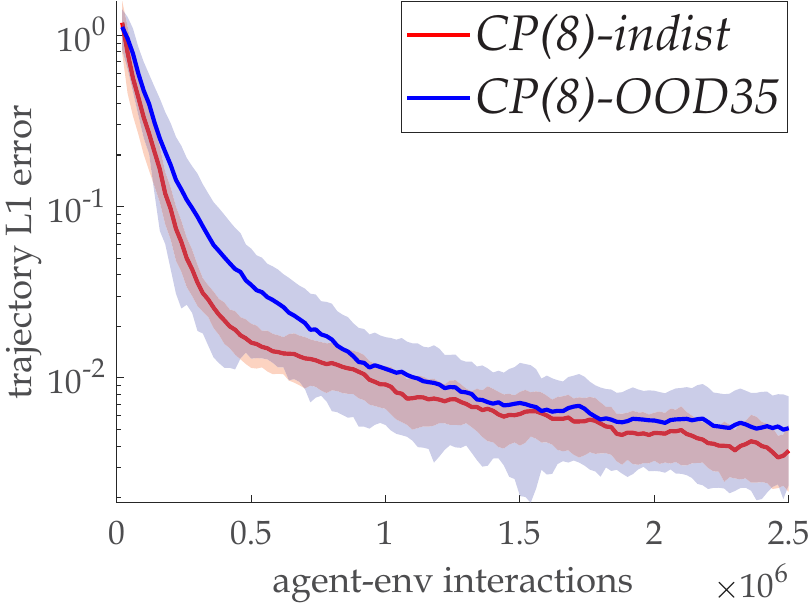}}

\caption[Ablation Results regarding Spatial Abstraction]{\textbf{Ablation Results regarding Spatial Abstraction via the Proposed Bottleneck Mechanism}: With $\delta = 0.35$, each error bar (95\%-CI) is obtained from $20$ independent seed runs.}
\label{fig:OOD35}
\end{figure}

With ablation studies, we validate the effectiveness of our design regarding the bottleneck and planning. We present the key results in Fig.~\ref{fig:OOD35}.


\subsubsection{Model Accuracy}
This set of experiments intends to understand how well the bottleneck sets capture the underlying dynamics of the environments. For each transition, with the help of DP, we partition the grid points into two classes: one containing all relevant objects that have an impact on reward or termination or changed during the transition; While, the other contains the remaining grid points. As a result, the dynamics errors are split into two terms which correspond to the accuracy of the model simulating the relevant and irrelevant objects, respectively.

Acknowledging the differences in the norm of the learned latent representations for each run, due to the stochasticity in the parameter initialization and optimization processes, we use the \textit{normalized} element-wise mean of $L_1$ (absolute value) difference between $\hat{s}_{t+1}$ and $s_{t+1}$. The residual is normalized by the element-wise mean $L_1$ norm of $s_{t+1}$. As a metric of model accuracy, we name such metric \textit{relative $L_1$}. This metric shows the degree of deviation in dynamics learning: the lower it is, the more consistent are the learned and observed dynamics.

\begin{figure}[htbp]

\subfloat[Split of Relative $L_1$ error]{
\captionsetup{justification = centering}
\includegraphics[width=0.48\textwidth]{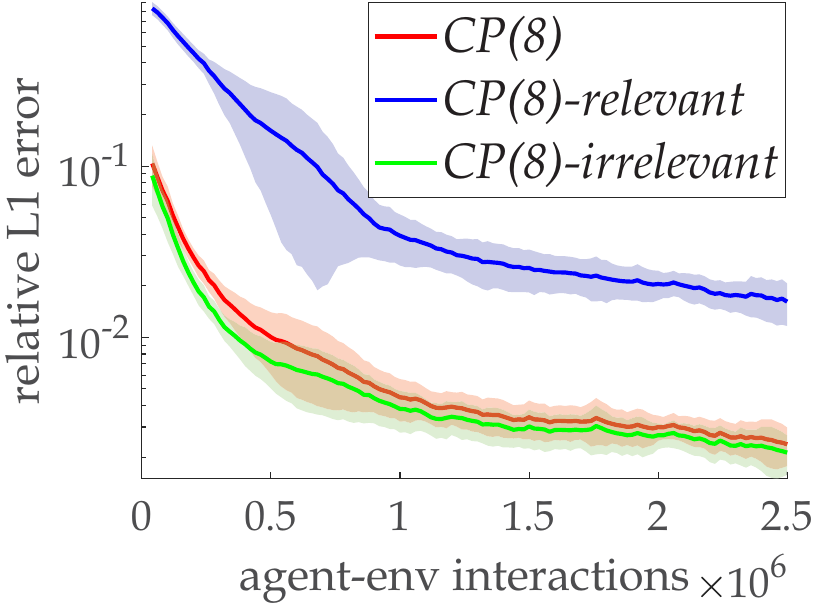}}
\hfill
\subfloat[Comparison of Relative $L_1$ error]{
\captionsetup{justification = centering}
\includegraphics[width=0.48\textwidth]{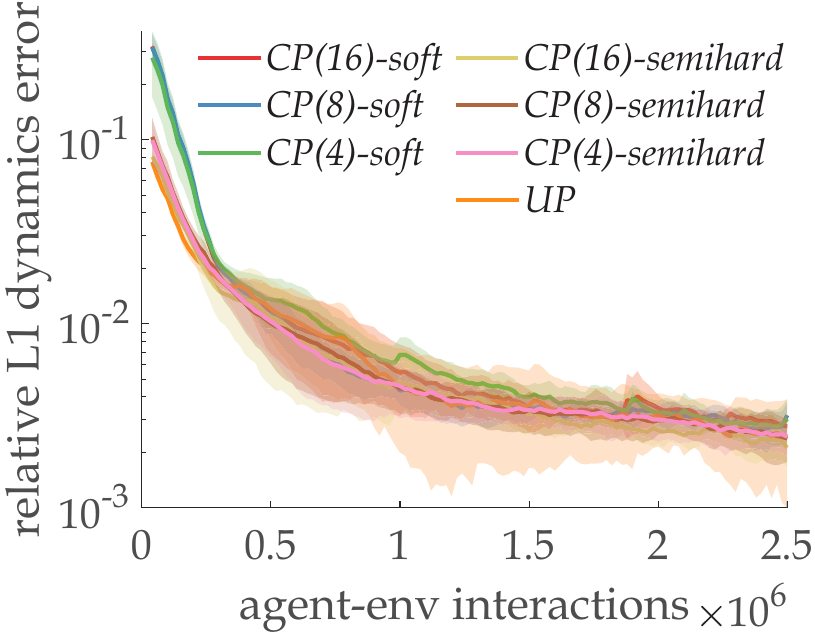}}

\caption[In-Distribution Model Performance]{In-Distribution Model Performance: Each band shows the mean curve (bold) and the 95\% confidence interval (shaded) obtained from $20$ independent seed runs. \textbf{a)}: Partitioning of the relative $L_1$ dynamics prediction errors into that of the relevant objects and the irrelevant ones: The difference in the errors shows that the bottleneck learns to ignore the irrelevance while prioritizing on the relevant parts of the state; \textbf{b)}: Comparison of the overall relative $L_1$ errors (not partitioned). For CP variants, the numbers in the parentheses correspond to the bottleneck sizes and the suffixes the types of attention for the bottleneck selection. Semi-hard attention learns more quickly than soft attention at early stages, but they both converge to similar accuracy levels. This is likely because semi-hard attention is forced to pick few objects and thus to ignore irrelevant objects even at early stages of training.}
\label{fig:in_dist_acc}
\end{figure}

Fig.~\ref{fig:in_dist_acc} \textbf{a)} presents the progression of \textit{relative $L_1$} error of the CP agent during the in-distribution learning process. With the help of the bottleneck, the error for the irrelevant parts converge very quickly while the model focuses on learning the relevant changes in the dynamics. We provide the model accuracy curves of the \textbf{WM} and \Dyna{} baselines in the later parts of the experiments.

For reward and termination estimations, our results show no significant difference in estimation accuracies, with varying bottleneck sizes. However, they do have significant impact on the quality of the learned dynamics. In Fig.~\ref{fig:in_dist_acc} \textbf{b)}, we present the convergence of the relative dynamics accuracy of different CP and UP variants. CP agents generally learn as fast as UPs, which indicates low overhead for learning the selection and integration. We will present a more detailed sensitivity analyses for the bottleneck size later in Sec.~\ref{sec:CP_exp_sensitivity}.

\subsubsection{Action Regularization}
\begin{SCfigure}[][htbp]
\includegraphics[width=0.45\textwidth]{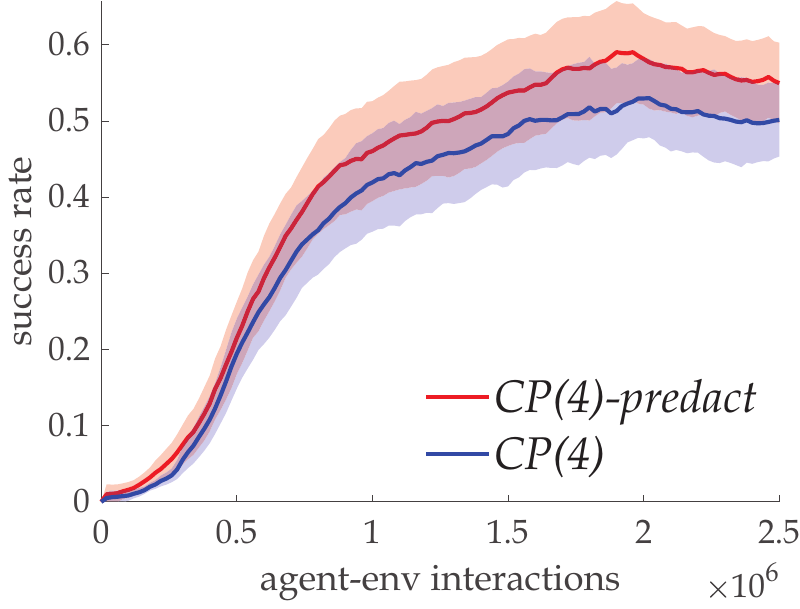}
\caption[Impact of Action Regularization Loss]{Impact of Action Regularization Loss: we present the results on an ablation study on the impact of the action regularization loss on the OOD evaluation success rates of CP(4) agents, with difficulty $\delta=0.35$ on \RDS{}. The ``predact'' configuration is by default enabled in all experiments. Each point of the band correspond to the mean and confidence interval are obtained from $20$ independent seed runs.}
\label{fig:predact}
\end{SCfigure}

For the stability of the CP agent, we applied an additional regulatory loss that predicts the action $a_t$ with $c_t$ and $\hat{c}_{t+1}$ as input, resembling an inverse model \citep{conant1970every}. The loss is implemented with categorical cross-entropy, similar to how we handled the  termination prediction loss. As shown in Fig.~\ref{fig:predact}, This additional training signal is shown in experiments to produce better OOD results, especially when the bottleneck is small.

\subsection{Sensitivity Studies}
\label{sec:CP_exp_sensitivity}

We investigate how sensitive the compared agents are to certain hyperparameters and to the sizes of the tasks.

\subsubsection{Bottleneck Size}

\begin{SCfigure}[][htbp]
\includegraphics[width=0.45\textwidth]{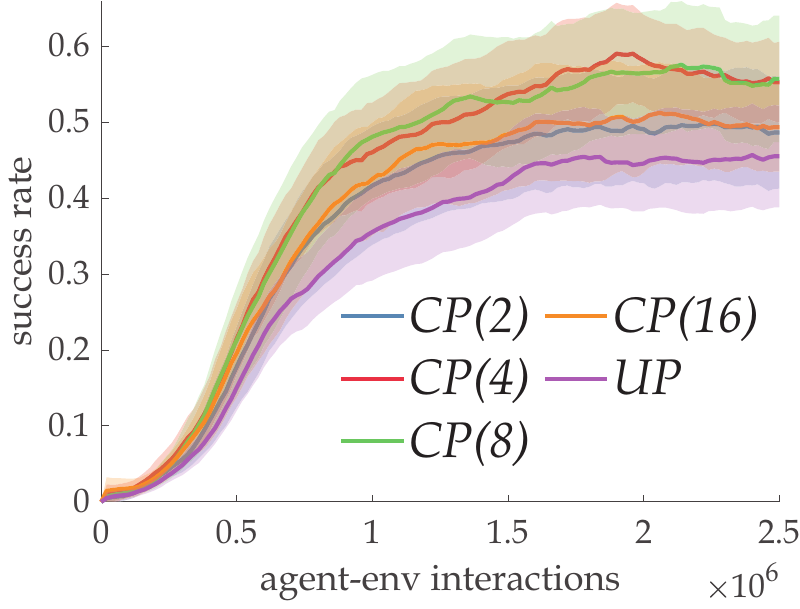}
\caption[Sensitivity to Bottleneck Sizes (controlled by k)]{\textbf{Sensitivity to Bottleneck Sizes (controlled by $k$)}: bottleneck sizes $k=4$ and $k=8$ perform similarly the best within $k \in \{2, 4, 8, 16\}$. Note that the performance of CP agents (agents with bottlenecks) is consistently better than those without (UP), showing the bottlenecks' effectiveness for generalization}
\label{fig:bottleneck_size}
\end{SCfigure}

Without a question, the most important hyperparameter of the CP agents is $k$, which controls the size of the bottleneck set. We present the results of a sensitivity study regarding the hyperparameter $k$ in Fig.~\ref{fig:bottleneck_size}.

\subsubsection{Planning Steps}
We investigate the number of planning steps allowed in each MPC session at each decision point. Intuitively, if the planning steps are too few, then the planning would have little gain over model-free methods. While, if the planning steps are too many, we suffer from cumulative planning errors and potentially prohibitive wall time.

\begin{SCfigure}[][htbp]
\includegraphics[width=0.45\textwidth]{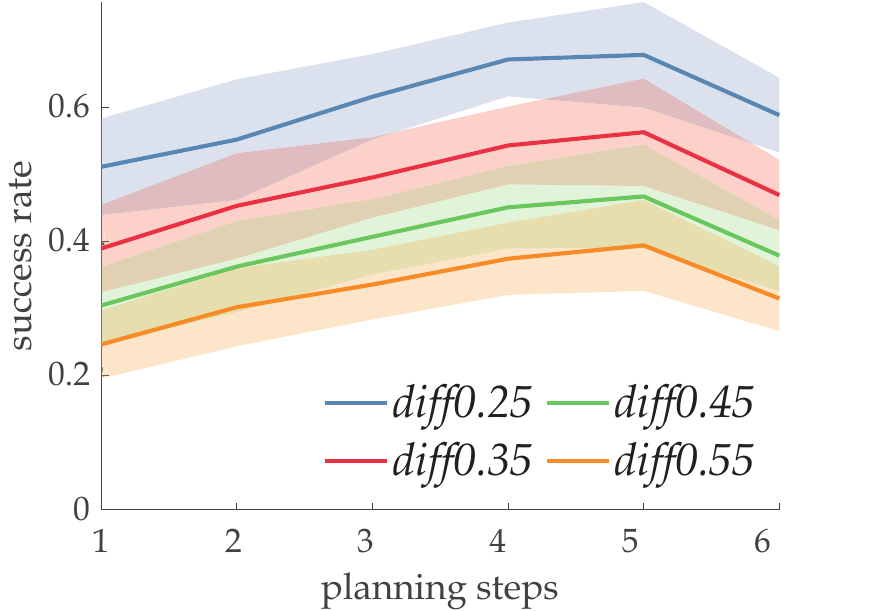}
\caption[Success Rates given Different Numbers of Planning Steps]{\textbf{Success Rates given Different Numbers of Planning Steps}: Success rates of CP(8) agent under OOD difficulty $\delta=0.35$ are presented. each data point is obtained by averaging from the last $20\%$ of $20$ independent seed runs.}
\label{fig:nshaped_steps}
\end{SCfigure}

Thus, we have a strong intuition that an appropriate value for the number of planning steps could potentially achieve a good tradeoff. To search for such good value, we tried different numbers of planning steps for CP(8) variant. Note that the planning steps during training and OOD evaluation are set to be the same, to ensure that the planning during evaluation would be carried out to the same extent during training. The results visualized in Fig.~\ref{fig:nshaped_steps} suggested that $5$ planning steps achieves the best performance in OOD with difficulty 0.35. Base on this, for all other experiments reported in this chapter, $5$ is set to be the default number of planning steps.

\subsubsection{World Sizes}
\label{sec:CP_world_sizes}

\begin{figure}[htbp]
\centering

\subfloat[OOD, $\delta = 0.25$]{
\captionsetup{justification = centering}
\includegraphics[width=0.242\textwidth]{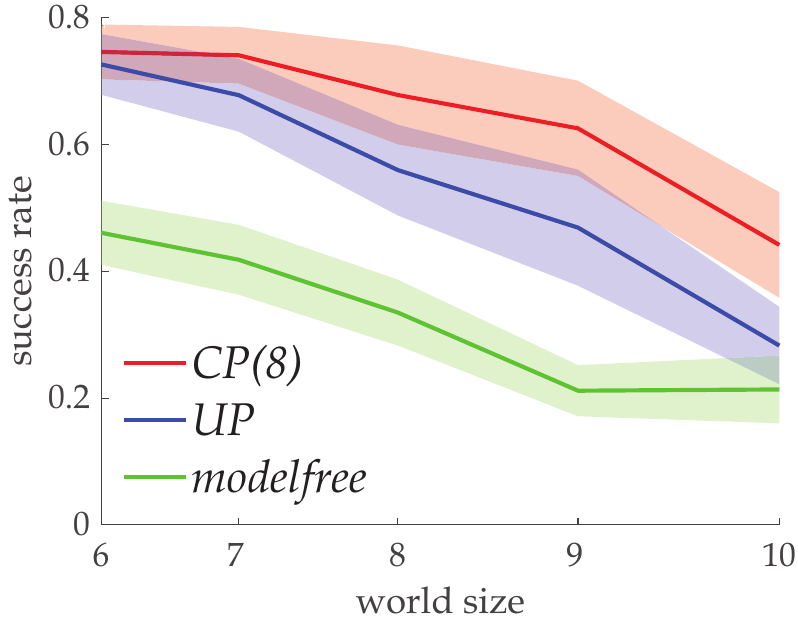}}
\hfill
\subfloat[OOD, $\delta = 0.35$]{
\captionsetup{justification = centering}
\includegraphics[width=0.242\textwidth]{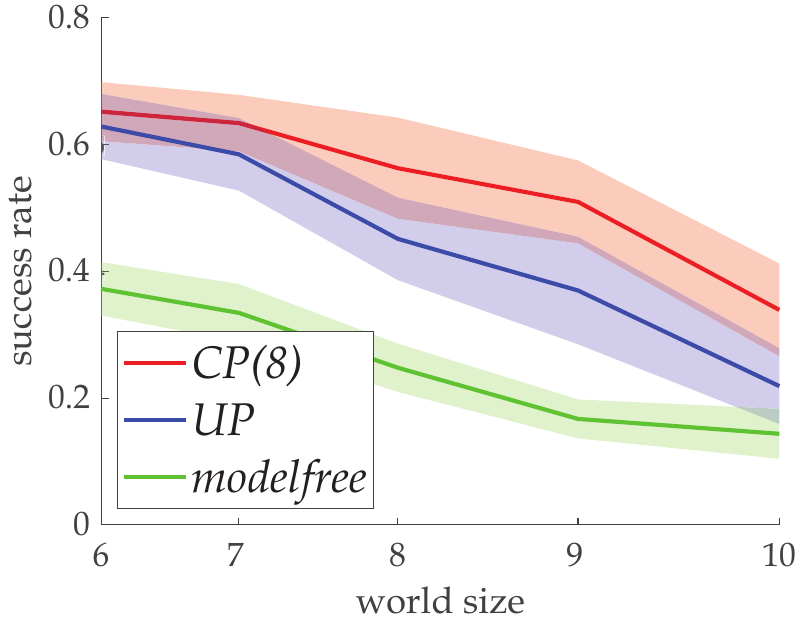}}
\hfill
\subfloat[OOD, $\delta = 0.45$]{
\captionsetup{justification = centering}
\includegraphics[width=0.242\textwidth]{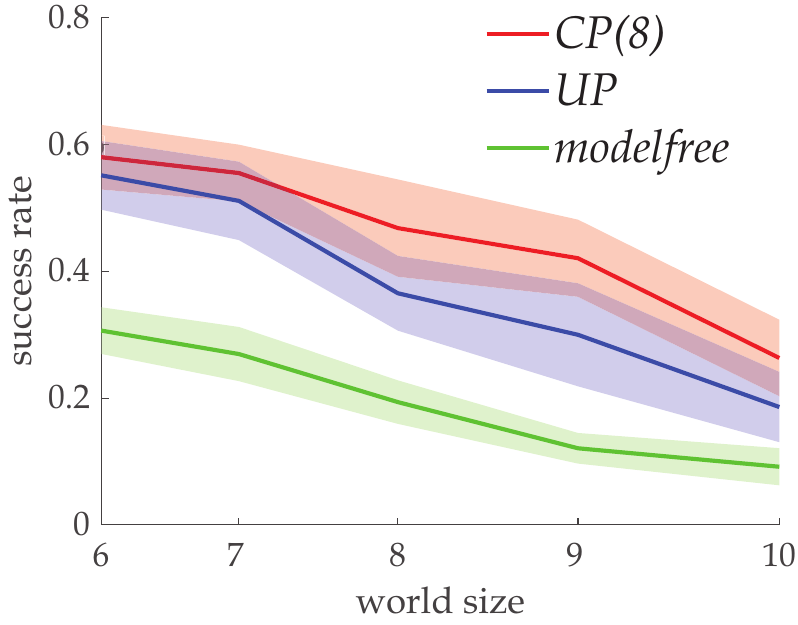}}
\hfill
\subfloat[OOD, $\delta = 0.55$]{
\captionsetup{justification = centering}
\includegraphics[width=0.242\textwidth]{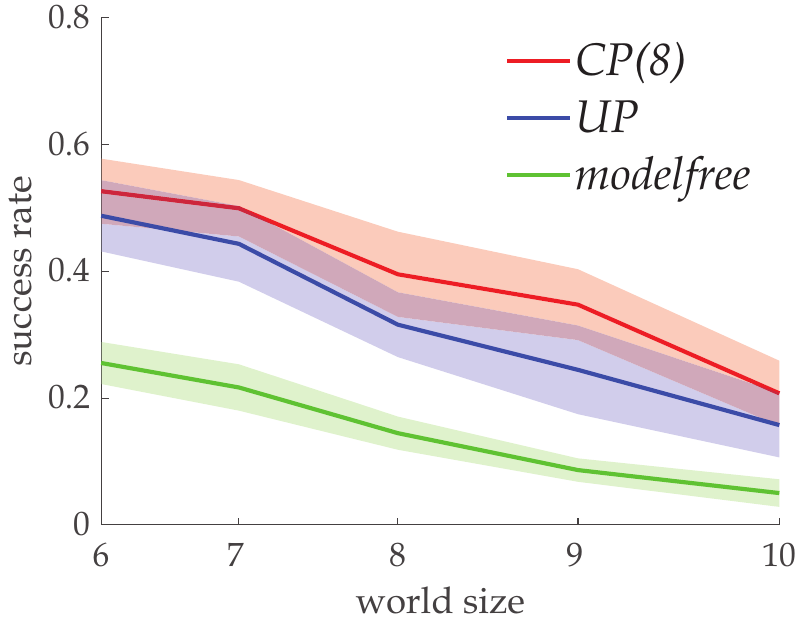}}

\caption[Scalability of OOD Performance under a Spectrum of Difficulties and World Sizes]{\textbf{Scalability of OOD Performance under a Spectrum of Difficulties and World Sizes}: The $x$ axes are ticked with \#grids in each gridworld size, representing the number of entities for in the state set, thus \textit{non-linear} \wrt{} the world sizes. We can observe that, generally, the smaller the world sizes, the better and the closer the performance of all 3 methods are. The fact that the CP(8) performance deteriorates slower than UP suggests that the bottleneck may contribute to more scalable performance in tasks with larger amount of entities. All error bars are obtained from $20$ independent seed runs.}
\label{fig:comparison_worldsizes}
\end{figure}

To inspect the scalability of the proposed method (in terms of the number of objects in the full state sets), we compare the methods CP(8), UP and model-free in a range of gridworld sizes (from $6\times6$ all the way to $10\times10$). The results are presented in Fig.~\ref{fig:comparison_worldsizes}.

\subsection{More Experiments, Discussions \& Failed Attempts}
\label{sec:CP_more_exp}

We go beyond the main experiments (``turn-or-forward'' \RDS{} instances) to gain some additional insights.

\subsubsection{Alternative Dynamics}
We want to know if the conclusions of previous experiments  would still hold, if the agents are given tasks with different dynamics. For this purpose, we modify the original tasks in the previous experiments by a new set of ``Turn-And-Forward'' dynamics: the action space is re-designed to include $4$ composite actions, each of which first turns to some directions (forward, left, right or back based on the agent's current facing direction in the environment) and then move forward if possible (remain in the boundaries of the grid world).

\begin{figure}[htbp]
\centering

\subfloat[OOD, $\delta = 0.25$]{
\captionsetup{justification = centering}
\includegraphics[width=0.242\textwidth]{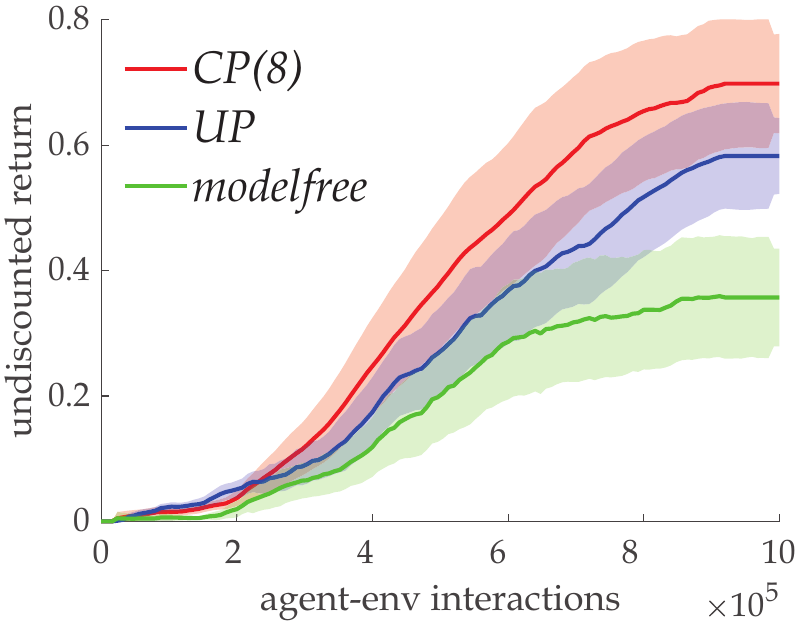}}
\hfill
\subfloat[OOD, $\delta = 0.35$]{
\captionsetup{justification = centering}
\includegraphics[width=0.242\textwidth]{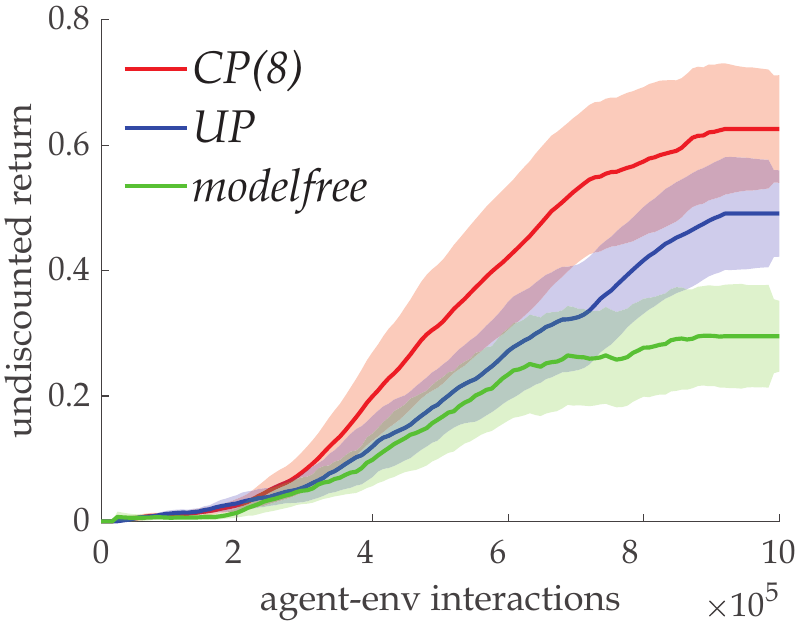}}
\hfill
\subfloat[OOD, $\delta = 0.45$]{
\captionsetup{justification = centering}
\includegraphics[width=0.242\textwidth]{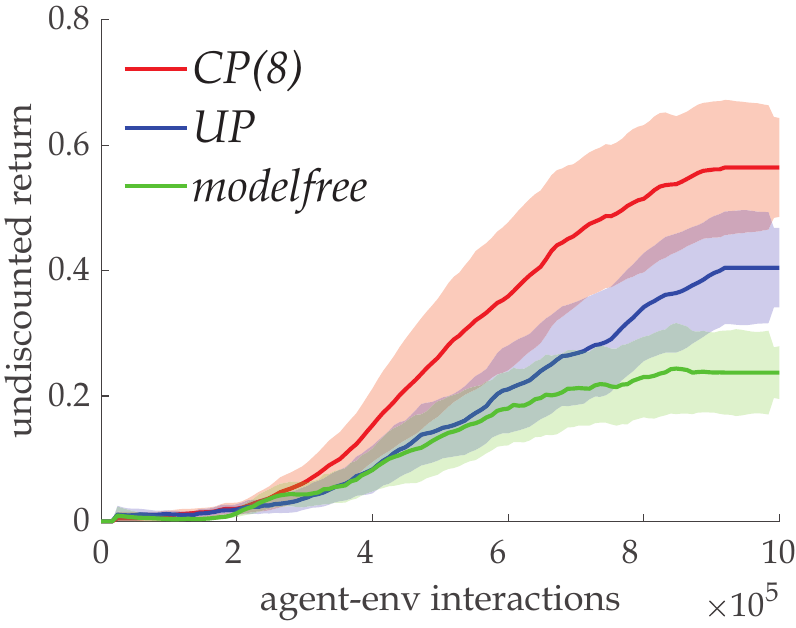}}
\hfill
\subfloat[OOD, $\delta = 0.55$]{
\captionsetup{justification = centering}
\includegraphics[width=0.242\textwidth]{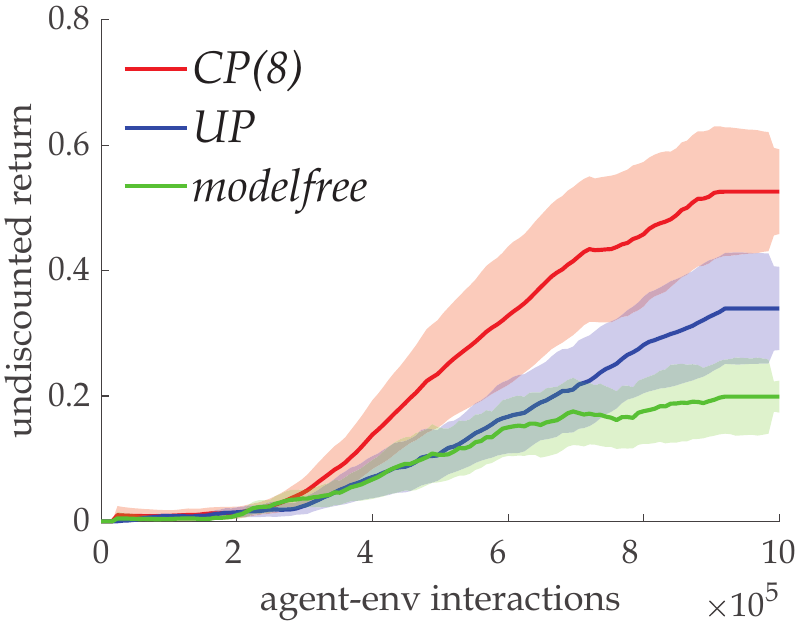}}

\caption[OOD Performance of Compared Agents  in ``Turn-and-Forward'' Tasks]{\textbf{OOD Performance of Compared Agents in ``Turn-and-Forward'' Tasks}: All error bars are obtained from $20$ independent seed runs.}
\label{fig:comparison_v3}
\end{figure}

This new set of environment dynamics (``turn-AND-forward'') can be seen as a composition of the original dynamics (``turn-OR-forward'') and hence produces shorter planning trajectories. In Fig.~\ref{fig:comparison_v3}, we observe that all three methods are performing better compared to the original tasks and the previous conclusions about experiments are again validated.

\subsubsection{Shaping Representations with Many Signals}
We have gathered more empirical evidence regarding the non-conflicting training of the state-representation based on the signals. This means, it is possible to train a state representation that can be used to predict several interesting quantities to the RL agent at the same time. In terms of the accuracy of model learning, according to our results of the \textbf{WM} baseline, removing the value estimation training signal would result in a poorer representation, but not in a lower accuracy when predicting other relevant signals; Similarly, removing termination signal would not impact the convergence of reward prediction accuracy or that of the dynamics prediction, despite that the RL performance would be impacted. Also, removing the reward signal or the next state dynamics prediction signal leads to collapse of the MPC based behavior policy. Yet, the convergence of the other remaining training signals is not significantly affected. With these observations, we would suggest that we have, at least in this task setting, learned a set-based representation capable of predicting all interesting quantities.

\subsubsection{Failed Effort: Straight-Through Hard Subset Selection with Gumbel}
We initially tried to use Gumbel subset selection \citep{xie2019differentiable} to implement a hard selection based bottleneck, but we later realized that such approach will not work. We expected the model to pick the right objects by generating a binary mask, and then use the masked objects as the bottleneck set. On the surface, this two-staged design would align more with the consciousness theories, and would yield clearer interpretability.

However, we came to realize that such method overlooked an implicit chicken-and-egg problem. This problem can be intuitively described as follows: to learn how to pick, the model should first understand the dynamics. Yet if the model does not pick the right objects frequently enough, the dynamics would never be understood. Mathematically, this problem is more recently described as the ``degeneracy'' of discrete bottlenecks. Our proposed semi-hard / soft approaches address such problem by essentially making the two staged selection and simulation as a whole for the gradient-based optimization.

\subsection{Details of Baselines}

\subsubsection{WM}
\label{sec:potential_WM}

Our \textbf{WM} baselines, \ie{} agents with World Model (WM) trained in stages, share the same architectures (and hyperparameters) as their CP or UP counterparts \citep{zhou2024dino}. The only difference is that \textbf{WM} adopts a $2$-staged training strategy: In the first $10^{6}$ agent-environment interactions, only the model is trained (together with the state representation encoder), and thus, the state representations are only shaped by the model's training losses (without the participation of RL signals). In the first stage, the agent relies on a uniformly random policy. After $10^{6}$ interactions, the agent freezes its state representation encoder as well as the model to carry out value estimator learning separately. Compared to CP or UP, the exploration scheme is delayed but unchanged.

\begin{figure}[htbp]
\centering

\subfloat[OOD, $\delta = 0.25$]{
\captionsetup{justification = centering}
\includegraphics[width=0.242\textwidth]{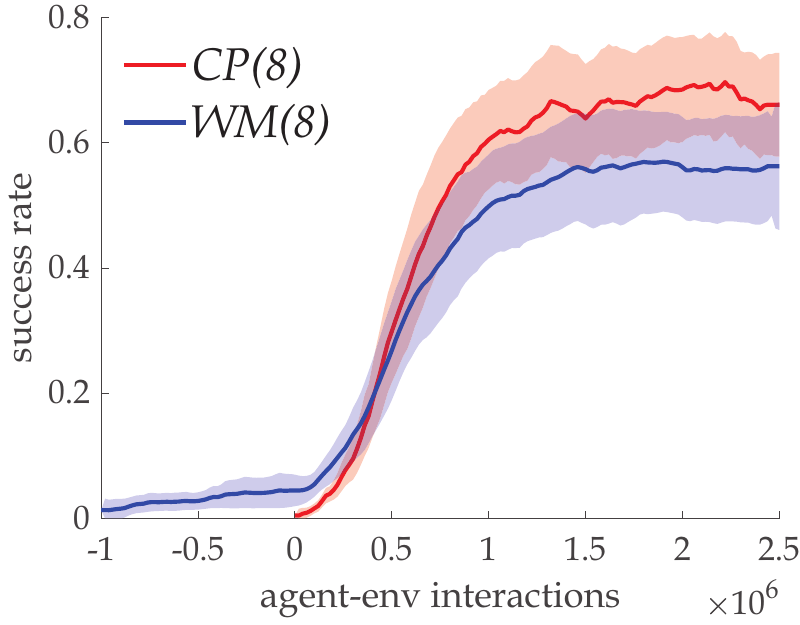}}
\hfill
\subfloat[OOD, $\delta = 0.35$]{
\captionsetup{justification = centering}
\includegraphics[width=0.242\textwidth]{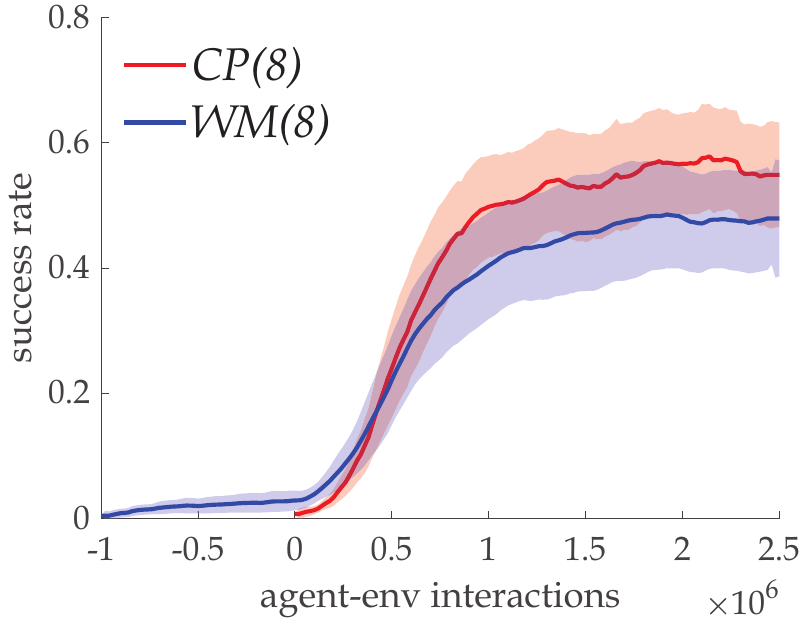}}
\hfill
\subfloat[OOD, $\delta = 0.45$]{
\captionsetup{justification = centering}
\includegraphics[width=0.242\textwidth]{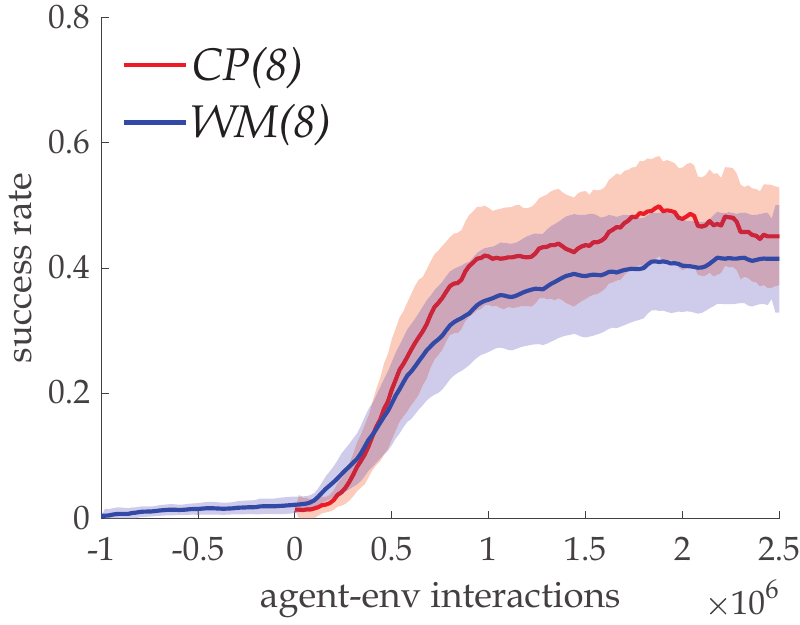}}
\hfill
\subfloat[OOD, $\delta = 0.55$]{
\captionsetup{justification = centering}
\includegraphics[width=0.242\textwidth]{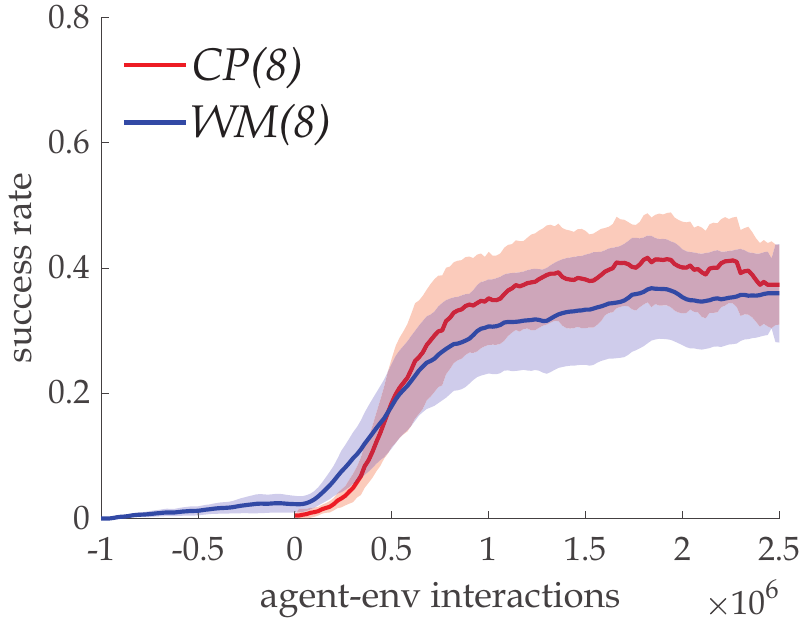}}

\caption[Extended Run to Demonstrate the OOD Performance Differences of CP and WM agents]{\textbf{Extended Run to Demonstrate the OOD Performance Differences of CP and WM agents}: for fair comparison, WM(8) is the WM variant which uses the same architecture as CP(8). Results of WM(8) are shifted along the $x$-axes for a free unsupervised world model learning phase of $10^6$ steps. All error bars are obtained from $20$ independent seed runs.}
\label{fig:free_unsupervised}
\end{figure}

We are also curious about how the WM baseline would evolve after the $2.5\times 10^{6}$-step cutoff. For this, we provide an additional set of experiments featuring a ``free'' unsupervised learning phase of $10^{6}$ agent-environment interactions, essentially prolonging the runs of \textbf{WM} baselines by extra $10^{6}$ steps. As presented in Fig.~\ref{fig:free_unsupervised}, results suggest that \textbf{WM} could not achieve similar performance as that of CP, likely because the state representation is not jointly shaped for value estimation and thus cannot perform well enough for such purpose. The results show promise of the methodology of representation learning with joint signals. However, this is not to deny the usability and applicability of the world model methodology in general.

\subsubsection{\texorpdfstring{\Dyna{}}{Dyna}}
As usual, for fair comparison, \Dyna{} baselines share the model-free part of the architecture as CP or UP. The models used by the \Dyna{} baselines are powered by the proposed set-to-set architecture, working on the observation-level. For each sampled batch of transitions, we feed the sampled current observations and actions to the \Dyna{} model to generate imagined next observations and train additionally on this second batch.

\subsubsection{\texorpdfstring{\NOSET{}}{NOSET}}
The \NOSET{} baseline utilizes traditional vectorized representations. We use the CNN feature extractor before the encoder, but instead of transforming the feature map into a set, we flatten it and then linearly project it to some specific dimensionality ($256$), which will be used as the vectorized state representation, similar to most existing DRL practices. Since all set-based operations would now be improper for the vectorized representation, they are substituted with $3$-layered, $512$-wide MLPs. The dimension of the vectorized state representation, the widths, and the depths of the FC layers are optimized through coarse grid tuning. We find that architectures exceeding the chosen size are hardly superior in terms of performance.

\begin{SCfigure}[][htbp]
\includegraphics[width=0.45\textwidth]{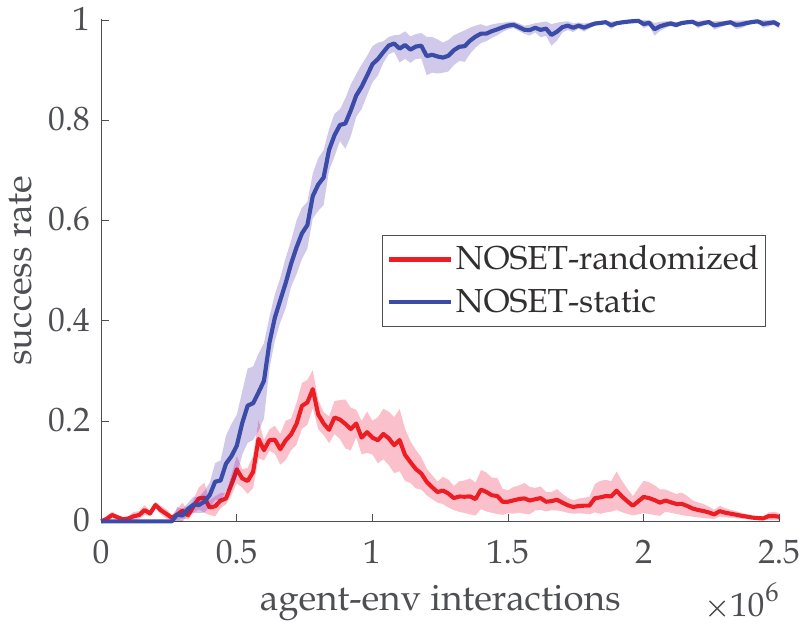}
\caption[NOSET Baseline Performance on Multitask and Monotask Settings]{\textbf{\NOSET{} Baseline Performance on Multitask and Monotask Settings}: Each band is consisted of the mean curve and the confidence interval shades obtained from $20$ independent seed runs.}
\label{fig:noset_static}
\end{SCfigure}

To maximize the performance of this baseline, we designed a  $2$-layered dynamics model, which employ a residual connection, with the expectation that the model might learn incremental changes in the dynamics. We first verified the competence of this baseline on a classical mono-task setting, as shown in Fig.~\ref{fig:noset_static}. However, to a degree as expected, in our experiments with randomly generated environments for each episode (the ``multi-task setting''), the \NOSET{} baseline performs miserably.

\subsection{Summary of Experiments}
The experimental results allow us to draw the following conclusions, within the scope of our experimental setting:

\begin{itemize}[leftmargin=*]
\item Set-based representations enable at least in-distribution generalization across different environment instances in our OOD-oriented multi-task setting, where the agents are forced to discover dynamics that are preserved across environments;
\item Model-free methods seem to rely more on memorization instead of understanding, as they face more difficulties in OOD evaluation scenarios, which emphasizes systematic generalization;
\item Decision-time MPC exhibits better performance than \Dyna{} in the tested OOD generalization settings;
\item Online joint training of the representation with all the relevant signals could bring benefits to RL, as suggested in \citet{jaderberg2016unreal}.
\item In accordance with our intuition, transition models with bottlenecks tend to learn dynamics better in our tests;
\item
From further experiments, we observe that bottleneck-equipped agents may also be less affected by larger size environments with more encoded objects, possibly due to their prioritized learning of interesting entities.
\end{itemize}

\section{Summary}

In this chapter, we introduced a consciousness-inspired bottleneck mechanism into model-based RL, facilitated by set-based representations, end-to-end learning and tree search MPC. In the multi-task systematic generalization-focused settings, the bottleneck allows selecting the relevant objects for planning and hence enables significant OOD performance.

The proposed bottleneck mechanism will be extended for spatial abstraction, a foundation of Chap.~\ref{cha:skipper}.

%% file: chapter_5_Skipper_main.tex
\chapter{Skipper Framework: Spatio-Temporal Abstractions for Planning}
\label{cha:skipper}

\minitoc

\section{Overview of This Thesis Milestone}

\textit{Inspired by the humans' abstract planning behaviors, my collaborators and I propose \Skipper{}, an RL / planning framework using both spatial and temporal abstractions organically to generalize better in novel situations. \Skipper{} automatically decomposes the given task into smaller, more manageable sub-tasks, and thus enables sparse decision-making and focused computation on the relevant parts of the environment. \Skipper{}'s task decomposition relies on the extraction of a ``proxy problem'', represented as a directed graph, in which vertices and edges are learned end-to-end from hindsight. Our theoretical analyses provide performance guarantees under appropriate assumptions and establish where our approach is expected to be helpful. Our carefully controlled experiments validate \Skipper{}'s significant advantage in zero-shot generalization, compared to some existing state-of-the-art hierarchical planning methods.}

Human abstract planning can attend to relevant aspects in both time and space. This kind of planning could competently break down long-horizon tasks into smaller scale and more manageable steps, each of which can be narrowed down even more. From the previously introduced perspective of consciousness in the first sense (C1) \citep{dehane2017consciousness}, this type of planning focuses attention on mostly the important decision points \citep{sutton1999between} and relevant environmental aspects \citep{tang2020neuroevolution}, thus \textit{operating abstractly both in time and in space}.

In contrast, existing RL agents either operate solely based on intuition (``system-1'', model-free methods) or are limited to reasoning over mostly relatively shortsighted plans (rudimentary ``system-2'', model-based methods, including the CP agent in the previous chapter) \citep{daniel2017thinking}. These intrinsic design limitations hinder the agents' real-world application, which demands robustness against generalization challenges \citep{mendonca2020meta}.

Building on intuitions gathered from our previous work on conscious planning (Chap. \ref{cha:CP}), we develop a planning framework, named \Skipper{}, that plans while skipping over unnecessary details in both space and time, seeking to decompose the complex task at hand into smaller sub-tasks, by constructing ``proxy problems''. A proxy problem can be viewed as a simplified version of the given task, and is represented as a graph where 1) the vertices consist of states imagined by a generative model, corresponding to sparse decision points; and 2) the edges, which define temporally-extended transitions, are constructed by focusing on a small amount of relevant information from the states, using an attention mechanism heavily investigated in the previous chapter. Once a proxy problem is constructed and the agent solves it to form a plan, each edge will define a new sub-problem, on which the agent will focus solving next. While granting agents the flexibility to construct necessary abstractions for the problem at hand, the divide-and-conquer strategy based on proxy problems allows constructing ``partial'' solutions that generalize better to novel situations. Our theoretical analyses tie the quality of the solution to the proxy problem to the solution of the overall problem, making the \Skipper{} framework distinguish itself as a principled approach.

For experiments, we use an implementation of the \Skipper{} framework to demonstrate its advantages in terms of out-of-training-distribution generalization. These experiments are conducted on a more refined multi-task, systematic generalization focused setting, where the agents are only trained on limited few task instances but are expected to be evaluated in a range of OOD scenarios in a zero-shot fashion. The detailed and controlled experiments that the proposed framework, in most cases performs significantly better in terms of zero-shot generalization, compared to the baselines and to some state-of-the-art hierarchical planning methods \citep{nasiriany2019planning,hafner2022deep}.

\section{Methodology: Proxy Problems}
\label{sec:skipper_proxy_problem}

\subsection{Definition}
Central to the temporal and spatial abstraction abilities of the proposed \Skipper{} framework, proxy problems are utilized to handle task decomposition. Representing simplified versions of the more complicated given tasks, proxy problems are proxies for solving the more complicated given tasks. Each solution of the proxy problem is equivalent to a plan that can be carried out in the original problem. While, each step of the plan is a sub-problem in the original problem.

\begin{figure}[htbp]
\centering
\subfloat[\RDS{} Instance]{
\captionsetup{justification = centering}
\includegraphics[height=0.4\textwidth]{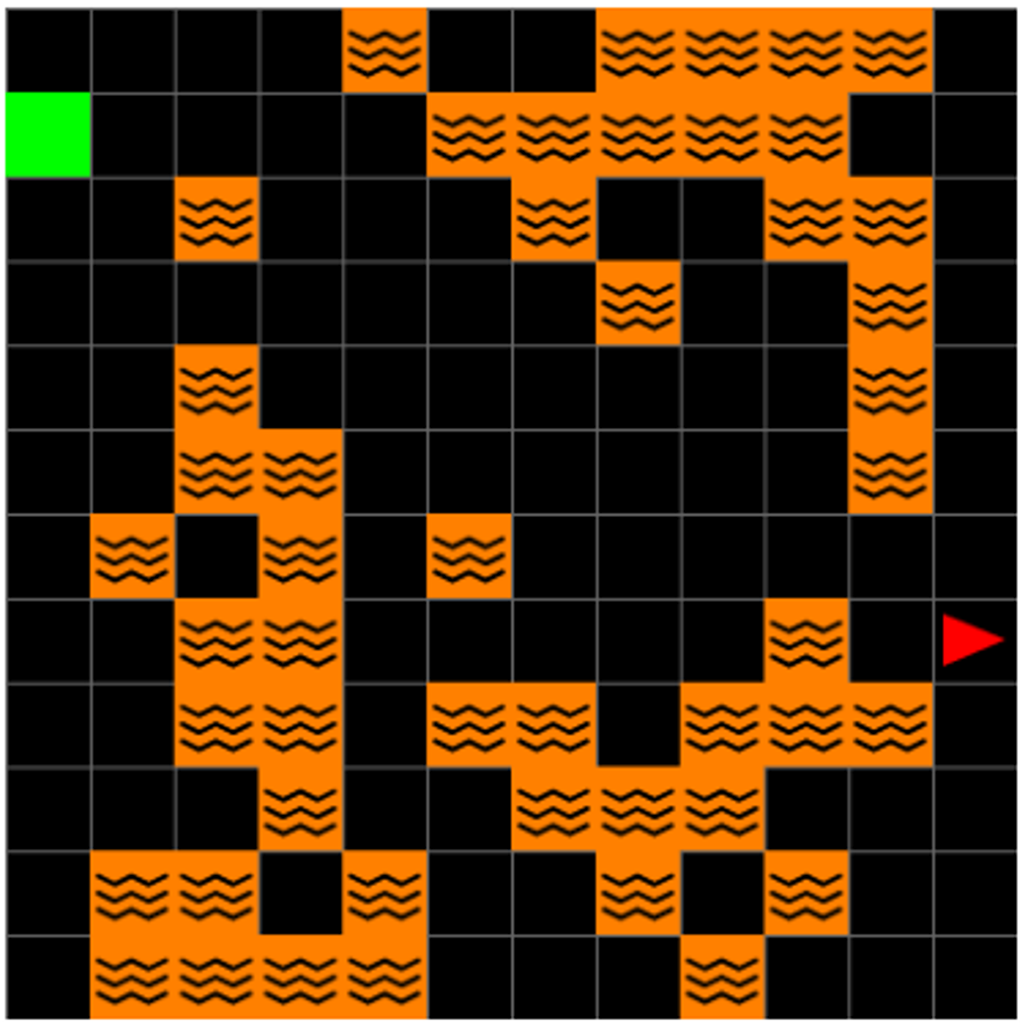}}
\subfloat[Proxy Problem]{
\captionsetup{justification = centering}
\includegraphics[height=0.4\textwidth]{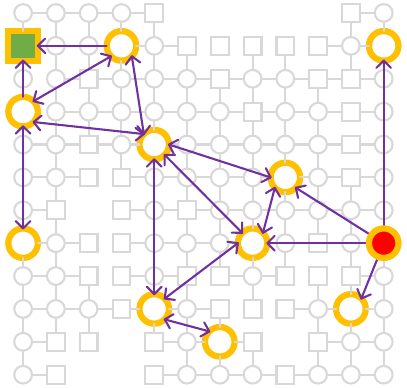}}

\caption[A Proxy Problem on an \RDS{} Task]{\textbf{A Proxy Problem Proposed by \Skipper{} on an \RDS{} Task}: the MDP of the \gray{original problem} is in colored gray and the terminal states (the lava grids and the goal) are marked with squares. The overall objective of the task is to navigate from the \red{red} position, to the goal (\darkgreen{green}). As shown, distant goals can be reached by leveraging a proxy problem, which can lead to a plan that involves hopping through a series of checkpoints (\orange{orange}).}
\label{fig:proxy_problem}
\end{figure}

Proxy problems are sparse graphs constructed at decision-time, whose vertices are states and whose (directed) edges estimate transition statistics between the vertices, as shown in Fig.~\ref{fig:proxy_problem}. For clarity, we call the states selected to be (or proposed by the generator to be) vertices of the proxy problems \textbf{checkpoints}, to differentiate them from other uninvolved states. Note that the current state should always be included as one of the vertices. At decision-time, during evaluation, the checkpoints are proposed by a generative model, a \textbf{generator}, and represent some states that the agent might experience in the current episode, often denoted as $S^\odot$ in this chapter. Each edge is annotated with estimates of the cumulative discount and reward associated with the transition between the connected checkpoints; these estimates are learned over the \textbf{relevant} aspects of the environment (``spatially-abstract'') and \textbf{depend} on the agent's capability (``capability-aware''). As the low-level policy responsible for solving the sub-problems (implementing the transitions from one checkpoint to the next) improves, the edges strengthen.

Planning in a proxy problem is by design temporally abstract, since the sub-problems are defined as the transitions between pairs of checkpoints, which are sparse decision points. Handling each checkpoint transition is also by design spatially abstract, as an option corresponding to such a task would base its decisions only on some aspects of the environment relevant to the sub-problem at hand \citep{konidaris2009efficient,bengio2017consciousness}, leading to improved generalization as well as computational efficiency \citep{zhao2021consciousness}. In the language of probabilistic modelling, the proxy problems alleviate the difficulties of long horizon planning by \textit{marginalizing the change of more irrelevant aspects} of the space and time within trajectories, and \textit{constraining the options} \st{}, the changes that matter to the objective are simpler to predict \citep{bengio2017consciousness}.

Proxy problems can be formally described in the language of SMDPs (Sec.~\ref{sec:SMDP}, Page.~\pageref{sec:SMDP}), where each directed edge is implemented as a checkpoint-conditioned option. Thus, edges in a proxy problem can be defined by the discount and reward matrices \blue{$\Gamma^\pi$} and \darkgreen{$V^\pi$}, where \blue{$\gamma^\pi_{ij}$} and \darkgreen{$v^\pi_{ij}$} are defined as:
\begin{align}
    \blue{\gamma^\pi_{ij}} &\coloneqq \mathbb{E}_\pi\left[\gamma^{T_\perp} | S_0=s_i, S_{T_\perp}=s_j\right] \\
    \darkgreen{v^\pi_{ij}} &\coloneqq \textstyle\mathbb{E}_\pi\left[\sum_{t=0}^{T_\perp} \gamma^t R_t | S_0=s_i, S_{T_\perp}=s_j\right]
\end{align}

\subsection{Potential of Proxy Problems}
\label{sec:perf_guarantee}

By planning with \blue{$\Gamma^\pi$} and \darkgreen{$V^\pi$}, \eg{}, using SMDP value iteration \citep{sutton1999between}, we can form a plan that hops through the checkpoints in the proxy problem, while simultaneously traveling among states in the original problem.

But why should we use this methodology? What could be the advantage? The following results establish that if the proxy problems can be estimated well (Condition \ref{eq:estimation_condition}), the obtained plan via the proxy problem will be a good quality solution for the original problem (Bound \ref{eq:composite_bound}):

\begin{coloredtheorem}{Performance under Proxy Problems}{proxyprobsperf}
Let $\mu$ be the SMDP policy (high-level planner that solves the proxy problems and produce a plan) and \red{$\pi$} be the low-level policy (that implements the sub-problems of transitioning from one checkpoint to another). Let \darkgreen{$\hat{V}^\pi$} and \blue{$\hat{\Gamma}^\pi$} denote learned estimates defining the edges of the proxy problem. If their estimation accuracy satisfies:
\begin{align}
\label{eq:estimation_condition}
    & |\darkgreen{v^\pi_{ij}}-\darkgreen{\hat{v}^\pi_{ij}}|<\epsilon_v  v_{\text{range}} \ll (1-\gamma)  v_{\text{range}} & \text{\textbf{and}}\\
    & \quad\quad |\blue{\gamma^\pi_{ij}} -\blue{\hat{\gamma}^\pi_{ij}}|<\epsilon_\gamma\ll (1-\gamma)^2 & \forall i,j. \nonumber
\end{align}
Then, the estimated value of the composite $\hat{v}_{\mu \circ \pi}(s)$ is accurate up to error terms linear in $\epsilon_v$ and $\epsilon_\gamma$:
\begin{align}
\label{eq:composite_bound}
    \hat{v}_{\mu \circ \pi}(s) &\coloneqq \sum_{k=0}^\infty \hat{v}_\pi(s_k^\odot |{s}_{k+1}^\odot ) \prod_{\ell=0}^{k-1}\hat{\gamma}_\pi (s_\ell^\odot | s_{\ell+1}^\odot ) = v_{\mu \circ \pi }(s) \pm \frac{\epsilon_v  v_{\text{range}}}{1-\gamma} \pm \frac{\epsilon_\gamma v_{\text{range}}}{(1-\gamma)^2}  + o(\epsilon_v+\epsilon_\gamma)
\end{align}
where \darkgreen{$\hat{v}_\pi(s_i | s_j )\equiv \hat{v}^\pi_{ij}$} and \blue{$\hat{\gamma}_\pi(s_i | s_j ) \equiv \hat{\gamma}^\pi_{ij}$}, and $v_{\text{range}} \coloneqq v_{\text{max}} - v_{\text{min}}$ denotes the range of values, \ie{}, the maximum possible value minus the minimum possible value.
\end{coloredtheorem}

\begin{proof}


Before anything, we need to acknowledge that we are in the presence of a hierarchical structure involving two policies. The high level policy $\mu$ determines (potentially) stochastically a sequence of checkpoints to form a plan. The low-level policy \red{$\pi$} is assumed to be evolving throughout training, and fixed for the moment. The composite policy $\mu\circ\pi$ is non-Markovian: it depends both on the current state and the current target checkpoint. Thus, there is no notion of a classical ``state value'', except when the agent arrives at a checkpoint, \ie{}, when a high level action (checkpoint selection) needs to be chosen.

To simplify the proof that proceeds further, we adopt the view that discounts can be implemented equivalently with sudden termination probabilities. In this equivalent view, $v_\pi$ denotes the undiscounted expected sum of reward before reaching the next checkpoint, and more interestingly $\gamma_\pi$ denotes the binomial random variable of non-termination during a sub-problem, \ie{}, the transition to the selected checkpoint.


With the above in mind, we can write the gain (in undiscounted cumulated rewards) $V$ from the initial checkpoint $S_0^{\odot}$ as:

\begin{align}
V = V(S_0^{\odot}|S_1^{\odot})+\Gamma(S_0^{\odot}|S_1^{\odot}) V_1 =  \sum_{k=0}^\infty V(S_k^{\odot}|S_{k+1}^{\odot}) \prod_{i=0}^{k-1}\Gamma(S_i^{\odot}|S_{i+1}^{\odot}),
\end{align}

where $S_{k+1} \sim \mu(\cdot|S_k)$, $V(S_k^{\odot}|S_{k+1}^{\odot})$ is the random variable of gain during the transition from $S_k^{\odot}$ to $S_{k+1}^{\odot}$, and the random variable $\Gamma(S_k^{\odot}|S_{k+1}^{\odot})$ can take a value of either $0$ or $1$, depending on whether the trajectory terminated prematurely or reached $S_{k+1}^{\odot}$.


It should be clear that the action space of $\mu$ is the a list of checkpoints $\{s^{\odot}_0=s_0, s^{\odot}_1, s^{\odot}_2, \cdots \}$ in the same proxy problem. 
Thus, for the expected (ground truth) value of $\mu\circ\pi$, taking the expectation of $V_0$, we have:
\begin{align}
    v_{\mu\circ\pi}(s_0) \coloneqq \mathbb{E}_{\mu\circ\pi}[V|S_0=s_0]=  \sum_{k=0}^\infty v_\pi(s^{\odot}_k|s^{\odot}_{k+1}) \prod_{i=0}^{k-1}\gamma_\pi (s^{\odot}_i|s^{\odot}_{i+1})  
\end{align}

Having obtained the ground truth value, in the following, we are going to consider the estimates with small error terms:
\begin{align}
    |v_\pi(\cdot)-\hat{v}_\pi(\cdot)|<\epsilon_v  v_{\text{max}} \ll (1-\gamma)  v_{\text{max}} \quad\quad\text{and}\quad\quad |\gamma_\pi(\cdot)-\hat{\gamma}_\pi(\cdot)|<\epsilon_\gamma\ll (1-\gamma)^2 
\end{align}

To simplify the algebra, we assume without loss of generality that all rewards are non-negative, \st{}, the values are guaranteed to be non-negative as well. Note that the assumption for the reward function to be non-negative is only a cheap trick to ensure we bound in the right direction of $\epsilon_\gamma$ errors in the discounting, but that the theorem would still stand if it were not the case: simply replace $v_{\text{max}}$ with $v_{\text{max}} - v_{\text{min}}$). With the assumption, we can write the upper bound as:

\begin{align}
    &\ \hat{v}_{\mu\circ\pi}(s) \coloneqq \sum_{k=0}^\infty \hat{v}_\pi(s^{\odot}_k|s^{\odot}_{k+1}) \prod_{i=0}^{k-1}\hat{\gamma}_\pi (s^{\odot}_i|s^{\odot}_{i+1})  \\
    &\leq \sum_{k=0}^\infty \left(v_\pi(s^{\odot}_k|s^{\odot}_{k+1})+\epsilon_v  v_{\text{max}}\right) \prod_{i=0}^{k-1}\left(\gamma_\pi(s^{\odot}_i|s^{\odot}_{i+1})+\epsilon_\gamma\right)  \\
    &\leq v_{\mu\circ\pi}(s) + \sum_{k=0}^\infty \epsilon_v  v_{\text{max}} \prod_{i=0}^{k-1}\left(\gamma_\pi(s^{\odot}_i|s^{\odot}_{i+1})+\epsilon_\gamma\right) + \sum_{k=0}^\infty \left(v_\pi(s^{\odot}_k|s^{\odot}_{k+1})+\epsilon_v v_{\text{max}}\right) k\epsilon_\gamma\gamma^k + o(\epsilon_v+\epsilon_\gamma)  \\
    &\leq v_{\mu\circ\pi}(s) + \epsilon_v  v_{\text{max}} \sum_{k=0}^\infty  \gamma^k + \epsilon_\gamma v_{\text{max}} \sum_{k=0}^\infty k\gamma^k  + o(\epsilon_v+\epsilon_\gamma) \\
    &\leq v_{\mu\circ\pi}(s) + \frac{\epsilon_v  v_{\text{max}}}{1-\gamma} + \frac{\epsilon_\gamma v_{\text{max}}}{(1-\gamma)^2}  + o(\epsilon_v+\epsilon_\gamma)
\end{align}

Similarly, we can derive a lower bound:
\begin{align}
    &\ \hat{v}_{\mu\circ\pi}(s) \coloneqq \sum_{k=0}^\infty \hat{v}_\pi(s^{\odot}_k|s^{\odot}_{k+1}) \prod_{i=0}^{k-1}\hat{\gamma}_\pi (s^{\odot}_i|s^{\odot}_{i+1})  \\
    &\geq \sum_{k=0}^\infty \left(v_\pi(s^{\odot}_k|s^{\odot}_{k+1})-\epsilon_v  v_{\text{max}}\right) \prod_{i=0}^{k-1}\left(\gamma_\pi(s^{\odot}_i|s^{\odot}_{i+1})-\epsilon_\gamma\right)  \\
    &\geq v_{\mu\circ\pi}(s) - \sum_{k=0}^\infty \epsilon_v  v_{\text{max}} \prod_{i=0}^{k-1}\left(\gamma_\pi(s^{\odot}_i|s^{\odot}_{i+1})-\epsilon_\gamma\right) - \sum_{k=0}^\infty \left(v_\pi(s^{\odot}_k|s^{\odot}_{k+1})-\epsilon_v v_{\text{max}}\right) k\epsilon_\gamma\gamma^k + o(\epsilon_v+\epsilon_\gamma)  \\
    &\geq v_{\mu\circ\pi}(s) - \epsilon_v  v_{\text{max}} \sum_{k=0}^\infty  \gamma^k - \epsilon_\gamma v_{\text{max}} \sum_{k=0}^\infty k\gamma^k  + o(\epsilon_v+\epsilon_\gamma) \\
    &\geq v_{\mu\circ\pi}(s) - \frac{\epsilon_v  v_{\text{max}}}{1-\gamma} - \frac{\epsilon_\gamma v_{\text{max}}}{(1-\gamma)^2}  + o(\epsilon_v+\epsilon_\gamma)
\end{align}


\end{proof}

Let us carefully interpret and discuss the implication of the proved theorem:

\subsubsection{No Assumption on Optimality}

The methodology of using proxy problems and the theorem above assume that it would be difficult to learn effectively a \red{$\pi$} for all sub-problems: when the target is too far, and that we would rather use a proxy to construct a path with shorter-distance transitions. Therefore, we never want to make any optimality assumption on \red{$\pi$}, otherwise our approach is pointless: 
If the low-level policy \red{$\pi$} is perfect, then the best high-level policy $\mu$ should always directly choose the task goal as target. A triangular inequality can be shown that with a perfect $\pi$ and a perfect estimate of $v_\pi$ and $\gamma_\pi$, the performance will always be optimal by selecting $s_1^\odot=s_g$. 

The basis of divide-and-conquer strategies, including proxy problems in the \Skipper{} framework, is that learning an all-capable policy is difficult in practice. The lack of assumption here is very appropriate in our view.

\subsubsection{Assumption on Checkpoint Feasibility}
The analyses assumed that the target checkpoints are all feasible for convenience. While in reality (in the approximate RL setting without access to all states), since checkpoints need to be generated, the checkpoints could end up being hallucinated targets which cannot be achieved. To be able to handle this more realistic setting, the estimators responsible for estimating $\blue{\gamma^\pi}$ need to produce the correct estimate $0$. Such a problem is not unique to \Skipper{} and I will discuss this problem in great detail in Chap.~\ref{cha:delusions}.

\subsubsection{Assumption of Estimation Accuracy \& Improving Low-Level Policy}

Thm.~\ref{thm:proxyprobsperf} provides guarantees on the solution to the overall problem, whose quality depends on both the quality of the estimates (distances/discounts, rewards) and the quality of the policy. Note that the theorem guarantees accuracy to the solution of the overall problem conditioned on a snapshot of the current policy, which should evolve towards optimal during training. We need to interpret this carefully: 1) bad policies with good estimation will lead to an accurate yet potentially bad overall solution; 2) No matter the quality of the policy, with a bad estimation, it may result in a poor estimate of solutions; 3) It is expected that a near-optimal policy with good estimations will lead to a near-optimal solution.

These amplify the necessity to be careful about if the learning rules for the estimates and the policy could converge to the correct values (Sec.~\ref{sec:justify_update_rules}, Page.~\pageref{sec:justify_update_rules}).

\subsubsection{Applicable Scenarios due to Assumptions}
Thm.~\ref{thm:proxyprobsperf} indicates that once the agent achieves high accuracy estimation of the model for the proxy problem and a near-optimal lower-level policy $\red{\pi}$, it converges toward optimal performance. 

Despite the theorem's generality, in the experiments, we limit ourselves to the scope where the accuracy assumption can be met non-trivially, \ie{}, while avoiding degenerate proxy problems whose checkpoint transitions involve no rewards, that is, in tasks with sparse terminal rewards (\eg{}, \RDS{} and \SSM{}), where the goals are included as permanent vertices in the proxy problems. This is a case where the accuracy assumption can be met non-trivially, \ie{}, while avoiding degenerate proxy problems whose edges involve no rewards. Following Thm.~\ref{thm:proxyprobsperf}, we design the \Skipper{} framework to implement the construction and planning over proxy problems.

In Sec.~\ref{sec:limitations_skipper} in Chap.~\ref{cha:conclusion} (Page.~\pageref{sec:limitations_skipper}), we will present some detailed discussion regarding the limitation brought by the assumptions used in the analyses.

\section{Methodology: \texorpdfstring{\Skipper{}}{Skipper} Framework - Spatially \& Temporally Abstract Planning}
We now look into the design of \Skipper{} - a framework that implements the task decomposition provided by \textbf{proxy} problems. The design of the framework has the following highlights:

\begin{itemize}[leftmargin=*]
    \item \textbf{Decision-time planning} is used for its ability to generalize better in novel situations;
    \item \textbf{Spatio-temporal abstraction}: temporal abstraction breaks down the given task into smaller ones, while spatial abstraction over the state features improves local generalization;
    \item \textbf{Higher quality proxies}: we introduce pruning techniques to improve the quality of sub-problems;
    \item \textbf{Learning end-to-end from hindsight, off-policy}: to maximize sample efficiency and the ease of training, we propose to use auxiliary (off-)policy methods for edge estimation, and learn a context-conditioned checkpoint generation, both from hindsight experience replay;
\end{itemize}

Designing \Skipper{} requires solving two big problems, which are explained as follows, corresponding to the edges and the vertices of the proxy problems, respectively.

\subsection{Problem 1: Edge Estimation}

First, we discuss how \Skipper{} provides the edges needed to assemble the proxy problems. In this sub-section, we assume that the vertices of the proxy problems are already given. We start the discussions on edges because it is easier to understand and can provide intuitions about the more difficult problem of how \Skipper{} can construct the vertices of the proxy problems.


\subsubsection{Spatial Abstraction}
\label{sec:skipper_spatial_abstraction}

Each edge in the proxy problem defines a sub-problem. Some sub-problems are good and desirable, and some are not. Sub-problems are characterized by their locality: transitioning from one checkpoint to another often only involves limited few aspects of the environmental state. It is thus crucial to use inductive biases that can take advantage of the locality instead of using generic neural network architectures.

To parallel with and differentiate from temporal abstraction, we call the behavior pattern of an agent to selectively focus on parts of the environment relevant for the sub-problem - ``spatial abstraction''. Spatial abstraction is thus core to the previous chapter.

Inspired by conscious information processing in brains \citep{dehane2017consciousness}, the intuitions gathered in the previous chapter, and existing approach in \citet{sylvain2019locality}, we introduce a local perceptive field selector, $\sigma$, consisting of an attention bottleneck that (soft-)selects the top-$k$ local segments of the full state (\eg{}, a feature map by a typical convolutional encoder); all segments of the state compete for the $k$ attention slots, \ie{}, irrelevant aspects of state are discouraged or discarded, forming a \textit{partial} state representation \citep{mott2019towards,tang2020neuroevolution,zhao2021consciousness,alver2023minimal}. More specifically, this soft-selection mechanism is implemented using a top-down attention mechanism that queries the set of local receptive fields extracted by the CNN. The $h \times w \times c$ feature map is ``sliced'' and transformed into a set containing $h \times w$ local unordered feature objects, each of dimension $c$, with each of these vectors being a local receptive field on top of a local patch of the observation (while in Chap.~\ref{cha:CP}, each vector maps to exactly one cell in the gridworld). The top-down mechanism conditions itself on the sub-problem, \ie{}, the intention to transition from the current state to the target state. The output of the soft-selection is a vector that aggregates the information from the top-$k$ most relevant local patches of the CNN features. For more details regarding top-$k$ and top-down attention mechanisms, please see Sec.~\ref{sec:attention} on Page.~\pageref{sec:attention}.

This method of selecting top-$k$ receptive fields can be viewed as a more relaxed approach, generalizing the top-down attention controlled bottleneck with set-based representations in Chap.~\ref{cha:CP} to be compatible with general visual inputs processed by Convolutional Neural Networks (CNNs) \citep{lecun1989backpropagation} without the need for set-based state representation encoding.

\begin{figure}
\centering
\includegraphics[width=0.95\textwidth]{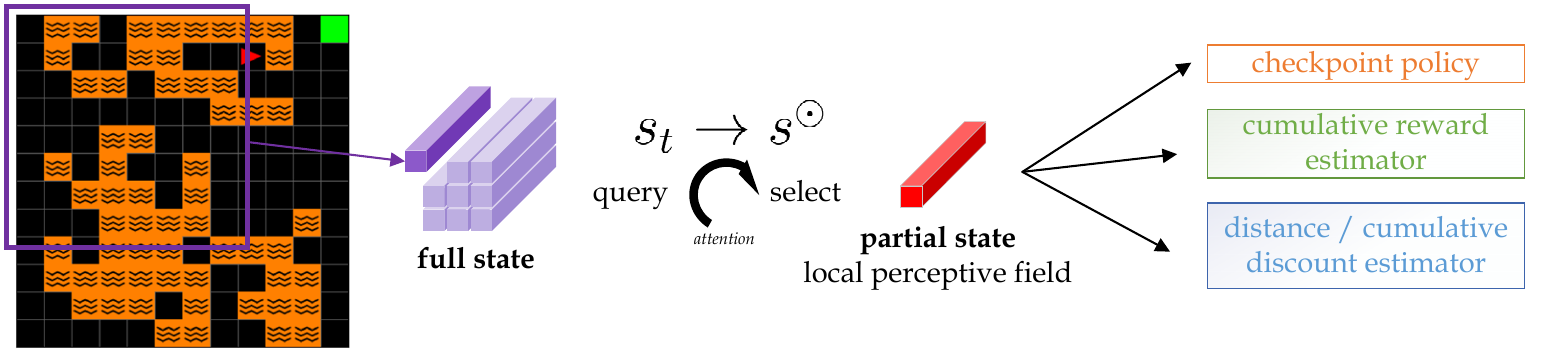}
\caption[Illustration of the Spatial Abstraction Mechanism]{\textbf{Illustration of the Spatial Abstraction Mechanism}: the top-down semi-hard attention is conditioned on an encoding of the checkpoint transition, established over two state-like representations.}
\label{fig:local_field}
\end{figure}

We provide a visualization of how spatial abstraction works in Fig.~\ref{fig:local_field}. Through $\sigma$, the auxiliary estimators, to be discussed soon, force the bottleneck mechanism to promote aspects relevant to the local estimation of connections between the checkpoints. The rewards and discounts are then estimated from the partial aspects of the environmental state relevant to the sub-problem at hand, conditioned on the agent's policy.

\subsubsection{Checkpoint-Achieving Policy}
The checkpoint-achieving policy implements \red{$\pi$}. We ask \Skipper{} to obtain the low-level policy $\red{\pi}$ by maximizing an intrinsic reward, which incentivize the achievement of a target checkpoint $S^{\odot}$. In this approach, the choice of intrinsic reward is flexible; for example, one could use a reward of $+1$ when $S_{t+1}$ is within a small radius of $S^{\odot}$
according to some distance metric, or use reward-respecting intrinsic rewards that enable more sophisticated behaviors, as in \citep{sutton2023reward}. In the following, for simplicity, we will denote the checkpoint-achievement condition with equality: $S_{t+1}=S^{\odot}$.

\subsubsection{Relationship Estimators}
\label{sec:skipper_update_rules}
Proxy problems require that the relationship estimates between checkpoints be capability-aware, \ie{}, conditioned on the low-level policy $\red{\pi}$. Thus, the learners must be off-policy compatible.

With the consideration in mind, we designed learning rules that estimate the relationship between checkpoints with auxiliary reward signals that need not be tuned for each application to different tasks \citep{zhao2020meta}. Notably, these estimates are learned with C51-style distributional RL \citep{bellemare2017distributional}, where the output of each estimator takes the form of a histogram over scalar support (Sec.~\ref{sec:distoutputs}, Page.~\pageref{sec:distoutputs}).

\paragraph{Cumulative Reward}
following the design of proxy problems, the first interesting edge estimate to learn is the cumulative-discounted task reward \darkgreen{$v_{ij}^\pi$} along the possible trajectories leading to the target checkpoint $s_j$ under $\red{\pi}$ from a starting state $s_i$. We learn this by policy evaluation on an auxiliary reward that is the same as the original task reward everywhere except when reaching the target. Given a hindsight sample $\langle x_t, a_t, r_{t+1}, x_{t+1}, x^{\odot} \rangle$ and the corresponding encoded sample $\langle s_t, a_t, r_{t+1}, s_{t+1}, s^{\odot} \rangle$, we train $\darkgreen{V_\pi}$ with KL-divergence as follows:
\begin{align}
\begin{split}
& \darkgreen{\hat{v}_\pi(s_t \to s^{\odot}, a_t)} \equiv \darkgreen{\hat{v}_\pi(\sigma(s_t| s_t \to s^{\odot}), a_t | \sigma(s^{\odot}| s_t \to s^{\odot}))} \\
& \gets \begin{cases}
R(s_t, a_t, s_{t+1}) + \gamma \darkgreen{\hat{v}_\pi(\sigma(s_{t+1}| s_{t+1} \to s^{\odot}), a_{t+1} | \sigma(s^{\odot}| s_{t+1} \to s^{\odot}))} &\text{if } s_{t+1} \neq s^{\odot}\\
R(s_t, a_t, s_{t+1}) &\text{if } s_{t+1} = s^{\odot}
\end{cases}
\end{split}
\label{eq:rule_learn_reward}
\end{align}
where $\sigma(s| s_{t} \to s^{\odot})$ is the spatially-abstracted from the full state $s$ given intention to go from $s$ to $s^{\odot}$, and $a_{t+1} \sim \red{\pi(\cdot | \sigma(s_{t+1}| s_{t+1} \to s^{\odot}), \sigma(s^{\odot}| s_{t+1} \to s^{\odot}))}$. Similar methods on estimating the reachability between states also exist in literature, however, without the considerations for spatial-abstraction.

\paragraph{Cumulative Distances / Discounts}
Besides $\darkgreen{v_\pi}$, the second important estimate for a checkpoint transition is the discounts $\blue{\gamma_\pi}$, which is needed for proxy problems.

Notably, learning the cumulative discount leading to $s_\odot$ under $\red{\pi}$ is numerically difficult, even with C51, because the target values are heavily skewed towards $1$ if $\gamma \approx 1$. Note that this is not to say that reward estimation in Eq.~\ref{eq:rule_learn_reward} is numerically simple, since reward functions can be diverse.



We propose a way to bypass the numerical difficulties by take a detour to instead effectively estimating cumulative (truncated) distances (or trajectory length) under \red{$\pi$}. Similar to how we learn the cumulative rewards $\darkgreen{\hat{v}_\pi}$, such distances can be learned also with policy evaluation, where the auxiliary reward is $+1$ on every transition, except at the targets:
\begin{align}
\begin{split}
& \blue{\hat{D}_\pi(s_t \to s^{\odot}, a_t)} \equiv \blue{\hat{D}_\pi(\sigma(s_t| s_{t} \to s^{\odot}), a_t | \sigma(s^{\odot}| s_{t} \to s^{\odot}))} \\
& \gets \begin{cases}
1 + \blue{\hat{D}_\pi(\sigma(s_{t+1}| s_{t+1} \to s^{\odot}), a_{t+1} | \sigma(s^{\odot}| s_{t+1} \to s^{\odot}))} &\text{if } s_{t+1} \neq s^{\odot}\\
1 &\text{if } s_{t+1} = s^{\odot}\\
\infty &\text{if } s_{t+1} \text{ is terminal and } s_{t+1} \neq s^{\odot}\\
\end{cases}
\end{split}
\label{eq:rule_learn_distance}
\end{align}

where $a_{t+1} \sim \red{\pi(\cdot | \sigma(s_{t+1}| s_{t+1} \to s^{\odot}), \sigma(s^{\odot}| s_{t+1} \to s^{\odot}))}$.

\begin{SCfigure}[][htbp]
\includegraphics[width=0.32\textwidth]{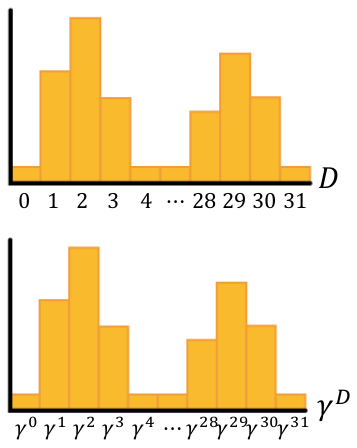}
\caption[Simultaneously Estimating Distributions of Discounts \& Distances]{\textbf{Simultaneously Estimating Distributions of Discounts \& Distances}: by swapping the support with the corresponding discount values, the distribution of the cumulative discount can be inferred from the output histogram of cumulative distances. Vice versa.}
\label{fig:support_override}
\end{SCfigure}

\paragraph{Recovering Discounts from Distances}
With the learned distribution of cumulative distances by a C51 architecture \citep{dabney2018distributional}, the cumulative discount can then be extracted by swapping the support of the output histogram with the corresponding discounts values. This process is illustrated in Fig.~\ref{fig:support_override}. This bypasses the problem of $\doubleE[\gamma^D] \neq \gamma^{\doubleE[D]}$, because the probability of having a trajectory length of 5 under policy $\pi$ from state $s_t$ to $s_\odot$ is the same as a trajectory having discount $\gamma ^ 5$.

The learned distances are not only used to recover the discounts needed for proxy problems, but are also used to prune unwanted checkpoints to simplify the proxy problem via $k$-medoids pruning (Sec.~\ref{sec:k_medoids}, Page.~\pageref{sec:k_medoids}), as well as pruning far-fetched edges (Sec.~\ref{sec:prune_edges}, Page.~\pageref{sec:prune_edges}), both to be introduced soon.

\subsubsection{Justifying Update Rules for Edge Estimation}
\label{sec:justify_update_rules}

As previously analyzed, the quality of the estimations of the cumulative rewards and the cumulative discounts during checkpoint transitions determine if effective solutions of the proxy problems can be found.

Thus, we want to understand if the update rules proposed previously would lead to high-quality estimates that the proxy problems need.

The cumulative reward and cumulative discount are estimated by applying policy evaluation given $\pi$, on the two sets of auxiliary reward signals, respectively.

The following equality for the cumulative discounted reward random variable can be shown:
\begin{align}
    V_\pi(s_t,a_t|s^{\odot}) = R(s_t,a_t,S_{t+1})+\gamma V_\pi(S_{t+1},A_{t+1}|s^{\odot}) = \sum_{\tau=t}^\infty \gamma^{\tau-t} R(S_\tau,A_\tau,S_{\tau+1}),
\end{align}
where $S_{t+1}\sim p(\cdot|s_t,a_t)$, $A_{t+1}\sim\pi(\cdot|S_{t+1},s^{\odot})$. Let $V_\pi(s|s^{\odot}) \coloneqq V_\pi(s,A|s^{\odot})$ with $A\sim\pi(\cdot|s,s^{\odot})$. Noticing that $V_\pi(S_{t+1},A_{t+1}|s^{\odot})=0$ if $S_{t+1}=s^{\odot}$, the equivalence is established towards Rule.~\ref{eq:rule_learn_reward} (Page.~\pageref{eq:rule_learn_reward}).
 
Similarly, we connect the cumulative discount random variable with Rule.~\ref{eq:rule_learn_distance} (Page.~\pageref{eq:rule_learn_distance}). Let $\Gamma_\pi(s_t|s^{\odot}) \coloneqq \Gamma_\pi(s_t,A_t|s^{\odot})$ with $A_{t+1}\sim\pi(\cdot|S_{t+1},s^{\odot})$:
\begin{align}
    \Gamma_\pi(S_t,A_t|s^{\odot}) =  \gamma\cdot\Gamma_\pi(S_{t+1},A_{t+1}|s^{\odot}) = \gamma^{T_\perp-t} \mathbb{I}{\{S_{T_\perp}=s^{\odot}\}}
\end{align}

where $T_\perp$ denotes the timestep when the trajectory terminates, and with $\Gamma_\pi(S_{t+1},A_{t+1}|s^{\odot})=1$ if $S_{t+1}=s^{\odot}$ and $\Gamma_\pi(S_{t+1},A_{t+1}|s^{\odot})=0$ if $S_{t+1}\neq s^{\odot}$ is terminal.


\subsection{Problem 2: Vertex Generation}
\label{sec:skipper_generator}

Proxy problems can only be constructed with both the vertices and the edges in hand. Since we do not assume the access to an oracle, to be able to use proxy problems, the vertices (checkpoint states) must be generated using a learned model.

Different from most model-based approaches that utilize a predictive model to simulate the changes of the environment, we instead propose a temporally abstract checkpoint generator which aims to directly imagine the distant future states \textit{without needing to know how exactly the agent might reach them nor worrying about if they are reachable}. The feasibility of checkpoint transitions will be handled by the connection estimates instead. In the language of some existing literature, the generator here can be called a ``goal generator''. We also discuss the implication of such approaches in target-assisted frameworks in Sec.~\ref{sec:target_directed_framework} on Page.~\pageref{sec:target_directed_framework}.

\subsubsection{Checkpoint Generator: Split Context \& Partial Descriptions}
We want to design a checkpoint generator which generalizes well across diverse tasks, \ie{}, propose useful checkpoints even in novel scenarios. This means the generator should be able to capture the underlying causal mechanisms across tasks (a challenge for existing model-based methods) \citep{zhang2020invariant}. For this, we propose that the checkpoint generator learns to split the state representation into two parts: 1) an episodic / task \textbf{context} and 2) a \textbf{partial description}.

We design the contexts and partial descriptions to coordinate as follows: in different contexts, the same partial description could correspond to different states. However, within a given context, the same partial description should always map to the same state. In other words, the partial description should capture the defining distinct feature of a certain state, while the combination of the context with the partial description recovers the information of a full state.

\begin{figure}[htbp]
\centering
\captionsetup{justification = centering}
\includegraphics[width=1.00\textwidth]{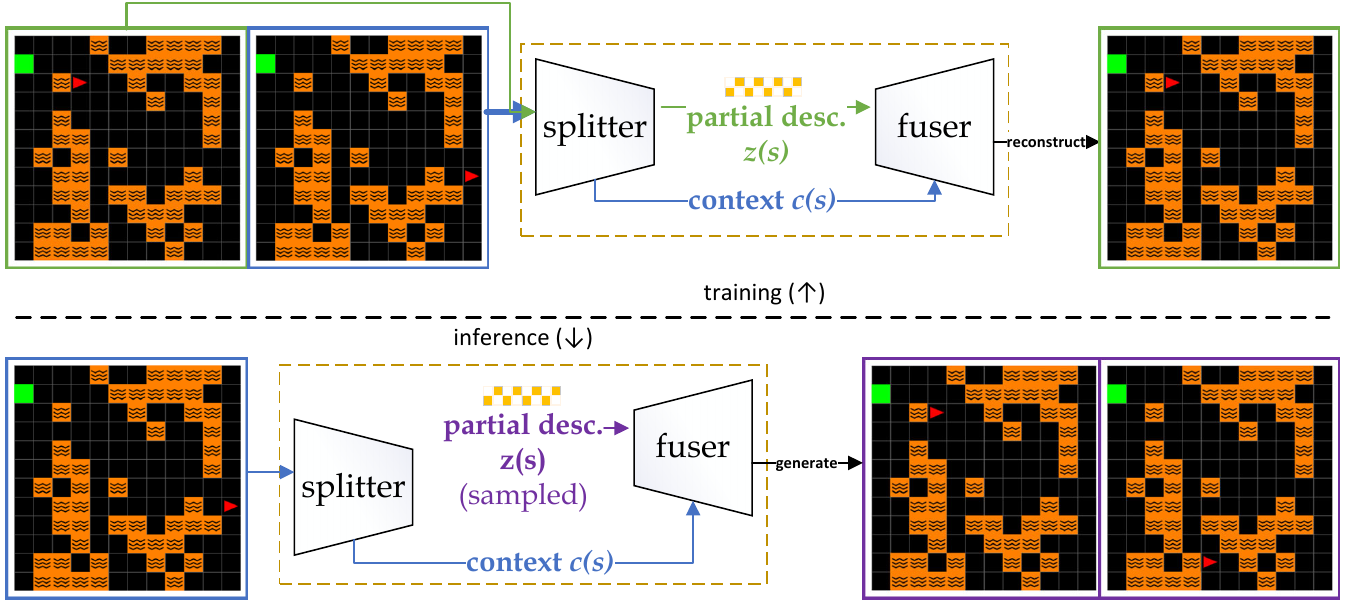}
\caption[Checkpoint Generator Training \& Inference]{\textbf{Checkpoint Generator Training \& Inference}: during training, both \blue{$x_t$} (current observation) and \darkgreen{$x^\odot$} (relabeled target observation) participate in the calculation of the training losses. The partial description is extracted from \darkgreen{$x^\odot$}, while the context is extracted from \blue{$x_t$}. Both are extracted by the same two-headed neural network, marked as the ``splitter''. The training losses seek to reconstruct \darkgreen{$x^\odot$}, by making the model learn that with a context from any current state, \eg{}, that of \blue{$x_t$}, \darkgreen{$x^\odot$} should be reconstructed as long as its corresponding partial description is used; during inference, the context from the current observation is used together with sampled partial descriptions to generate checkpoints.}
\label{fig:generator_training_inference}
\end{figure}

In a navigation problem, for example, as in Fig.~\ref{fig:generator_training_inference}, a context could be a representation of the layout of the gridworld (of a specific task instantiation), and the partial description could represent the 2D-coordinates of the agent's location.

Note that the representations of partial descriptions do  not necessarily capture the relevant transitional information between two states. This is why the design of partial description / context inside the generator should not be confused with the partial states created by spatial abstraction in Sec.~\ref{sec:skipper_spatial_abstraction} on Page.~\pageref{sec:skipper_spatial_abstraction}.

As shown in Fig.~\ref{fig:generator_training_inference}, the proposed information split is achieved by an inductive bias of two parameterized trainable functions: the \textit{splitter} $\scriptE_{CZ}$, mapping the input $s$ into a representation of a context $c(s)$ and a partial description $z(s)$, as well as the \textit{fuser} $\bigoplus$ that recovers $s$ from the extracted $\langle c, z \rangle$. To achieve consistent context extraction across states in the same episode, at training time, we force the context to be extracted from other states in the same episode, instead of the input (upper pass in Fig.~\ref{fig:generator_training_inference}).

Crucial for grounding the two representations into reality and stabilizing the hierarchical planning framework that is \Skipper{}, we rely on a reconstruction loss, despite having previously criticized this methodology in Sec.~\ref{sec:dynamics_predictors} and having avoided it in the previous work. As previously discussed, in \Skipper{}, we rely on a generative model instead of a predictive model, which relies more on techniques such as reconstruction. A reconstruction loss can be established on either the observation-level or a learned state level, depending on the need of the application. With full-observability, when the observation space is simpler than the state space to learn, we would recommend the reconstruction there, given the environment is fully observable.

The two representations are learned by the generator, powered by a conditional Variational Auto-Encoder (CVAE) architecture (Sec.~\ref{sec:VAE}, Page.~\pageref{sec:VAE}) from hindsight-labeled targets (Sec.~\ref{sec:source_target_pair_hindsight_relabeling}, Page.~\pageref{sec:source_target_pair_hindsight_relabeling}). The generator learns a distribution $p(s^{\odot} | C(s_t)) = \sum_z{p(s^{\odot} | C(s_t), z) p(z | C(s_t))}$, where $C(s_t)$ is the extracted context from $s_t$ and $z$s are the partial descriptions. We train the generator by minimizing the evidence lower bound on $\langle s_t, s^{\odot} \rangle$ pairs chosen with HER. The processes of training and inference with the proposed checkpoint generator are illustrated in Fig.~\ref{fig:generator_training_inference}.

With hindsight relabeling, we may need a more careful control over the distribution of the future checkpoints, which we seek to learn with the generator. This distribution, unlike the outcome of $1$-step transition, could be more diverse and non-stationary given a changing behavior policy, depending on the relabeling strategies \citep{andrychowicz2017hindsight}. In our implementation, the targets are labeled with the \episodestr{} relabeling strategy. Improper hindsight relabeling, such as using \episodestr{} exclusively as here, can cause delusional behaviors in target-assisted agents such as \Skipper{}. We will discuss this issue in detail in the next chapter.

\phantomsection
\label{sec:skipper_discrete_partial_desc}

Similar to the discrete and categorical architectures used in \citet{hafner2023mastering,hansen2023td}, we constrain the partial description in the forms of bundles of binary variables, which requires the techniques of straight-through gradient estimators \citep{bengio2013estimating}. By design, these partial description latents can be easily sampled or composed during inference, \ie{}, decision-time planning. Compared to models such as that in \Director{} \citep{hafner2022deep}, which generates intermediate sub-goals on the on-policy trajectory, our proposed generator can generate and handle a more diverse distribution of states, beneficial for generalization in novel scenarios.

The simple representations of the partial descriptions also granted us the simplicity to represent the sub-problems with a simple concatenation of the partial descriptions of two checkpoints. This is used in the top-down attentions for the spatial abstraction mechanism.

\subsubsection{Pruning with $k$-medoids}
\label{sec:k_medoids}

In this chapter, the proposed conditional generation model is limited in the sense that they are return-unaware. Without learning how to be return-aware, the proxy problem can instead be improved by making it more sparse, and making the proxy problem vertices more evenly spread in state space. To achieve this, we propose a pruning process based on the classical $k$-medoids clustering algorithm \citep{kaufman1990medoids}, which only requires pairwise distance between states as input. During the construction of a proxy problem, a larger number of checkpoints are first generated. Then, these checkpoints are clustered using $k$-medoids, after which all checkpoints but the cluster centers are discarded. The cluster centers are always real generated checkpoints instead of imaginary weighted sums of state representations that will be produced by clustering algorithms such as $k$-means \citep{lloyd1982least}.

Notably, for sparse reward tasks, the generator cannot guarantee the presence of the rewarding checkpoints, such as the task goal, to be in the proposed proxy problem. We could remedy this by explicitly learning the generation of the rewarding states with another conditional generator. These rewarding states should be kept as vertices (immune from pruning).

We present the pseudocode of the modified $k$-medoids algorithm for pruning overcrowded checkpoints in Alg.~\ref{alg:k_medoids}. The changes upon the original $k$-medoids algorithm are marked in \purple{purple}, which correspond to a forced preservation of the special input data points as cluster centers: when $k$-medoids is called after the unpruned graph is constructed, $\scriptS_\vee$ is set to be the set containing the goal state only. This is intended to make the remaining cluster centers (selected checkpoints) span more uniformly in the state space, while preserving the goal. This is particularly proper if the proxy problems are going to be conserved throughout the episode and reused.

There are some additional technical details we need to pay attention to, particularly because $k$-medoids was designed to handle undirected distances. Let the estimated distance matrix be $D$, where each element $d_{ij}$ represents the estimated trajectory length it takes for $\pi$ to fulfill the transition from checkpoint $i$ to checkpoint $j$. Since $k$-medoids cannot handle infinite distances (\eg{}, from a terminal state to another state), to still maximize for the span, the distance matrix $D$ is truncated, and then we take the elementwise minimum between the truncated $D$ and $D^T$ to preserve the one-way distances. The matrix containing the elementwise minimums would be the input of the pruning algorithm.

\begin{algorithm*}[htbp]
\caption{Checkpoint Pruning with $k$-medoids}
\label{alg:k_medoids}

\SetAlgoNlRelativeSize{-1}
\KwData{$X = \{x_1, x_2, \ldots, x_n\}$ (state indices), $D$ (estimated distance matrix), $\scriptS_{\vee}$ (states that must be kept), $k$ (\#checkpoints to keep)\\
}
\KwResult{$\scriptS_\odot \equiv \{M_1, M_2, \ldots, M_k \}$ (checkpoints kept)}
\BlankLine

Initialize $\scriptS_\odot \equiv \{M_1, M_2, \ldots, M_k \}$ randomly from $X$\\

\purple{make sure $\scriptS_\vee \subset \scriptS_\odot$}\\

\Repeat{Convergence (no cost improvement)}{
    Assign each data point $x_i$ to the nearest medoid $M_j$, forming clusters $C_1, C_2, \ldots, C_k$\;
    \ForEach{medoid $M_j$}{
        Calculate the cost $J_j$ of $M_j$ as the sum of distances between $M_j$ and the data points in $C_j$\;
    }
    Find the medoid $M_j$ with the lowest cost $J_j$\;
    \If{$M_j$ changes}{
        \purple{make sure $\scriptS_\vee \subset \scriptS_\odot$}\\
        Replace $M_j$ with the data point in $C_j$ that minimizes the total cost\;
    }
}
\end{algorithm*}

\phantomsection
\label{sec:prune_edges}

In addition to vertex pruning, we also prune the edges according to a distance threshold, \ie{}, all edges with estimated distance over the threshold are deleted from the complete graph of the pruned vertices. This is enabled by how we interchangeably learned the cumulative discounts and distances. Such approach biases the plans, \ie{}, the solutions of the proxy problems, towards shorter-length, smaller-scale sub-problems, as distant checkpoints are difficult for \red{$\pi$} to achieve.

\subsection{\texorpdfstring{\Skipper{}}{Skipper} Framework: Assemble Everything}

\begin{figure}[htbp]
\centering
\captionsetup{justification = centering}
\includegraphics[width=1.00\textwidth]{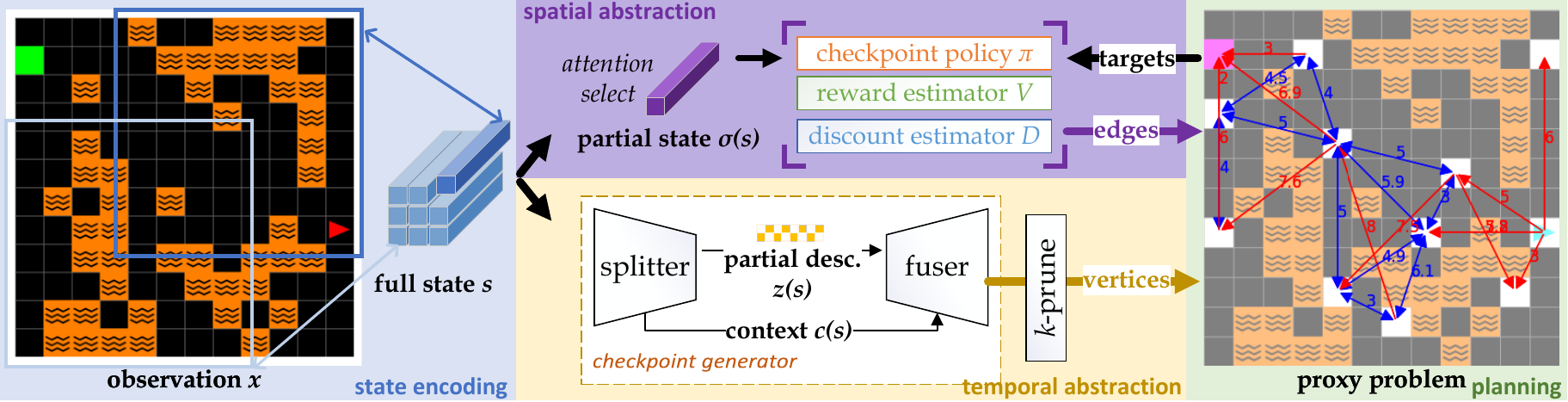}
\caption[\Skipper{} Framework]{\textbf{\Skipper{} Framework}: 1) Partial states, used for low-level policy and edge estimation, consist of a few local perceptive fields, soft-selected via top-$k$ attention \citep{gupta2021memory}. \Skipper{}'s edge estimations and low-level behaviors \red{$\pi$} are based on the partial states. 2) The checkpoint generator learns by splitting the full states into contexts and partial descriptions, and fusing them to reconstruct the input. It imagines checkpoints by sampling partial descriptions and combining them with the episodic contexts; 3) We prune the vertices and edges of the denser graphs to extract sparser proxy problems. Value iteration can be used to solve the proxy problems and the solutions are equivalent to plans in the original problem. Each step in the plan becomes a sub-problem. In the proxy problem example, blue edges are estimated to be bidirectional and red edges have the other direction pruned.}
\label{fig:overall}
\end{figure}

An overall view of the proposed \Skipper{} framework with all components assembled is illustrated in Fig.~\ref{fig:overall}. The pseudocode of \Skipper{} is provided in Alg.~\ref{alg:skipper}, together with the hyperparameters used in later experiments.

\begin{algorithm*}[htbp]
\caption{\Skipper{} with Random Checkpoints \purple{(implementation choice in purple)}}
\label{alg:skipper}

\For{each episode}{
    \darkgreen{// --- start of the \textbf{subroutine} to construct the proxy problem} \\
    
    generate more than necessary \purple{(32)} checkpoints by sampling from the partial descriptions given the extracted context from the initial state; \\

    \purple{$(k=12)$}-medoids pruning upon estimated distances among all checkpoints; \darkgreen{// prune vertices} \\

    use estimators to annotate the edges between the nodes \purple{(including a terminal state estimator to correct the estimates)}; \\

    prune edges that are too far-fetched according to distance estimations \purple{(threshold set to be $8$, same as replan interval)}; \darkgreen{// prune edges} \\

    \darkgreen{// --- end of the \textbf{subroutine} to construct the proxy problem} \\

    \For{each agent-environment interaction step until termination of episode}{
        \If{decided to explore \purple{(\DQN{}-style annealing $\epsilon$-greedy)}}{ 
            take a random action; \\    
        }
        \Else{
            \If{abstract problem just constructed \textbf{or} a checkpoint / timeout reached \purple{($\geq8$ steps since last planned)}}{
                [\textbf{OPTIONAL}] call the \textbf{subroutine} above for \textbf{\Skipper{}-regen}; \\
                run value iteration \purple{(for $5$ iterations)} on the proxy problem, select the target checkpoint; \\
            }
            follow the action suggested by the checkpoint-achieving policy; \\ 
        }
        \If{time to train \purple{(every $4$ actions)}}{
            sample hindsight-relabelled transitions and train checkpoint-achieving policy, checkpoint generator \& estimators \purple{(including a terminal state estimator)}; \\

        }
        save interaction into the trajectory experience replay; \\
    }
    convert trajectory into HER samples \purple{(relabel $4$ random states as additional goals)}; \\
}
\end{algorithm*}

\section{Discussions of Methodologies}
\label{sec:skipper_related_work}

\subsection{About Proxy Problems}
The representation of a proxy problem draws similarity to an early work on landmark-based approximate value iteration \citep{mann2015approximate}. 

Concurrently, \citep{zadem2024reconciling} introduced a three-level hierarchical planning method aimed at decomposing an overall task into abstract goal regions, with sub-steps defined over goal transitions; \citet{lo2024goal} proposed to decompose the task by goal-space MDPs, essentially proxy problems, and conduct background planning (Sec.~\ref{sec:background_planning}, Page.~\pageref{sec:background_planning}) to improve performance. 

Similar to \Skipper{}, whole sets of relationship estimation learning, \eg{}, for reachability, between states and goals are proposed and analyzed in both works. Different from \Skipper{}, both works were not motivated from OOD generalization and focused on a classical single-task RL setting.

For Goal Space Planning (GSP) from \citet{lo2024goal} specifically, extensive studies were done on multiple classical single-training task environments, spanning from tabular methods to deep RL. \citet{lo2024goal} was motivated by the fact that background planning methods often behave worse than the model-free baselines due to the generation of invalid states (a behavior named \textbf{hallucination}, to be discussed in Sec.~\ref{sec:problematic_targets} on Page.~\pageref{sec:problematic_targets}). GSP sought to address the performance issue by constraining background planning to a given set of (abstract) subgoals. The method applied can be viewed from the lens of target-assisted frameworks and is highly similar to our mitigation strategies in Chap.~\ref{cha:delusions} on Page.~\pageref{cha:delusions}.

For methods that abstract the original MDP with representative states, such as proxy problems, planning can be efficient, while the resulting policy could be suboptimal. \citet{lo2024goal} argued that the sub-optimality issue forces a trade-off between increasing the size of the abstract MDP (to increase the policy's expressivity) and increasing the computational cost of decision-time planning. We acknowledge such limitation to the \Skipper{} framework (and so for other decision-time methods such as  \citet{zadem2024reconciling}) and agree it may be a good motivation for using background planning in a classical single-task focused RL setting. On a high-level, proxy problems indeed trade optimality for robustness. We presented the framework's sensitivity to the number of checkpoints in proxy problems in Sec.~\ref{sec:sensitivity_num_checkpoints} on Page.~\pageref{sec:sensitivity_num_checkpoints}. While acknowledging the shortcomings of the methodology of \Skipper{}, we have to point out that background planning methods, including GSP, assumes that a know-all do-all policy can be effectively learned, which is rarely the case in practice. In the analyses in Sec.~\ref{sec:perf_guarantee} (Page.~\pageref{sec:perf_guarantee}), we explicitly discussed why we assumed no such condition.

While both \citet{zadem2024reconciling} and \citet{lo2024goal} require oracles to work, \Skipper{} operates at mainly the state-level and is able to generate the checkpoints, \ie{}, the subgoals, autonomously.

\subsection{About Task Decomposition with Selected States / Sub-Goals}

Some early works such as \citet{mcgovern2001automatic} suggested to use bottleneck states to abstract given tasks into manageable steps. In \citet{kim2021landmarkguided}, promising states to explore are generated and selected with shortest-path algorithms. Similar ideas were attempted for guided exploration \citep{erraqabi2022temporal,kulkarni2016hierarchical}.

Similar to \citet{hafner2022deep}, \citet{czechowski2021subgoal} generate fixed-steps ahead subgoals, while \citet{bagaria2021skill} augments the search graph by states reached fixed-steps ahead. \citet{nasiriany2019planning,xie2020latent,shi2022skill} employ CEM to plan a chain of subgoals towards the task goal \citep{rubinstein1997optimization}.

Similar to our approach, \citet{nair2018visual,florensa2018automatic} use models to imagine subgoals while \citet{eysenbach2019search} search directly on the experience replay. \Skipper{} utilizes proxy problems to abstract the given tasks via spatio-temporal abstractions \citep{bagaria2021skill}. Checkpoints can be seen as sub-goals that generalize the notion of ``landmarks" or ``waypoints'' in \citet{sutton1999between,dietterich2000hierarchical,savinov2018semi}. \citet{zhang2021world} used latent landmark graphs as high-level guidance, where the landmarks are sparsified with weighted sums in the latent space to compose subgoals. In comparison, our checkpoint pruning selects a subset of generated states, which is less prone to issues created by weighted sums.

\subsection{About Temporal Abstraction}

Somewhat similar to the idea behind attention, choosing a checkpoint target is a selection towards certain decision points in the dimension of time, \ie{}, a form of temporal abstraction. Constraining options, \Skipper{} learns the options targeting certain ``outcomes'', which dodges the difficulties of option collapse \citep{bacon2017option} and option outcome modeling by design. It should be acknowledged that the constraints shift the difficulties to generator learning \citep{silver2012compositional,tang2019multiple}. We expect this shift to bring benefits in cases where states are easy to learn and generate, and / or in stochastic environments where the outcomes of unconstrained options are difficult to learn. Constraining options was also investigated in unsupervised settings \citep{sharma2019dynamics}.

\subsection{About Spatial Abstraction}
\label{sec:related_spatial_abstraction}

Spatial abstraction is a special, dynamic kind of ``state abstraction'' \citep{sacerdoti1974planning,knoblock1994automatically}. When state abstraction is being done in a non-dynamic way, it roughly equates to a synonym for state space partitioning due to the focus  \citep{li2006towards}. Spatial abstraction, characterized to capture the behavior of conscious planning in the spatial dimension, focuses on the \textbf{intra-state} partial selection of the environmental state for decision-making. It corresponds naturally to the intuition that state representations should contain useful aspects of the environment, while not all aspects are useful for a particular intent (sub-problem). Efforts toward spatial abstraction are traceable to early hand-coded proof-of-concepts proposed in \citet{dietterich2000hierarchical}. In \citet{zadaianchuk2020self,fu2021learning,shah2021value}, $3$ more recent model-based approaches, spatial abstractions are attempted to remove visual distractors.

Until only recently, attention mechanisms had primarily been used to construct state representations in model-free agents for sample efficiency purposes, overlooking their potentials for generalization \citep{mott2019towards,manchin2019reinforcement,tang2020neuroevolution}. Emphasizing on generalization, our previous work \citep{zhao2021consciousness} used spatially-abstract partial states in decision-time planning, as introduced in Chap.~\ref{cha:CP}. We proposed an attention bottleneck to dynamically select a subset of environmental entities during the atomic-step forward simulation, without explicit goals provided as in \citet{zadaianchuk2020self}. \Skipper{}'s checkpoint transition is a step-up from our old approach, where we now show that spatial abstraction, an overlooked missing flavor, is as crucial for longer-term planning, arguably as important as temporal abstraction \citep{konidaris2009efficient}. 

Concurrently, GSP sought to constrain background planning to a given set of (abstract) subgoals and learning only local, subgoal-conditioned models \citep{lo2024goal}. To a degree, the claimed compatibility to learn only local and sub-goal conditioned models is only materialized with our implemented spatial abstraction mechanisms, in this chapter. 


\subsection{About Planning Estimates}
Hierarchical planning must be established over estimates that can guide the low-level behaviors. \citet{zhang2021world} proposed a distance estimate with an explicit regression. With Temporal-Difference Models (TDMs) \citep{pong2018temporal}, \LEAP{} \citep{nasiriany2019planning} embraces a sparse intrinsic reward based on distances to the goal states. Contrasting with our distance estimates, there is no empirical evidence of TDMs' compatibility with stochasticity and terminal states. \citet{eysenbach2019search} employs a similar distance learning technique to extract the shortest path distance between states present in the experience replay; while our proposed estimators learn the distance conditioned on evolving policies. Similar aspects were also investigated in \citet{nachum2018dataefficient}.

\input{chapter_5_Skipper_exp}

\section{Summary}

In this chapter, we proposed, analyzed and validated \Skipper{}, which builds combined spatio-temporal abstractions and generalizes its learned skills better than other comparable methods.

%% file: chapter_5_Skipper_exp.tex
\section{Research Findings: Experiments}
\label{sec:skipper_exp}
The primary goal of our experiments is to understand how the proposed \Skipper{} framework can bring advantages of (zero-shot) generalization. To fully understand the results, we must have precise control of the difficulty of the training and evaluation tasks. Also, to validate if the empirical performance of our agents matches the formal analyses (Thm.~\ref{thm:proxyprobsperf}), we need to know how close to the (optimal) ground truth our edge estimations and checkpoint policies are. These objectives lead to the need for environments whose ground truth information (optimal policies, true distances between checkpoints, \etc{}) can be computed. For these reasons, we chose to employ \RDS{} as our main experiment environments.

Despite that the experiments in this chapter employ essentially the same kind of environment as the previous chapter, the experimental settings are quite different. 

Across all experiments, we sample training tasks from an environment distribution of difficulty $\delta = 0.4$: each cell in the field has probability $0.4$ to be filled with lava while guaranteeing a path from the initial position to the goal. The evaluation tasks are sampled from a gradient of OOD difficulties - $0.25$, $0.35$, $0.45$ and $0.55$, where the training difficulty acts as mean\footnote{This is an updated setting from that used in Chap.~\ref{cha:CP} (Sec.~\ref{sec:CP_experiments}, Page.~\pageref{sec:CP_experiments}). The training difficulty of $\delta = 0.4$ now is OOD from those of evaluation and is right in the middle of the OOD evaluation difficulty range. The transposed OOD evaluation instances are no longer used because without it 1) the OOD setting becomes closer to the distribution shfit setting commonly used; and 2) the performance discrepancy between the evaluation curves on the training tasks and the evaluation tasks can directly reflect the generalization gap. See Sec.~\ref{sec:OOD_settings} on Page.~\pageref{sec:OOD_settings} for more discussions.}. 

We have taken measures to accelerate the experiments and isolate challenges from the aspects of exploration. During evaluation, the agent is always spawned at the opposite side from the goals. During training, the agent's position is uniformly initialized to speed up training. We provide results for non-uniform training initialization in the later experiments. This change means that the agents can be trained for only $1.5 \times 10^{6}$ interactions to achieve convergence. 

Meanwhile, to increase generalization difficulty for long(er) term planning, we conduct experiments done on large, $12 \times 12$ maze sizes, compared to the maximum $10 \times 10$ used in Chap.~\ref{cha:CP}. The compared agents include:

\begin{itemize}[label={},leftmargin=-0mm]
    \item \textbf{\Skipper{}-once}: A \Skipper{} variant that generates one proxy problem at the start of the episode, and the replanning (choosing a checkpoint target based on the existing proxy problem) only triggers a quick re-selection of the immediate checkpoint target;
    \item \textbf{\Skipper{}-regen}: A variant that re-generates a proxy problem when replanning is triggered;
    \item \textbf{modelfree}: A model-free baseline agent sharing the same base architecture with the \Skipper{} variants - a prioritized distributional \DDQN{} \citep{dabney2018distributional,hasselt2015double};
    \item \textbf{\Director{}}: An adapted \Director{} agent \citep{hafner2022deep}. Since \Director{} discards trajectories that are not long enough for training purposes, we give it extra agent-environment interaction budget to make sure that the same amount of training data is gathered as for the other agents;
    \item \LEAP{}: An adapted \LEAP{} agent for discrete action spaces. We waived the interaction costs for its generator pretraining stage, only showing the second stage of RL pretraining.
\end{itemize}

Please refer to Sec.~\ref{sec:skipper_aux} on Page.~\pageref{sec:skipper_aux} for more details regarding implementation of these agents.

\subsection{Generalization Performance with \texorpdfstring{$50$}{50} Training Environments}
The main set of experiments is focused on how well the compared agents could generate after training on $50$ tasks, a representative configuration of different numbers of training tasks including $\{1, 5, 25, 50, 100, \infty \}$\footnote{$\infty$ training tasks mean that an agent is trained on a different task for each episode. In reality, this may lead to prohibitive costs in creating the training environment.}

\begin{figure}[htbp]
\centering
\subfloat[training, $\delta = 0.4$]{
\captionsetup{justification = centering}
\includegraphics[height=0.22\textwidth]{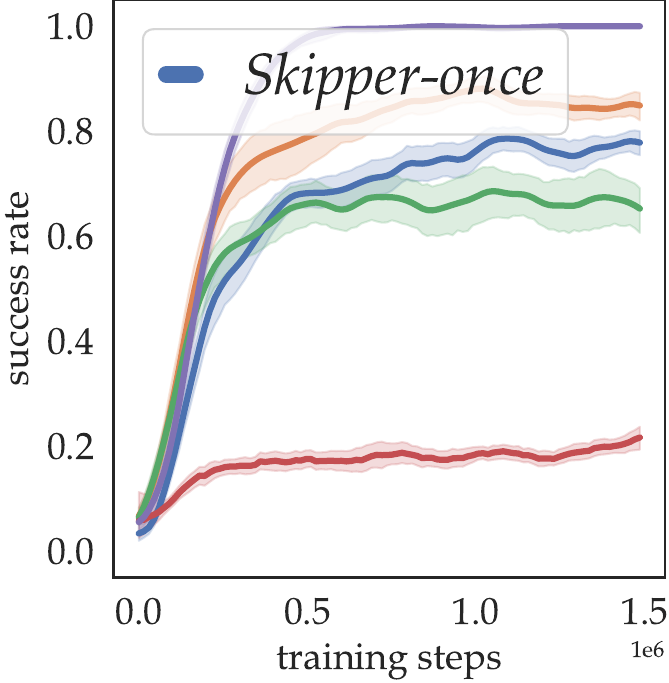}}
\hfill
\subfloat[OOD, $\delta = 0.25$]{
\captionsetup{justification = centering}
\includegraphics[height=0.22\textwidth]{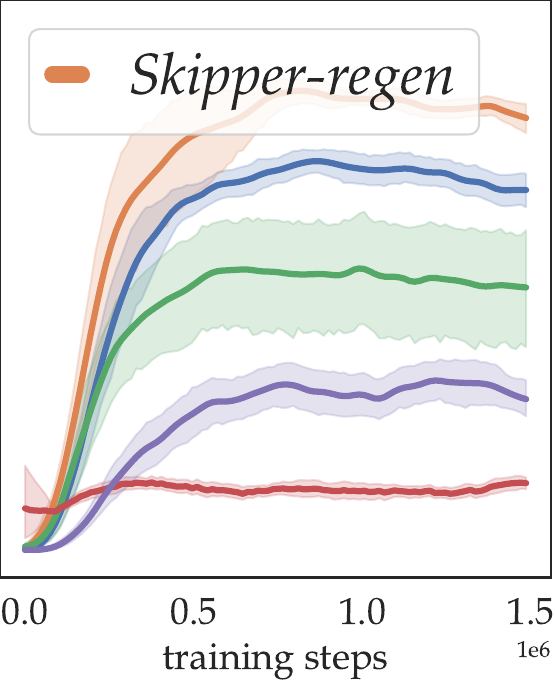}}
\hfill
\subfloat[OOD, $\delta = 0.35$]{
\captionsetup{justification = centering}
\includegraphics[height=0.22\textwidth]{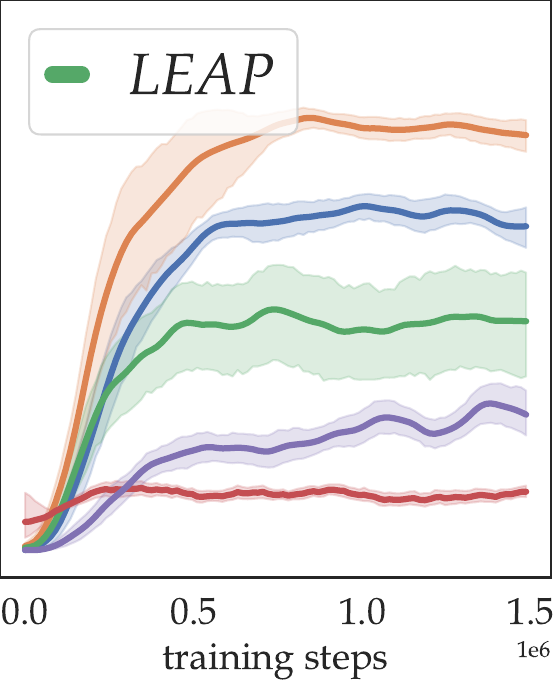}}
\hfill
\subfloat[OOD, $\delta = 0.45$]{
\captionsetup{justification = centering}
\includegraphics[height=0.22\textwidth]{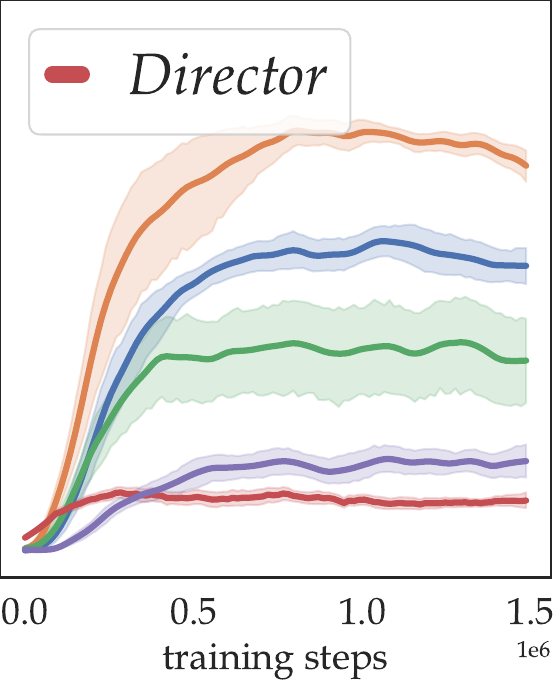}}
\hfill
\subfloat[OOD, $\delta = 0.55$]{
\captionsetup{justification = centering}
\includegraphics[height=0.22\textwidth]{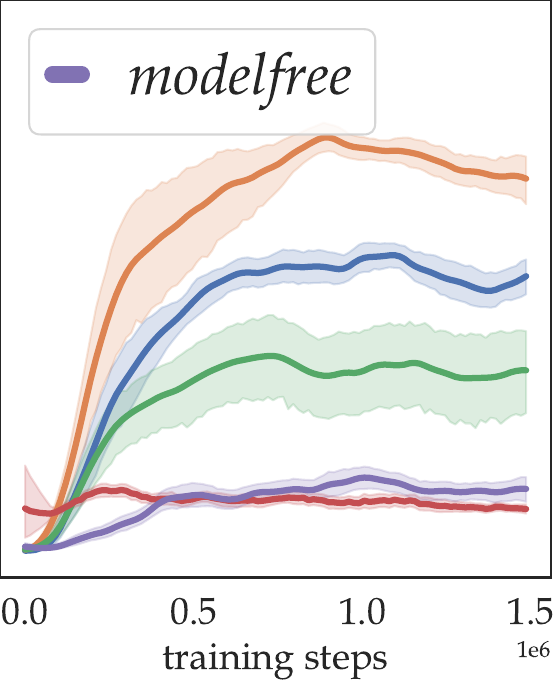}}

\caption[Evaluation Performance Evolution of Compared Agents during Training]{\textbf{Evaluation Performance Evolution of Compared Agents during Training}: the $x$-axes correspond to training progress, while the aligned $y$-axes represent the success rate of episodes (optimal is 1.0). Each agent is trained with $50$ tasks. Each data point is the average success rate over $20$ evaluation episodes, and each error bar (95\% confidence interval) is processed from $20$ independent seed runs. Training tasks performance is shown in a) while OOD evaluation performance is shown in b), c), d), e). }
\label{fig:50_envs}
\end{figure}

Fig.~\ref{fig:50_envs} shows how the agents' performance (measured by the episodic task success rates) evolves during training. These results are obtained with $50$ fixed training tasks (different $50$ task instances for each independent seed run, with each task instance generated at $\delta = 0.4$). In Fig.~\ref{fig:50_envs} \textbf{a)}, we observe how well an agent performs on its training tasks (note the differences in initialization between an evaluation episode and a training episode). If an agent performs well here but badly in \textbf{b)}, \textbf{c)}, \textbf{d)} and \textbf{e)}, like the \textbf{modelfree} baseline, then we suspect that it overfitted on training tasks, likely indicating a reliance on memorization \citep{cobbe2019procgen}. This was also discussed in the previous chapter.

Due to the non-overlapping confidence intervals, we observe a statistically significant advantage in the generalization performance of the \Skipper{} agents throughout training, compared to the other methods. It is worth noting that the \textbf{regen} variant exhibits significant performance advantages over all other methods, including the \textbf{once} variant. This is likely because, the frequent reconstruction of the graph makes the agent less prone to being trapped in a low-quality proxy problem, and thus provides extra adaptability in novel scenarios. 

During training, it should be acknowledged that \Skipper{} variants behave less optimally than expected, despite the strong generalization on evaluation tasks. As our later ablation results show, concurring with our previous theoretical analyses, such a phenomenon is an outcome of inaccuracies both in the proxy problem and the checkpoint policy. One major symptom of an inaccurate proxy problem is that the agent would chase delusional targets. We address this behavior in the next chapter.

Performing better than the \textbf{modelfree} baseline, \LEAP{} obtains reasonable generalization performance, that is, if we are willing to waive the extra budget it needed for pretraining. In the next chapter, we show that \LEAP{} benefits largely from removing hallucinated state targets, which validates that optimizing for a path in the latent space may be prone to errors caused by delusional subgoals. Lastly, we see that the \Director{} agents suffer in these experiments despite their good performance in the single environment experimental settings reported by \citet{hafner2022deep}. One may be surprised by how badly a state-of-the-art hierarchical planning methods such as \Director{} performed. For this, we will present later additional experiments to show that \Director{} is certainly strong in the more classical mono-task experimental setting, but not strong in a multi-task, generalization-focused setting (the one we adopted in this chapter): \Director{} performs well in single environment configurations as expected, but its performance deteriorates fast with more training tasks. Such results indicate poor scalability in terms of generalization, and thus a limitation to its real-world applications.

\subsection{Scalability Studies: Number of Training Tasks}
Inspired by \citet{cobbe2019procgen}, we want to investigate the scalability of the agents' generalization abilities \wrt{} the number of training tasks. A method is favorable if it could achieve the highest generalization performance with the least amounts of training tasks.

\begin{figure}[htbp]
\centering
\subfloat[training, $\delta = 0.4$]{
\captionsetup{justification = centering}
\includegraphics[height=0.22\textwidth]{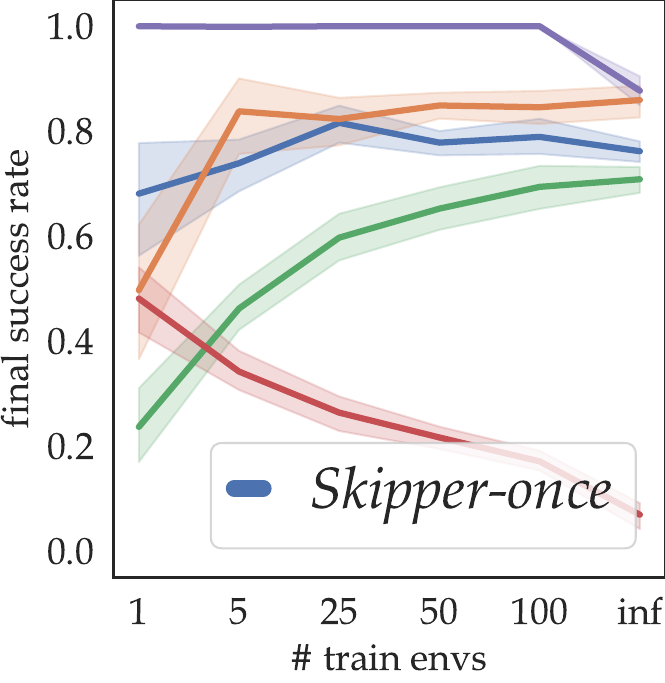}}
\hfill
\subfloat[OOD, $\delta = 0.25$]{
\captionsetup{justification = centering}
\includegraphics[height=0.22\textwidth]{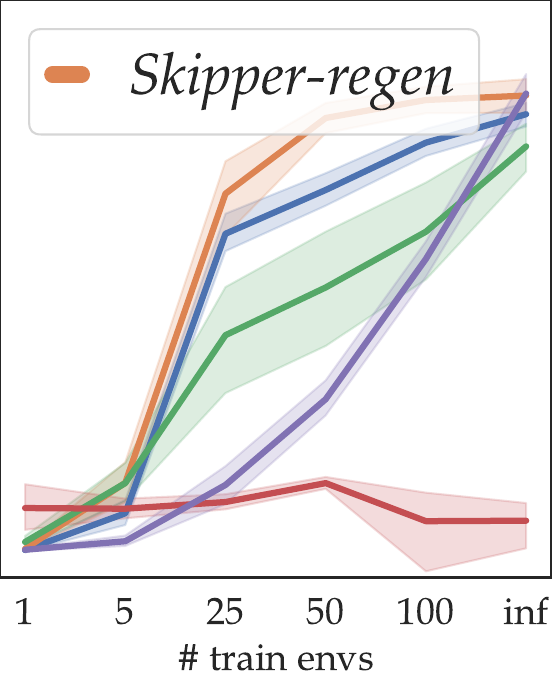}}
\hfill
\subfloat[OOD, $\delta = 0.35$]{
\captionsetup{justification = centering}
\includegraphics[height=0.22\textwidth]{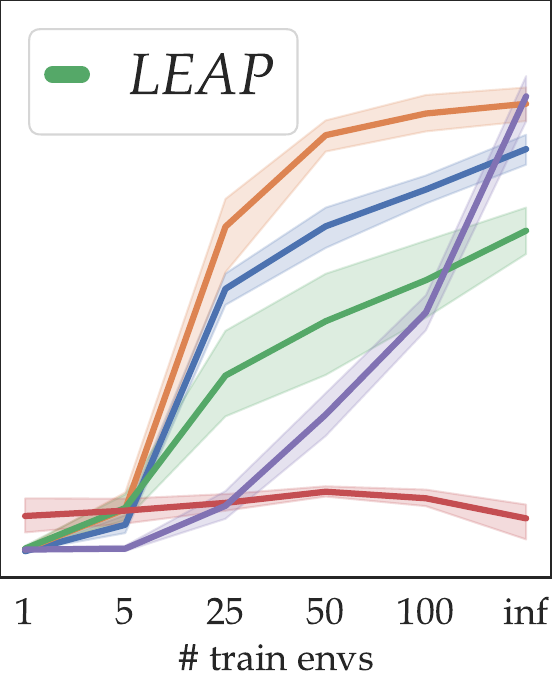}}
\hfill
\subfloat[OOD, $\delta = 0.45$]{
\captionsetup{justification = centering}
\includegraphics[height=0.22\textwidth]{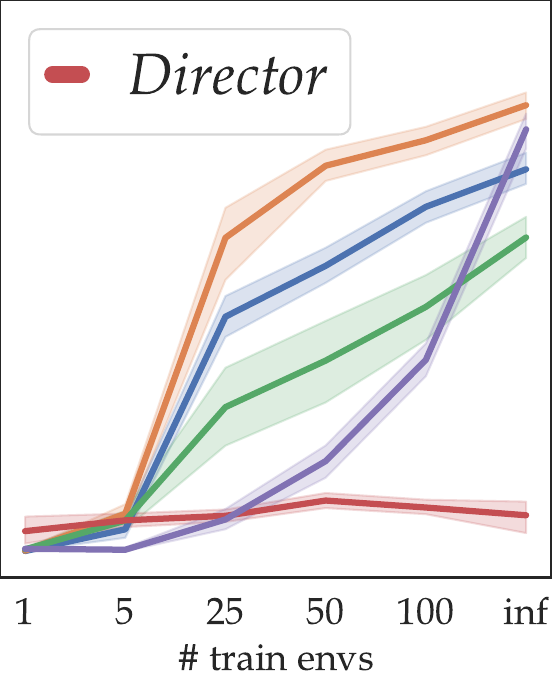}}
\hfill
\subfloat[OOD, $\delta = 0.55$]{
\captionsetup{justification = centering}
\includegraphics[height=0.22\textwidth]{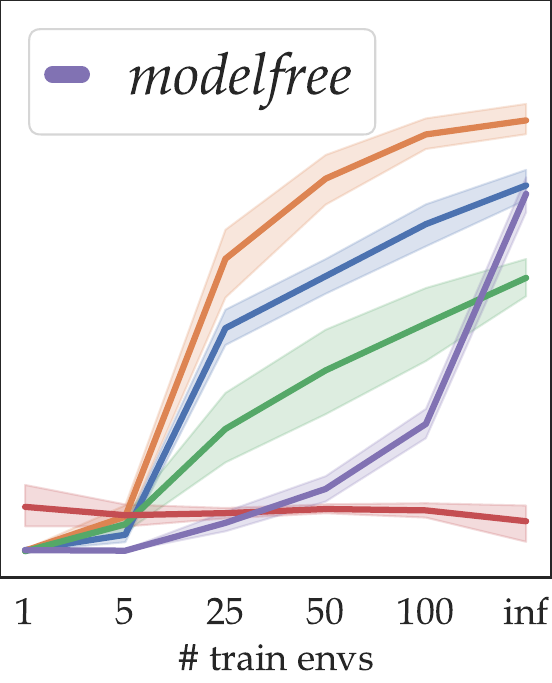}}
\caption[Evaluation Performance of Agents with Different Numbers of Training Tasks]{\textbf{Evaluation Performance of Agents with Different Numbers of Training Tasks}: each data point and corresponding error bar (95\% confidence interval) are based on the final performance from $20$ independent seed runs. Training task performance is shown in a) while OOD performance is shown in b), c), d), e). Notably, the \Skipper{} agents as well as the adapted \LEAP{} behave poorly during training when being trained on only one task. This is likely because the split of context and partial information in the checkpoint generator cannot be achieved. Training on one task invalidates the purpose of the proposed generalization-focused checkpoint generator.}
\label{fig:num_envs_all}
\end{figure}

To this end, in Fig.~\ref{fig:num_envs_all}, we present the results of the agents' final evaluation performance after training over different numbers of training tasks.

We can observe that, with few or more training tasks, \Skipper{}s and the baseline show consistent improvements in generalization performance. While both \LEAP{} and \Director{} behave similarly as in the previous $50$-task experiments. Notably, the \textbf{modelfree} baseline can reach similar performance as \Skipper{}, but only when trained on a different task in each episode, which is generally infeasible in the real world beyond simulator-based environments.

\subsubsection{\texorpdfstring{\Skipper{}}{Skipper}-once Scalability}
\begin{figure}[htbp]
\centering
\subfloat[training, $\delta = 0.4$]{
\captionsetup{justification = centering}
\includegraphics[height=0.22\textwidth]{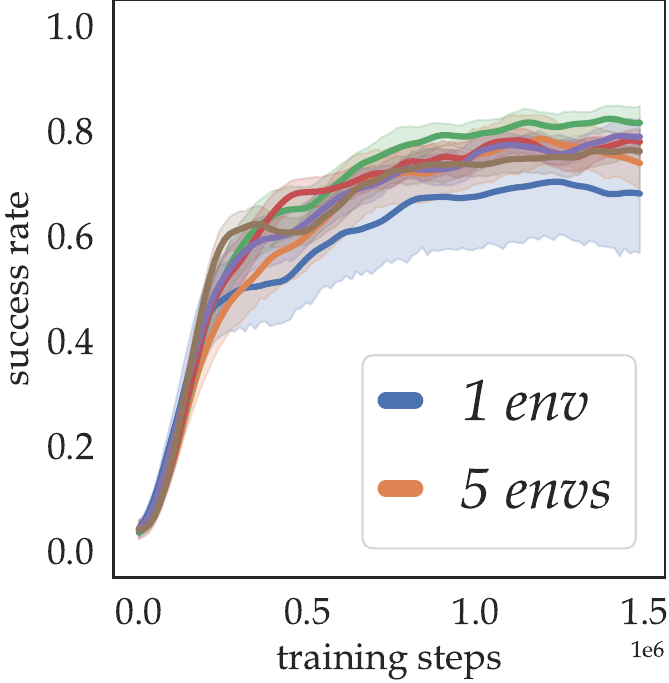}}
\hfill
\subfloat[OOD, $\delta = 0.25$]{
\captionsetup{justification = centering}
\includegraphics[height=0.22\textwidth]{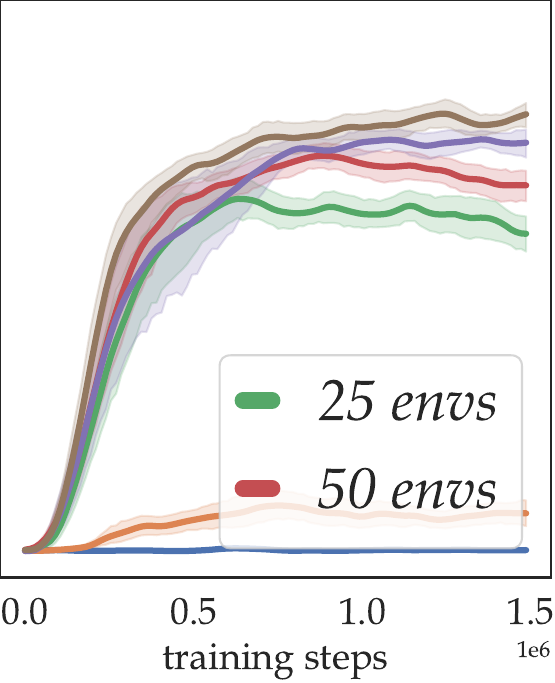}}
\hfill
\subfloat[OOD, $\delta = 0.35$]{
\captionsetup{justification = centering}
\includegraphics[height=0.22\textwidth]{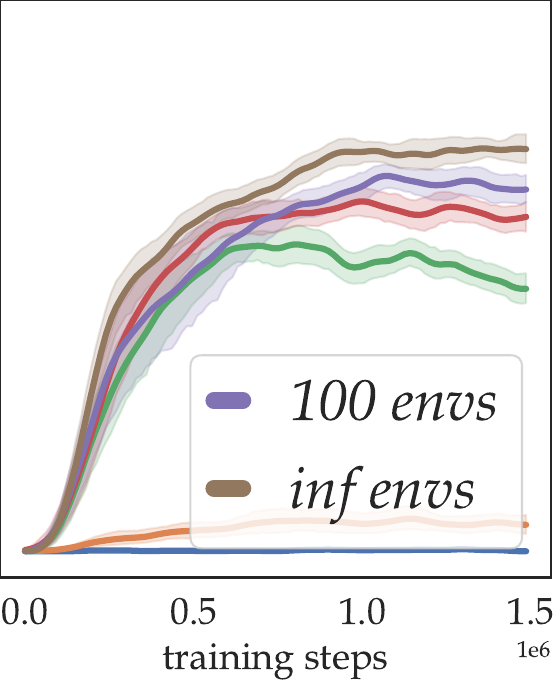}}
\hfill
\subfloat[OOD, $\delta = 0.45$]{
\captionsetup{justification = centering}
\includegraphics[height=0.22\textwidth]{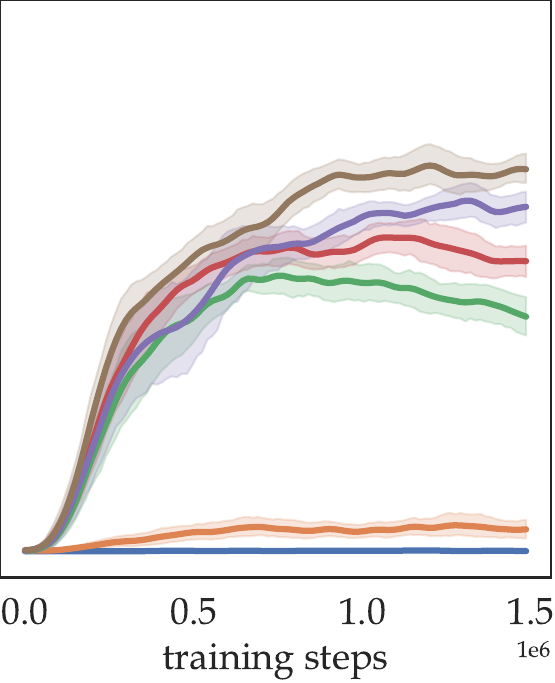}}
\hfill
\subfloat[OOD, $\delta = 0.55$]{
\captionsetup{justification = centering}
\includegraphics[height=0.22\textwidth]{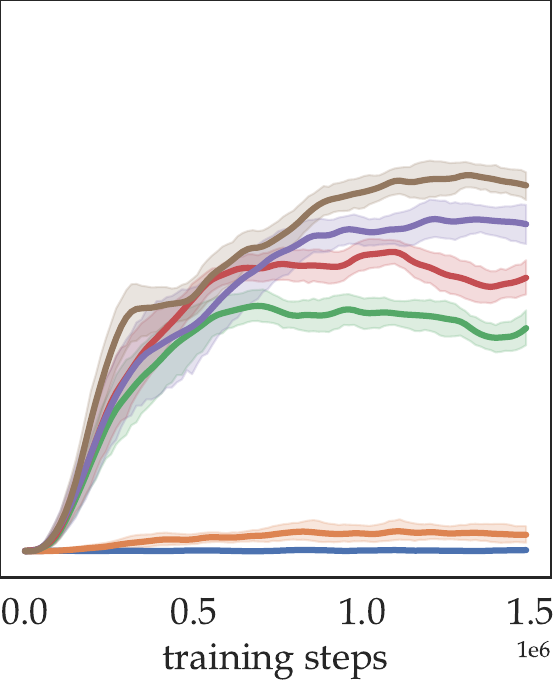}}

\caption[Evaluation Performance Evolution of \Skipper{}-once with Different Numbers of Training Tasks]{\textbf{Evaluation Performance Evolution of \Skipper{}-once with Different Numbers of Training Tasks}: each error bar (95\% CI) was obtained from $20$ seed runs.}
\label{fig:once_num_envs}
\end{figure}

We present the performance evolution (throughout training) of \textbf{\Skipper{}-once} with different numbers of training tasks, in Fig.~\ref{fig:once_num_envs}.

\subsubsection{\texorpdfstring{\Skipper{}}{Skipper}-regen Scalability}
\begin{figure}[htbp]
\centering
\subfloat[training, $\delta = 0.4$]{
\captionsetup{justification = centering}
\includegraphics[height=0.22\textwidth]{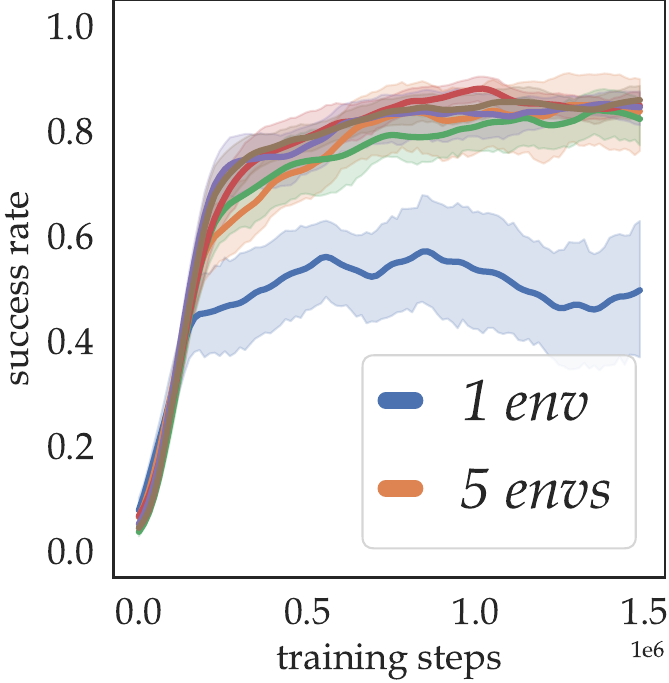}}
\hfill
\subfloat[OOD, $\delta = 0.25$]{
\captionsetup{justification = centering}
\includegraphics[height=0.22\textwidth]{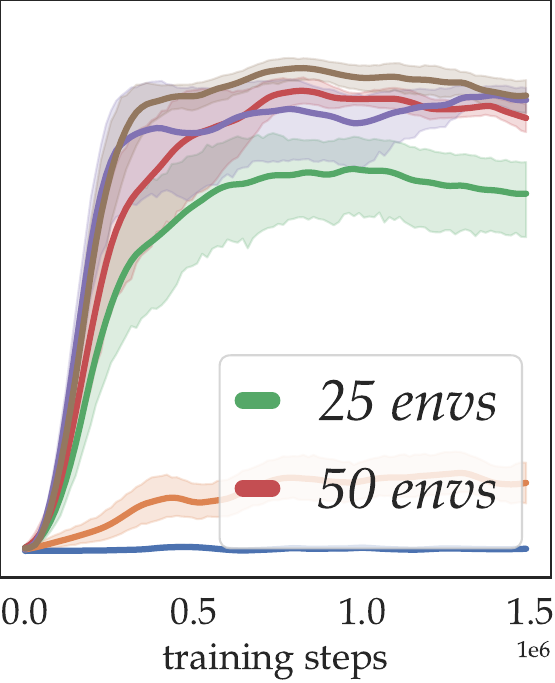}}
\hfill
\subfloat[OOD, $\delta = 0.35$]{
\captionsetup{justification = centering}
\includegraphics[height=0.22\textwidth]{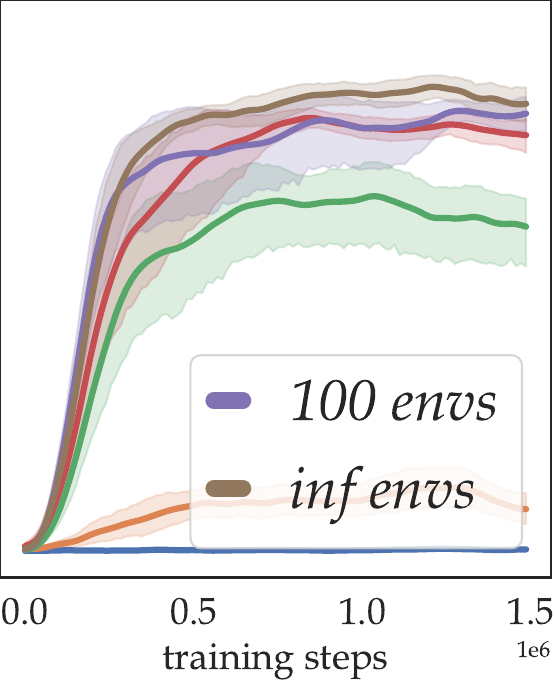}}
\hfill
\subfloat[OOD, $\delta = 0.45$]{
\captionsetup{justification = centering}
\includegraphics[height=0.22\textwidth]{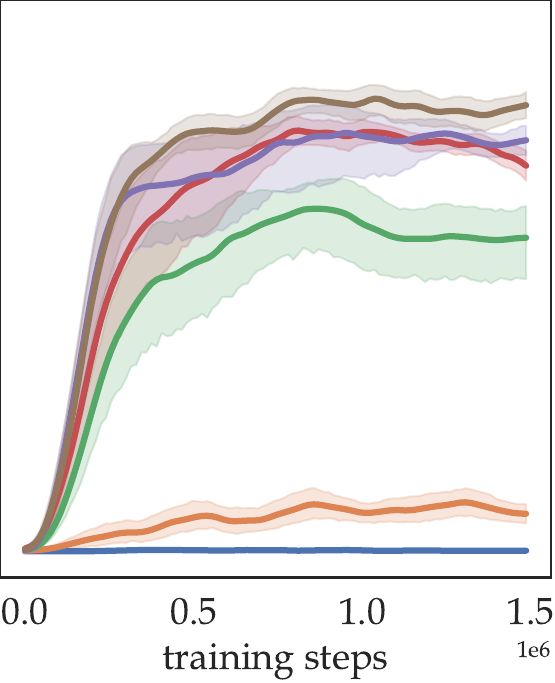}}
\hfill
\subfloat[OOD, $\delta = 0.55$]{
\captionsetup{justification = centering}
\includegraphics[height=0.22\textwidth]{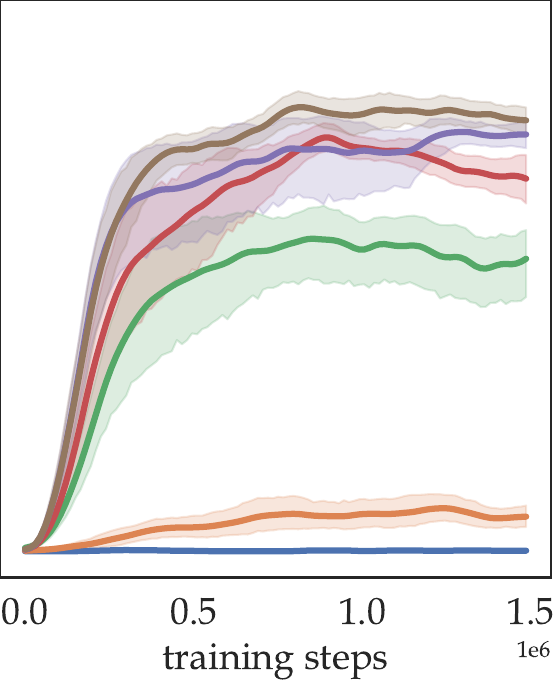}}

\caption[Performance of \Skipper{}-regen with Different Numbers of Training Tasks]{\textbf{Performance of \Skipper{}-regen with Different Numbers of Training Tasks}: each error bar (95\% CI) was obtained from $20$ seed runs.}
\label{fig:regen_num_envs}
\end{figure}

We present the performance evolution (throughout training) of \textbf{\Skipper{}-regen} with different numbers of training tasks, in Fig.~\ref{fig:regen_num_envs}.

\subsubsection{\textbf{modelfree} Scalability}
\begin{figure}[htbp]
\centering
\subfloat[training, $\delta = 0.4$]{
\captionsetup{justification = centering}
\includegraphics[height=0.22\textwidth]{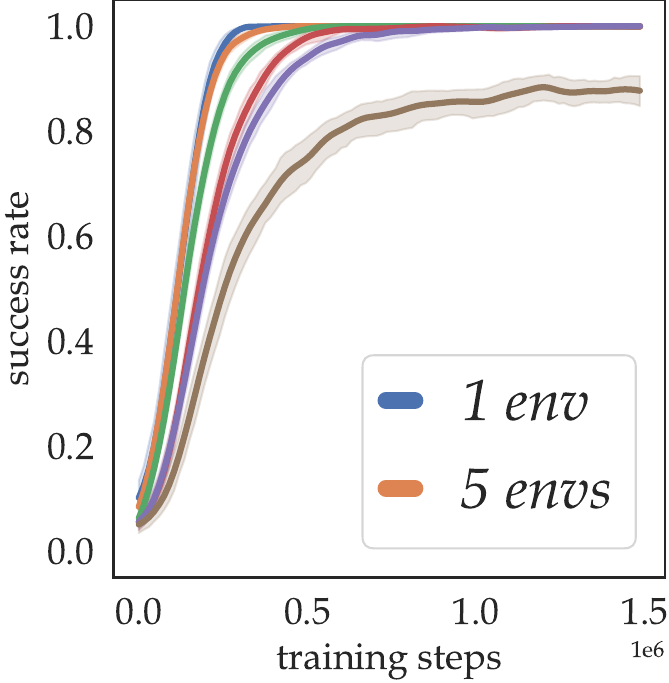}}
\hfill
\subfloat[OOD, $\delta = 0.25$]{
\captionsetup{justification = centering}
\includegraphics[height=0.22\textwidth]{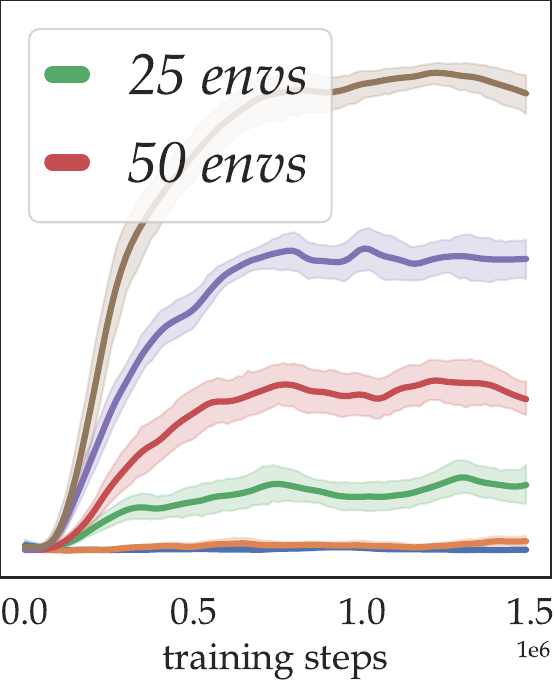}}
\hfill
\subfloat[OOD, $\delta = 0.35$]{
\captionsetup{justification = centering}
\includegraphics[height=0.22\textwidth]{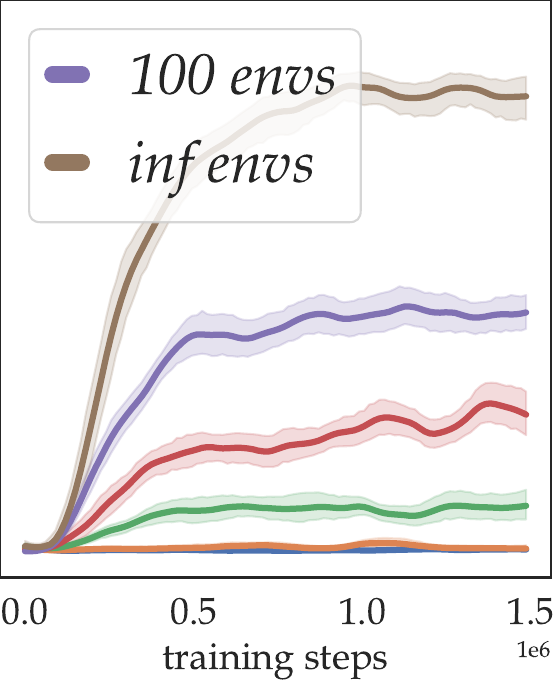}}
\hfill
\subfloat[OOD, $\delta = 0.45$]{
\captionsetup{justification = centering}
\includegraphics[height=0.22\textwidth]{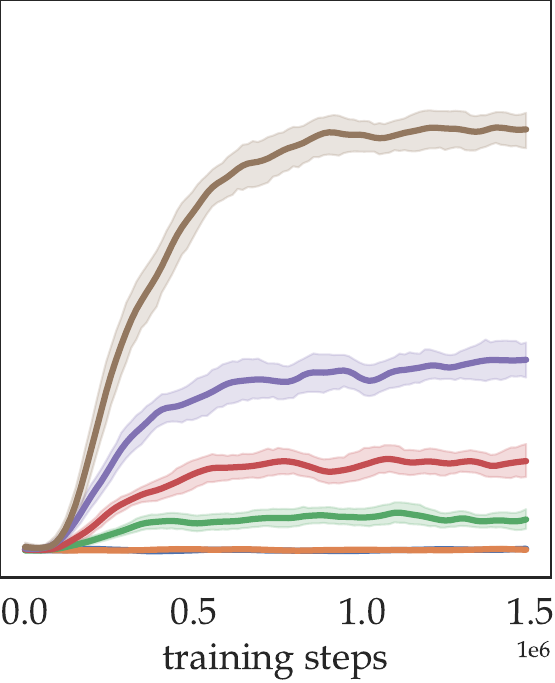}}
\hfill
\subfloat[OOD, $\delta = 0.55$]{
\captionsetup{justification = centering}
\includegraphics[height=0.22\textwidth]{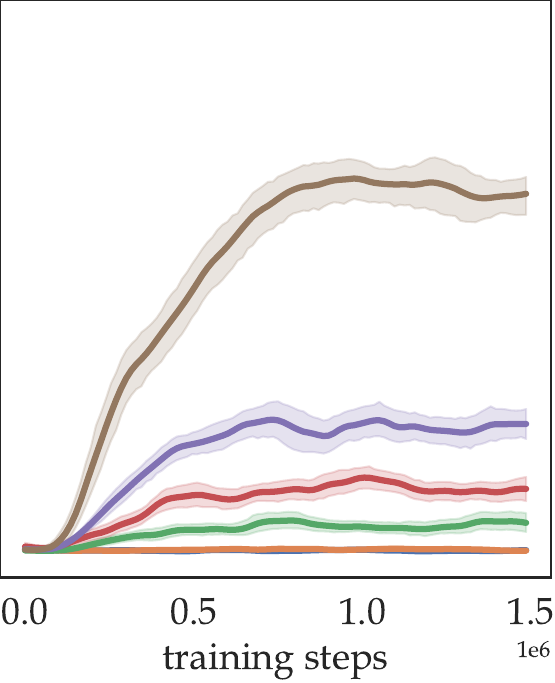}}

\caption[Evaluation Performance Evolution of \textbf{modelfree} with Different Numbers of Training Tasks]{\textbf{Evaluation Performance Evolution of \textbf{modelfree} with Different Numbers of Training Tasks}: each error bar (95\% CI) was obtained from $20$ seed runs.}
\label{fig:modelfree_num_envs}
\end{figure}

We present the performance evolution (throughout training) of the \textbf{modelfree} with different numbers of training tasks, in Fig.~\ref{fig:modelfree_num_envs}.

\subsubsection{\texorpdfstring{\LEAP{}}{LEAP} Scalability}
\begin{figure}[htbp]
\centering
\subfloat[training, $\delta = 0.4$]{
\captionsetup{justification = centering}
\includegraphics[height=0.22\textwidth]{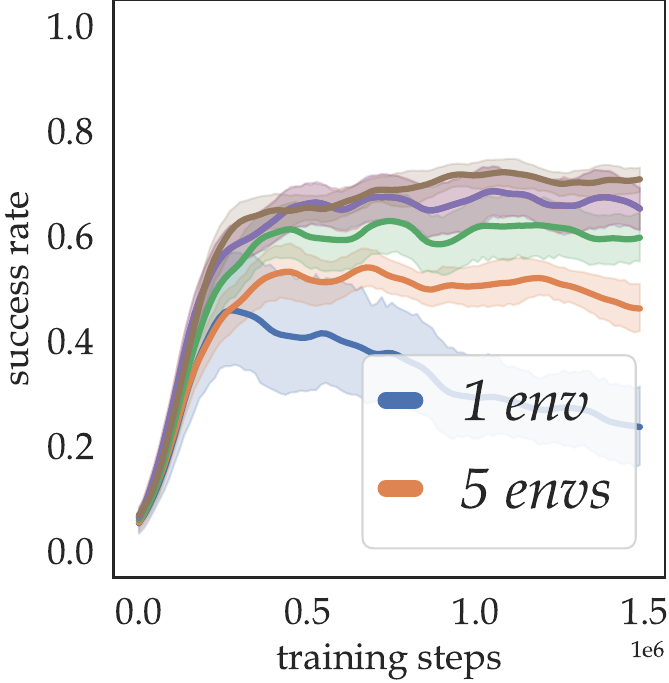}}
\hfill
\subfloat[OOD, $\delta = 0.25$]{
\captionsetup{justification = centering}
\includegraphics[height=0.22\textwidth]{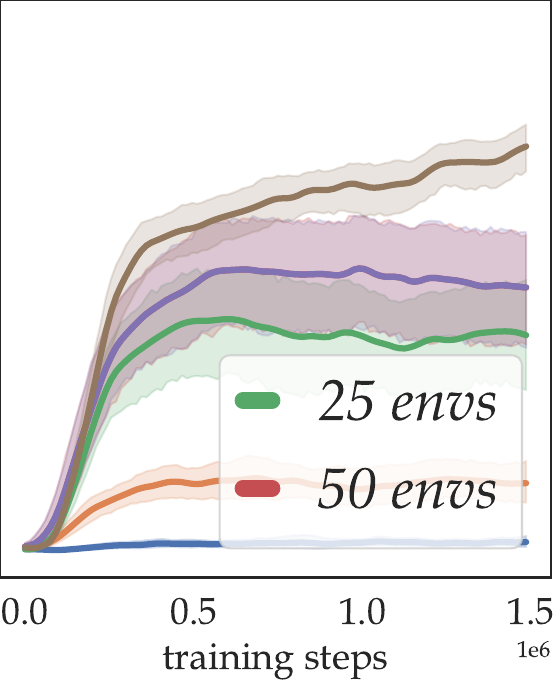}}
\hfill
\subfloat[OOD, $\delta = 0.35$]{
\captionsetup{justification = centering}
\includegraphics[height=0.22\textwidth]{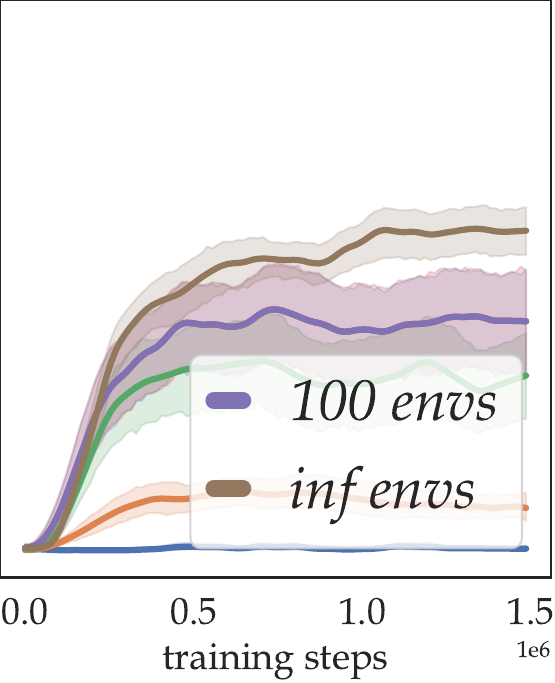}}
\hfill
\subfloat[OOD, $\delta = 0.45$]{
\captionsetup{justification = centering}
\includegraphics[height=0.22\textwidth]{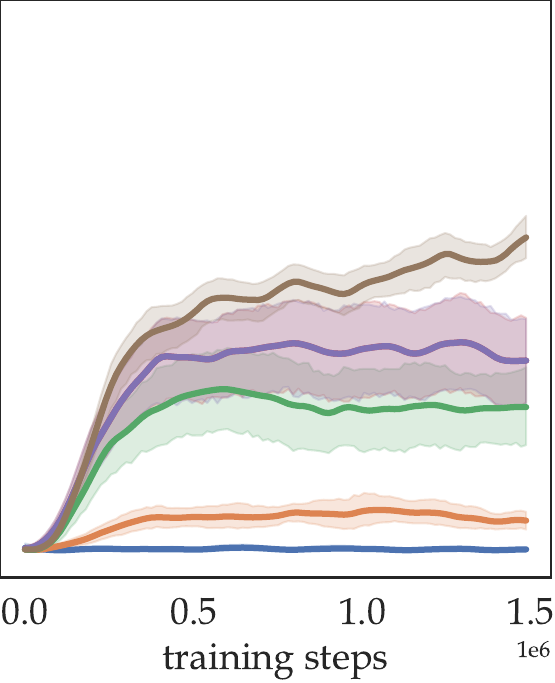}}
\hfill
\subfloat[OOD, $\delta = 0.55$]{
\captionsetup{justification = centering}
\includegraphics[height=0.22\textwidth]{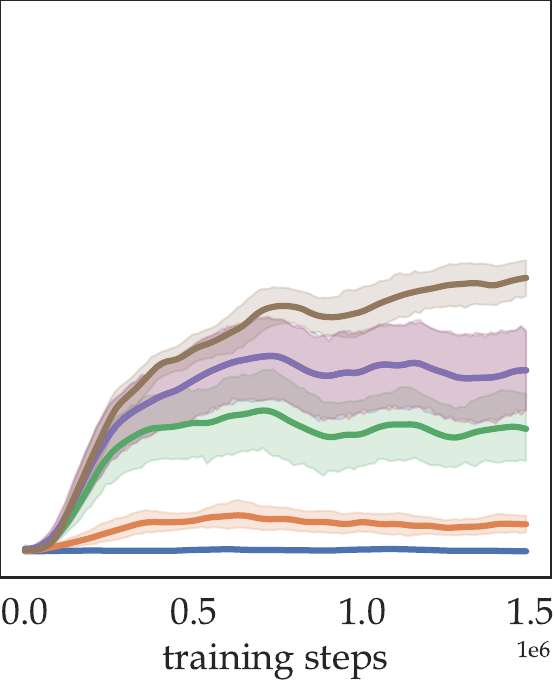}}

\caption[Evaluation Performance Evolution of \LEAP{} with Different Numbers of Training Tasks]{\textbf{Evaluation Performance Evolution of \LEAP{} with Different Numbers of Training Tasks}: each error bar (95\% CI) was obtained from $20$ seed runs.}
\label{fig:leap_num_envs}
\end{figure}

We present the performance evolution (throughout training) of the adapted \LEAP{} baseline with different numbers of training tasks, in Fig.~\ref{fig:leap_num_envs}.

\subsubsection{\texorpdfstring{\Director{}}{Director} Scalability}
\begin{figure}[htbp]
\centering
\subfloat[training, $\delta = 0.4$]{
\captionsetup{justification = centering}
\includegraphics[height=0.22\textwidth]{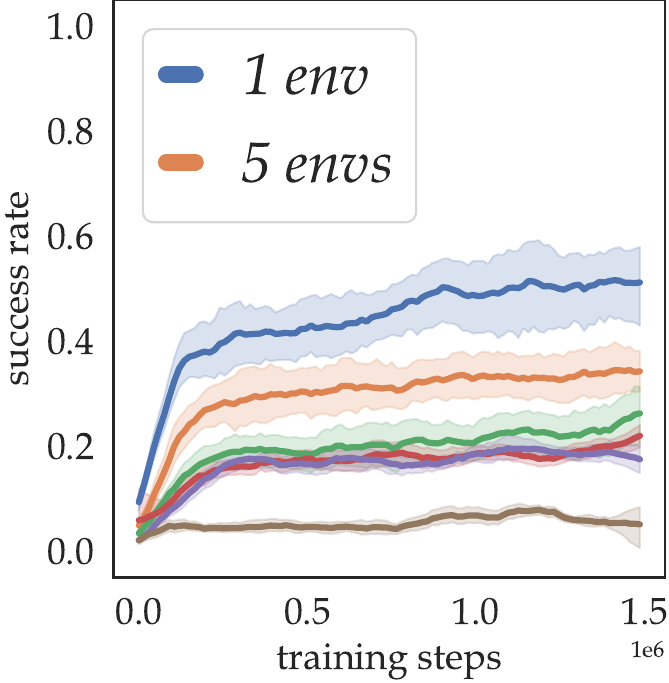}}
\hfill
\subfloat[OOD, $\delta = 0.25$]{
\captionsetup{justification = centering}
\includegraphics[height=0.22\textwidth]{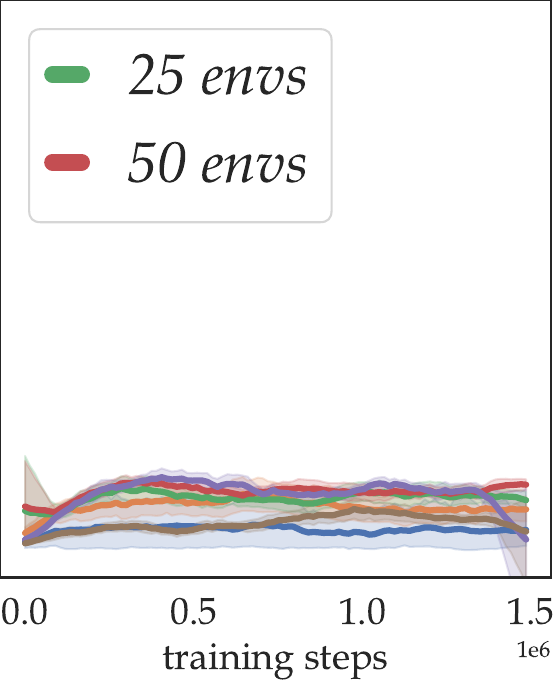}}
\hfill
\subfloat[OOD, $\delta = 0.35$]{
\captionsetup{justification = centering}
\includegraphics[height=0.22\textwidth]{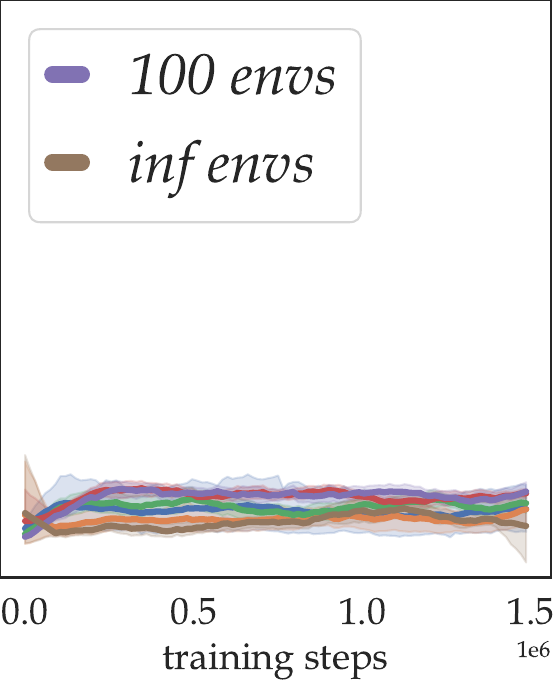}}
\hfill
\subfloat[OOD, $\delta = 0.45$]{
\captionsetup{justification = centering}
\includegraphics[height=0.22\textwidth]{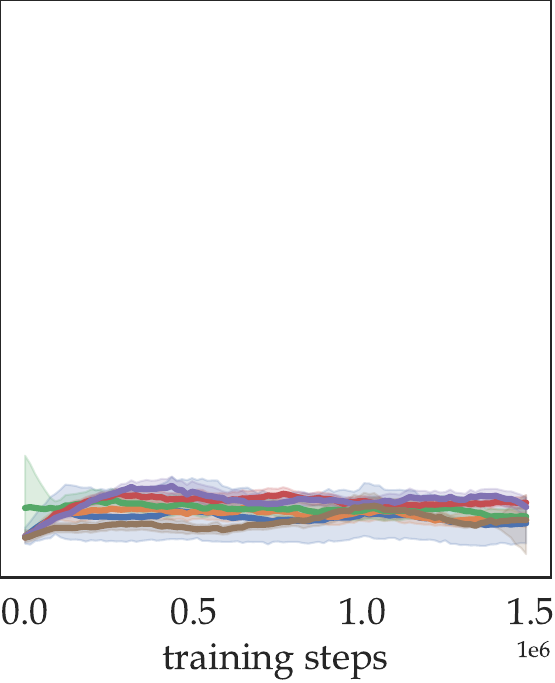}}
\hfill
\subfloat[OOD, $\delta = 0.55$]{
\captionsetup{justification = centering}
\includegraphics[height=0.22\textwidth]{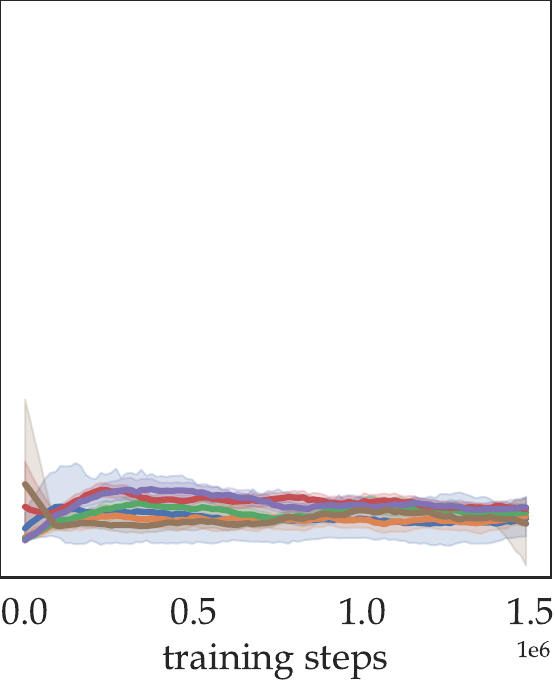}}

\caption[Evaluation Performance Evolution of \Director{} with Different Numbers of Training Tasks]{\textbf{Evaluation Performance Evolution of \Director{} with Different Numbers of Training Tasks}: each error bar (95\% CI) was obtained from $20$ seed runs.}
\label{fig:director_num_envs}
\end{figure}

We present the performance evolution (throughout training) of the adapted \Director{} baseline on different numbers of training tasks, in Fig.~\ref{fig:director_num_envs}.

\subsubsection{Detailed OOD Performance with Different Numbers of Training Tasks}
\begin{figure}[htbp]
\centering
\subfloat[training, $\delta = 0.4$]{
\captionsetup{justification = centering}
\includegraphics[height=0.22\textwidth]{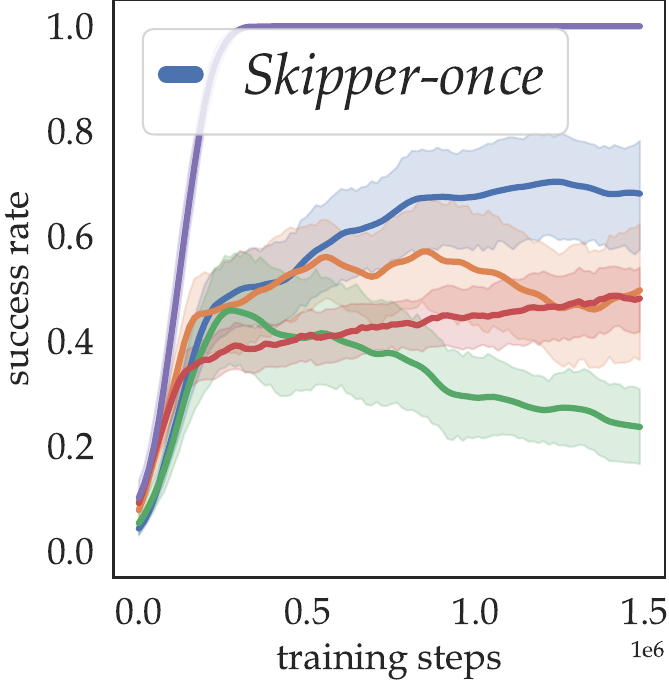}}
\hfill
\subfloat[OOD, $\delta = 0.25$]{
\captionsetup{justification = centering}
\includegraphics[height=0.22\textwidth]{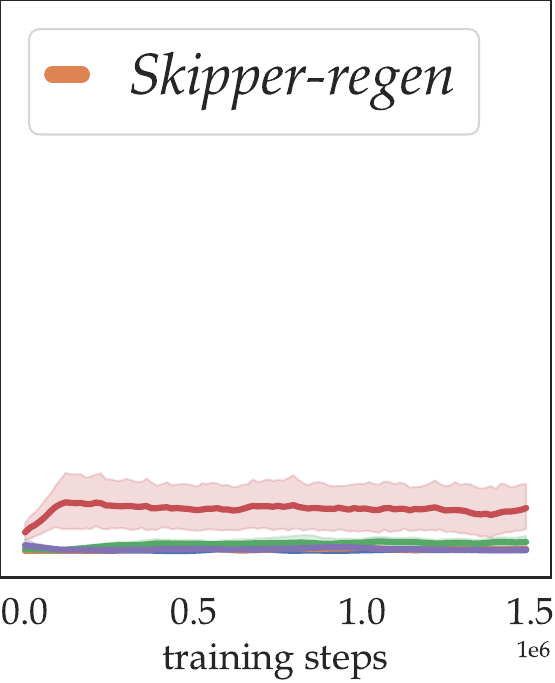}}
\hfill
\subfloat[OOD, $\delta = 0.35$]{
\captionsetup{justification = centering}
\includegraphics[height=0.22\textwidth]{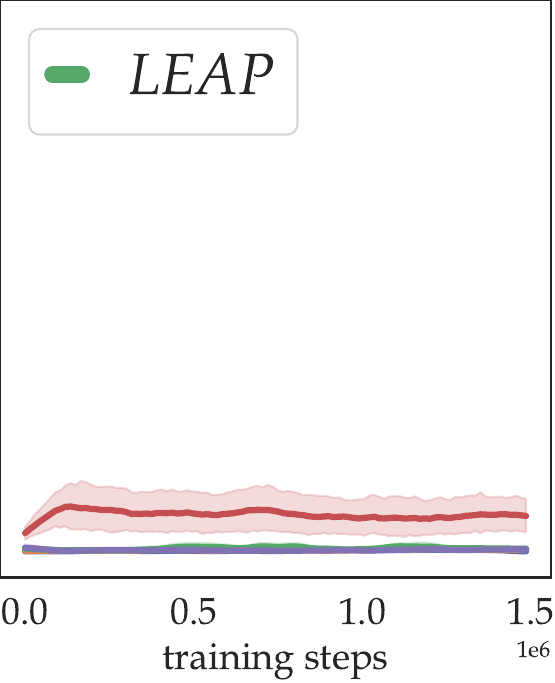}}
\hfill
\subfloat[OOD, $\delta = 0.45$]{
\captionsetup{justification = centering}
\includegraphics[height=0.22\textwidth]{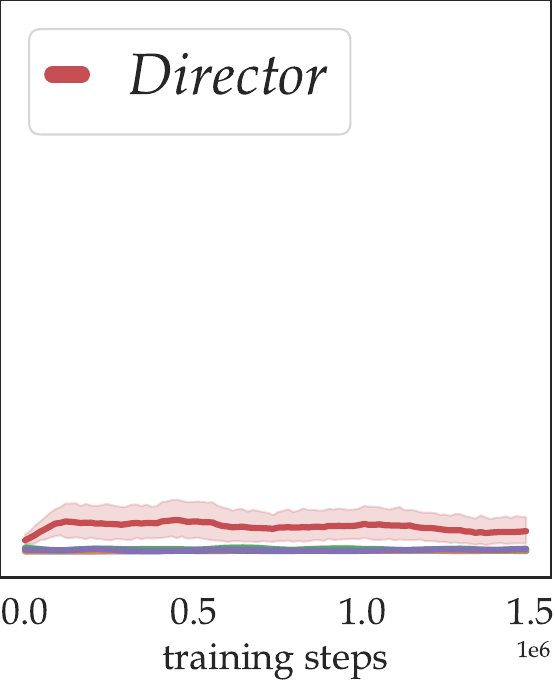}}
\hfill
\subfloat[OOD, $\delta = 0.55$]{
\captionsetup{justification = centering}
\includegraphics[height=0.22\textwidth]{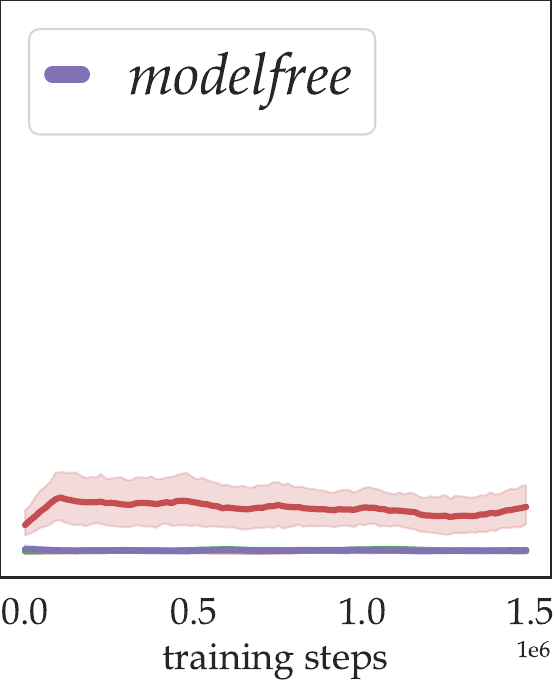}}

\caption[Evaluation Performance of Agents Trained on $1$ Task]{\textbf{Evaluation Performance of Agents Trained on $1$ Task}: each error bar (95\% CI) was obtained from $20$ seed runs.}
\label{fig:1_envs}
\end{figure}

\begin{figure}[htbp]
\centering
\subfloat[training, $\delta = 0.4$]{
\captionsetup{justification = centering}
\includegraphics[height=0.22\textwidth]{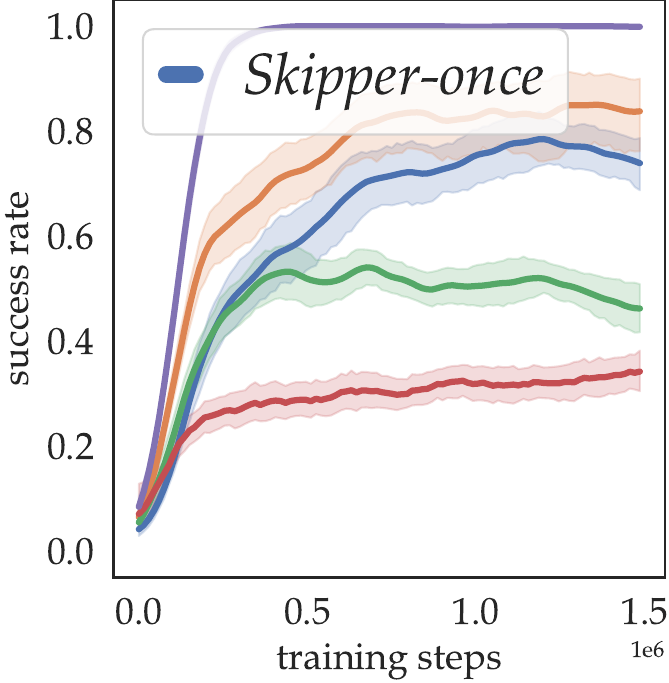}}
\hfill
\subfloat[OOD, $\delta = 0.25$]{
\captionsetup{justification = centering}
\includegraphics[height=0.22\textwidth]{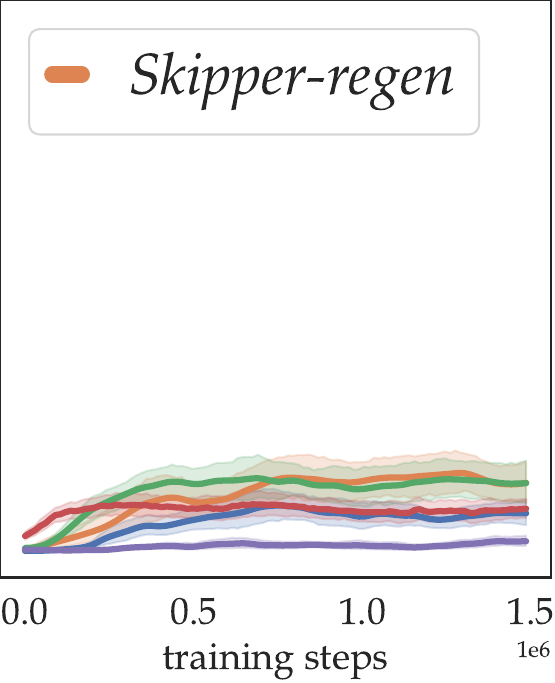}}
\hfill
\subfloat[OOD, $\delta = 0.35$]{
\captionsetup{justification = centering}
\includegraphics[height=0.22\textwidth]{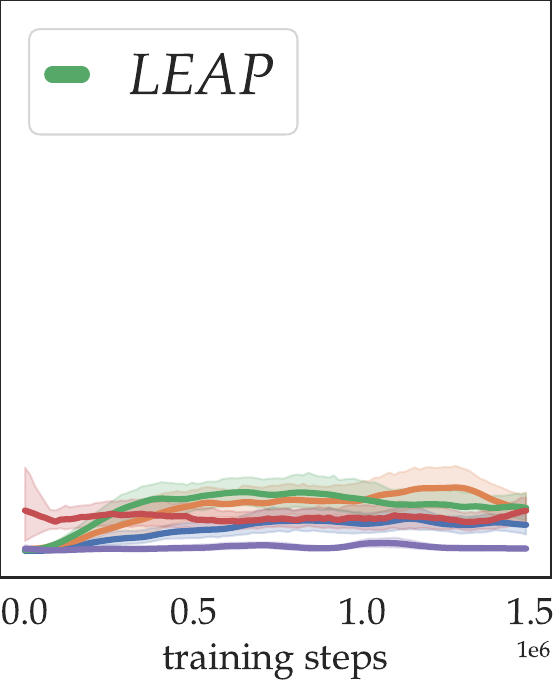}}
\hfill
\subfloat[OOD, $\delta = 0.45$]{
\captionsetup{justification = centering}
\includegraphics[height=0.22\textwidth]{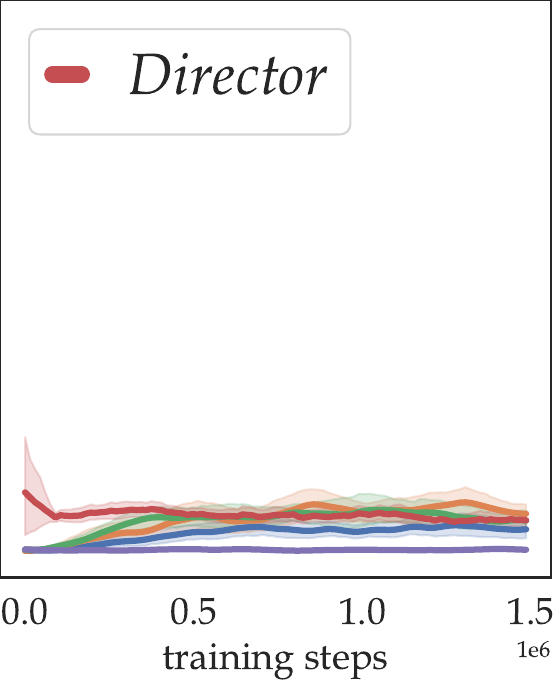}}
\hfill
\subfloat[OOD, $\delta = 0.55$]{
\captionsetup{justification = centering}
\includegraphics[height=0.22\textwidth]{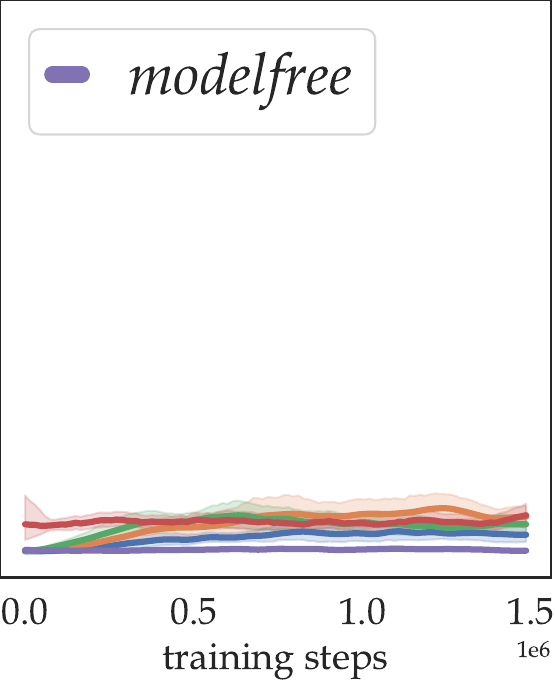}}

\caption[Evaluation Performance of Agents Trained on $5$ Tasks]{\textbf{Evaluation Performance of Agents Trained on $5$ Tasks}: each error bar (95\% CI) was obtained from $20$ seed runs.}
\label{fig:5_envs}
\end{figure}

\begin{figure}[htbp]
\centering
\subfloat[training, $\delta = 0.4$]{
\captionsetup{justification = centering}
\includegraphics[height=0.22\textwidth]{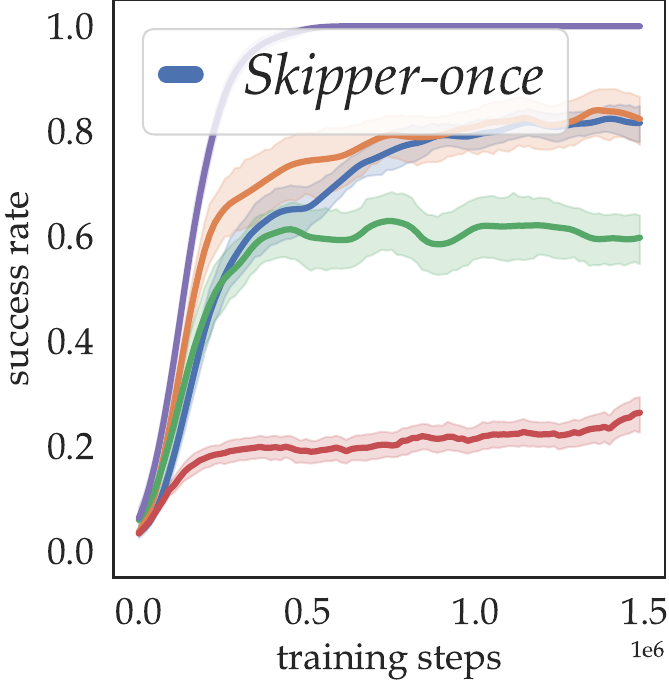}}
\hfill
\subfloat[OOD, $\delta = 0.25$]{
\captionsetup{justification = centering}
\includegraphics[height=0.22\textwidth]{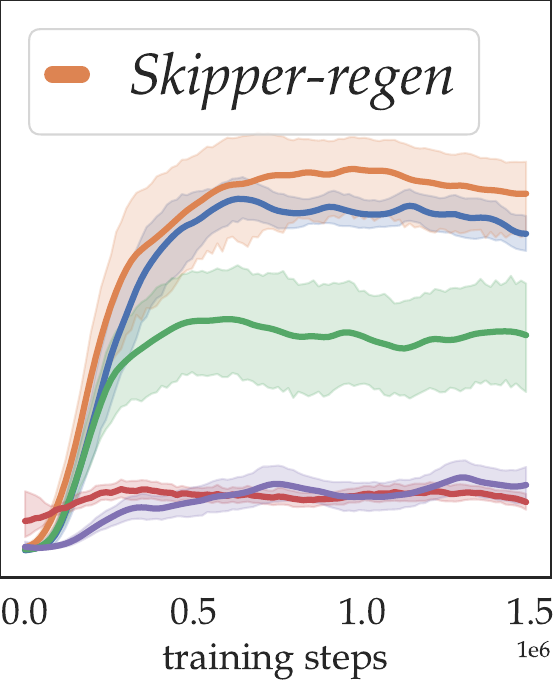}}
\hfill
\subfloat[OOD, $\delta = 0.35$]{
\captionsetup{justification = centering}
\includegraphics[height=0.22\textwidth]{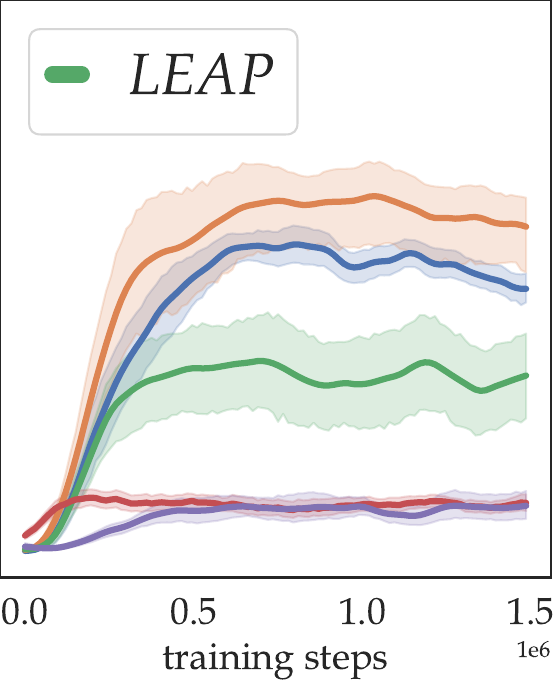}}
\hfill
\subfloat[OOD, $\delta = 0.45$]{
\captionsetup{justification = centering}
\includegraphics[height=0.22\textwidth]{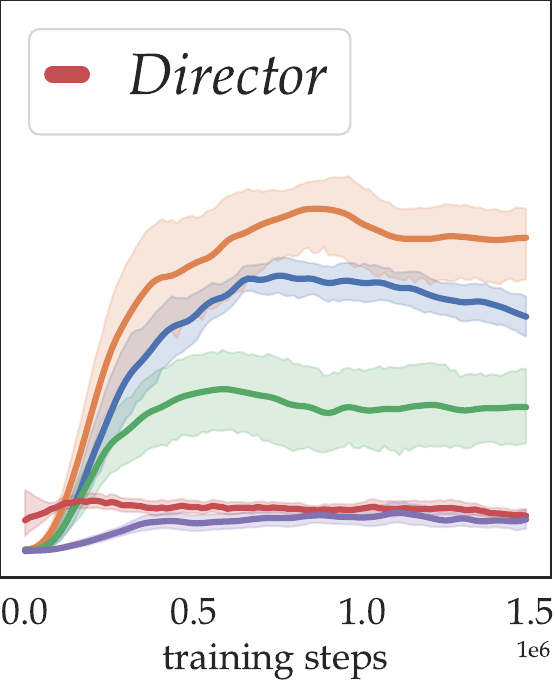}}
\hfill
\subfloat[OOD, $\delta = 0.55$]{
\captionsetup{justification = centering}
\includegraphics[height=0.22\textwidth]{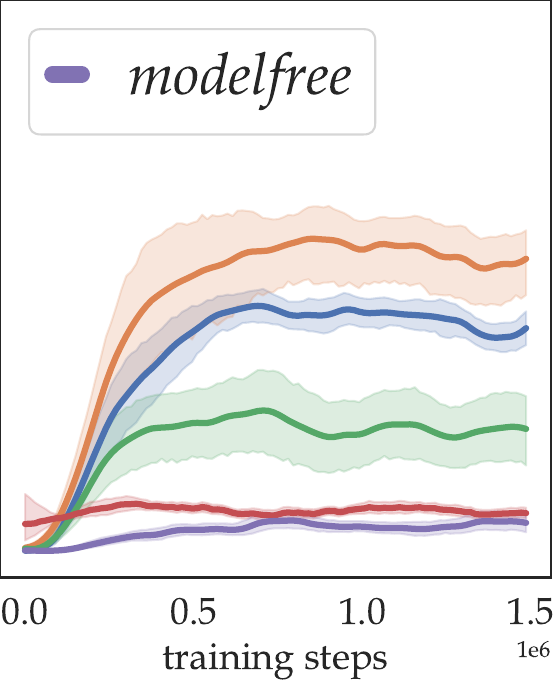}}

\caption[Evaluation Performance of Agents Trained on $25$ Tasks]{\textbf{Evaluation Performance of Agents Trained on $25$ Tasks}: each error bar (95\% CI) was obtained from $20$ seed runs.}
\label{fig:25_envs}
\end{figure}

\begin{figure}[htbp]
\centering
\subfloat[training, $\delta = 0.4$]{
\captionsetup{justification = centering}
\includegraphics[height=0.22\textwidth]{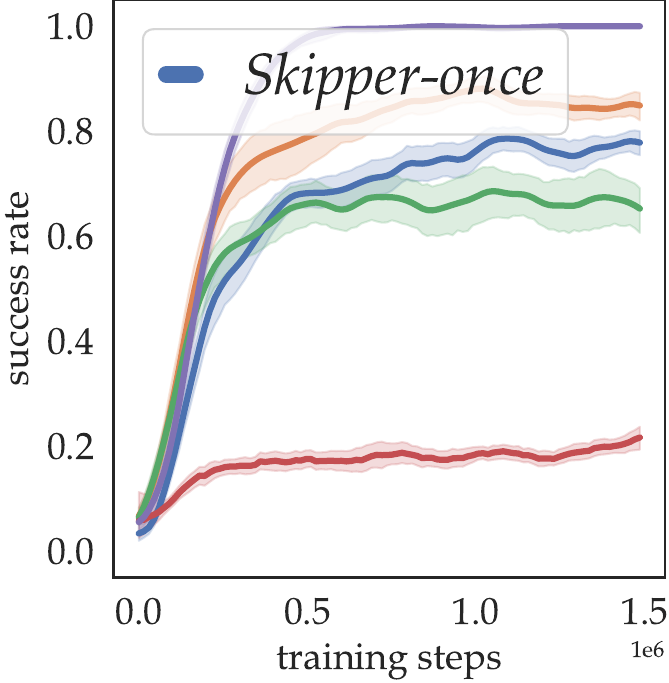}}
\hfill
\subfloat[OOD, $\delta = 0.25$]{
\captionsetup{justification = centering}
\includegraphics[height=0.22\textwidth]{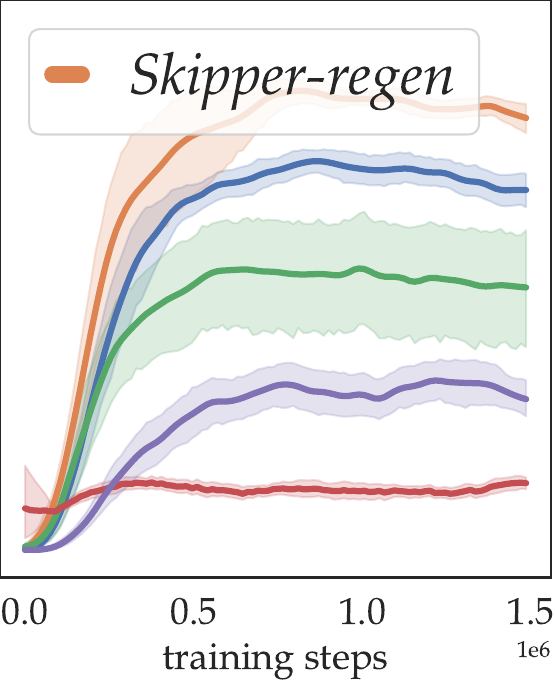}}
\hfill
\subfloat[OOD, $\delta = 0.35$]{
\captionsetup{justification = centering}
\includegraphics[height=0.22\textwidth]{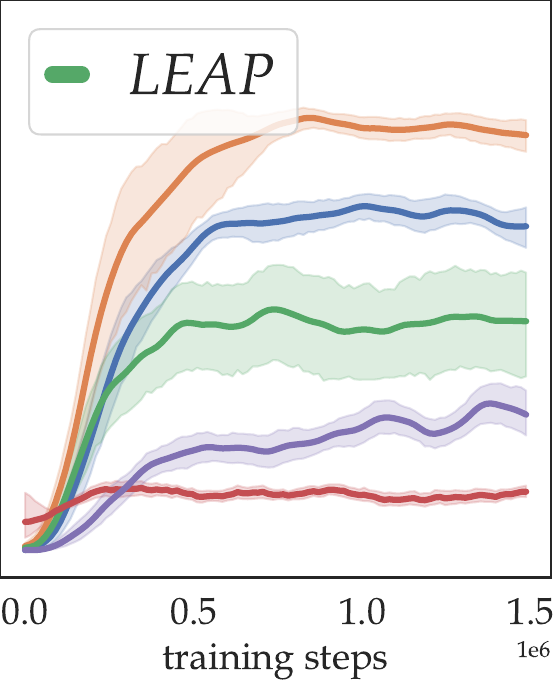}}
\hfill
\subfloat[OOD, $\delta = 0.45$]{
\captionsetup{justification = centering}
\includegraphics[height=0.22\textwidth]{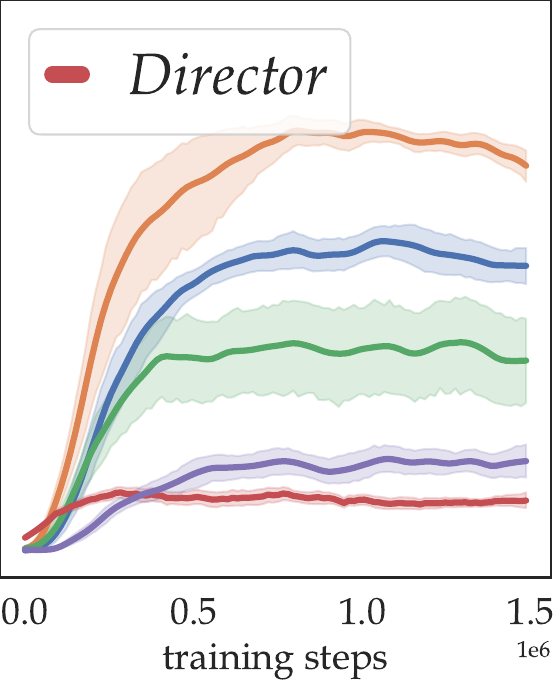}}
\hfill
\subfloat[OOD, $\delta = 0.55$]{
\captionsetup{justification = centering}
\includegraphics[height=0.22\textwidth]{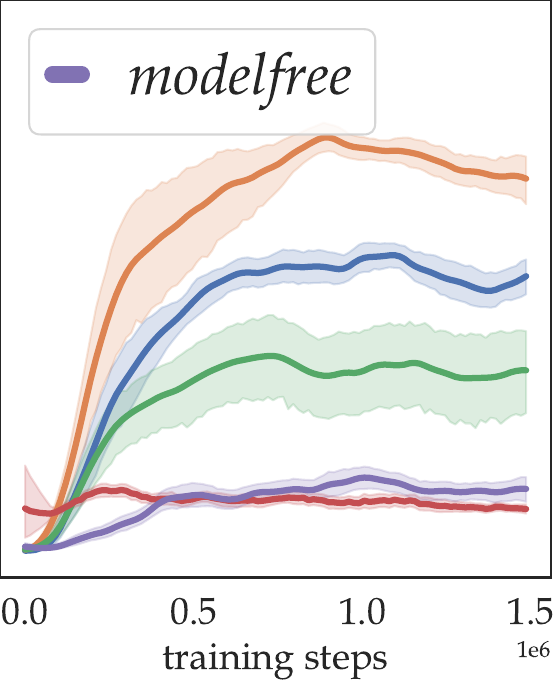}}

\caption[Evaluation Performance of Agents Trained on $50$ Tasks]{\textbf{Evaluation Performance of Agents Trained on $50$ Tasks}: each error bar (95\% CI) was obtained from $20$ seed runs.}
\label{fig:50_envs_app}
\end{figure}

\begin{figure}[htbp]
\centering
\subfloat[training, $\delta = 0.4$]{
\captionsetup{justification = centering}
\includegraphics[height=0.22\textwidth]{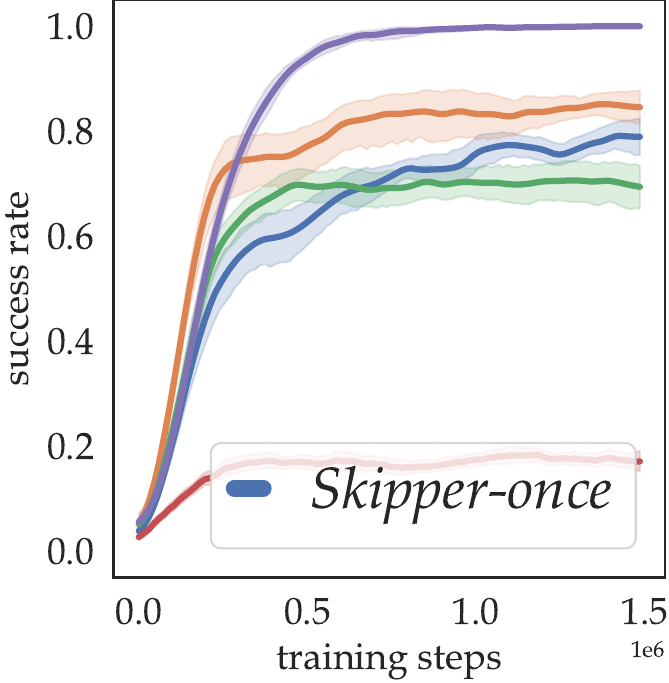}}
\hfill
\subfloat[OOD, $\delta = 0.25$]{
\captionsetup{justification = centering}
\includegraphics[height=0.22\textwidth]{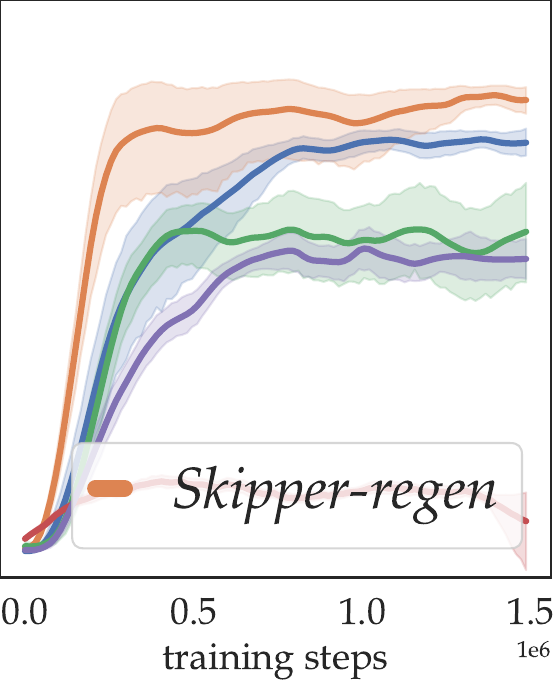}}
\hfill
\subfloat[OOD, $\delta = 0.35$]{
\captionsetup{justification = centering}
\includegraphics[height=0.22\textwidth]{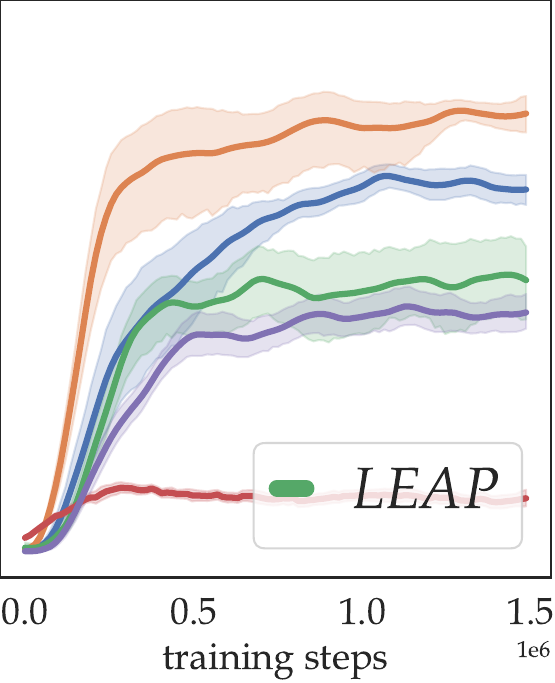}}
\hfill
\subfloat[OOD, $\delta = 0.45$]{
\captionsetup{justification = centering}
\includegraphics[height=0.22\textwidth]{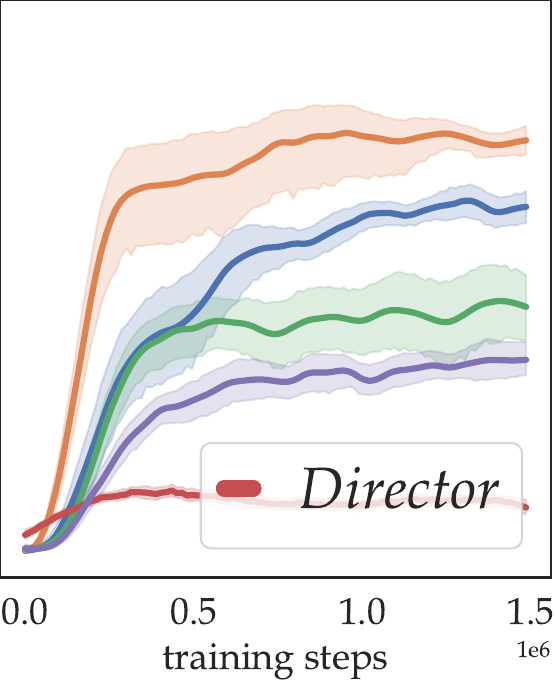}}
\hfill
\subfloat[OOD, $\delta = 0.55$]{
\captionsetup{justification = centering}
\includegraphics[height=0.22\textwidth]{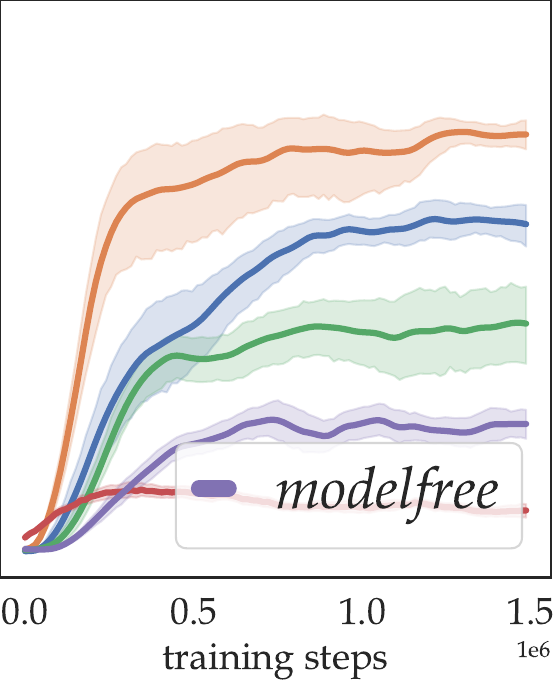}}

\caption[Evaluation Performance of Agents Trained on $100$ Tasks]{\textbf{Evaluation Performance of Agents Trained on $100$ Tasks}: each error bar (95\% CI) was obtained from $20$ seed runs.}
\label{fig:100_envs}
\end{figure}

\begin{figure}[htbp]
\centering
\subfloat[training, $\delta = 0.4$]{
\captionsetup{justification = centering}
\includegraphics[height=0.22\textwidth]{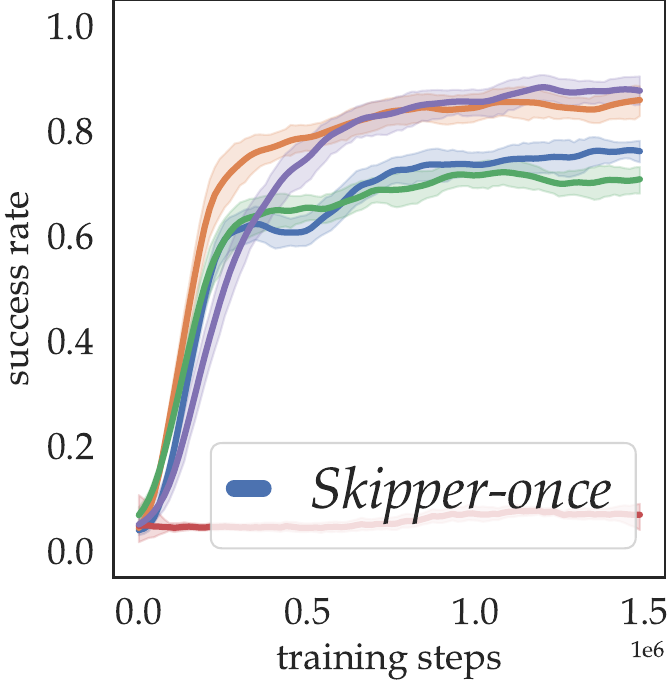}}
\hfill
\subfloat[OOD, $\delta = 0.25$]{
\captionsetup{justification = centering}
\includegraphics[height=0.22\textwidth]{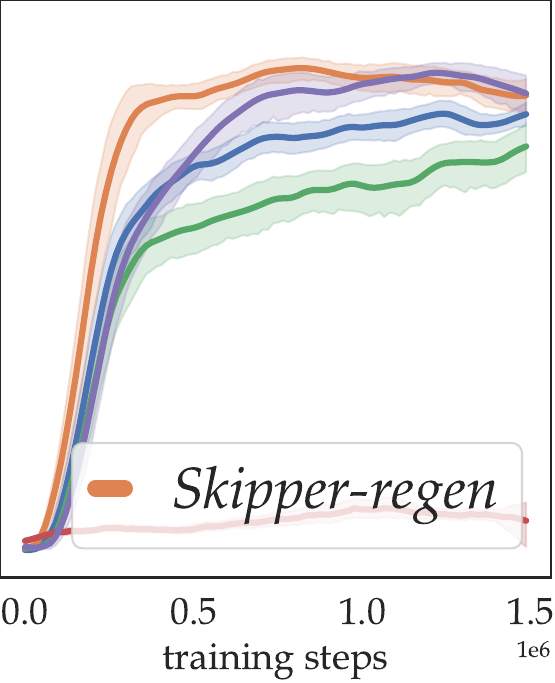}}
\hfill
\subfloat[OOD, $\delta = 0.35$]{
\captionsetup{justification = centering}
\includegraphics[height=0.22\textwidth]{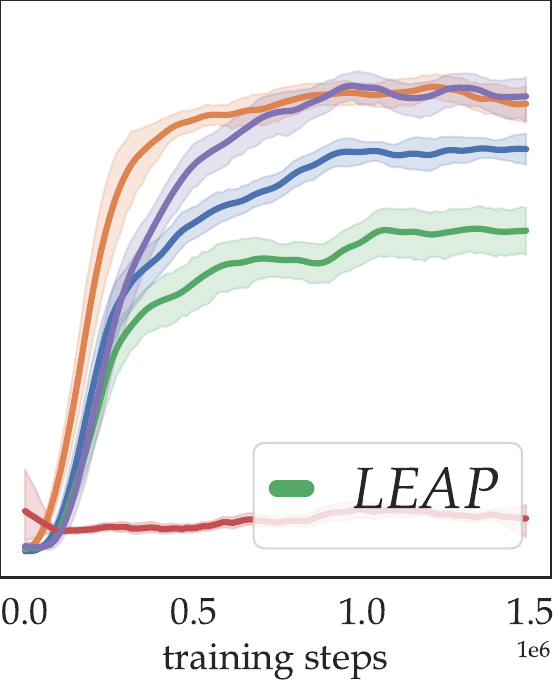}}
\hfill
\subfloat[OOD, $\delta = 0.45$]{
\captionsetup{justification = centering}
\includegraphics[height=0.22\textwidth]{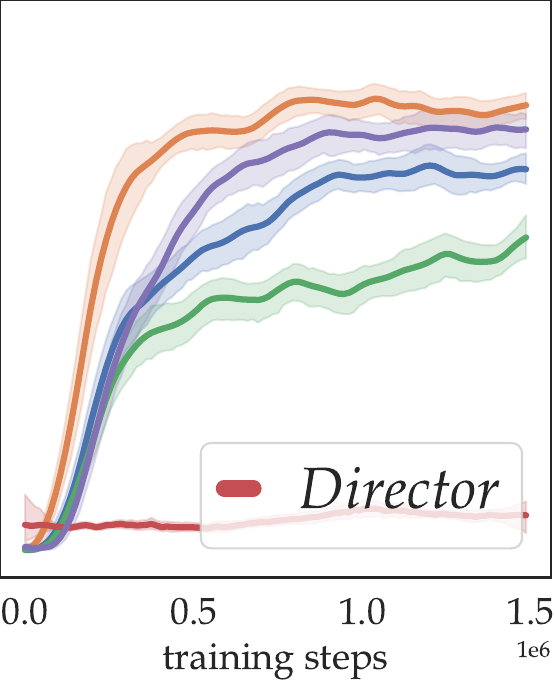}}
\hfill
\subfloat[OOD, $\delta = 0.55$]{
\captionsetup{justification = centering}
\includegraphics[height=0.22\textwidth]{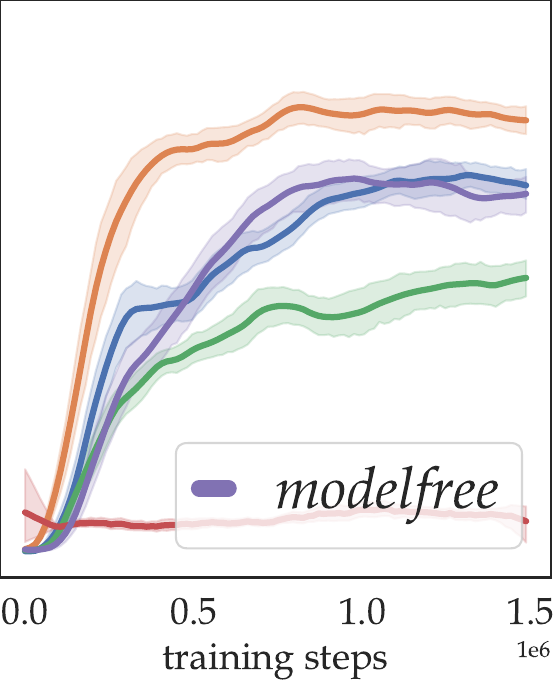}}

\caption[Evaluation Performance of Agents Trained on $\infty$ Tasks]{\textbf{Evaluation Performance of Agents Trained on $\infty$ Tasks (a new task each training episode)}: each error bar (95\% CI) was obtained from $20$ seed runs.}
\label{fig:inf_envs}
\end{figure}

The performance of all agents on all training configurations, \ie{}, different numbers of training tasks, are presented in Fig.~\ref{fig:1_envs}, Fig.~\ref{fig:5_envs}, Fig.~\ref{fig:25_envs}, Fig.~\ref{fig:50_envs_app}, Fig.~\ref{fig:100_envs} \& Fig.~\ref{fig:inf_envs}.

\subsection{Validation of Claims}
This sub-section focuses on validating the two important claims of advantages brought by the \Skipper{} framework: 1) this framework is compatible with stochasticity and 2) empirically validate if the agent's performance is as described as the theorem.

\subsubsection{Validation of Effectiveness on Stochastic Environments}
\begin{figure}[htbp]
\centering
\subfloat[training, $\delta = 0.4$]{
\captionsetup{justification = centering}
\includegraphics[height=0.22\textwidth]{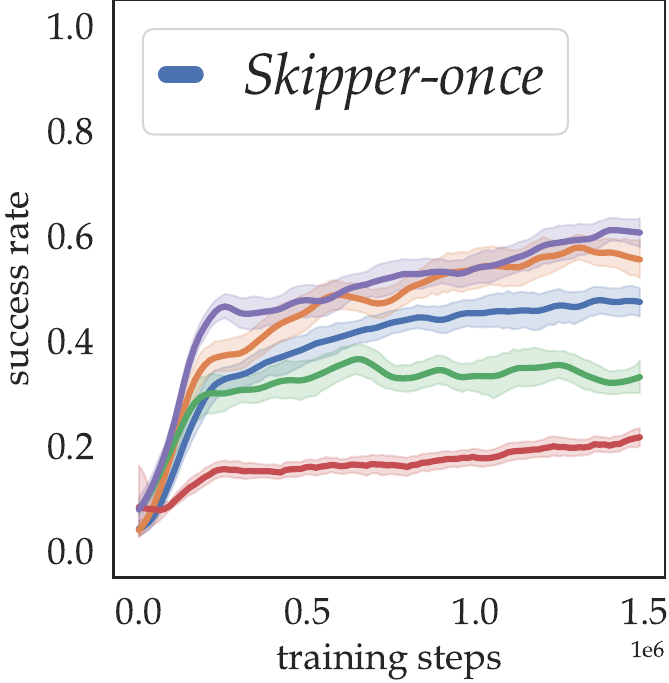}}
\hfill
\subfloat[OOD, $\delta = 0.25$]{
\captionsetup{justification = centering}
\includegraphics[height=0.22\textwidth]{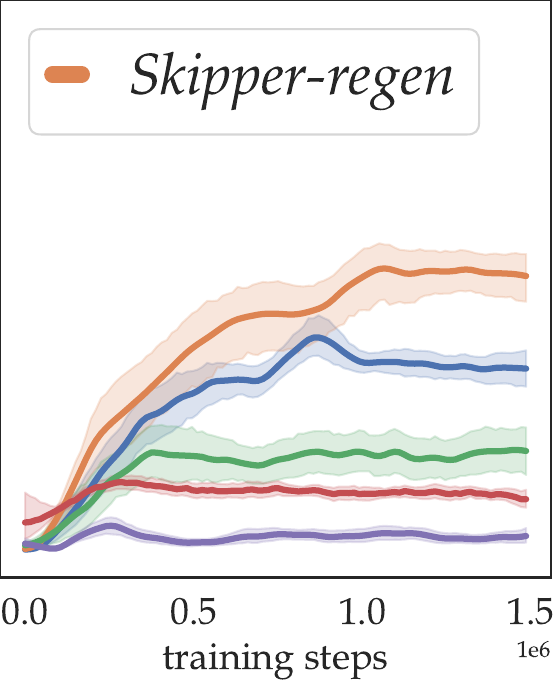}}
\hfill
\subfloat[OOD, $\delta = 0.35$]{
\captionsetup{justification = centering}
\includegraphics[height=0.22\textwidth]{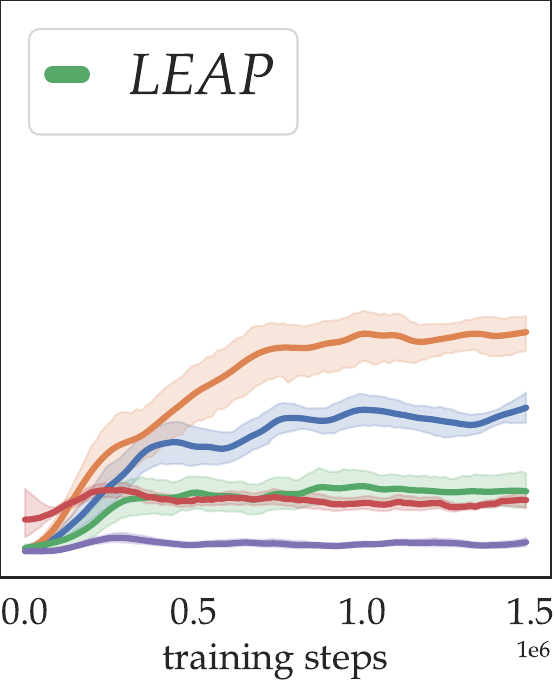}}
\hfill
\subfloat[OOD, $\delta = 0.45$]{
\captionsetup{justification = centering}
\includegraphics[height=0.22\textwidth]{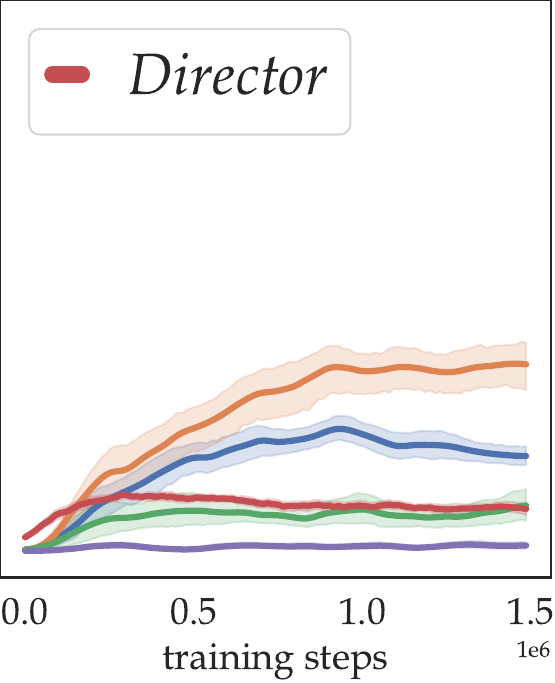}}
\hfill
\subfloat[OOD, $\delta = 0.55$]{
\captionsetup{justification = centering}
\includegraphics[height=0.22\textwidth]{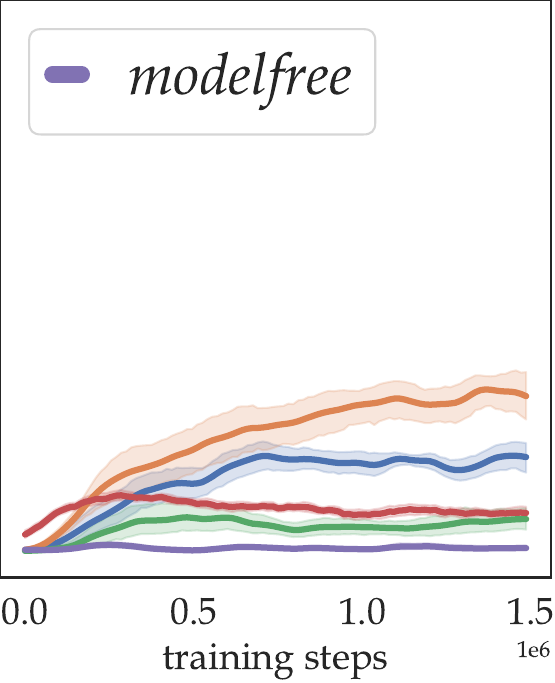}}

\caption[Evaluation Performance of Agents in \textbf{Stochastic} Environments]{\textbf{Evaluation Performance of Agents in \textbf{Stochastic} Environments}: $\epsilon$-greedy style randomness is added to each primitive action with $\epsilon=0.1$. Each agent is trained with $50$ tasks and each curve is processed from $20$ seed runs.}
\label{fig:50_envs_stoch}
\end{figure}

We present the performance evolution (throughout training) of the agents in stochastic variants of \RDS{}. We modify \RDS{} dynamics by implementing $\epsilon$-greedy actions, \ie{}, with probability $\epsilon = 0.1$, each action is changed into a random action. We present the $50$-training tasks performance evolution in Fig.~\ref{fig:50_envs_stoch}. The results validate the compatibility of our agents with stochasticity in environmental dynamics. Notably, the performance of the baseline deteriorated to worse than even \Director{} with the injected stochasticity.

\subsubsection{Accuracy of Proxy Problems \& Checkpoint Policies}

\begin{figure}[htbp]
\centering
\subfloat[training, $\delta = 0.4$]{
\captionsetup{justification = centering}
\includegraphics[height=0.22\textwidth]{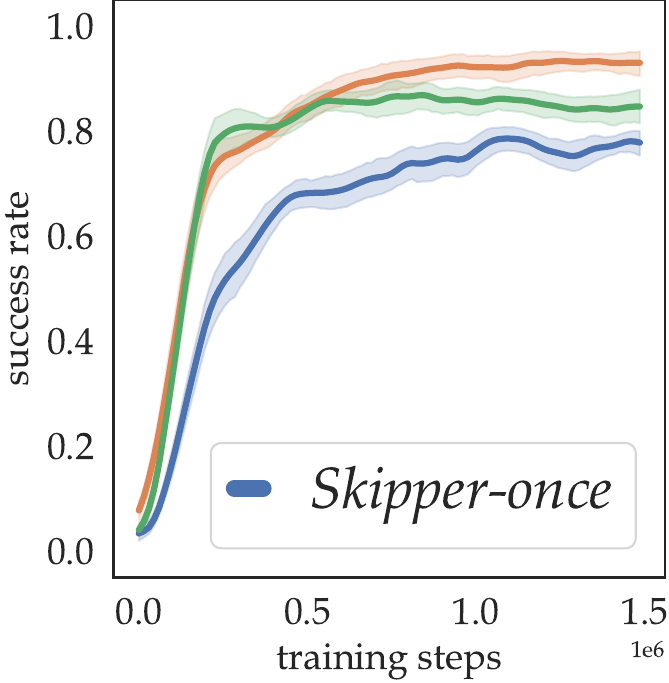}}
\hfill
\subfloat[OOD, $\delta = 0.25$]{
\captionsetup{justification = centering}
\includegraphics[height=0.22\textwidth]{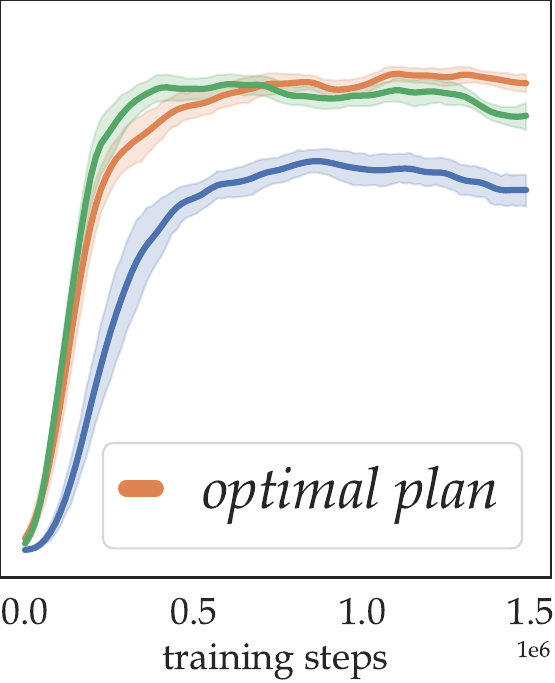}}
\hfill
\subfloat[OOD, $\delta = 0.35$]{
\captionsetup{justification = centering}
\includegraphics[height=0.22\textwidth]{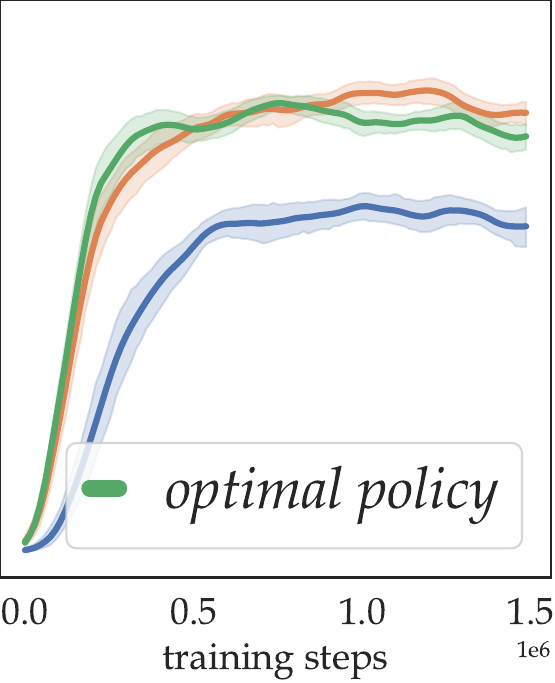}}
\hfill
\subfloat[OOD, $\delta = 0.45$]{
\captionsetup{justification = centering}
\includegraphics[height=0.22\textwidth]{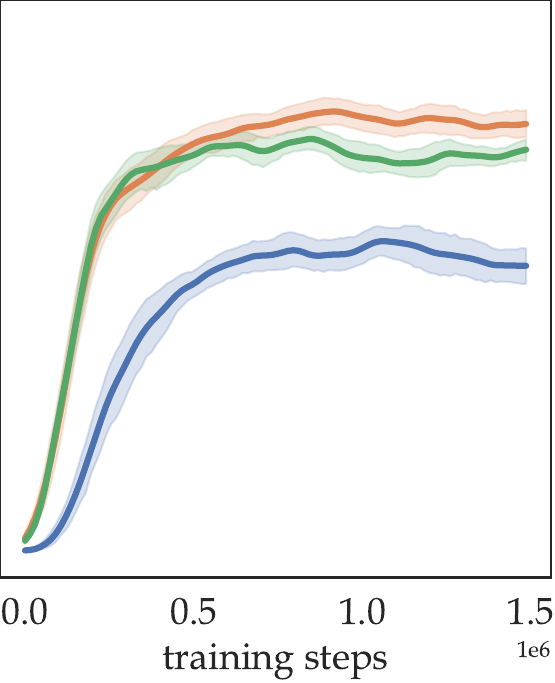}}
\hfill
\subfloat[OOD, $\delta = 0.55$]{
\captionsetup{justification = centering}
\includegraphics[height=0.22\textwidth]{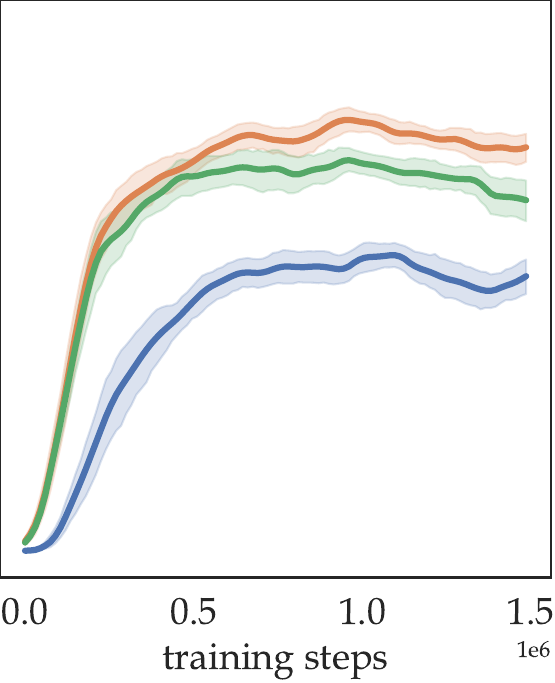}}

\caption[Evaluation Performance of \Skipper{}-once \vs{} Oracle Agents]{\textbf{Evaluation Performance of \Skipper{}-once \vs{} Oracle Agents}: both the \textit{optimal policy} and \textit{optimal plan} variants are assisted by ground truths solved by DP. The default deterministic setting induces the fact that combining optimal policy and optimal plan results in $1.0$ success rate. The figures suggest that the learned agent is limited by errors both in the proxy problem estimation and the checkpoint policy $\pi$. Each agent is trained with $50$ tasks and each curve is processed from $20$ seed runs. }
\label{fig:50_envs_GT}
\end{figure}

We present in Fig.~\ref{fig:50_envs_GT} the results on the accuracy of proxy problems as well as the checkpoint policies of the \textbf{\Skipper{}-once} agents, trained with $50$ tasks. We created two variants of \textbf{\Skipper{}-once} based on the DP-solved ground truths.

Concurring with our previous theoretical analyses, these results indicate that the performance of \Skipper{} was indeed bottlenecked by the accuracy of the proxy problem estimation on the high-level and the optimality of the checkpoint policy on the lower level. 

Notably, the results for the generalization performance across training tasks, as in \textbf{a)} of \ref{fig:50_envs_GT}, indicate that the lower-than-expected performance is an outcome composed of errors in the two levels, the sub-optimality of the low-level policy as well as the inaccuracies of the proxy problem estimates.

\subsection{Ablation Studies}
\label{sec:skipper_ablation}

We use the following ablation studies to verify the effectiveness of each proposed component of the \Skipper{} framework as well as certain experimental settings. These experiments are mainly done with the more lightweight \textbf{\Skipper{}-once} variant on $50$ training tasks.

\subsubsection{Spatial Abstraction}
\begin{figure}[htbp]
\centering
\subfloat[training, $\delta = 0.4$]{
\captionsetup{justification = centering}
\includegraphics[height=0.22\textwidth]{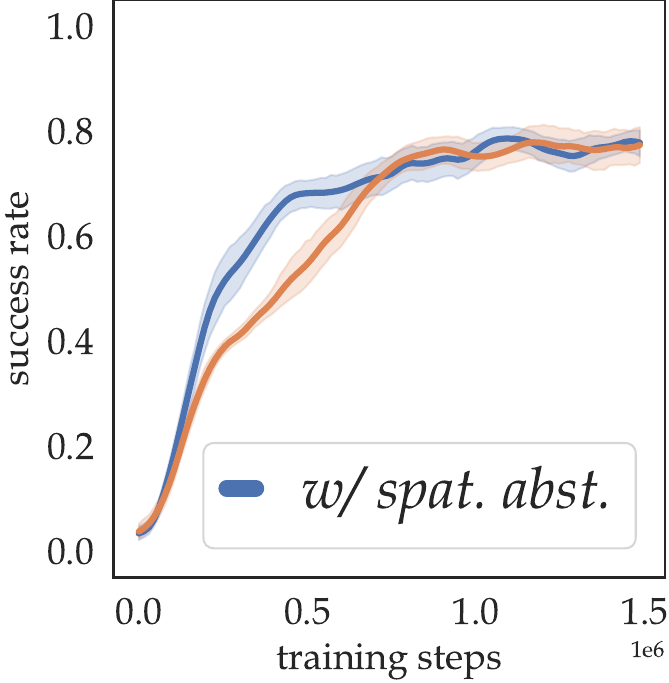}}
\hfill
\subfloat[OOD, $\delta = 0.25$]{
\captionsetup{justification = centering}
\includegraphics[height=0.22\textwidth]{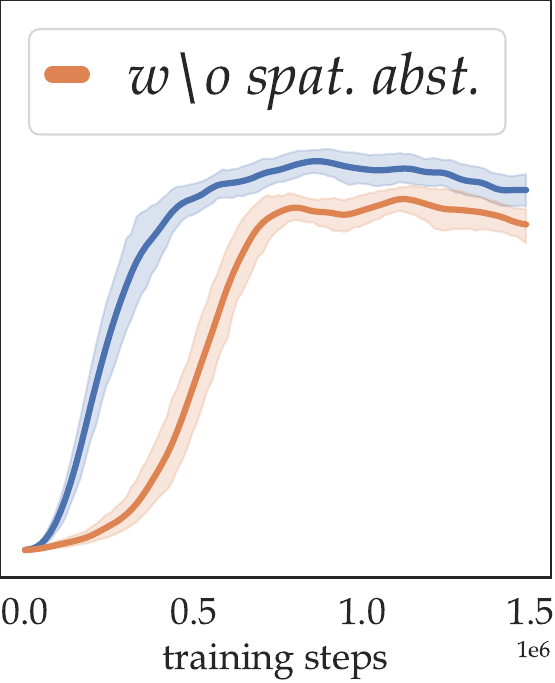}}
\hfill
\subfloat[OOD, $\delta = 0.35$]{
\captionsetup{justification = centering}
\includegraphics[height=0.22\textwidth]{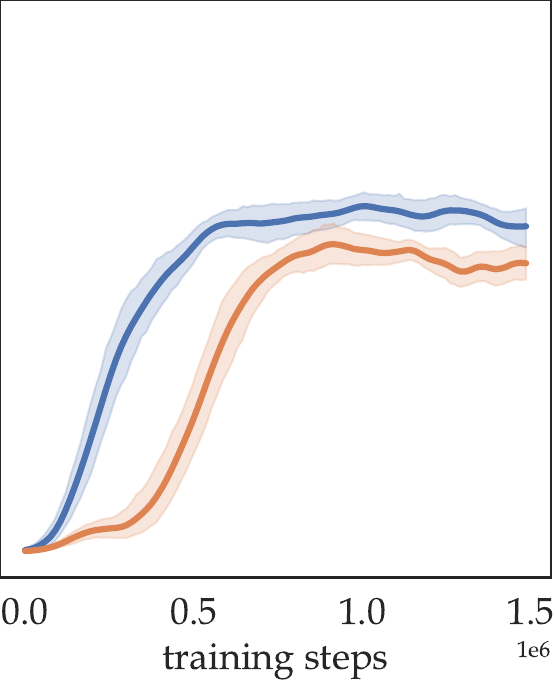}}
\hfill
\subfloat[OOD, $\delta = 0.45$]{
\captionsetup{justification = centering}
\includegraphics[height=0.22\textwidth]{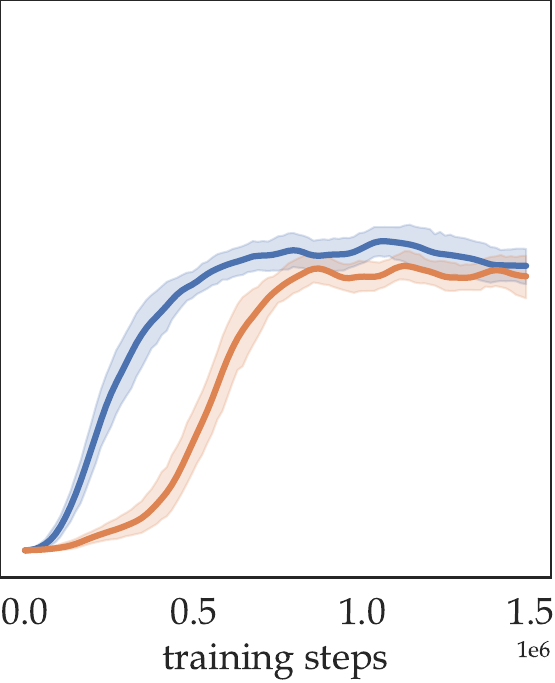}}
\hfill
\subfloat[OOD, $\delta = 0.55$]{
\captionsetup{justification = centering}
\includegraphics[height=0.22\textwidth]{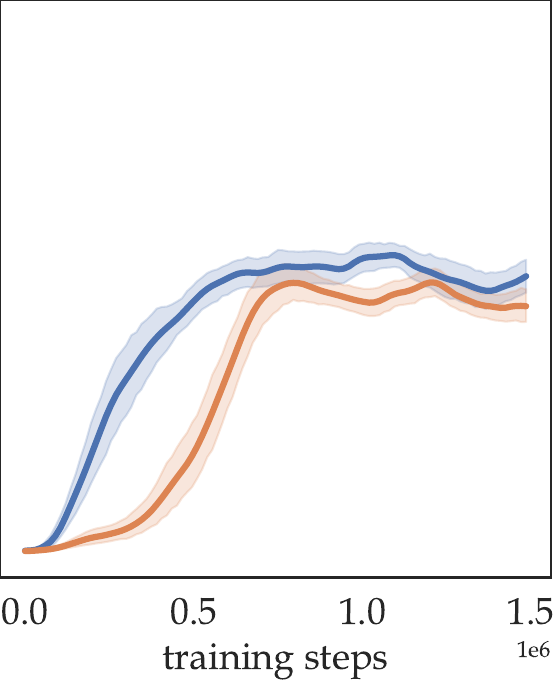}}

\caption[Ablation for Spatial Abstraction on \Skipper{}-once agent]{\textbf{Ablation for Spatial Abstraction on \Skipper{}-once agent}: each compared method is trained with $50$ environments and each curve is processed from $20$ seed runs.}
\label{fig:50_envs_once_local}
\end{figure}

We present in Fig.~\ref{fig:50_envs_once_local} the ablation results on the spatial abstraction component with \textbf{\Skipper{}-once} variant, trained with $50$ tasks. The alternative component, without the spatial abstraction, is an MLP on a flattened full state. The results confirm significant advantage in terms of generalization performance as well as sample efficiency in training, introduced by spatial abstraction.

\subsubsection{Training Initialization: uniform \vs{} same as evaluation}
\begin{figure}[htbp]
\centering
\subfloat[training, $\delta = 0.4$]{
\captionsetup{justification = centering}
\includegraphics[height=0.22\textwidth]{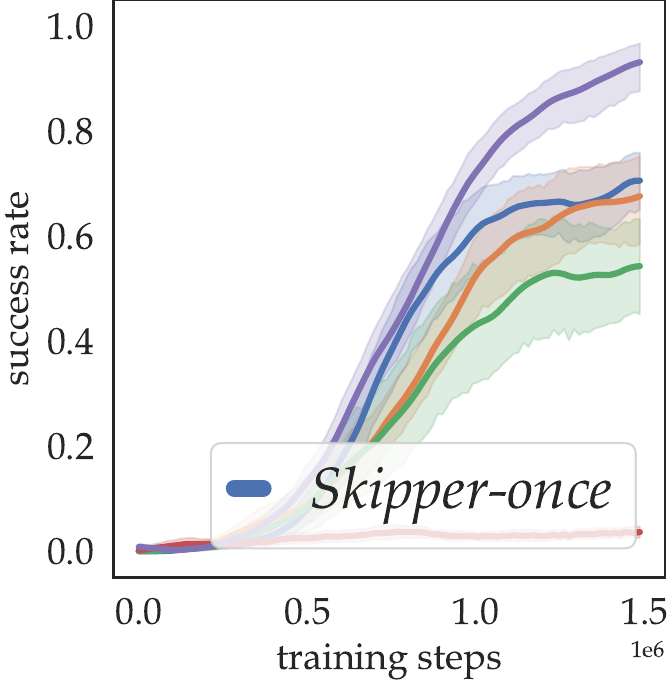}}
\hfill
\subfloat[OOD, $\delta = 0.25$]{
\captionsetup{justification = centering}
\includegraphics[height=0.22\textwidth]{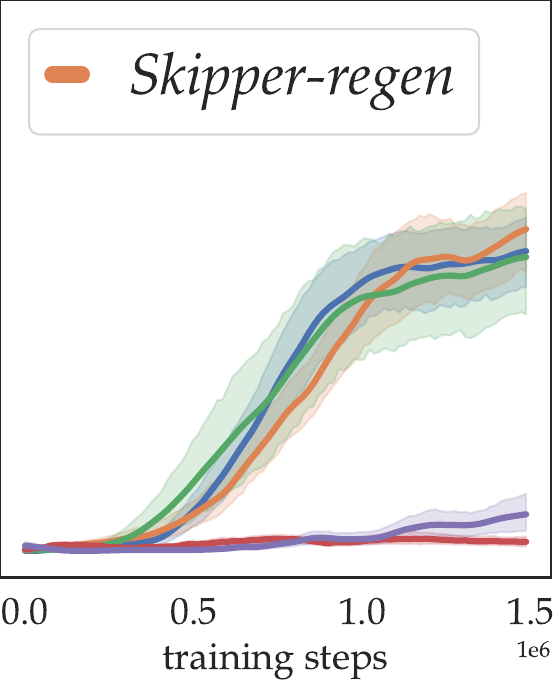}}
\hfill
\subfloat[OOD, $\delta = 0.35$]{
\captionsetup{justification = centering}
\includegraphics[height=0.22\textwidth]{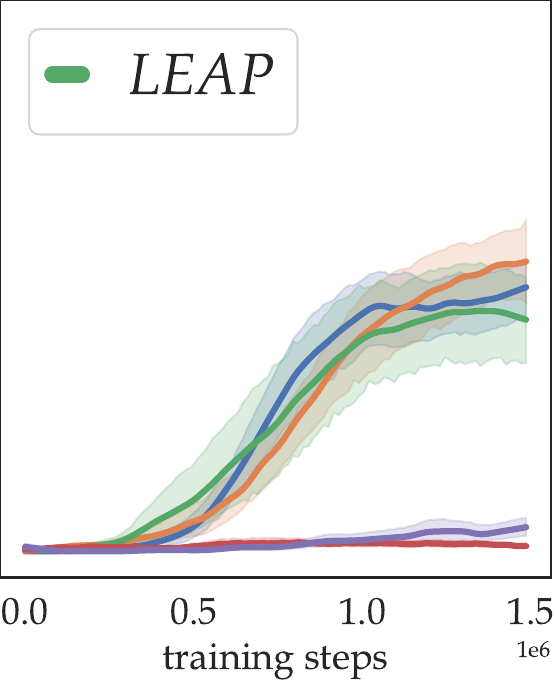}}
\hfill
\subfloat[OOD, $\delta = 0.45$]{
\captionsetup{justification = centering}
\includegraphics[height=0.22\textwidth]{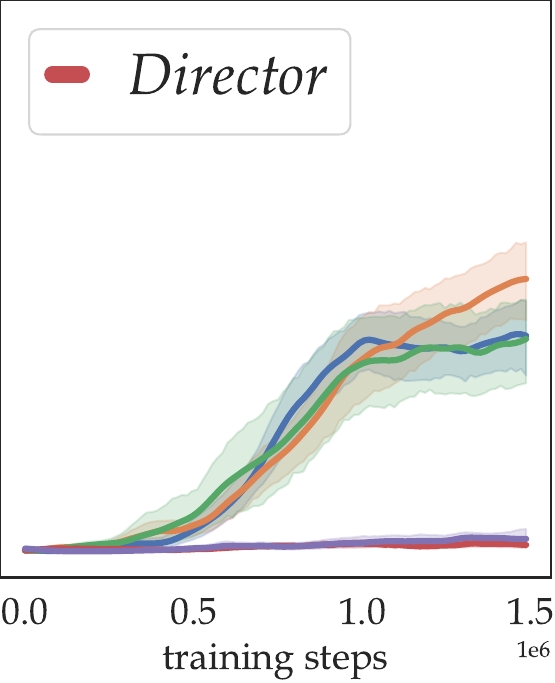}}
\hfill
\subfloat[OOD, $\delta = 0.55$]{
\captionsetup{justification = centering}
\includegraphics[height=0.22\textwidth]{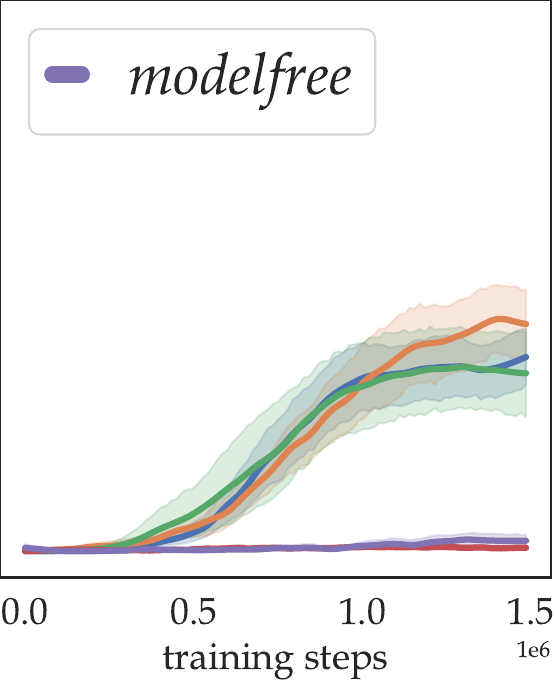}}

\caption[Ablation Results on $50$ Training Tasks without Uniform Initial State Distributions]{\textbf{Ablation Results on $50$ Training Tasks without Uniform Initial State Distributions}: each curve is processed from $20$ seed runs.}
\label{fig:50_envs_non_uniform}
\end{figure}

This set of experiments is used to justify the changes in initialization, that we previously introduced to accelerate the experiments. We compare the agents' performance with and without uniform initial state distribution. The non-uniform starting state distributions introduce additional difficulties in terms of exploration. As Presented in Fig.~\ref{fig:50_envs_non_uniform}, these results are obtained from training on $50$ tasks. We conclude that given similar computational budget, using non-uniform initialization only slows down the learning curves without introducing significant changes to our conclusions, and thus we use the new initialization setting throughout this chapter by default.

\subsubsection{Planning over Proxy Problems}
\begin{figure}[htbp]
\centering
\subfloat[training, $\delta = 0.4$]{
\captionsetup{justification = centering}
\includegraphics[height=0.22\textwidth]{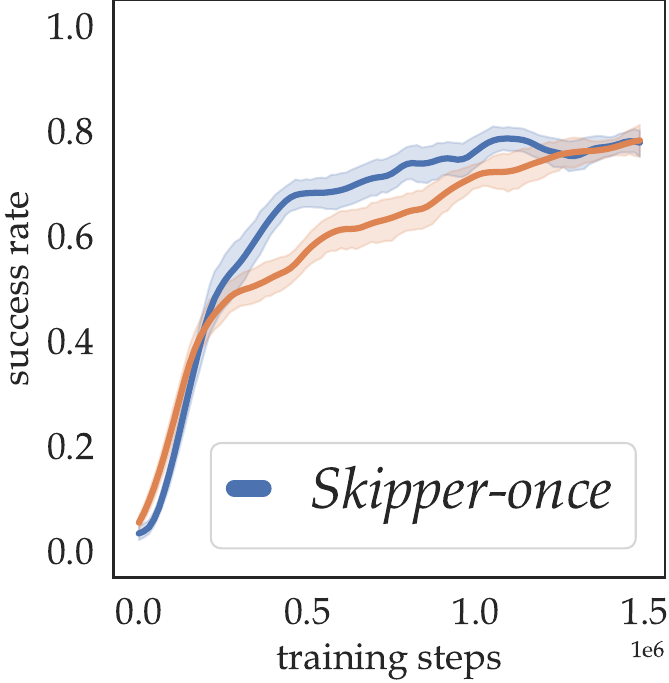}}
\hfill
\subfloat[OOD, $\delta = 0.25$]{
\captionsetup{justification = centering}
\includegraphics[height=0.22\textwidth]{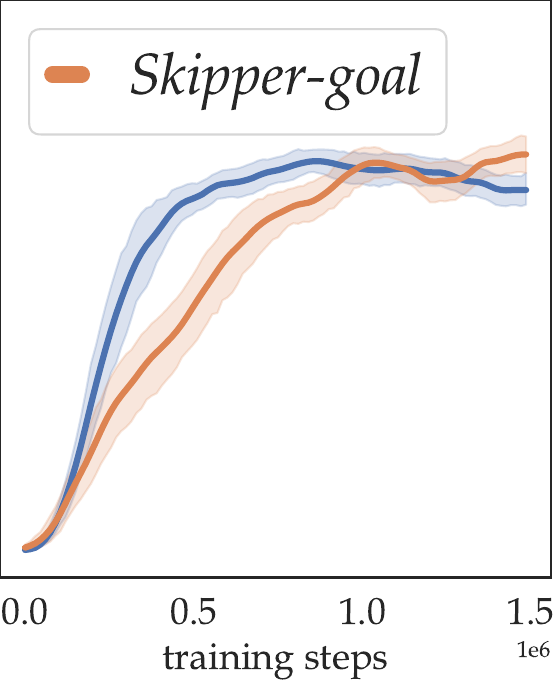}}
\hfill
\subfloat[OOD, $\delta = 0.35$]{
\captionsetup{justification = centering}
\includegraphics[height=0.22\textwidth]{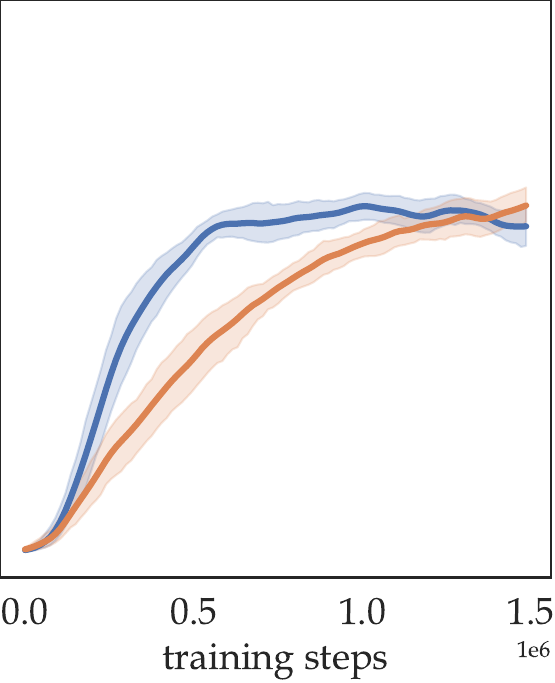}}
\hfill
\subfloat[OOD, $\delta = 0.45$]{
\captionsetup{justification = centering}
\includegraphics[height=0.22\textwidth]{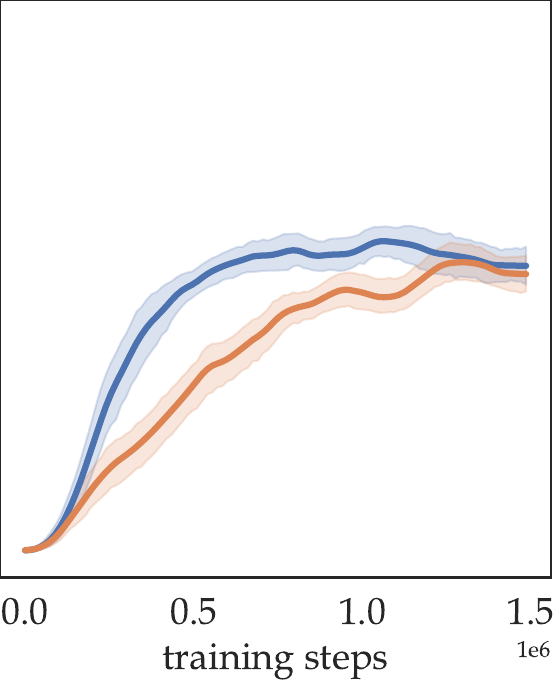}}
\hfill
\subfloat[OOD, $\delta = 0.55$]{
\captionsetup{justification = centering}
\includegraphics[height=0.22\textwidth]{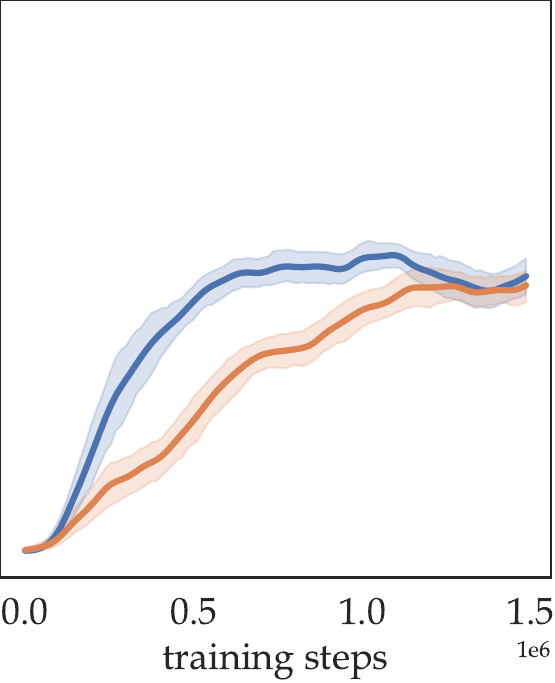}}

\caption[Ablation Result for the Effectiveness of Proxy Problems]{\textbf{Ablation Result for the Effectiveness of Proxy Problems}: Each agent is trained with $50$ tasks and each curve is processed from $20$ seed runs.}
\label{fig:50_envs_once_always_goal}
\end{figure}

We provide additional results for the readers to intuitively understand and validate the effectiveness of planning over proxy problems, which is the core to this chapter, This is done by comparing the results of \textbf{\Skipper{}-once} with a baseline \textbf{\Skipper{}-goal} that blindly selects the task goal as its target all the time. We present the results based on $50$ training tasks in Fig.~\ref{fig:50_envs_once_always_goal}. Concurring with our vision on temporal abstraction, we can see that by utilizing proxy problems and solving more manageable sub-problems leads to faster convergence. The \textbf{\Skipper{}-goal} variant catches up later when the policy slowly improves to be capable of solving longer distance navigation.

\subsubsection{Vertex Pruning}
We want to validate if the proposed $k$-medoids based vertex pruning technique is useful and could produce better generalization abilities by improving the quality of the proxy problems.

\begin{figure}[htbp]
\centering
\subfloat[training, $\delta = 0.4$]{
\captionsetup{justification = centering}
\includegraphics[height=0.22\textwidth]{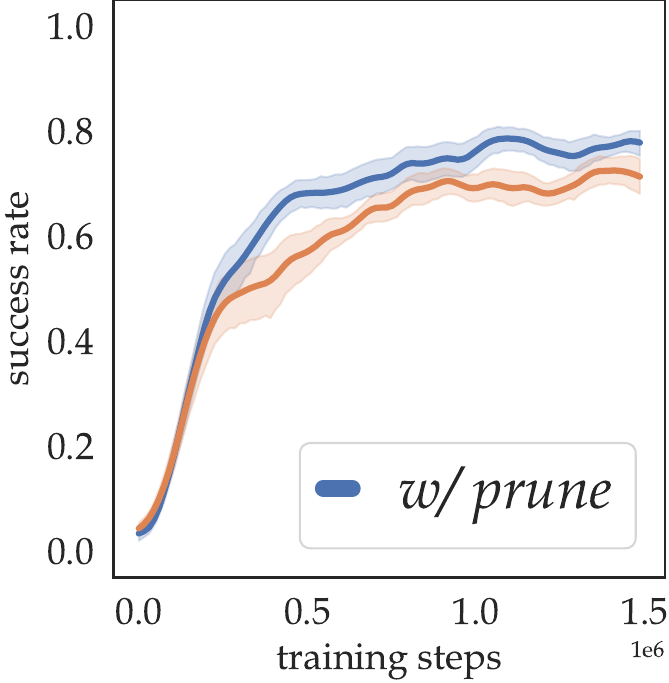}}
\hfill
\subfloat[OOD, $\delta = 0.25$]{
\captionsetup{justification = centering}
\includegraphics[height=0.22\textwidth]{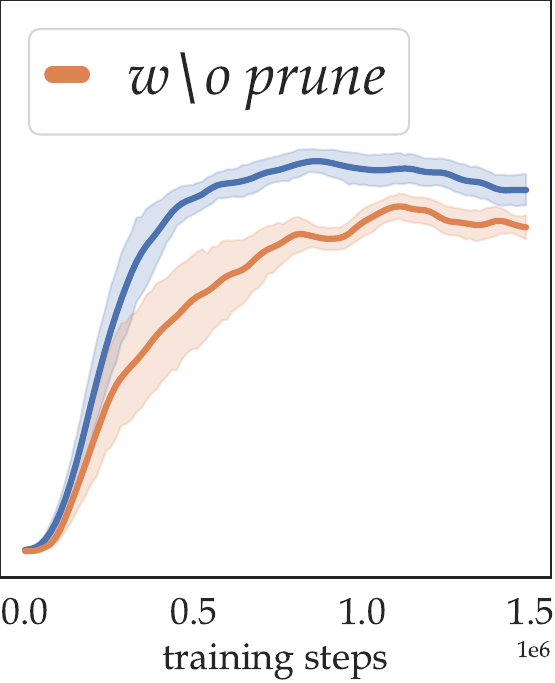}}
\hfill
\subfloat[OOD, $\delta = 0.35$]{
\captionsetup{justification = centering}
\includegraphics[height=0.22\textwidth]{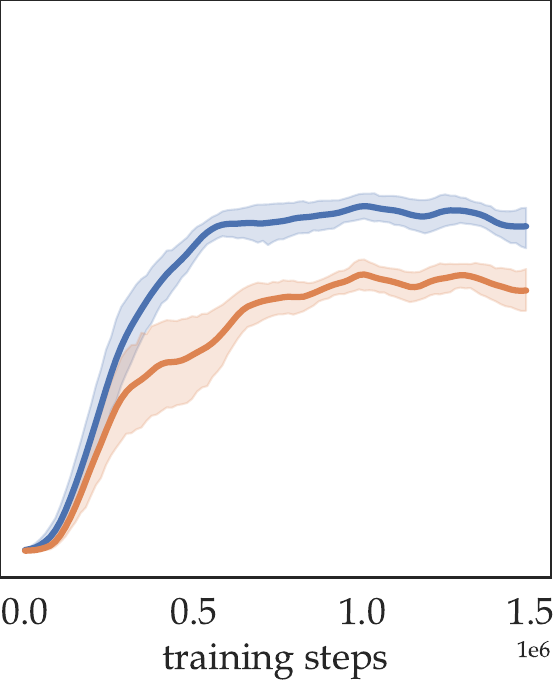}}
\hfill
\subfloat[OOD, $\delta = 0.45$]{
\captionsetup{justification = centering}
\includegraphics[height=0.22\textwidth]{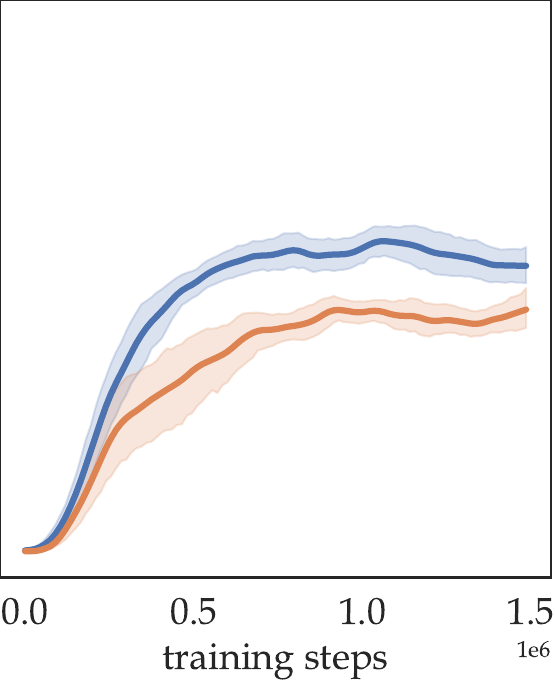}}
\hfill
\subfloat[OOD, $\delta = 0.55$]{
\captionsetup{justification = centering}
\includegraphics[height=0.22\textwidth]{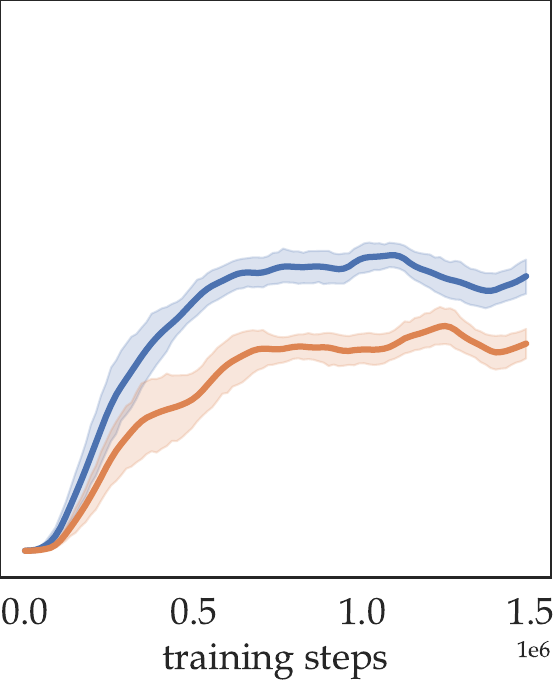}}

\caption[Ablation Results on $50$ Training Tasks for $k$-medoids Vertex Pruning]{\textbf{Ablation Results on $50$ Training Tasks for $k$-medoids Vertex Pruning}: each curve is processed from $20$ seed runs.}
\label{fig:50_envs_once_no_prune}
\end{figure}

In all previous experiments, each proxy problem is reduced from $32$ vertices to $12$ with $k$-medoids. We compare the performance of the default $32\to12$ configuration against a baseline that generates $12$-vertex proxy problems without pruning and present the results in Fig.~\ref{fig:50_envs_once_no_prune}.

From these results, we can observe that generating more checkpoints than needed then pruning produce higher quality proxy problems than directly generating the exact number of checkpoints without pruning. And thus, we can deduce that the proposed pruning process not only increases the generalization but also the stability of performance. 

\subsection{Sensitivity Studies}
\label{sec:skipper_sensitivity}
In this sub-secction, we conduct experimental studies to understand how sensitive the proposed \Skipper{} framework is to certain important hyperparameters.

\subsubsection{Number of Checkpoints in Proxy Problem}
\label{sec:sensitivity_num_checkpoints}

\begin{figure}[htbp]
\centering
\subfloat[training, $\delta = 0.4$]{
\captionsetup{justification = centering}
\includegraphics[height=0.22\textwidth]{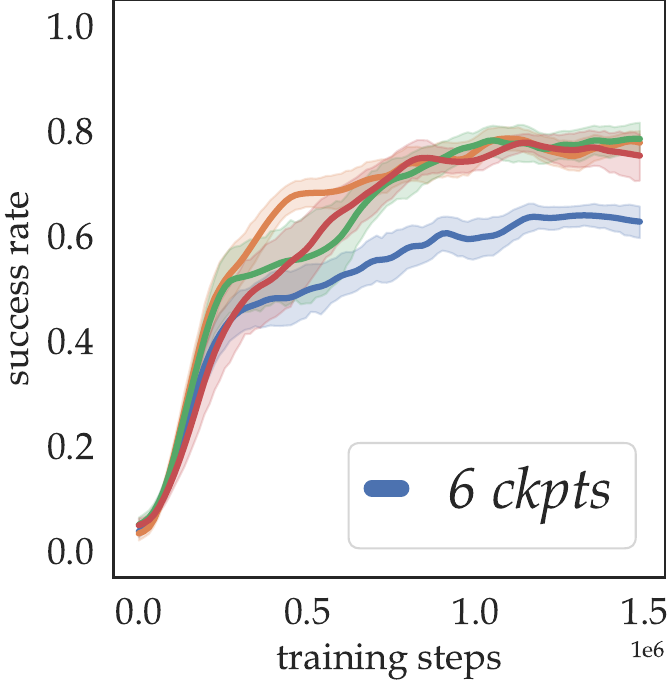}}
\hfill
\subfloat[OOD, $\delta = 0.25$]{
\captionsetup{justification = centering}
\includegraphics[height=0.22\textwidth]{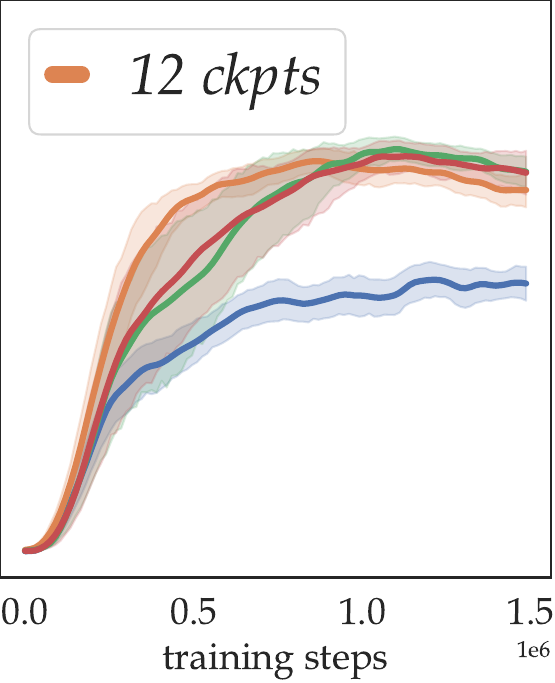}}
\hfill
\subfloat[OOD, $\delta = 0.35$]{
\captionsetup{justification = centering}
\includegraphics[height=0.22\textwidth]{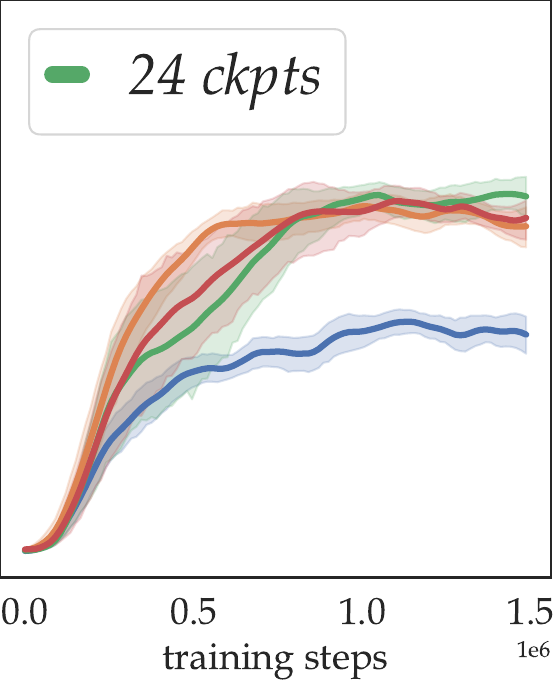}}
\hfill
\subfloat[OOD, $\delta = 0.45$]{
\captionsetup{justification = centering}
\includegraphics[height=0.22\textwidth]{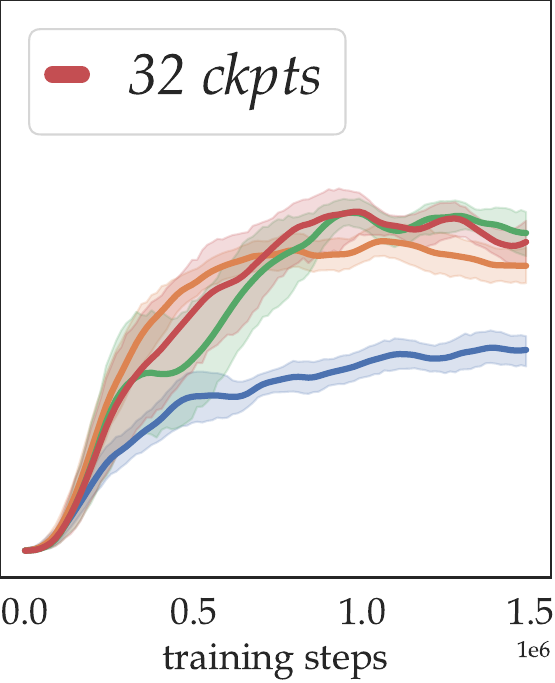}}
\hfill
\subfloat[OOD, $\delta = 0.55$]{
\captionsetup{justification = centering}
\includegraphics[height=0.22\textwidth]{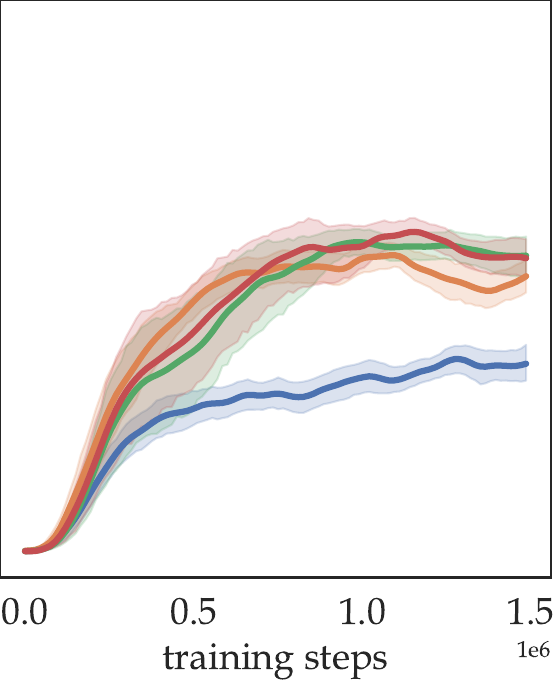}}

\caption[Sensitivity of \Skipper{}-once to Number of Checkpoints in Proxy Problem]{\textbf{Sensitivity of \Skipper{}-once to Number of Checkpoints in Proxy Problem}: Each agent is trained with $50$ tasks. All curves are processed from $20$ independent seed runs. The error bars are $95\%$ confidence intervals.}
\label{fig:50_envs_once_ckpts}
\end{figure}

We now conduct a sensitivity study on the number of checkpoints (number of vertices) in each proxy problem. We present the results of \textbf{\Skipper{}-once} on $50$ training tasks with different numbers of post-pruning checkpoints (all reduced from $32$ by pruning), in Fig.~\ref{fig:50_envs_once_ckpts}. From the results, we can see that as long as the number of checkpoints is above $6$, \Skipper{} exhibits good performance. We therefore chose $12$, the one with a rather small computation cost, as the default hyperparameter. 

As a default, for our experiments, we adopted the episodic success rate as the performance metric, because using discounted return would cause high variance in the curves without intuitive upper bounds depicting the upper limit of the agents' performances. This choice of metric induces a tradeoff, since this metric does not differentiate the agents taking longer timesteps to succeed in tasks. This acknowledgement is also relevant for this sensitivity study.

When there are too many overcrowded checkpoints each potentially contributing to planning errors, more sub-optimal decisions emerge. Thus, we expect with even more checkpoints, the performance will actually decrease.

\subsection{Summary of Experiments}
Within the scope of the previous experiments, we conclude that \Skipper{} provides significant benefits for generalization; And it can achieve even better generalization when exposed to more training tasks;

From the content presented above, we can deduce additionally that: 

\begin{itemize}[leftmargin=*]
\item{} Spatial abstraction based on the local perception field is crucial for the generalization abilities;
\item{} \Skipper{} performs well by reliably decomposing the given tasks, and achieving the sub-tasks robustly. Its performance is bottleneck-ed by the accuracy of the estimated proxy problems as well as the checkpoint policies. This matches well with our theory;
\item{} \LEAP{} in its original form fails to generalize well within its original form and can generalize better when combined with the ideas proposed in this project; We believe that \Director{} may generalize better only in domains where long and informative trajectory collection is possible;
\item{} We verified empirically that, \Skipper{} is compatible with stochasticity, as expected.
\end{itemize}

%% file: chapter_6_Delusions.tex
\chapter{Rejecting Hallucinations: Addressing Delusional Planning Behaviors}
\label{cha:delusions}

\minitoc

\section{Overview of This Thesis Milestone}
\label{sec:delusions_intro}

\textit{In planning processes of computational decision-making agents, generative or predictive models are often used as ``generators'' to propose ``targets'' representing sets of expected or desirable states. Unfortunately, learned models inevitably hallucinate infeasible targets that can cause delusional behaviors and safety concerns. We first investigate the kinds of infeasible targets that generators can hallucinate. Then, we devise a strategy to identify and reject infeasible targets by learning a target feasibility evaluator. To ensure that the evaluator is robust and non-delusional, we adopted a design choice combining off-policy compatible learning rule, distributional architecture, and data augmentation based on hindsight relabeling. Attaching to a planning agent, the designed evaluator learns by observing the agent's interactions with the environment and the targets produced by its generator, without the need to change the agent or its generator. Our controlled experiments show significant reductions in delusional behaviors and performance improvements for various kinds of existing agents.}

The advent of computational modeling has spurred advancements in computational decision making agents, most notable of which is, model-based RL. This work is focused on those models that play the role of \textit{generators}, which imagine or specify future outcomes for agents. Some generators imagine next states or observations, while others specify subgoals (sets of states) to accomplish. No matter how they are represented, we refer to these generator outputs as \textit{targets} and methods that make such use of generators \textit{Target-Assisted Planning (TAP)}.

TAP is a new perspective for unifying many planning / reasoning agents with wildly different behaviors. For instance, some rollout-based model-based RL methods are TAP, such as those following the classic \Dyna{} or Monte-Carlo tree-search frameworks, utilize fixed-horizon transition models to simulate experiences \citep{sutton1991dyna,schrittwieser2019mastering,kaiser2019model}; While, TAP also encompasses methods that directly generate arbitrarily distant targets acting as candidate sub-goals to divide-and-conquer the tasks into smaller, more manageable steps \citep{zadem2024reconciling,zhao2024consciousness,lo2024goal}.

It is the defining characteristic of TAP - the usage of generators, that brings us to the topic of this work: an often unstated assumption is that generated targets are always feasible. However, the desired generalization abilities of generative models are inevitably accompanied by \textit{hallucinations} \citep{xu2024hallucination,zhang2024generalization,xing2024mitigating,jesson2024estimating,aithal2024understanding} - the ``dark side'' that produces infeasible targets that can never be experienced by anyhow. Hallucinations impact TAP agents differently based on their planning behaviors. In \textit{decision-time} TAP methods ~\citep{alver2022understanding}, where models are used to make an immediate decision on what to do next, hallucinated targets can lead to delusional plans that compromise performance and safety \citep{langosco2022goal,bengio2024managing}. For \textit{background} TAP agents that train on simulated experiences constructed with generated targets, delusional values estimated from hallucinated targets can be catastrophically destabilizing \citep{jafferjee2020hallucinating,lo2024goal}.

\begin{figure}[tbp]
\centering
\includegraphics[width=0.7\textwidth]{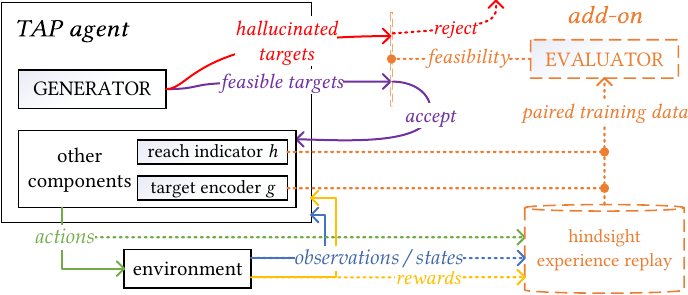}
\caption[Target Evaluator attached to a Target-Assisted Planning (TAP) Agent]{\textbf{Target Evaluator attached to a Target-Assisted Planning (TAP) Agent}:  An abstracted framework that encompasses, but is not limited to, methods listed in Tab.~\ref{tab:methods}. The \textit{generator} proposes candidate target embeddings $\bm{g}^\odot$. Our proposed \textit{evaluator} can be attached to learn to reject infeasible targets (related parts marked in dashed lines).
}
\label{fig:TAP_framework}
\end{figure}

Hallucination and creativity are the two-sides of the same coin that is the generalization ability of the learned models. In psychiatry, it is widely accepted that human brains reject infeasible intentions hallucinated by the belief formation system (that can often hallucinate, similar to the generators in TAP agents) using a belief evaluation system, which acts like a firewall \citep{kiran2009understanding}. Inspired by such discussions, we propose to assist existing TAP agents with a target evaluator, which can be used to reject infeasible targets and in turn delusional behaviors. To maximize compatibility, the evaluator is designed as a minimally-intrusive add-on to attach to existing TAP agents \citet{zhao2020meta}: it learns by observing the TAP agent's interactions with the environment and the targets produced, without the need to change the agent's behaviors or the architecture.. Our main contributions are as follows: 

\begin{enumerate}[leftmargin=*]
\item We systematically categorized and characterized infeasible targets \wrt{} time horizons
\item We discussed the desiderata of learning a minimally-intrusive non-delusional target evaluator
\item We proposed a combination of a) off-policy compatible update rule that enables the evaluator to learn by observing, b) an evaluator architecture compatible with different time horizons that enables a unified solution for most TAP agents and c) two assistive hindsight relabeling strategies performing data augmentation to provide training data beyond those collected via interactions, which ensure that the evaluator itself does not produce delusional evaluations
\item We implemented the solution, as illustrated in Fig.~\ref{fig:TAP_framework}, for several types of existing TAP agents, as discussed in Tab.~\ref{tab:methods}, and showed that agents can better manage generated targets, reduce delusional behaviors and significantly improve performance
\end{enumerate}

\begin{table*}[htbp]
\centering
\aboverulesep=0ex
\belowrulesep=0ex
\caption[Discussed Methods, Properties \& How to use the Feasibility Evaluator]{\textbf{Discussed Methods, Properties \& How to use the Feasibility Evaluator}}
 \resizebox{\textwidth}{!}{
 \begin{NiceTabular}{m[c]{0.1\textwidth}|m[c]{0.16\textwidth}|m[l]{0.3\textwidth}|m[l]{0.35\textwidth}|m[l]{0.4\textwidth}}
 
\toprule

\textbf{Method} & \textbf{TAP Category} & \textbf{Delusional Planning Behaviors} & \textbf{How Our Solution Helps} & \textbf{Implementation Details \& Challenges} \\ \hline
\Dyna{} \citep{sutton1991dyna} & Fixed-Horizon Background Planning & The imagined transitions could contain hallucinated (next) observations / states, whose delusional value estimates could destabilize the bootstrapping-based TD learning & Cancel updates involving rejected next states (not evaluated to be reachable within $1$ timestep). & \cellcolor{blue!25} \textbf{Implemented} (for 1-step Dyna): If the output histogram of the evaluator has significant density on the bin corresponding to $t=1$, then accept the generation, or else, reject \\ \hline
\Dreamer{} \citep{hafner2023mastering} & Fixed-Horizon Background Planning & The imagined trajectories could contain infeasible, hallucinated states & Do not let the rejected imagined states participate in the construction of update targets for the actor-critic system (See Sec.\ref{sec:appendix_dreamer}). & \cellcolor{blue!10} \textbf{Implemented} (insufficient compute for results, Sec.~\ref{sec:appendix_dreamer}): Use the deterministic state $\bm{s}$ as the state representation to feed to the evaluator (also for imagined future target states). Establish $h$ with Mahalanobis distance on the state representations and use the discount, reward and value predictions to force behavioral realism. Truncate $\lambda$-returns until infeasible imagined target states. \\ \hline
\Director{} \citep{hafner2022deep} & Fixed-Horizon Decision-Time Planning (mainly) & The internally sampled goals may be unreachable & Reject unreachable goals and re-sample reachable ones & \cellcolor{blue!10} Similar to our implementation for \Dreamer{}. \\ \hline
\MuZero{} \citep{schrittwieser2019mastering} & Fixed-Horizon Decision-Time Planning & The predicted states in the tree search could be unreachable hallucinations & Reject hallucinated state generations, regenerate node in tree search if necessary & \cellcolor{blue!10} Similar to our implementation for 1-step \Dyna{} \\ \hline
\SimPLe{} \citep{kaiser2019model} & Fixed-Horizon Background Planning & The predicted next observation could be an unreachable hallucination & Reject learning against the delusional estimates (potential) & \cellcolor{blue!10} Similar to our implementation for 1-step \Dyna{} \\ \hline
\Skipper{} \citep{zhao2024consciousness} & Arbitrary-Horizon Decision-Time Planning & Hallucinated subgoals could lead to decision-time planning committing to them, leading to unsafe behaviors & Use an evaluator to learn that the expected cumulative discount is $0$ when aiming to reach the hallucinated subgoals. This disconnects the hallucinated subgoals from the current state in the planning & \cellcolor{green!25} \textbf{Implemented}: diversify the source-target pairs with \generatestr{} and \pertaskstr{} mixtures. $\scriptG{}$ is discrete and $h$ is a trivial comparison. \\ \hline
\GSP{} \citep{lo2024goal} & Arbitrary-Horizon Background Planning & Hallucinated subgoals could lead to value estimation destabilization, like in \Dyna{}. & Use output histogram of the add-on evaluator to correct the delusions by \GSP{}'s own estimators. Use the ``support swap'' technique. & \cellcolor{green!10} Similar to our implementation for \Skipper{} \\ \hline
\LEAP{} \citep{nasiriany2019planning} & Arbitrary-Horizon Decision-Time Planning & Hallucinated subgoals could help fake a sequence of subgoals that is too good to be true and committed to during planning & Use an evaluator to learn that the expected cumulative distance is infinite when aiming to reach the hallucinated subgoals. This makes sure that subgoal sequences containing hallucinated subgoals will not be favored & \cellcolor{red!25} \textbf{Implemented}: pay attention to the representation space of the sub-goals.  \\ \hline
\PlaNet{} \citep{hafner2019learning} & Arbitrary-Horizon Decision-Time Planning & Hallucinated subgoals could help fake a sequence of subgoals that is too good to be true and committed to during planning & Reject the delusional subgoals and therefore reject the delusional subgoal sequences & \cellcolor{red!10} Same as our implementation for \LEAP{} (both uses CEM for planning \citep{rubinstein1997optimization}) \\ \hline
\bottomrule
 \end{NiceTabular}%
 }
\small \textit{Similar colors are used to denote similar implementations for the solution proposed in this work.}

\label{tab:methods}
\end{table*}

\section{Methodology: Target Evaluation \& Relabeling}
\label{sec:delusions_prelim}

\subsection{Target-Assisted Planning \& Targets}
\label{sec:target_directed_framework}
In this work, for generality, we consider a target to be an embedding representing a set of states. Each target $\bm{g}^\odot \mapsto \{ s^\odot \}$ is paired with an indicator function $h$ outputting $h(s',\bm{g}^\odot)=1$ if $s' \in  \{ s^\odot \}$ and $0$ otherwise. For the interest of time horizons, we also introduce $\tau$ - the maximum number of time steps an agent is allowed to reach a state in $\bm{g}^\odot$.

Let $D_\pi(s,\bm{g})$ represents \textit{\nth{1}} timestep $t$ \st{} $h(s_t,\bm{g}^\odot) = 1$, if $\pi$ (conditional or not) is followed from state $s$. We define \textbf{$\tau$-feasibility} of $\bm{g}^\odot$ from state $s$ under $\pi$ as $p(D_\pi(s, \bm{g}^\odot) \leq \tau) \coloneqq \sum_{t=1}^{\tau} p(D_\pi(s, \bm{g}^\odot) = t)$. $\bm{g}^\odot$ is called \textbf{$\tau$-feasible} if $p(D_\pi(s, \bm{g}^\odot) \leq \tau) > 0$, and \textbf{$\tau$-infeasible} otherwise. Notably, $\tau$-feasibility is an intrinsic property of a state-policy-target tuple, not a heuristic measure.

Aligning with the RL objective of maximizing returns, a target is good if it leads to rewarding outcomes, \ie{}:
\begin{align}\label{eq:utility}
\begin{split}
& \scriptU_{\pi, \mu} (s, \bm{g}^\odot, \tau) \coloneqq \\
& r_{\pi}(s, \bm{g}^\odot, \tau) + \gamma_\pi(s, \bm{g}^\odot, \tau) \cdot V_\mu(s_{\min(D_\pi(s,\bm{g}^\odot), \tau)})
\end{split}
\end{align}
where $\min(D_\pi(s,\bm{g}^\odot), \tau)$ denotes the timestep when the commitment to $\bm{g}^\odot$ is terminated (by $h$ or $\tau$), $s_{\min(D_\pi(s,\bm{g}^\odot), \tau)}$ is the state the agent ended up in, $r_{\pi}(s, \bm{g}^\odot, \tau) \coloneqq \sum_{t=1}^{\min(D_\pi(s,\bm{g}^\odot), \tau)}{\gamma^{t-1} r_t}$ is the cumulative discounted reward along the way, $ \gamma_\pi(s, \bm{g}^\odot, \tau) \coloneqq \gamma^{\min(D_\pi(s,\bm{g}^\odot), \tau)}$ is the cumulative discount, and $V_\mu(\cdots)$ is the future value for following $\mu$ afterwards.

Eq.~\ref{eq:utility} shows that if $\bm{g}^\odot$ is $\tau$-infeasible, \ie{}, $s_{min(D_\pi, \tau)} \notin \bm{g}^\odot$, then TAP methods blindly using $\bm{g}^\odot$ to determine $V_\mu$ will produce delusional evaluations - the cause of delusional planning behaviors. For example, feasibility-unaware methods, \eg{}, \citet{sutton1991dyna,schrittwieser2019mastering,hafner2023mastering}, assume that targets are always reachable as long as they can be generated. Meanwhile, planned trajectories involving infeasible targets are delusional; There are also some feasibility-aware methods, \eg{} \citet{nasiriany2019planning,zhao2024consciousness,lo2024goal}, in which agents estimate certain metrics to decide if a target is feasible. However, as we will discuss later, they often produce incorrect estimates, thus may still favor infeasible targets. In later sections, we propose an evaluator that simultaneously estimates the $\tau$-feasibility and $D_\pi$ of the proposed targets, where these estimations are used to decide if the evaluation of a target should be trusted or if the target should be rejected.

\subsection{Source-Target Pairs \& Hindsight Relabeling}

To learn the feasibility of a target from a given state, ``source-target pairs'' are needed, which are tuples involving a source state and a target embedding. The quality of these paired training data is critical for the training outcome \citep{dai2021diversity,moro2022goaldirected,davchev2021wish}. Hindsight Experience Replay (HER) can be seen as a way to enhance the diversity of the pairs, by re-using targets that happened to have been achieved on existing trajectories, and pretending that they were the chosen targets during the interactions \citep{andrychowicz2017hindsight}. \textbf{Relabeling strategies}, corresponding to how $s^{\odot}$ is obtained, are critical for HER's performance \citep{shams2022addressing,eysenbach2021clearning}. Most existing relabeling strategies are \textit{trajectory-level}, meaning that $s^{\odot}$ comes from the same trajectory as $s_t$. These include \futurestr{}, where $s^{\odot} \leftarrow s_{t'}$ with $t' > t$, and \episodestr{}, with $0 \leq t' \leq T_\perp$. HER greatly enhanced the sample efficiency of learning about experienced targets. Meanwhile, its limitations to learning about only experienced targets planted a hidden risk of delusions towards hallucinated targets for TAP agents, to be discussed later. For more discussions regarding Hindsight Experience Replay (HER) and hindsight relabeling strategies, please see Sec.~\ref{sec:source_target_pair_hindsight_relabeling} on Page.~\pageref{sec:source_target_pair_hindsight_relabeling}.

\section{Methodology: Understanding Hallucinated State Targets}
\label{sec:problematic_targets}

Categorizing targets proposed by the generator helps inform us about how to correctly learn the feasibility of targets, \st{} hallucinated targets can be properly rejected.

Let us warm-up with singleton targets, \ie{}, $\bm{g}^\odot$ has a single element $\hat{s}^\odot$, and propose a characterization of singleton targets into $3$ \textit{disjoint} categories, which we call \Gzero{}, \Gone{} and \Gtwo{}. Then, we will extend to the non-singleton targets composed of these $3$ ``ingredients''.

\gdelusion{delusion:g0}
\subsection{\Gzero{}: \texorpdfstring{$\infty$-Feasible}{Feasible}}
Given source state $s$, a generated singleton target $\bm{g}^\odot$ is called \Gzero{} if it maps to one state which is $\infty$-feasible from $s$, with some policy $\pi$. Note that \Gzero{} includes $\tau$-infeasible states for given finite values of $\tau$.


\gdelusion{delusion:g1}
\subsection{\Gone{} - Permanently Infeasible (Hallucinated)}
\Gone{} includes generated ``states'' that do not belong to the MDP at all, \ie{}, a target ``state'' $\hat{s}^\odot$ is \Gone{} if $\forall s,\pi,~ p(D_\pi(s, \hat{s}^\odot) < \infty) = 0$. 


\gdelusion{delusion:g2}
\subsection{\Gtwo{} - Temporarily Infeasible (Hallucinated)}
This type includes those MDP states that are \textit{currently infeasible from state $s$}. Unlike \Gone{}, \Gtwo{} states could be \Gzero{} if they were evaluated from a different source state. \Gtwo{}s can often be overlooked, not only because hallucinations are mostly studied in contexts that do consider the source state $s$, but also because they only exist in some special MDPs.

\subsection{Examples}

To provide intuition about these concepts, we use the MiniGrid platform to create a set of fully-observable environments, minimizing extraneous factors to focus on the targets \citep{chevalierboisvert2023minigrid}. We call this environment \SSMfull{} (\SSM{} for short); In \SSM{}, agents navigate by moving one step at a time in one of four directions across fields of randomly placed, episode-terminating lava-traps, while searching for both a sword and a shield to defeat a monster with a terminal reward. The lava-traps' density is controlled by a difficulty parameter $\delta$, but there is always a feasible path to success. Approaching the monster without both the randomly placed sword and shield ends the episode. Once acquired, either of the two items cannot be dropped, leading to a state space where not all states are accessible from the others. Thus, \SSM{} states are partitioned into $4$ semantic classes, defined by $2$ indicators for sword and shield possession. For example, $\langle 0, 1 \rangle$ denotes ``sword not acquired, shield acquired''.

\begin{figure}[htbp]
\captionsetup[subfigure]{labelformat=empty}
\centering
\subfloat[\Gone{}-induced delusional behavior]{
\captionsetup{justification = centering}
\includegraphics[width=0.238\textwidth]{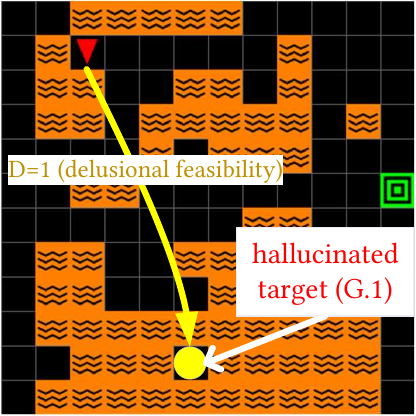}}
\hfill
\subfloat[]{
\captionsetup{justification = centering}
\includegraphics[width=0.23\textwidth]{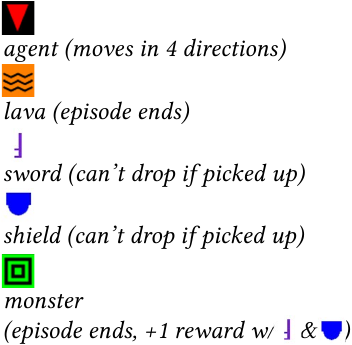}}
\subfloat[\Gtwo{}-induced delusional behavior ($s$ in $\langle 1, 1 \rangle$, $s^\odot$ in $\langle 0, 0 \rangle$)]{
\captionsetup{justification = centering}
\includegraphics[width=0.48\textwidth]{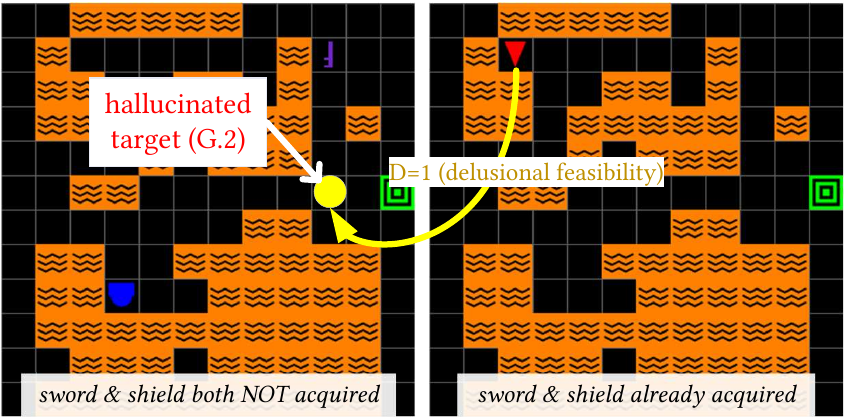}}
\caption[Delusional Plans in SSM]{\textbf{Delusional Plans in \SSM{}}: In both cases, a TAP agent, lacking understanding about the hallucinated targets (yellow dots), mis-evaluates their feasibility, leading to delusional plans which suggest that the task goal can be achieved via the infeasible targets.
}
\label{fig:example_delusions}
\end{figure}

\Gone{} generations in this environment may be semantically valid, \eg{}, an \SSM{} ``state'' with the agent surrounded by lava, as in Fig.~\ref{fig:example_delusions} (top row), or totally absurd, \eg{}, an \SSM{} observation without an agent.

\Gtwo{} states can be once \Gzero{} but are now blocked due to a past transition, \eg{}, after acquiring the sword in \SSM{}, the agent transitions from class $\langle 0, 0\rangle$ to $\langle 1, 0\rangle$, sealing off access to $\langle 0, 0\rangle$ or $\langle 0, 1\rangle$; \Gtwo{} can also appear due to the initial state distribution $d$: some states can only be accessed from specific initial states, \eg{}, an agent spawned in $\langle 1, 0\rangle$ cannot reach $\langle 0, 0\rangle$ or $\langle 0, 1\rangle$. An example of delusional behavior caused by a \Gtwo{} target is provided in Fig.~\ref{fig:example_delusions} (bottom row).

Despite rising concerns regarding the safety of TAP agents \citep{bengio2024managing}, their delusional behaviors remain under-investigated, largely due to the \textbf{lack of access} to ground truths needed to identify hallucinations and their resulting delusional behaviors. Thus, it is critical to analyze with clear examples and conduct rigorous controlled experiments where the ground truth of targets could be solved with Dynamic Programming (DP) \citep{howard1960dynamic}, which is why we created \SSM{} and used it later for experiments.

\subsection{Non-Singleton Targets}

For the general case, where a generated target embedding $\bm{g}^\odot$ potentially corresponds to a set of ``states'' $\{\hat{s}^\odot\}$, the elements of the associated set may span all categories, \ie{}, the target can correspond to a mixture of \Gzero{}, \Gone{} \& \Gtwo{} states. Tab.~\ref{tab:birdseye} summarizes the possible target compositions and their properties\footnote{There may be no explicit mapping from a target embedding to a set of ``states'' and thus any target can always map to arbitrarily many \Gone{} ``states''. Thm.~\ref{thm:feasibility_reduction} explains that this problem is benign.}.

\begin{table*}[htbp]
\centering
\aboverulesep=0ex
\belowrulesep=0ex
\caption[Categorization of Targets based on Composition, Characteristics, Risks \& Delusion Mitigation Strategies]{\textbf{Categorization of Targets based on Composition, Characteristics, Risks \& Delusion Mitigation Strategies}}
 \resizebox{\textwidth}{!}{
 \begin{NiceTabular}{m[c]{0.12\textwidth}|m[l]{0.17\textwidth}|m[c]{0.17\textwidth}|m[l]{0.32\textwidth}|m[l]{0.32\textwidth}}
 
\toprule

\textbf{Target Composition} & \textbf{State Correspondence} & \textbf{$\infty$-Feasibility} $p(D_\pi(s, \bm{g}^\odot) < \infty)$ & \textbf{Feasibility Errors \& Resulting Delusional Planning Behaviors} & \textbf{Data Augmentation (Relabeling) Strategies against Feasibility Delusions} \\
\hline
\textbf{Only or Single \Gzero{}} & non-hallucinated feasible states from $s$ & $>0$ & \edelusion{delusion:e0} \textbf{\Ezero{}}: May think \Gzero{} states are infeasible, thus turn to riskier alternatives, \eg{}, \Gone{} or \Gtwo{} & \episodestr{} for \Gzero{} (and \Gtwo{}) states in the same episode + \pertaskstr{} for \Gzero{} (and \Gtwo{}) beyond the episode \\
\hline
\textbf{Only or Single \Gone{}} & hallucinated ``states'' not belonging to the MDP & should $=0$ & \edelusion{delusion:e1} \textbf{\Eone{}}: May think \Gone{} states are favorable, thus commit to them. Impacted by ill-defined  $V_\mu(\cdots)$ & \generatestr{} for \Gone{} (and \Gzero{} \& \Gtwo{}) states, to be proposed by the generator \\
\hline
\textbf{Only or Single \Gtwo{}} & hallucinated MDP states infeasible from $s$ & should $=0$ & \edelusion{delusion:e2} \textbf{\Etwo{}}: May think \Gtwo{} states are favorable, thus commit to them & \pertaskstr{} for \Gtwo{} (and \Gzero{}) beyond episode + \episodestr{} for \Gtwo{} (and \Gzero{}) states in the same episode \\
\bottomrule
\textbf{Some \Gzero{}} & at least one non-hallucinated state from $s$ & $=$ \newline $p(D_\pi(s, \bm{g}_{-}^\odot) < \infty)$ \newline $>0$ (Thm.~\ref{thm:feasibility_reduction}) & \textbf{\Ezero{}} & \episodestr{} + \pertaskstr{}\\ 
\hline
\textbf{Only \Gone{} \& \Gtwo{}} & set of ONLY hallucinated states & should $=0$ & \textbf{\Eone{}} \& \textbf{\Etwo{}} & \generatestr{} or \generatestr{} + \pertaskstr{} \\
\bottomrule
 \end{NiceTabular}%
 }

\label{tab:birdseye}
\end{table*}

\section{Methodology: Evaluating Targets Correctly and Robustly}
\label{sec:delusions_solutions}

Knowing that hallucinations cannot be eradicated in general, we intend to lower their risks by adopting the brain-inspired solution - to reject infeasible targets post-generation. If done effectively, the negative impact of hallucinated targets becomes limited to the resource cost of generating and rejecting targets, to be discussed in detail. This approach is in contrast with directly trying to address hallucinations in the generators case-by-case, which we deem to have an unbreakable glass ceiling and not versatile enough to be generalized to generic TAP methods. In contrast, agents without evaluators accept proposed targets unconditionally and thus are at risk of delusional planning behaviors.

For a feasibility evaluator to be effectively differentiate the proposed targets, it should correctly estimate the feasibility of targets which maps to all kinds of states (\Gzero{}, \Gone{} \& \Gtwo{}). However, learning to estimate feasibility is not as trivial as it seems, because improper training could naturally lead to delusional feasibility estimations, which cannot be simply addressed by scaling up training. If the evaluator has delusions of feasibility, then its incorporation becomes futile, as hallucinated targets could still be favored.

For estimation errors, we similarly warm up with those of the singleton targets. For clarity, we use matching identifiers \Ezero{}, \Eone{}, and \Etwo{} to denote the estimation errors of feasibility towards \Gzero{}, \Gone{}, and \Gtwo{} ``states'', respectively. These discussions are presented in Tab.~\ref{tab:birdseye}.

When targets correspond to general sets of states, we have:

\begin{result}
    Let $\bm{g}^\odot$ be a target embedding. Its feasibility from state $s$ satisfies:
    \vspace*{-2mm}
    $$\forall \pi, p(D_\pi(s,\bm{g}^\odot) \leq \tau ) = p(D_\pi(s, \bm{g}_{-}^\odot) \leq \tau )$$
    where $\bm{g}^\odot_{-}$ is a target that correspond to the set of states of $\bm{g}^\odot$ with all infeasible states (\Gone{} \& \Gtwo{}) removed.
    \label{thm:feasibility_reduction}
\end{result}

This result indicates that a target is infeasible if and only if it consists entirely of infeasible states, allowing us to focus on learning processes that identify such cases.

\subsection{Desiderata for Evaluator}

We used the following important considerations to guide our design for an appropriate feasibility evaluator. 

\begin{itemize}[leftmargin=*]
\item \textbf{[automatic]} the evaluator must learn to automatically differentiate the feasibility of all kinds of targets without pre-labeling: \textit{we need to exploit $h$}

\item \textbf{[minimally intrusive]} the evaluator should be generally applicable to existing TAP agents, without changing the agents too much to disturb the generally-sensitive RL components: \textit{we need to ensure its behavior as an add-on and it can be conditioned on the policy $\pi$ of the agent, to learn alongside the agent by merely observing}

\item \textbf{[unified]} the evaluator should have a unified behavior compatible with different $\tau$s: \textit{we can design it in a way to learn the $\tau$-feasibilities for many $\tau$ values simultaneously}

\end{itemize}

\subsection{Learning Rule \& Architecture for Feasibility}
\label{sec:learning_rules}

Following the considerations, we propose to use the following learning rule to \textit{indirectly} learn the targets' feasibility by \textit{directly} learning the distribution of $D_\pi(s, \bm{g}^\odot)$.

\begin{align}\label{eq:rule_feasibility_gamma}
& D_{\pi}(s, \bm{g}^\odot) \leftarrow 1 + D_{\pi}(s', \bm{g}^\odot)~\text{, with}~ \\
\begin{split}\nonumber
&\left\{\begin{array}{l}
 D_{\pi}(s, \bm{g}^\odot) \equiv D_{\pi}(s, a, \bm{g}^\odot), a \sim \pi(\cdot|s, \bm{g}^\odot) \\
D_{\pi}(s', \bm{g}^\odot) \coloneqq \infty ~\text{if}~ s' \text{is terminal and } h(s',\bm{g}^\odot)=0 \\
D_{\pi}(s', \bm{g}^\odot) \coloneqq 0 ~\text{if}~ h(s',\bm{g}^\odot)=1
\end{array}\right.
\end{split}
\end{align}

This results in an off-policy compatible policy evaluation process over a parallel MDP almost-identical to the task MDP, but adapted for $\bm{g}^\odot$, where all transitions yield ``reward'' $+1$ and states satisfying $\bm{g}^\odot$ are changed to terminal with state value $0$. Every time an infeasible target embedding is sampled for training, the update rule will gradually push the estimate towards $\infty$, for all sampled source state $s$.

Our design only learns $D_\pi$ in a way that can lead to $\tau$-feasibilities $p(D_\pi(s, \bm{g}^\odot) \leq \tau)$. For this purpose and the consideration for a unified design, we propose to use Eq.~\ref{eq:rule_feasibility_gamma} in conjunction with a C51-style distributional architecture \citep{bellemare2017distributional}, which outputs a distribution represented by a histogram over pre-defined supports. When we set the support of the estimated $D_\pi(s, \bm{g}^\odot)$ to be $[1, 2, \cdots, T]$ with $T$ sufficiently large, the learned histogram bins via Eq.~\ref{eq:rule_feasibility_gamma} will correspond to the probabilities of $p(D_\pi(s, \bm{g}^\odot) = t)$ for all $t \in \{1,\dots,T - 1\}$. This technique of using C51 distributional learning enables the extraction of $\tau$-feasibility $p(D_\pi(s, \bm{g}^\odot) \leq \tau)$ from a learned $T$-feasibility with $p(D_\pi(s, \bm{g}^\odot) = t)$ over $t \in \{1, \dots, T\}$, thus learning all $\tau$-feasibility with $\tau < T$ \textit{simultaneously}. Take the example of the $1$-step \Dyna{} agent we implemented for experiments (Sec.~\ref{sec:exp_main_dyna}): if the estimated histogram has little probability density for $p(D_\pi(s, \bm{g}^\odot) = 1)$, then the target (simulated next state) is likely hallucinated and should be rejected, avoiding a potential delusional value update.

Note that the C51 architecture also allows us to extract the distribution of $\gamma_\pi(s, \bm{g}^\odot,\tau)$, which, as defined in Sec.~\ref{sec:delusions_prelim} (Page.~\pageref{sec:delusions_prelim}), the cumulative discount with a chosen target. This is done via transplanting the output histogram of $D_\pi(s, \bm{g}^\odot)$ over $[1, 2, \dots, \tau, \tau+1, \tau+2, \dots]$ onto the changed support of $[\gamma^1, \gamma^2, \cdots, \gamma^\tau,\gamma^\tau,\gamma^\tau,\dots]$

\subsection{Training Data for Feasibility}
\label{sec:training_data}
With the proper learning rule and architecture, we now need to ensure that the evaluator have proper training data and does not become delusional. In Sec.~\ref{sec:delusions_prelim}, we mentioned the incompleteness of the relabeling strategies, which will be discussed in detail now: \textbf{1)} Certain relabeling strategies naturally create exposure issues, even for \Gzero{} targets. For instance, \futurestr{} only relabels with future observations, thus only exposes a learner to future feasible targets, confusing the evaluator when a ``past'' target is proposed during planning; \textbf{2)} Trajectory-level relabeling is, by design, limited. Short trajectories, common in many training procedures, cover limited portions of the state space and prevent evaluators from learning about distant targets, risking delusions when such distant targets are proposed. Short trajectories can be the product of experimental design (initial state distributions, maximum episode lengths \citep{erraqabi2022temporal}, or environment characteristics, \eg{}, density of terminal states).

Avoiding feasibility delusions requires learning from all kinds of targets, including those that can never be experienced. This is to counter the exposure bias - the discrepancy between (most existing) TAP agents' behaviors (involving all targets that can be generated) and training (learning from only experienced targets), identified in \citet{talvitie2014model}.

We introduce $2$ data augmentation (relabeling) strategies to expand training source-target pair distributions.

\subsubsection{\generatestr{}: Expose Candidates Targets (to be generated)}

The first strategy, named \generatestr{}, is to \textit{expose the targets that will be proposed during planning to the evaluator}, so that it can figure out if these targets are infeasible.

We can implement this as a Just-In-Time (JIT) relabeling strategy that relabels a sampled (un-relabeled) transition for training with a generated target (provided by the generator). We can expect \generatestr{} to be effective, as evaluators will get exposed to hallucinated targets that the generator could offer. Note that \generatestr{} requires the use of the generator, thus it incurs additional computational burden, depending on the complexity of target generation. The JIT-compatibility lowers the need for  storage and provides timely coverage of the generators’ changing outputs, especially helpful for non-pretrained generators. The idea behind \generatestr{} can be traced back to \citet{talvitie2014model}.

\subsubsection{\pertaskstr{}: Expose Experienced Targets Beyond the Episode}
The second strategy, named \pertaskstr{}, is to \textit{expose the evaluator to \textbf{all} targets $g(s^\odot)$ experienced before}, so that it could realize if some previously achieved targets not present in the current episode are infeasible from the current state.

\begin{figure}[htbp]
\centering
\captionsetup{justification = centering}
\includegraphics[width=0.8\textwidth]{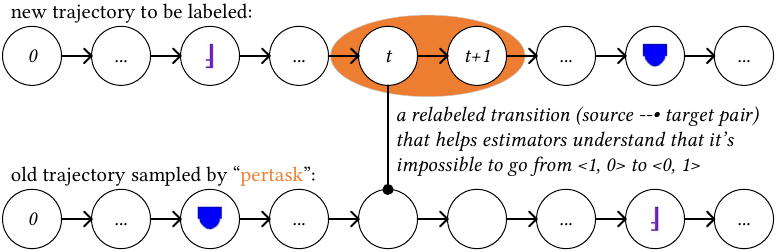}
\caption[An Example of How \pertaskstr{} Reduces E.2 Errors by Sampling Across Episodes]{\textbf{An Example of How \pertaskstr{} Reduces \Etwo{} Errors by Sampling Across Episodes}: The new trajectory acquired the sword first and the shield later, while the old trajectory acquired the shield first and then the sword. When relabeling a transition in the new trajectory (in $\langle 1, 0 \rangle$), a target observation in the old trajectory (in $\langle 0, 1 \rangle$) can be paired to make an agent realize the infeasibility of the relabeled target, thus reducing \Etwo{} delusions.
}
\label{fig:pertask_E2_example}
\end{figure}

We implement \pertaskstr{} by relabeling transitions with (the target embedding of) observations from the same training task, sampled across the entire replay. \pertaskstr{} can be seen as an generalization of the ``random'' strategy in \citet{andrychowicz2017hindsight} to multi-task training settings. Importantly, \pertaskstr{} exposes the evaluator to \Etwo{} delusions and to long-distance \Ezero{} caused by trajectory-level relabeling on short trajectories. An example is shown in Fig.~\ref{fig:pertask_E2_example}.

\subsubsection{Applicability}
\pertaskstr{} cannot address \Eone{} delusions. Meanwhile, \generatestr{} can be used against \textbf{some} \Gtwo{} targets that the generator hallucinates. \pertaskstr{} is a specialized and computationally efficient strategy to reduce feasibility delusions towards \textit{all} experienced \Gtwo{} target states and importantly also the long-distance \Ezero{} errors that \generatestr{} cannot handle. \pertaskstr{} is expected to be more effective than \generatestr{} in generalization-focused scenarios, where the distribution of \Gzero{} \& \Gtwo{} targets proposed by the generator during evaluation can go beyond those trained under \generatestr{}.

Importantly, relabeling strategies such as \futurestr{}, \episodestr{} and \pertaskstr{} rely on the existence of $g$ that maps a state into a target embedding, which is commonly found in TAP agents \citep{andrychowicz2017hindsight}. However, if only the target set indicator function $h$ is available, we may need to accumulate $\langle s, \bm{g} \rangle $ tuples for which $h(s,\bm{g})=1$, and the use them to train a $g$. Or, in the cases where feasibility is only used for rejection, such as when dealing with simulated experiences and tree search, we could also rely on only \generatestr{}, which does not require $g$; Sometimes, it is $h$ that needs to be constructed. We provide detailed discussions for applying our solution on \Dreamer{}V2 in Sec.~\ref{sec:appendix_dreamer}, with a focus on how to construct a proper $h$.

\subsubsection{Mixtures}
\label{sec:mixtures}

Both \generatestr{} \& \pertaskstr{} bias the training data distribution, making the evaluator spread out its learning efforts to the source-target pairs possibly distant from each other. Despite increasing training data diversity, distant pairs are less likely to contribute to better evaluation compared to the closer in-episode ones offered by \episodestr{}, as close-proximity \Gzero{} targets matter the most.

\begin{figure*}[htbp]
\centering
\captionsetup{justification = centering}
\includegraphics[width=0.9\textwidth]{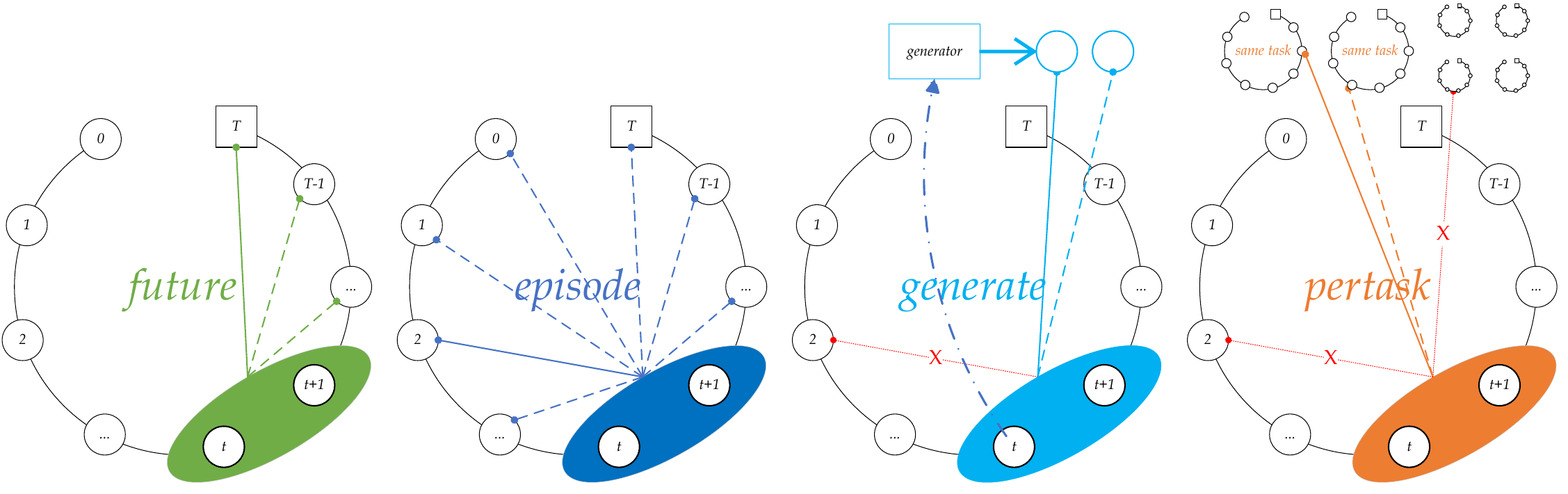}
\caption[Representative Atomic Hindsight Relabeling Strategies \& Newly Proposed Ones]{\textbf{Representative Atomic Hindsight Relabeling Strategies \& Newly Proposed Ones}: \textit{The first two strategies, \futurestr{} and \episodestr{}, are widely used as they create relabeled transitions that help evaluators efficiently handle \Gzero{} target states during planning. The last two, \generatestr{} and \pertaskstr{}, are effective at addressing delusions, making them useful in specific scenarios.}
}
\label{fig:atomic_strategies}
\end{figure*}

Creating a mixture of sources of training data can increase the diversity of source-target combinations. For HER specifically, each atomic strategy, enumerated in Tab.~\ref{tab:atom_hindsight_strategies} and illustrated in Fig.~\ref{fig:atomic_strategies}, exhibits different estimation accuracies for different types of source-target pairs, including short-distance and long-distance ones involving all of \Gzero{}, \Gone{} and \Gtwo{}.

When the training budget is fixed, \ie{}, training frequency, batch sizes, \etc{}, stay unchanged, the mixing proportions of strategies pose a tradeoff to the learning of different kinds of source-target pairs. In the experiments, we mainly used \FEPG{}, which is a mixture of $2/3$ \episodestr{} and $1/3$ \pertaskstr{}, with $1/4$ chance using \generatestr{} JIT, resulting in a mixture of $50\%$ \episodestr{}, $25\%$ \pertaskstr{} and $25\%$ \generatestr{}. \FEPG{} exploits the fact that assisting \episodestr{} with \generatestr{} and \pertaskstr{} often results in better performance in evaluator training, striking a balance between the investment of training budgets for the feasible and infeasible targets \citep{nasiriany2019planning,yang2021mher}.

\subsection{Computational Overhead}
The portion of overhead for the evaluation of targets is straightforward, as each target will be fed into the neural networks (paired with a source state) for a forward pass at inference time. This portion of the overhead depends on the complexities of evaluator’s architecture and of the state / target representations. We can expect fast evaluations with lightweight evaluators.

It is the strategy post evaluation that determines the overall overhead, which depends on the planning behavior of the TAP agent that the evaluator is attached to. For background TAP agents that generate batches of targets, the improper ones can be rejected and the whole batch can be all rejected without problem (no \Dyna{} update this time). For decision-time TAP agents, targets act as subgoals and when they are rejected, the agent can either re-generate or commit to more random explorations.

\begin{table*}[htbp]
\centering
\aboverulesep=0ex
\belowrulesep=0ex
 \resizebox{\textwidth}{!}{
 \begin{NiceTabular}{c|m[l]{0.25\textwidth}|m[l]{0.5\textwidth}|m[l]{0.3\textwidth}}
 \toprule
 \toprule
 \textbf{Strategies} & \multicolumn{1}{c|}{\textbf{Advantages}} & \multicolumn{1}{c|}{\textbf{Disadvantages}} & \multicolumn{1}{c}{\textbf{Gist}} \\
 \midrule
 \episodestr{} & Efficient for evaluator to learn close-proximity relationships & When used exclusively to train evaluator, 1) cannot handle \Etwo{} and 2) prone to \Ezero{} - cannot learn well from short trajectories; Can cause \Gtwo{} target states when used to train generators & 
 Creates training data with source-target pairs sampled from the same episodes \\
 \bottomrule
 \futurestr{} & Can be used to learn a conditional generator with temporal abstractions & In addition to the shortcomings of \episodestr{} (those for evaluators only), this additionally causes \Ezero{} when used as the exclusive strategy for evaluator training & Creates training data with temporally ordered source-target pairs from the same episodes \\
 \bottomrule
 \generatestr{} & Addresses \Eone{} with data diversity (also \Etwo{} when generator produces \Gtwo{}) & Relies on the generator with additional computational costs; Potentially low efficiency in reducing \Ezero{}. & Augments training data to include candidate targets proposed at decision time \\
 \bottomrule
 \pertaskstr{} & Addresses evaluator delusions (\Etwo{} \& \Ezero{} for long-distance pairs) & low efficiency in learning close-proximity source-target relationships & Augments training data to include targets that were experienced \\
 \bottomrule
 \bottomrule
 \end{NiceTabular}%
 }
\caption[Detailed Comparison of Atomic Hindsight Relabeling Strategies]{\textbf{Detailed Comparison of Atomic Hindsight Relabeling Strategies}: \textit{\episodestr{} and \futurestr{} are widely used as they increase sample efficiency towards \Gzero{} states significantly; \generatestr{} and \pertaskstr{}, proposed in this work, should be applied against delusions in relevant scenarios.}
}
\label{tab:atom_hindsight_strategies}
\end{table*}

\section{Discussions of Methodologies}
\label{sec:delusions_related_work}

\subsection{About TAP Agents}
Most rollout-based TAP methods are oblivious to model hallucinations and utilize all generated targets without question. These include fixed-step background methods such as  \citet{sutton1991dyna,kaiser2019model,yun2024guided,lee2024gta} and decision-time methods based on tree-search, such as \citet{schrittwieser2019mastering,hafner2019learning,zhao2021consciousness,zhang2024focus}; TAP methods compatible with arbitrarily distant targets ($\tau=\infty$) often struggle to produce non-delusional feasibility-like estimates for hallucinated targets. Thus, they cannot properly reject infeasible targets despite having their own ``evaluators''. These include background methods such as \citet{lo2024goal} and decision-time methods for path planning \citep{nasiriany2019planning,yu2024trajectory,duan2024learning}, OOD generalization \citep{zhao2024consciousness}, and task decomposition \citep{zadem2024reconciling}.

\subsection{About Hindsight Relabeling}
\label{sec:related_work_HER}

Hindsight relabeling is highlighted for its improved sample efficiency towards \Gzero{} targets, around which most follow-up works revolved as well \citep{andrychowicz2017hindsight,dai2021diversity}. However, sample efficiency is not the only concern in TAP agents, as delusions toward  generated targets can cause delusional behaviors leading to other failure modes.
\citet{shams2022addressing} studied the sample efficiency of atomic strategies, without looking into their failure modes. \citet{deshpande2018improvements} detailed experimental techniques in sparse reward settings using \futurestr{}. In \citep{yang2021mher}, a mixture strategy similar to \generatestr{} improved estimation of feasible targets, though its impact on hallucinated targets was not investigated. Note that the performance of existing HER-trained agents is often limited by their reliance on \futurestr{} or \episodestr{}, whose delusions this paper intends to address.

Learning feasibility of targets has long been on the list of many research agendas for its wide applicability \citep{tian2020model,bae2024tldr,myers2024learning}. Our way of using hindsight relabeling to create source-target pairs (instead of using the next states / observations as targets) circumvents issues of learning Monte-Carlo distance functions, which were heavily investigated in \citet{akella2023distributional}. The mixture strategies also circumvent the failure modes of hindsight relabeling that were presented in \citet{eysenbach2021clearning}.

\subsection{About Delusions \& Delusional Behaviors}
Delusional value estimates of hallucinated states are hypothesized to plague background planning \citep{jafferjee2020hallucinating}. \citet{lo2024goal} introduced a temporally-abstract background TAP method to limit temporal-difference updates to only a few trustworthy targets. \citet{langosco2022goal} classified goal mis-generalization, a behavior describing when an agent competently pursues a problematic target, a form of delusional planning behaviors. \citet{talvitie2017self} tried to trains the model to correct itself when error is produced. \citet{zhao2024consciousness} gave first examples of delusional behaviors caused by hallucinated targets in decision-time TAP agents. Recently, when I watched David Silver's keynote presentation during the first Reinforcement Learning Conference (RLC), I came to know that previously his AlphaGo team struggled with delusions in the Shoji game. He claimed that his solution was ``trial and error'' and I suspect that it would be approximately a simulation-based variant of \generatestr{}. Previously, there were method-specific approaches proposed against delusional planning behaviors, such as by constraining to certain probabilistic models \citep{deisenroth2011pilco,chua2018deep}, or training a target evaluator separately on a collected dataset. The proposed solution in this work, however, is an add-on auxiliary learner that does not seek to introduce additional constraints or changes to the generative models.

\subsection{About Goal Mis-Generalization}
\label{sec:goal_misgen}

\citet{langosco2022goal} proposed a dichotomy of two dimensions to analyze the failure modes for goal-conditioned RL agents: capability mis-generalization \vs{} goal mis-generalization, corresponding to \textit{the competence of implementing targets} and \textit{the validity of targets}.

While such a perspective is a sensible abstraction from a behavioral standpoint, our work was not established on this perspective. Nevertheless, we are happy to discuss the connections between our work and \citet{langosco2022goal}:

Directly reducing hallucination (which is not the strategy of this work) directly corresponds to reducing goal mis-generalization. However, in TAP frameworks, goal mis-generalization is also reduced by the estimator that can reject ``mis-generalized targets''.

Reducing feasibility delusions about targets (with our proposed strategies) reduces both capability mis-generalization and goal mis-generalization. To see both sides, we must point out that 1) evaluating a proposed target requires capability generalization, since it includes the understanding of if the agent can reach certain target and how the agent would reach it (tied to the agent’s capabilities); and 2) fewer feasibility delusions about targets in TAP frameworks means problematic targets are less likely favored.

\section{Research Findings: Experiments}
\label{sec:delusions_exp}

To investigate the effectiveness and generality of our proposed solution against delusional behaviors caused by hallucinated targets, we use a simple and unified implementation of our evaluator (3-layers of \codeword{ReLU} activated \codeword{MLP} with output bin $T=16$) for $8$ sets of experiments, encompassing decision-time \vs{} background planning, TAP methods compatible with arbitrary $\tau$s and fixed $\tau$s, singleton and non-singleton targets, on controlled environments with respective emphases on \Gone{} and \Gtwo{} difficulties. The implementations of our solutions for these experiments can be extended to various existing TAP methods, as suggested in Tab.~\ref{tab:methods}.

\begin{enumerate}[itemsep=0mm,leftmargin=*,start=1,label={\bfseries Exp.$\nicefrac{\arabic*}{8}$:}]
\item \Skipper{} (a decision-time TAP method with singleton targets, compatible with arbitrary $\tau$) on \SSM{}
\item \LEAP{} (a decision-time TAP method with singleton targets, compatible with arbitrary $\tau$) on \SSM{}
\item \Skipper{} on \RDS{}
\item \LEAP{} on \RDS{}
\item \Dyna{} (a background TAP method with singleton targets, $\tau=1$) on \SSM{}
\item \Dyna{} on \RDS{}
\item Feasibility estimation of \textit{non-singleton} targets with arbitrary $\tau$ on \SSM{}
\item Feasibility of \textit{non-singleton} targets with arbitrary $\tau$ on \RDS{}
\end{enumerate}

\subsubsection{Data Augmentation Strategies}:
In the following sub-sections, we will focus on a particular implementation of evaluator which utilizes \FEPG{} for data augmentation. Since we have the full degrees of freedom in deciding the mixture ratios of the involved relabeling strategies, \ie{}, \episodestr{}, \generatestr{} \& \pertaskstr{}, we will provide more results that encompass more relabeling strategies. These results could not only provide the readers with more understanding of the empirical characteristics of the relabeling strategies but also can serve as an ablation test for the two alternative relabeling strategies, \ie{}, \generatestr{} \& \pertaskstr{}. The variant relabeling strategies are as follows:

\begin{itemize}[leftmargin=*]
\item \textbf{\FEG{}} - a mixture against \Eone{}. \episodestr{} with $50\%$ chance using \generatestr{} JIT, resulting in a half-half mixture of \episodestr{} \& \generatestr{}
\item \textbf{\FEP{}} -  against \Etwo{}. Half \episodestr{} \& half \pertaskstr{}
\end{itemize}

We can see that \FEPG{} is the middle ground between \FEG{} and \FEP{}, with a comprehensive coverage for both \Gzero{}, \Gone{} and \Gtwo{} cases. This is why we have chosen \FEPG{} as the default for our evaluator, since we do not wish to assume access to the state space structures of the environments. Note that for the readers' convenience, we have used consistent colors for each variant throughout.

\subsection{Rejecting Infeasible Goals in Decision-Time Planning \texorpdfstring{(Exp.~$\nicefrac{1}{8}$ - $\nicefrac{4}{8}$)}{(Exp.~1/8 - 4/8)}}

For decision-time TAP agents, we are interested in understanding how rejecting hallucinated targets can influence their abilities to generalize their learned skills after learning from a limited number of training tasks. This also means, the evaluator is expected to learn to generalize its identification of infeasible targets in novel situations (with consistent dynamics), by identifying the patterns of the infeasible targets.

For such experimental purpose, we use distributional shifts provided in \SSM{} to simulate real-world OOD systematic generalization scenarios in evaluation tasks \citep{frank2009connectionist}. For each seed run on \SSM{}, we sample and preserve $50$ training tasks of size $12\times12$ and difficulty $\delta=0.4$. For each episode, one of the $50$ tasks is sampled for training. Agents are trained for $1.5 \times 10^{6}$ interactions in total. To speed up training, we make the initial state distributions span all the non-terminal states in each training task, making trajectory-level relabeling even more problematic.

\subsubsection{Methods}
To demonstrate the generality of our proposed solution against hallucinated targets for decision-time TAP, we apply it onto two methods which utilize targets differently:

\paragraph{\Skipper{}} \citep{zhao2024consciousness}: generates candidate target states that, together with the current state, constitute the vertices of a directed graph for task decomposition, while the edges are pairwise estimations of cumulative rewards and discounts, under its evolving policy. A target is chosen after applying \textit{value iteration}, \ie{}, the values of targets are the $\scriptU$ values of the planned paths.\footnote{As shown in Tab.~\ref{tab:methods}, our adaptation for \Skipper{} can be extended to methods utilizing arbitrarily distant targets, including background TAP methods such as \GSP{} \citep{lo2024goal}}

\paragraph{\LEAP{}} \citep{nasiriany2019planning}: \LEAP{} uses the cross-entropy method to evolve the shortest sequences of sub-goals leading to the task goal \citep{rubinstein1997optimization}. The immediate sub-goal of the elitist sequence is then used to condition a lower-level policy. Compared to \Skipper{}, \LEAP{} is more prone to delusional behaviors, since one hallucinated sub-goal can render a whole sub-goal sequence delusional.\footnote{As shown in Tab.~\ref{tab:methods}, our implementation for \LEAP{} can be extended to planning methods proposing sub-goal sequences, such as \PlaNet{} \citep{hafner2019learning}}

For Exp.~$\nicefrac{1}{8}$ - $\nicefrac{4}{8}$, targets are observation-like generations, where \Gone{} \& \Gtwo{} can be clearly identified. See Chap.~\ref{cha:appendix} on Page.~\pageref{sec:details_implement} for more implementation details.

\subsubsection{Evaluative Metrics}
\paragraph{Feasibility Errors} At each evaluation timing, we use the average errors of $\hat{\doubleE}[D_\pi]$ against the ground truths as a proxy to understand the convergence of the evaluators' estimated feasibility of targets.

\paragraph{Delusional Behavior Frequencies} We monitor the frequency of a hallucinated target (made of \Gone{} and \Gtwo{}) being chosen by the agents (as the next sub-goal for \Skipper{}, as a part of the sub-goal chain for \LEAP{}), \ie{}, delusional planning behaviors.

\paragraph{OOD Generalization Performance}
We analyze the changes in agents' OOD generalization performance. The evaluation tasks (targeting systematic generalization) are sampled from a gradient of OOD difficulties - $0.25$, $0.35$, $0.45$ and $0.55$. Because of the lack of space, we present the ``aggregated'' OOD performance, such as in Fig.~\ref{fig:Skipper_SSM} \textbf{d)}, by sampling $20$ task instances from each of the $4$ OOD difficulties, and combine the performance across all $80$ episodes, which have a mean difficulty matching the training tasks. To maximize the evaluation difficulty, the initial state is fixed in each evaluation task instance: the agents are not only spawned to be at the furthest edge to the monster, but also in semantic class $\langle 0, 0 \rangle$, \ie{}, with neither the sword nor the shield in hand.

\subsubsection{\Skipper{} on \SSM{} \texorpdfstring{(Exp.~$\nicefrac{1}{8}$)}{(Exp.~1/8)}}

\begin{figure*}[htbp]
\centering

\subfloat[\textit{Evolution of \Eone{} Errors}]{
\captionsetup{justification = centering}
\includegraphics[height=0.31\textwidth]{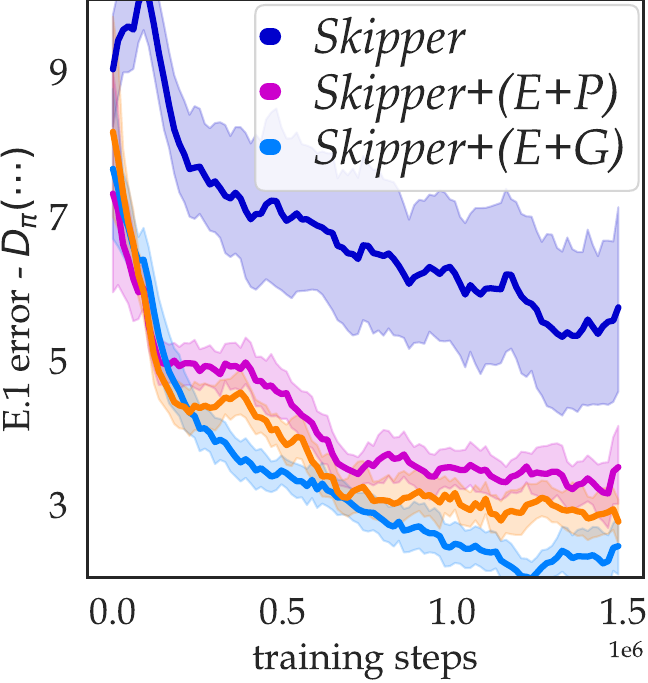}}
\hfill
\subfloat[\textit{Evolution of \Etwo{} Errors}]{
\captionsetup{justification = centering}
\includegraphics[height=0.31\textwidth]{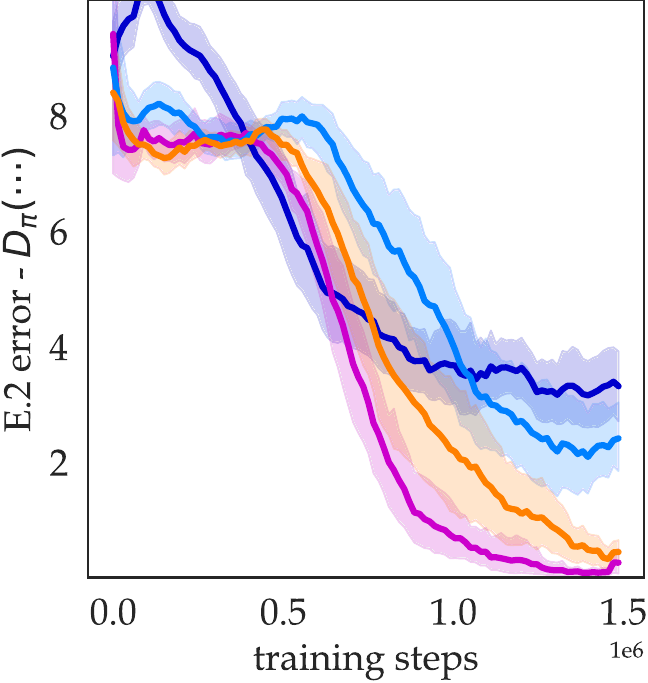}}
\hfill
\subfloat[\textit{Final \Ezero{} Errors by Distance}]{
\captionsetup{justification = centering}
\includegraphics[height=0.31\textwidth]{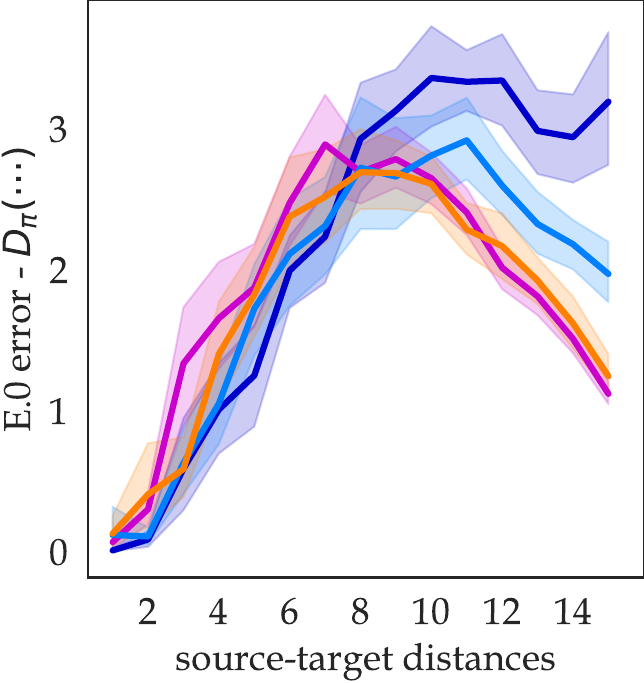}}

\subfloat[\textit{Evolution of \Eone{} Behavior Ratio}]{
\captionsetup{justification = centering}
\includegraphics[height=0.31\textwidth]{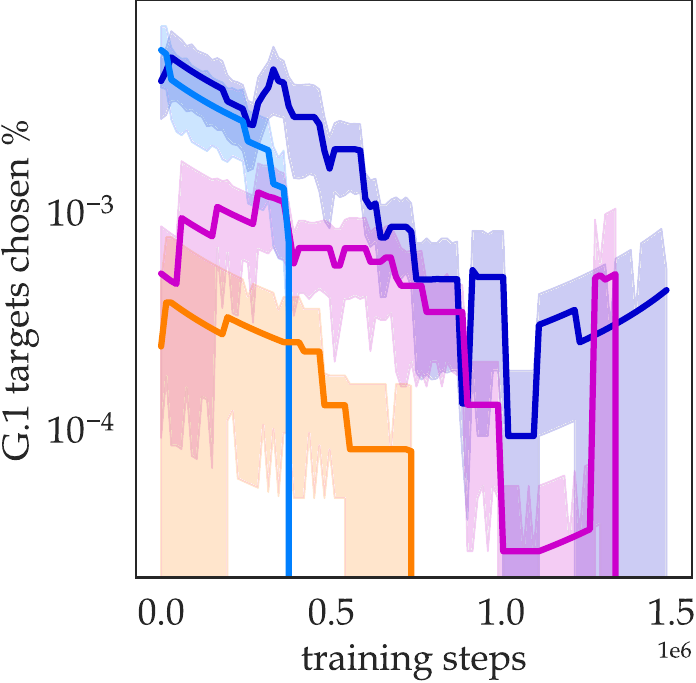}}
\hfill
\subfloat[\textit{Evolution of \Etwo{} Behavior Ratio}]{
\captionsetup{justification = centering}
\includegraphics[height=0.31\textwidth]{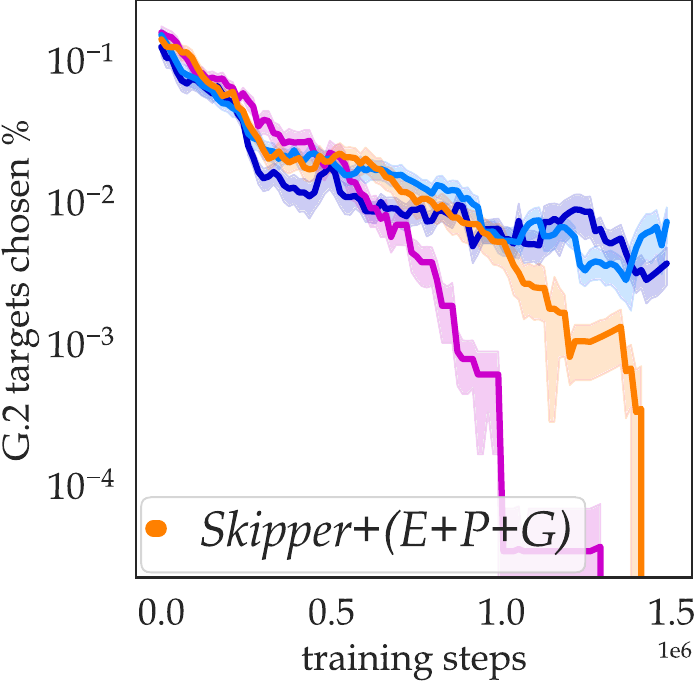}}
\hfill
\subfloat[\textit{Aggregated OOD Performance}]{
\captionsetup{justification = centering}
\includegraphics[height=0.31\textwidth]{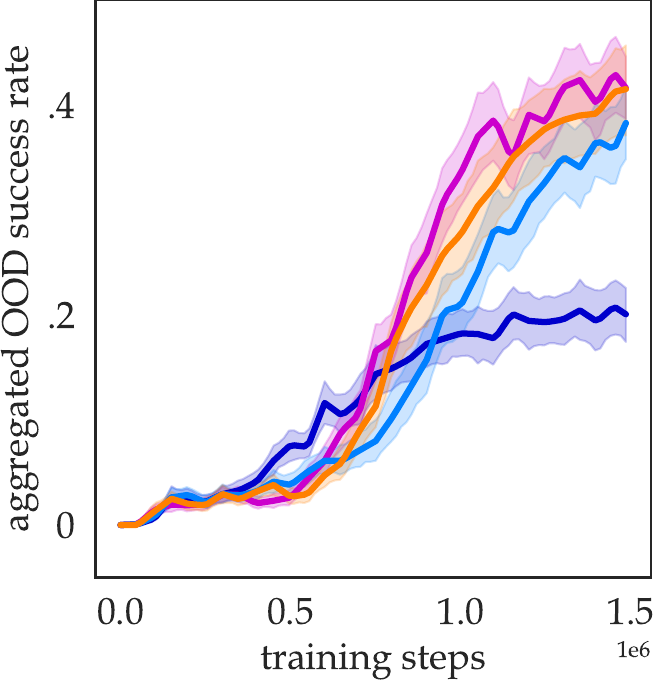}}

\caption[Details of Skipper's Performance on SSM]{\textbf{Details of \Skipper{}'s Performance on \SSM{}}: All error bars (95\%-CI) are established over $20$ seed runs. We compare the original form of \Skipper{}, which learns its own feasibility estimates of target states in its own way, against three \Skipper{}+ variants, which have our proposed evaluator injected to assist the evaluation of the feasibility of targets, powered by the \FEG{}, \FEP{} \& \FEPG{} relabeling strategies, respectively. \textbf{a)}: Final \Ezero{} errors separated across a range of ground truth distances. Both estimated and true distances are conditioned on the evolving policies; \textbf{b)}: \Eone{} errors measured as $L_1$ error in estimated (clipped) distance throughout training; \textbf{c)}: \Gtwo{}-counterparts of \textbf{b}; \textbf{d)} \& \textbf{e)}: the changes of frequencies in delusional behaviors throughout training, for \Gone{} and \Gtwo{} composed targets, respectively. The curves denote the frequencies of \Gone{} and \Gtwo{} targets becoming the imminent subgoals that \Skipper{} seeks to achieve next; \textbf{f)}: Each data point represents OOD evaluation performance aggregated over $4 \times 20$ newly generated tasks, with mean difficulty matching training. The decomposed results for each OOD difficulty are presented in Fig.~\ref{fig:SSM_perfevol}.
}
\label{fig:Skipper_SSM}
\end{figure*}

We compare the original form of \Skipper{} with \SkipperPlus{}, a variant that is assisted by the proposed evaluator. Details of the variants are shown in the captions of Fig.~\ref{fig:Skipper_SSM}.

\begin{figure*}[htbp]
\centering

\subfloat[\textit{\Gone{} Candidate Ratio in \SSM{}}]{
\captionsetup{justification = centering}
\includegraphics[height=0.32\textwidth]{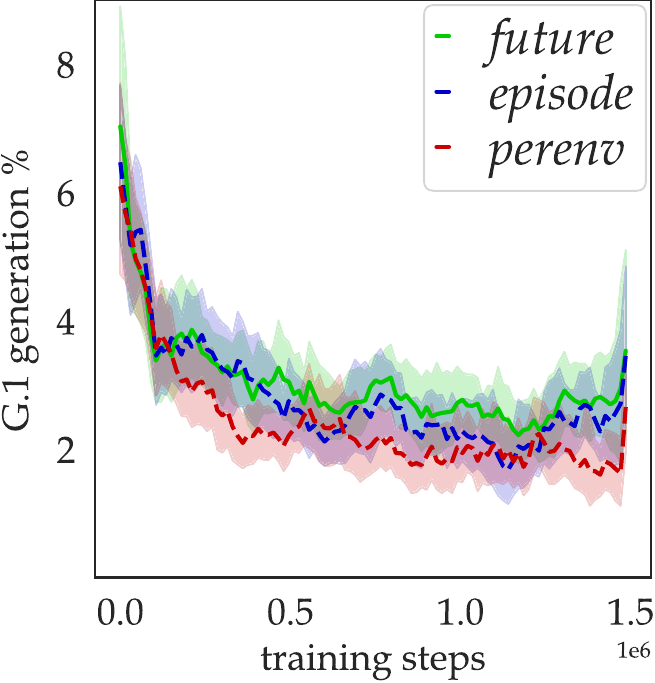}}
\hfill
\subfloat[\textit{\Gtwo{} Candidate Ratio in \SSM{}}]{
\captionsetup{justification = centering}
\includegraphics[height=0.32\textwidth]{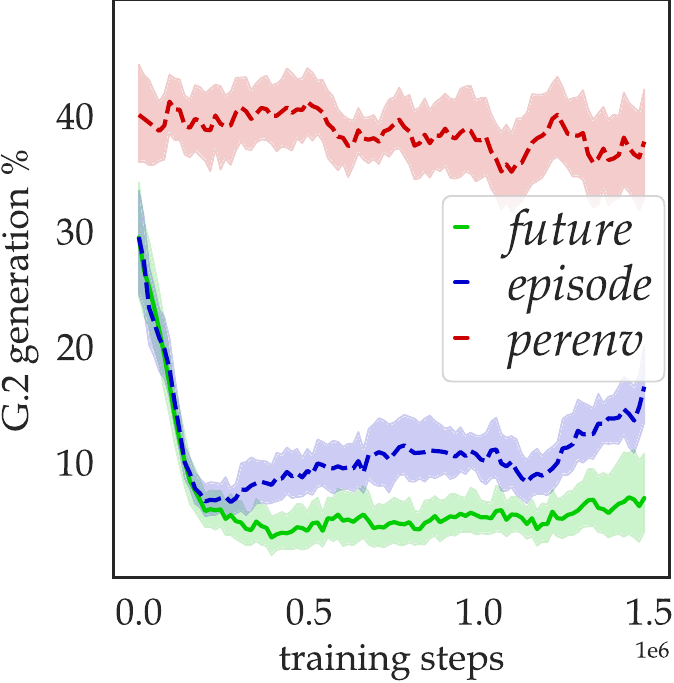}}
\hfill
\subfloat[\textit{\Gone{} Candidate Ratio in \RDS{}}]{
\captionsetup{justification = centering}
\includegraphics[height=0.32\textwidth]{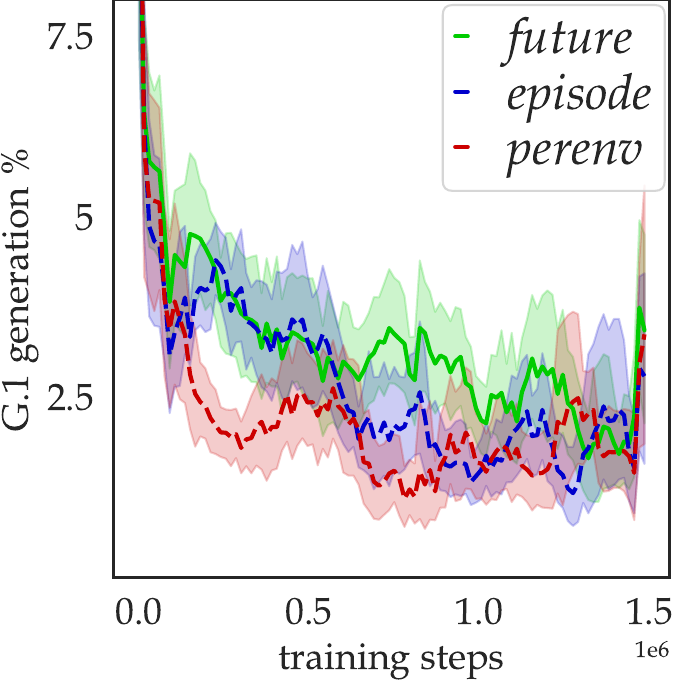}}

\caption[Hallucination Frequencies]{\textbf{Hallucination Frequencies}: All error bars (95\%-CI) are established over $20$ seed runs. \textbf{a)} Evolving ratio of \Gone{} ``states'' among all candidates at each target selection, throughout training; Subfigure \textbf{b)} is the \Etwo{}-counterpart of \textbf{a)} on \SSM{}; Subfigure \textbf{c)} is the \RDS{}-counterpart of \textbf{a)}.
}
\label{fig:hallucination_frequencies}
\end{figure*}

\phantomsection
\label{sec:hallucination_freqs}
\paragraph{Hallucination} We first investigate generator's rates of hallucinations. As shown in Fig.~\ref{fig:hallucination_frequencies}, the generator produces targets that correspond to \Gone{} and \Gtwo{} with the rate of around $3\%$ and $5\%$, respectively.

We use hindsight-relabeled transitions to train the generators in the two methods, to demonstrate how different ways of training the generator could affect the rates of hallucinations. \Gtwo{} can appear more frequently if the generator is trained to imagine more diverse kinds of targets than needed. For example, a conditional target generator which learns from \episodestr{} will be more likely to produce \Gtwo{} target states (compared to \futurestr{}). This was why we mostly used \futurestr{} to train the generators in the related experiments.

For the HER-trained generators, Fig.~\ref{fig:hallucination_frequencies} \textbf{a)}, shows that different training targets for the generator could lead to different degrees of hallucinations, in terms of \Gone{} and \Gtwo{}, but not $0$. Importantly, Fig.~\ref{fig:hallucination_frequencies} \textbf{b)} indicates that, \futurestr{} generates \Gtwo{} target states significantly less frequently than \episodestr{} and \pertaskstr{}, as the other two wasted training budget on \Gtwo{} target states, especially \pertaskstr{} that brings in more problematic training samples from long distances. \textit{In all other experiments, we only compare variants with \futurestr{} for the generator training.}

The generator is consistently used for both \Skipper{} and \LEAP{} in these 4 sets of experiments.

\paragraph{Feasibility Errors}
\Skipper{} relies on a built-in cumulative discount estimator whose estimations can be converted to feasibility estimates that our evaluator seeks to learn. Thus, we can examine the errors of the feasibility estimates corrected by the injected evaluator to understand how the proposed evaluator could reduce feasibility delusions of arbitrary-horizon TAP methods. From Fig.~\ref{fig:Skipper_SSM} \textbf{b}) and \textbf{c)}, we can see that feasibility estimates corrected by our evaluator have significantly less errors compared to the original, towards both \Gone{} and \Gtwo{} targets. As a perk for \SkipperPlus{}'s utilization of \pertaskstr{} for \Etwo{} delusions (included in \FEPG{}), its positive effect on far-away \Gzero{} targets are also shown in Fig.~\ref{fig:Skipper_SSM} \textbf{a)}. It can be seen that the evaluator is generally helpful for \Skipper{} to understand the feasibility of all \Gzero{}, \Gone{} and \Gtwo{} targets.

\paragraph{Frequency of Delusional Plans}
The purpose of identifying infeasible targets is to reduce delusional plans that involve them. We provide detailed results on this in Fig.~\ref{fig:Skipper_SSM}, where we observed significant reduction in delusional plans involving both \Gone{} and \Gtwo{} targets.

\paragraph{Generalization}
Comparing \Skipper{} and \SkipperPlus{}, we can deduce from Fig.~\ref{fig:Skipper_SSM} that generally, lower \Etwo{} errors (\textbf{c}) lead to less frequent delusional behaviors (\textbf{d} \& \textbf{e})), which in turn improves the OOD performance in \textbf{f)}. This indicates that rejecting infeasible targets can help decision-time TAP agents in systematic OOD generalization.

\begin{figure}[htbp]
\centering

\subfloat[training, $\delta=0.4$]{
\captionsetup{justification = centering}
\includegraphics[height=0.22\textwidth]{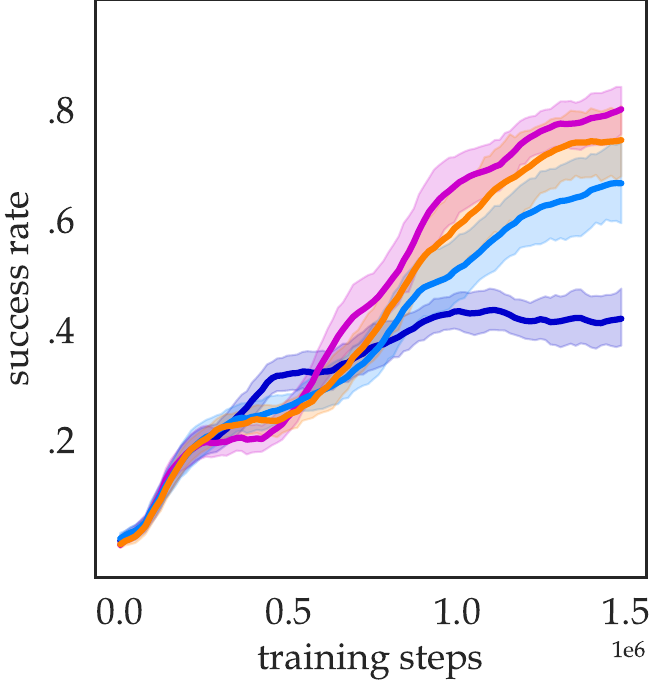}}
\hfill
\subfloat[OOD evaluation, $\delta=0.25$]{
\captionsetup{justification = centering}
\includegraphics[height=0.22\textwidth]{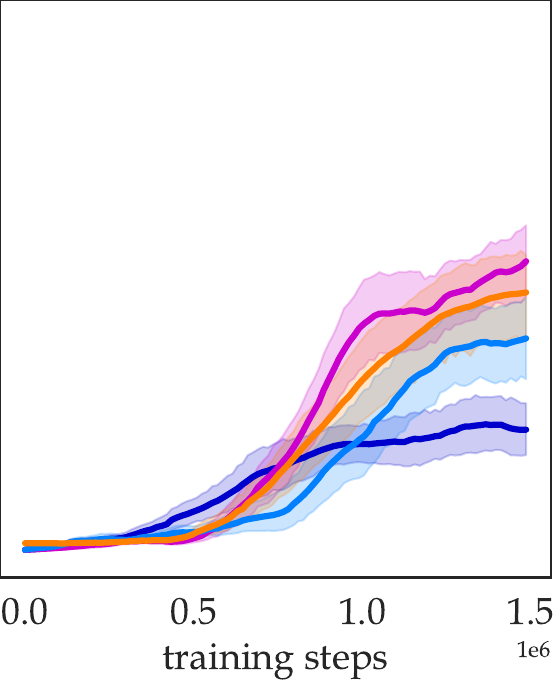}}
\hfill
\subfloat[OOD evaluation, $\delta=0.35$]{
\captionsetup{justification = centering}
\includegraphics[height=0.22\textwidth]{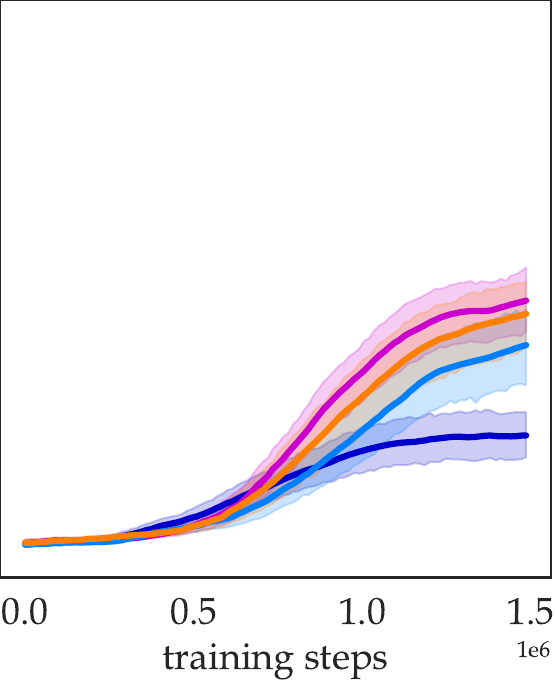}}
\hfill
\subfloat[OOD evaluation, $\delta=0.45$]{
\captionsetup{justification = centering}
\includegraphics[height=0.22\textwidth]{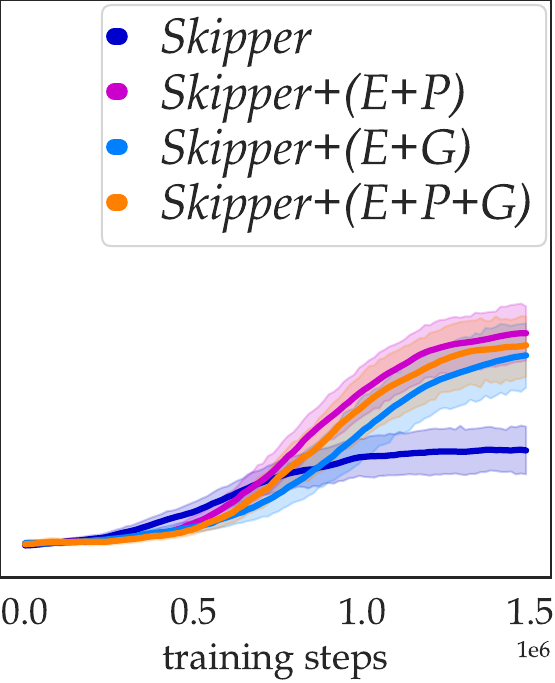}}
\hfill
\subfloat[OOD evaluation, $\delta=0.55$]{
\captionsetup{justification = centering}
\includegraphics[height=0.22\textwidth]{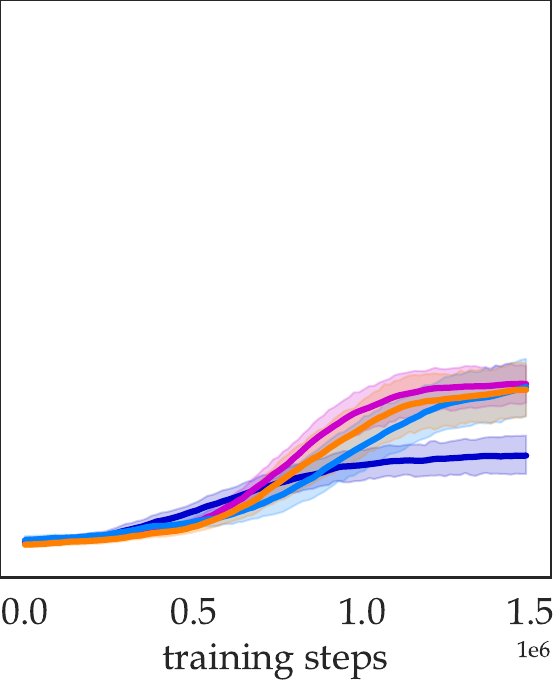}}
\caption[Evolution of OOD Performance of \Skipper{} Variants on \SSM{}]{\textbf{Evolution of OOD Performance of \Skipper{} Variants on \SSM{}}: All error bars (95\%-CI) are established over $20$ seed runs. 
}
\label{fig:SSM_perfevol}
\end{figure}

\paragraph{Breakdown of Task Performance}
In Fig.~\ref{fig:SSM_perfevol}, we present the evolution of \Skipper{} variants' performance on the training tasks as well as the OOD evaluation tasks throughout the training process. Note that Fig.~\ref{fig:Skipper_SSM} \textbf{h)} is an aggregation of all $4$ sources of OOD performance in Fig.~\ref{fig:SSM_perfevol} \textbf{b-e)}.

From the performance advantages of the hybrid variants (in both training and evaluation tasks), we can see that learning to address delusions during training brings better understanding for novel situations posed in OOD tasks.

\subsubsection{\LEAP{} on \SSM{} \texorpdfstring{(Exp.~$\nicefrac{2}{8}$)}{(Exp.~2/8)}}
This set of experiments seeks to demonstrate that the proposed feasibility evaluator is applicable to other decision-time TAP agents, utilizing their generators in different ways. For this purpose, we study \LEAP{} performance on \SSM{}, with or without the help of the target rejection provided by the feasibility estimators.

\LEAP{} is different from \Skipper{}, as its decision-time planning process constructs a singular sequence of subgoals leading to the task goal. Due to a lack of backup subgoals, even if one among them is problematic, the whole resulting plan would be delusional, making \LEAP{} much more prone to failures compared to \Skipper{}, where candidate targets can still be reused if deviation from the original plan occurred.

\SSM{} has a relatively large state space that requires more intermediate subgoals for \LEAP{}'s plans. However, an increment of the number of subgoals also dramatically increases the frequencies of delusional plans, damaging the agents' performance. Because of this, our experimental results of \LEAP{} on \SSM{} with size $12 \times 12$ became difficult to analyze because of the rampant failures. We chose instead to present the results on \SSM{} with size $8 \times 8$ here.

\begin{figure*}[htbp]
\centering

\subfloat[\textit{\Gone{} Ratio Planned}]{
\captionsetup{justification = centering}
\includegraphics[height=0.19\textwidth]{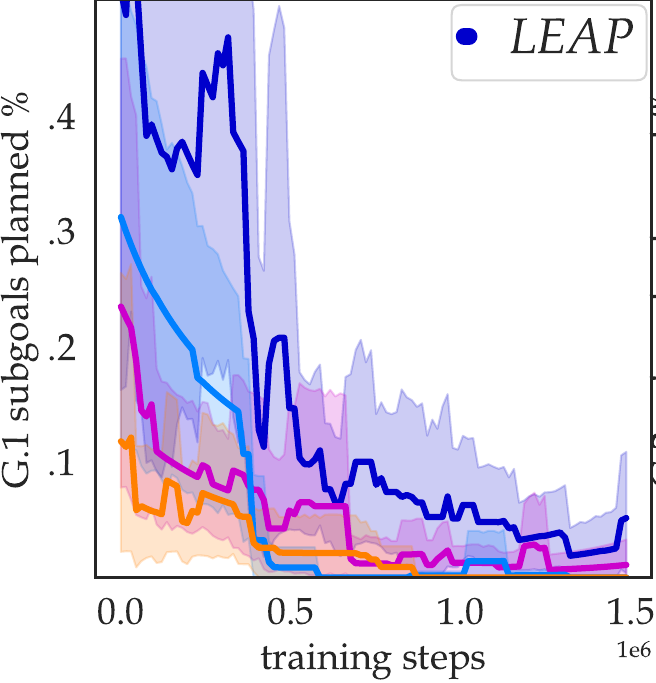}}
\hfill
\subfloat[\textit{\Gtwo{} Ratio Planned}]{
\captionsetup{justification = centering}
\includegraphics[height=0.19\textwidth]{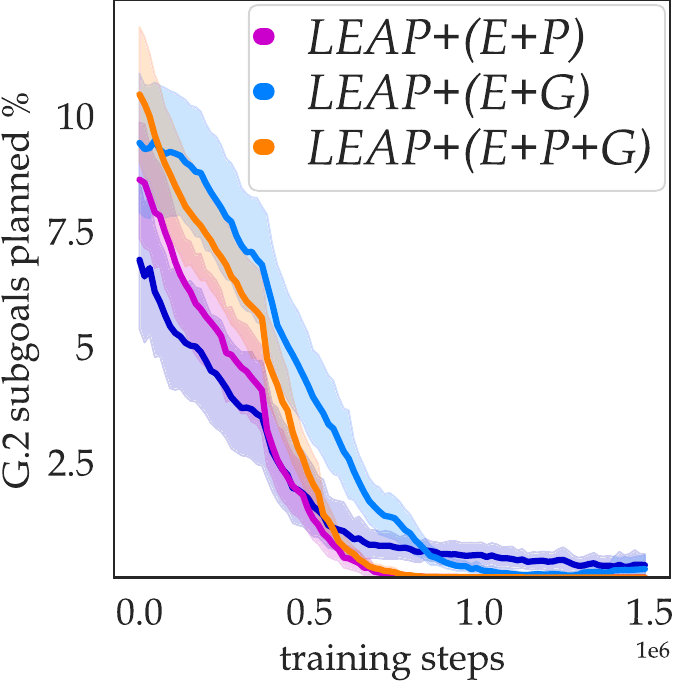}}
\hfill
\subfloat[\textit{Delusional Plan Ratio}]{
\captionsetup{justification = centering}
\includegraphics[height=0.19\textwidth]{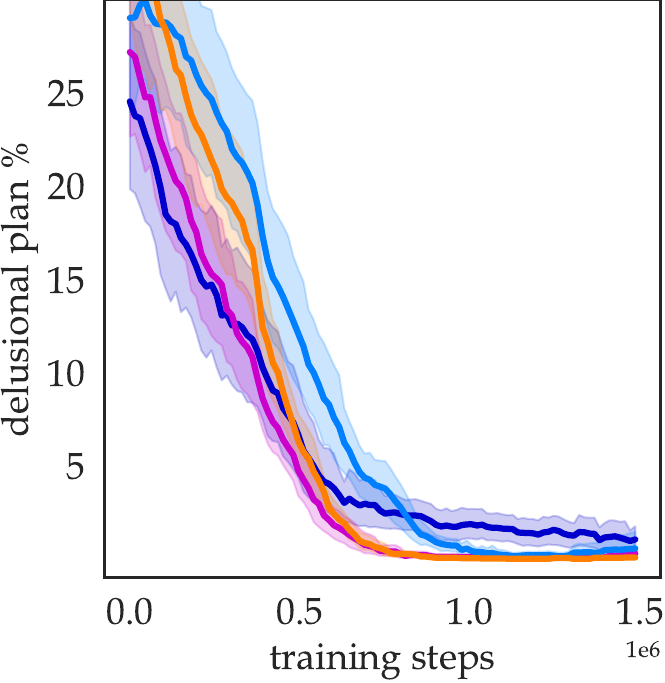}}
\hfill
\subfloat[\textit{\Ezero{} Errors}]{
\captionsetup{justification = centering}
\includegraphics[height=0.19\textwidth]{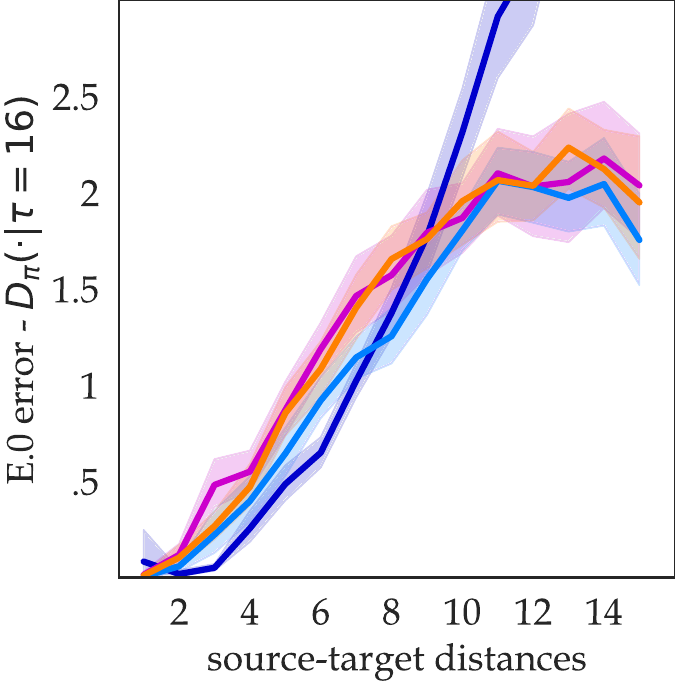}}
\hfill
\subfloat[\textit{Aggregated OOD Perf.}]{
\captionsetup{justification = centering}
\includegraphics[height=0.19\textwidth]{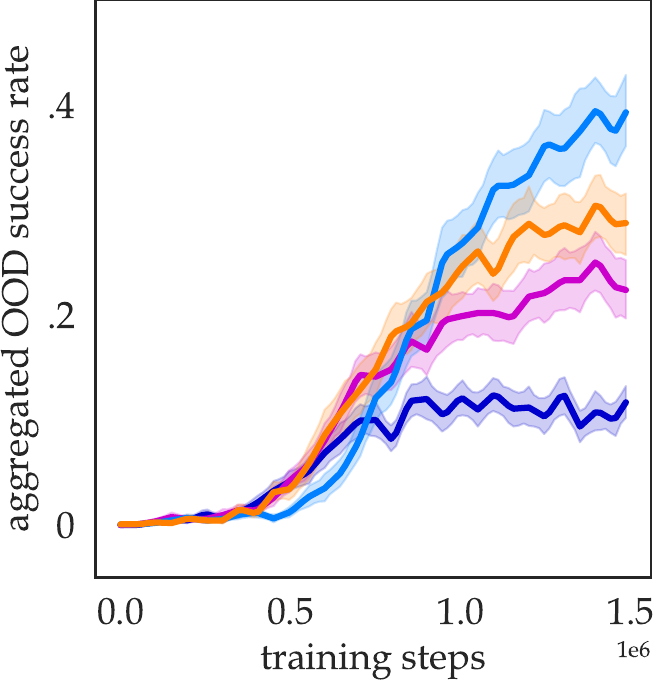}}

\caption[LEAP on SSM]{\textbf{\LEAP{} on \SSM{}}: All error bars (95\%-CI) are established over $20$ seed runs. \textbf{a)} Ratio of \Gone{} subgoals among the planned sequences; \textbf{b)} Ratio of \Gtwo{} subgoals in the planned sequences; \textbf{c)} Ratio of evolved sequences containing at least one \Gone{} or \Gtwo{} target; \textbf{d)} The final estimation accuracies towards \Gzero{} targets after training completed, across a spectrum of ground truth distances. In this figure, both distances (estimation and ground truth) are conditioned on the final version of the evolving policies; \textbf{e)} Each data point represents OOD evaluation performance aggregated over $4 \times 20$ newly generated tasks, with mean difficulty matching the training tasks.
}

\label{fig:LEAP_SSM}
\end{figure*}

For \LEAP{}, we use some different metrics to analyze the effectiveness of the proposed strategies in addressing delusions. This is because, if \LEAP{}'s evaluator successfully addressed delusions and learned not to favor the problematic targets (\Gone{} and \Gtwo{}), then they will not be selected in the evolved elitist sequence of subgoals. This makes it inconvenient for us to use the distance error in the delusional source-target pairs during decision-time as a metric to analyze the reduction of delusional estimates, because of their growing scarcity.

As we can see from Fig. \ref{fig:LEAP_SSM}, similar arguments about the effectiveness of the proposed hybrid strategies can be made, to those with \Skipper{}. The hybrids with more investment in addressing \Eone{}, \ie{}, \FEG{} and \FEPG{}, exhibit the lowest \Eone{} errors (\textbf{a)}). Similarly, \FEP{} and \FEPG{} achieve the lowest \Etwo{} errors (\textbf{b)}). In \textbf{e)}, we see that the $3$ hybrid variants achieve better OOD performance than the baseline \FE{}. Specifically, \FEG{} achieved the best performance. This is likely because that it induced the highest sample efficiency in terms of learning the estimations towards \Gzero{} subgoals, as shown in \textbf{d)}. Assistive strategies such as \generatestr{} and \pertaskstr{} do not only induce problematic targets, but also \Gzero{} ones that can shift the training distribution towards higher sample efficiencies in the traditional sense.

\begin{figure*}[htbp]
\centering

\subfloat[training, $\delta=0.4$]{
\captionsetup{justification = centering}
\includegraphics[height=0.22\textwidth]{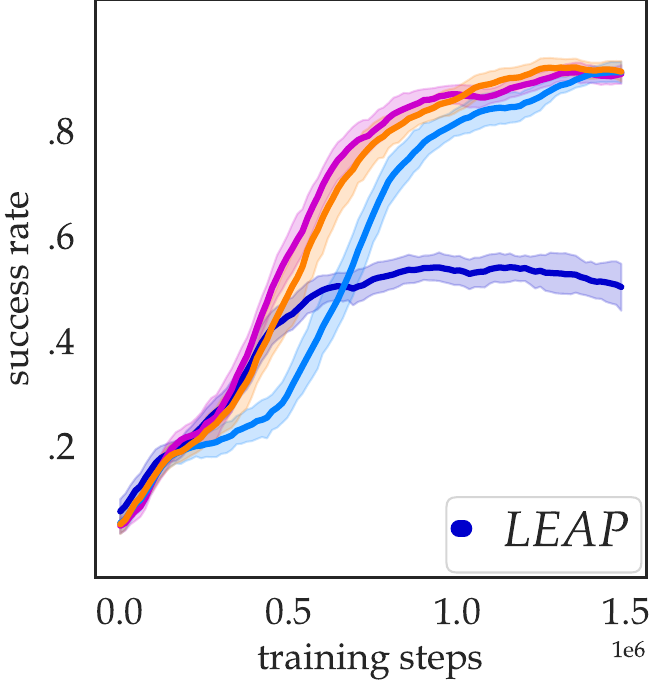}}
\hfill
\subfloat[OOD evaluation, $\delta=0.25$]{
\captionsetup{justification = centering}
\includegraphics[height=0.22\textwidth]{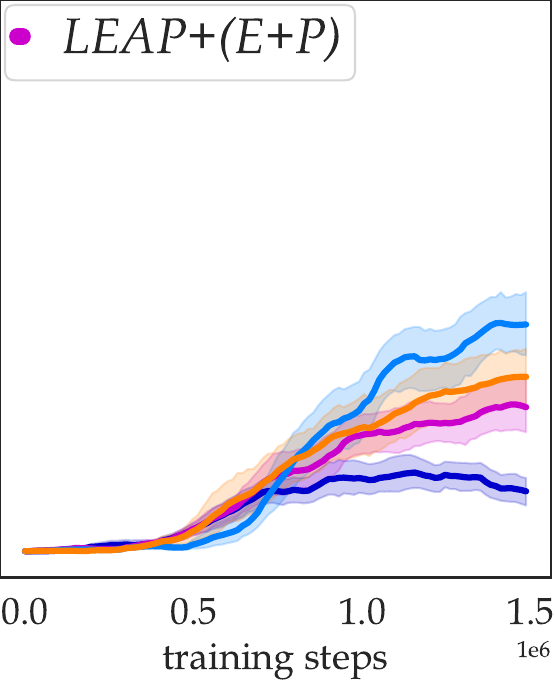}}
\hfill
\subfloat[OOD evaluation, $\delta=0.35$]{
\captionsetup{justification = centering}
\includegraphics[height=0.22\textwidth]{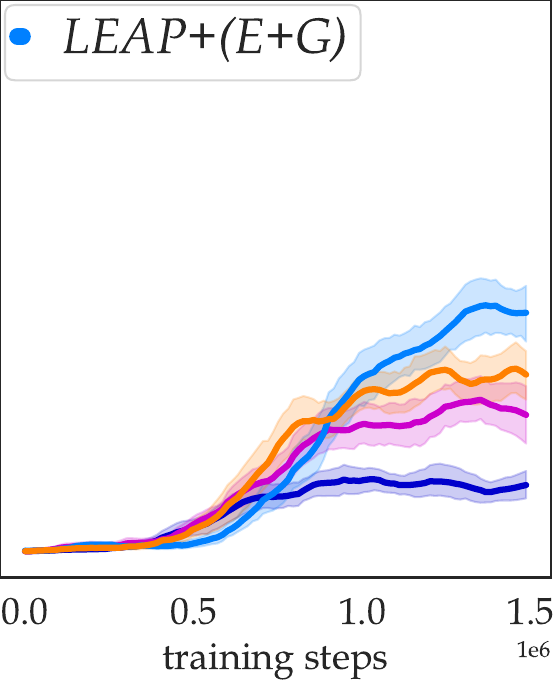}}
\hfill
\subfloat[OOD evaluation, $\delta=0.45$]{
\captionsetup{justification = centering}
\includegraphics[height=0.22\textwidth]{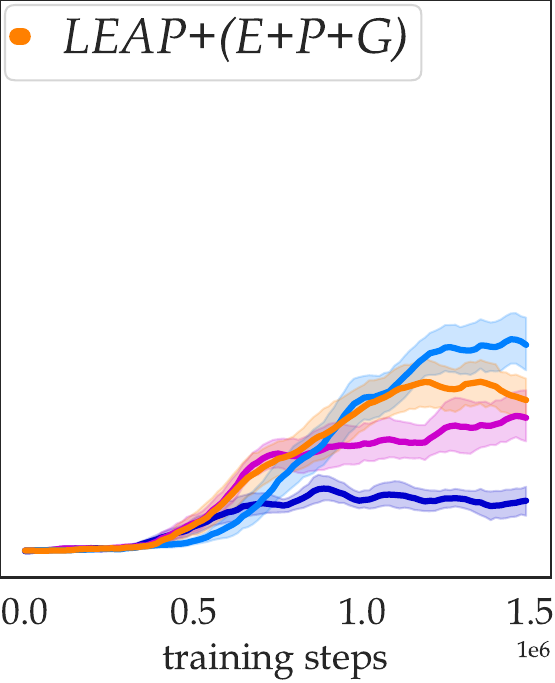}}
\hfill
\subfloat[OOD evaluation, $\delta=0.55$]{
\captionsetup{justification = centering}
\includegraphics[height=0.22\textwidth]{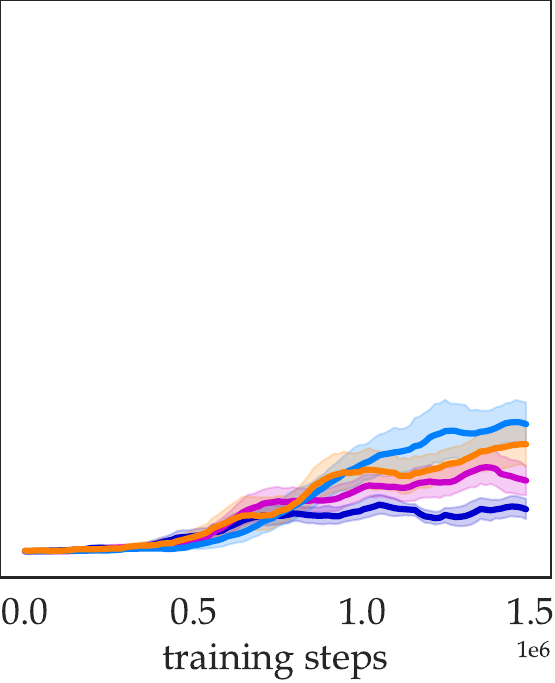}}

\caption[Evolution of OOD Performance of LEAP Variants on SSM]{\textbf{Evolution of OOD Performance of \LEAP{} Variants on \SSM{}}: All error bars (95\%-CI) are established over $20$ seed runs. 
}

\label{fig:LEAP_SSM_perfevol}
\end{figure*}

\paragraph{Breakdown of Task Performance}
In Fig. \ref{fig:LEAP_SSM_perfevol}, we present the evolution of \LEAP{} variants' performance on the training tasks as well as the OOD evaluation tasks throughout the training process. Note that Fig. \ref{fig:LEAP_SSM} \textbf{e)} is an aggregation of all $4$ sources of OOD performance in Fig. \ref{fig:LEAP_SSM_perfevol} \textbf{b-e)}.

\subsubsection{\Skipper{} on \RDS{} \texorpdfstring{(Exp.~$\nicefrac{3}{8}$)}{(Exp.~3/8)}}

\begin{figure*}[htbp]
\centering

\subfloat[\textit{\Eone{} Estimation Errors}]{
\captionsetup{justification = centering}
\includegraphics[height=0.24\textwidth]{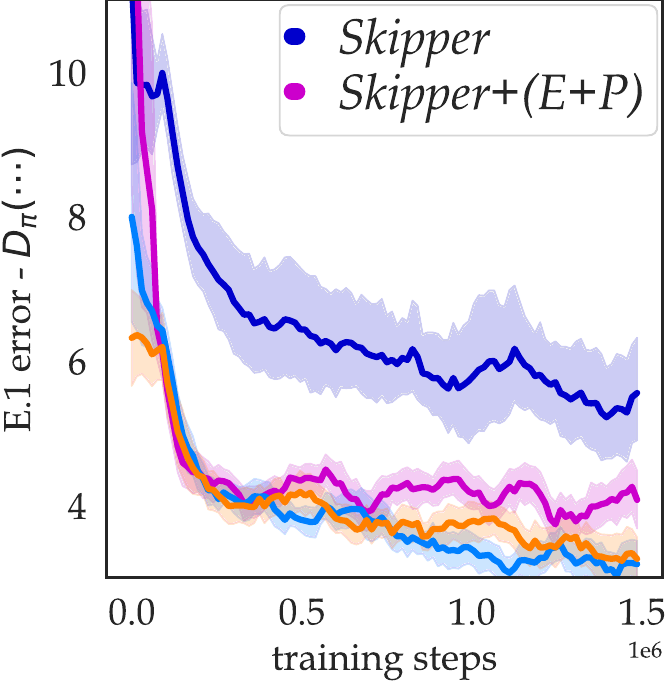}}
\hfill
\subfloat[\textit{\Gone{} + \Eone{} Behavior Ratio}]{
\captionsetup{justification = centering}
\includegraphics[height=0.24\textwidth]{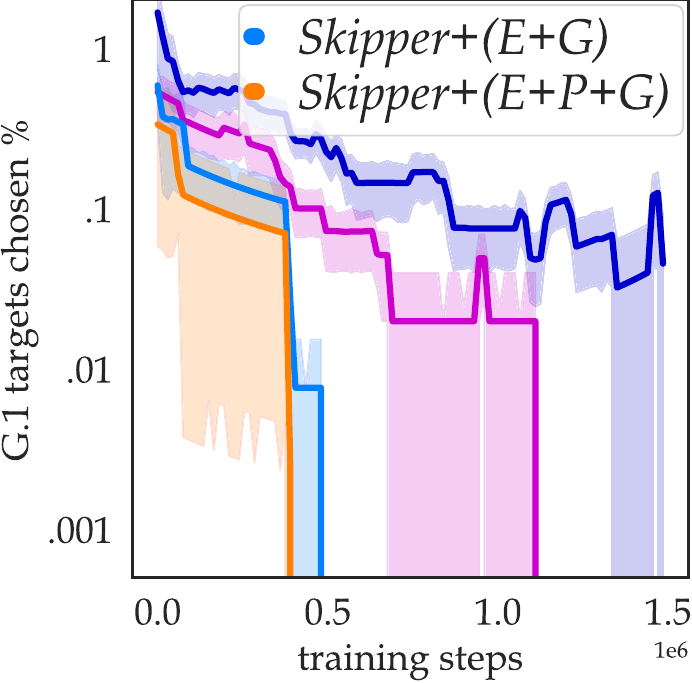}}
\hfill
\subfloat[\textit{\Ezero{} Errors}]{
\captionsetup{justification = centering}
\includegraphics[height=0.24\textwidth]{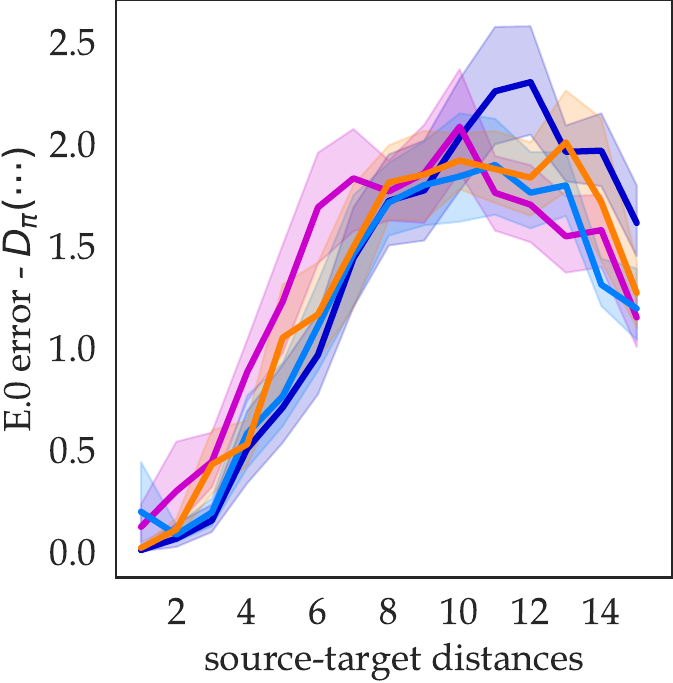}}
\hfill
\subfloat[\textit{Aggregated OOD Perf.}]{
\captionsetup{justification = centering}
\includegraphics[height=0.24\textwidth]{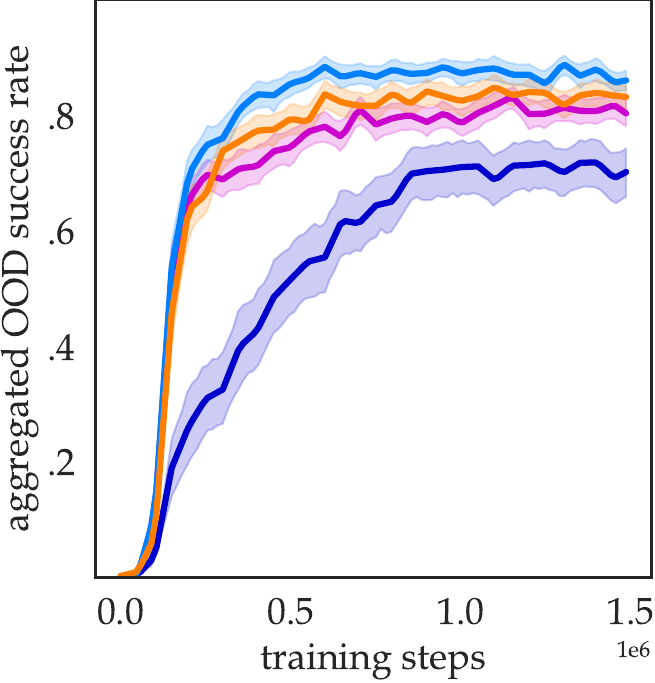}}

\caption[Skipper's Performance on RDS]{\textbf{\Skipper{}'s Performance on \RDS{}}: All error bars (95\%-CI) are established over $20$ seed runs. \textbf{a)} \Eone{} delusions in terms of $L_1$ error in estimated distance is visualized, throughout the training process. \textbf{b)} The curves represent the frequencies of choosing \Gone{} ``states'' whenever a selection of targets is initiated; \textbf{c)} The final estimation accuracies towards \Gzero{} target states after training completed, across a spectrum of ground truth distances. In this figure, both distances (estimation and ground truth) are conditioned on the final version of the evolving policies; The state structure of \RDS{} does not allow \Gtwo{} target states and the corresponding \Etwo{} delusions; \textbf{d)} Each data point represents OOD evaluation performance aggregated over $4 \times 20$ newly generated tasks, with mean difficulty matching the training tasks.
}

\label{fig:main_RDS}
\end{figure*}

This set of experiments focus on the feasibility evaluator's abilities in the face of \Gone{} challenges. We present \Skipper{}'s evaluative curves in Fig. \ref{fig:main_RDS}.

From Fig. \ref{fig:main_RDS} \textbf{d)}, we can see that, probably because of the lack of dominant \Gtwo{} + \Etwo{} cases, the OOD performance of even the most basic \episodestr{} variant is high, despite the hybrid variants perform even better. \FEG{}, \ie{} the hybrid with the most investment in \generatestr{} (aiming at \Eone{}), performs the best both in terms of \Eone{} delusion suppression (\textbf{a)}), and OOD generalization (\textbf{d)}), as expected. In \RDS{}, the short-distance \Ezero{} estimation accuracy as well as the OOD performance of \FP{} are not as bad as in \SSM{}. This is possibly because \RDS{} has much smaller state spaces, where \episodestr{} and \pertaskstr{} produce more similar results (than in large state spaces of \SSM{}).

\begin{figure*}[htbp]
\centering

\subfloat[training, $\delta=0.4$]{
\captionsetup{justification = centering}
\includegraphics[height=0.22\textwidth]{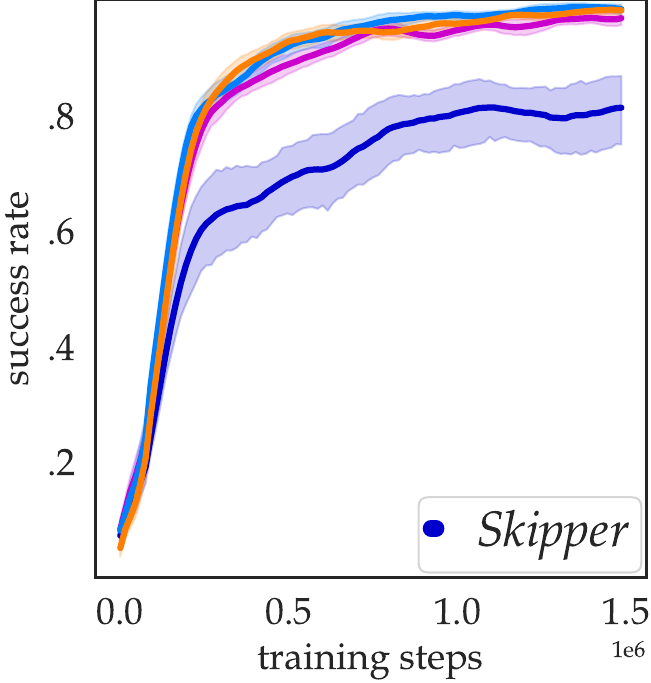}}
\hfill
\subfloat[OOD evaluation, $\delta=0.25$]{
\captionsetup{justification = centering}
\includegraphics[height=0.22\textwidth]{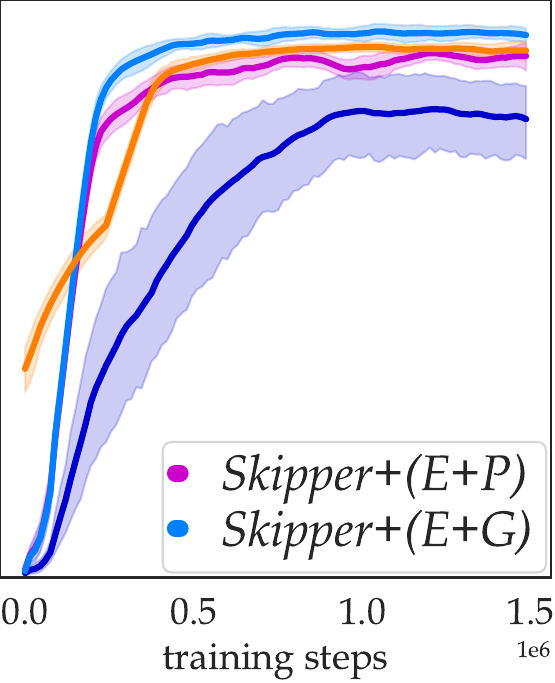}}
\hfill
\subfloat[OOD evaluation, $\delta=0.35$]{
\captionsetup{justification = centering}
\includegraphics[height=0.22\textwidth]{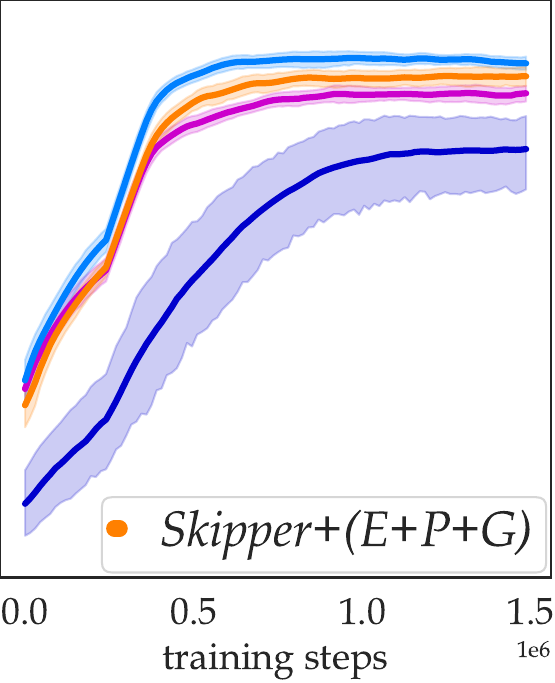}}
\hfill
\subfloat[OOD evaluation, $\delta=0.45$]{
\captionsetup{justification = centering}
\includegraphics[height=0.22\textwidth]{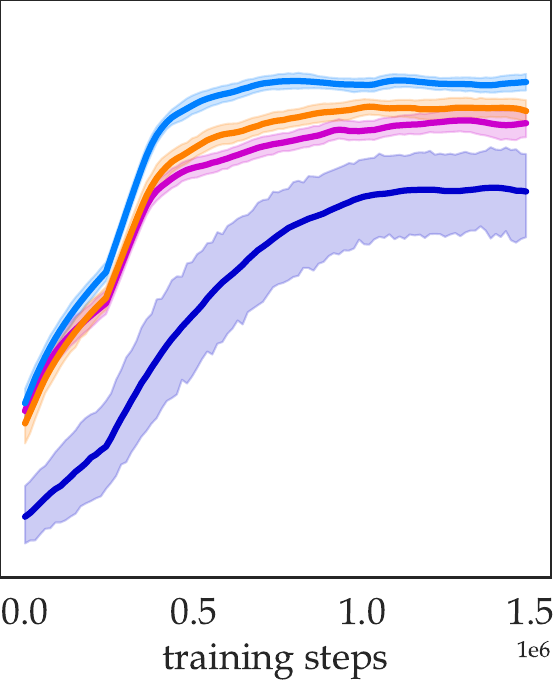}}
\hfill
\subfloat[OOD evaluation, $\delta=0.55$]{
\captionsetup{justification = centering}
\includegraphics[height=0.22\textwidth]{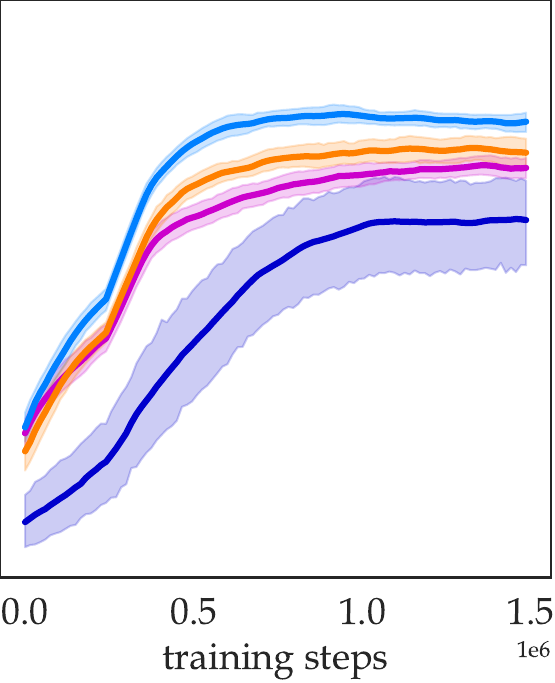}}

\caption[Evolution of OOD Performance of Skipper Variants on RDS]{\textbf{Evolution of OOD Performance of \Skipper{} Variants on \RDS{}}: All error bars (95\%-CI) are established over $20$ seed runs.
}

\label{fig:RDS_perfevol}
\end{figure*}

\paragraph{Breakdown of Task Performance}
In Fig. \ref{fig:RDS_perfevol}, we present the evolution of \Skipper{} variants' performance on the training tasks as well as the OOD evaluation tasks throughout the training process. Note that Fig. \ref{fig:main_RDS} \textbf{d)} is an aggregation of all $4$ sources of OOD performance in Fig. \ref{fig:RDS_perfevol} \textbf{b-e)}.

\subsubsection{\LEAP{} on \RDS{} \texorpdfstring{(Exp.~$\nicefrac{4}{8}$)}{(Exp.~4/8)}}

\begin{figure*}[htbp]
\centering

\subfloat[\textit{\Gone{} Ratio in Planned Sequence}]{
\captionsetup{justification = centering}
\includegraphics[height=0.24\textwidth]{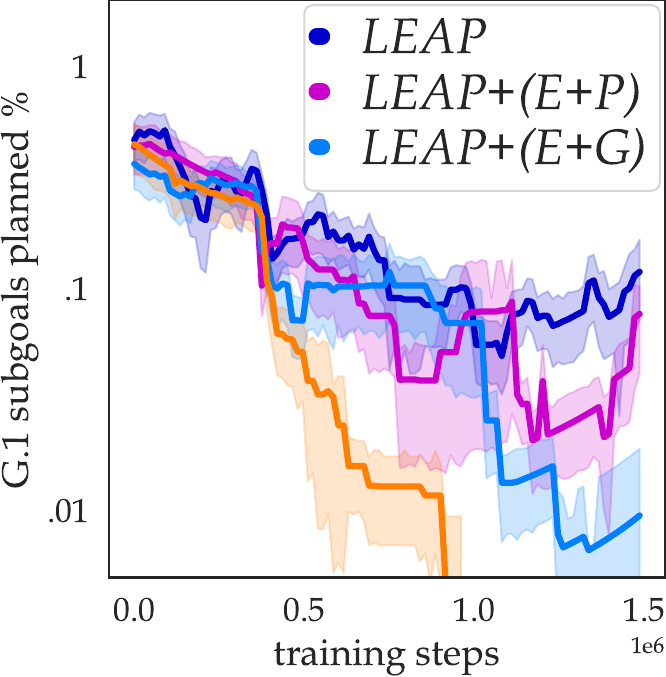}}
\hfill
\subfloat[\textit{Delusional Plan Ratio}]{
\captionsetup{justification = centering}
\includegraphics[height=0.24\textwidth]{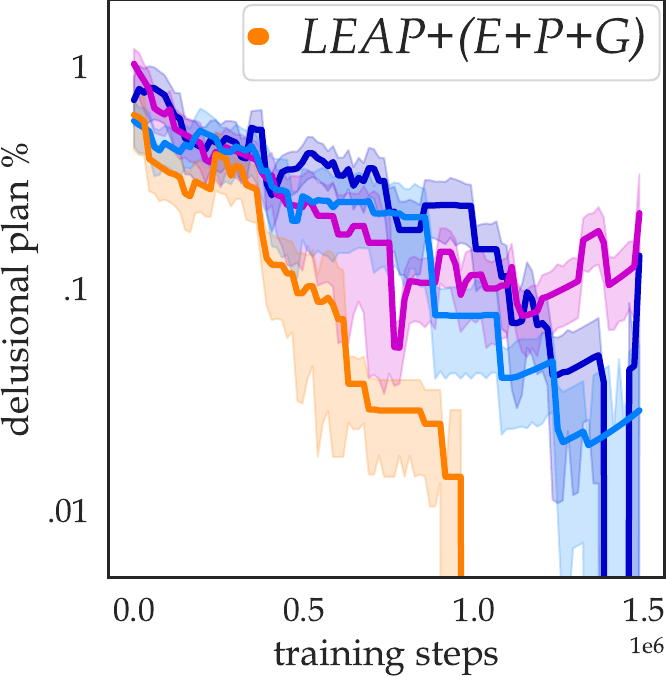}}
\hfill
\subfloat[\textit{\Ezero{} Estim. Errors}]{
\captionsetup{justification = centering}
\includegraphics[height=0.24\textwidth]{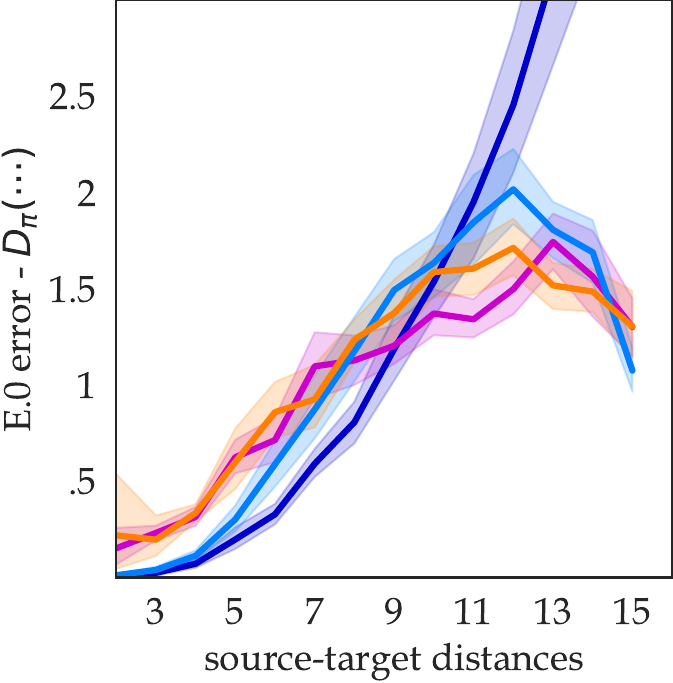}}
\hfill
\subfloat[\textit{Aggregated OOD Performance}]{
\captionsetup{justification = centering}
\includegraphics[height=0.24\textwidth]{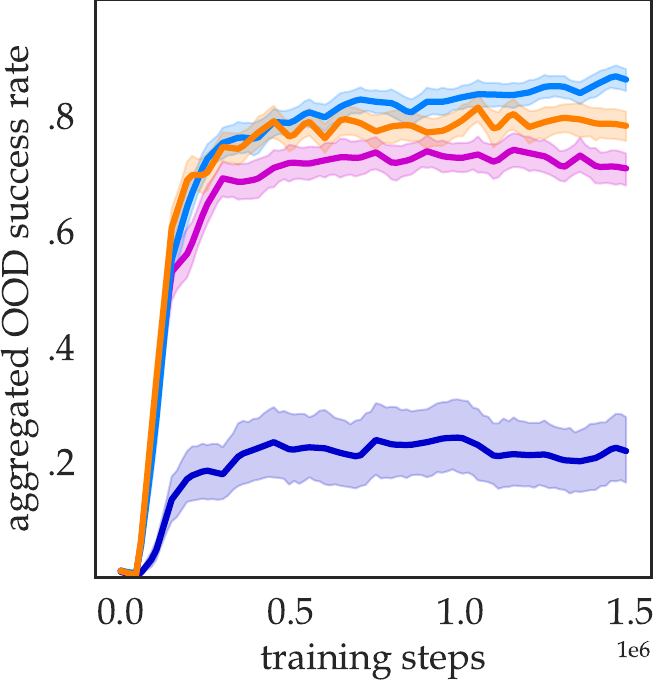}}

\caption[LEAP's Performance on RDS]{\textbf{\LEAP{}'s Performance on \RDS{}}: All error bars (95\%-CI) are established over $20$ seed runs. \textbf{a)} Ratio of \Gone{} subgoals among the planned sequences; \textbf{b)} Ratio of planned sequences containing at least one \Gone{} target; \textbf{c)} The final estimation accuracies towards \Gzero{} target states after training completed, across a range of ground truth distances. In this figure, both distances (estimation and ground truth) are conditioned on the final version of the learned policies; \textbf{d)} Each data point represents OOD evaluation performance aggregated over $4 \times 20$ newly generated tasks, with mean difficulty matching the training tasks.
}
\label{fig:LEAP_RDS}
\end{figure*}

This set of experiments focus on \LEAP{}'s performance on \RDS{}. Similarly, we present the evaluative metrics in Fig. \ref{fig:LEAP_RDS}.

The conclusions are similar, despite that the OOD performance gain by addressing delusions is significantly higher than in \SSM{}.

\begin{figure*}[htbp]
\centering

\subfloat[training, $\delta=0.4$]{
\captionsetup{justification = centering}
\includegraphics[height=0.22\textwidth]{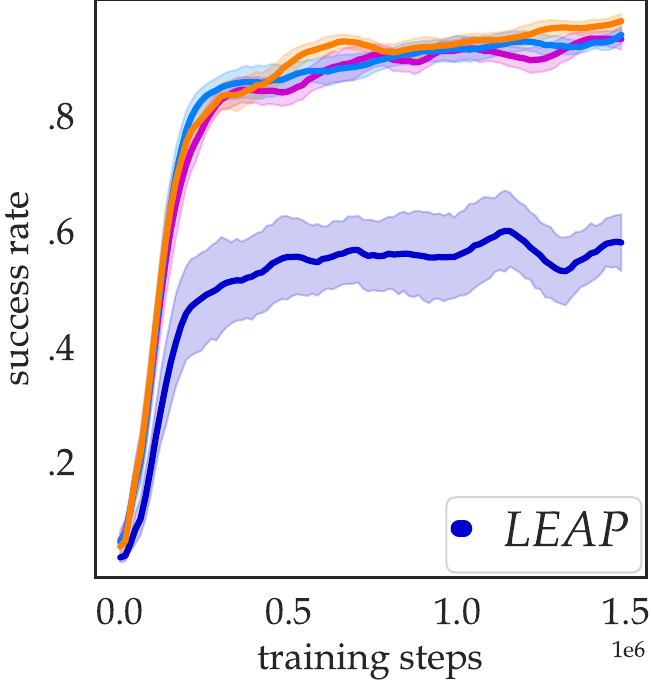}}
\hfill
\subfloat[OOD evaluation, $\delta=0.25$]{
\captionsetup{justification = centering}
\includegraphics[height=0.22\textwidth]{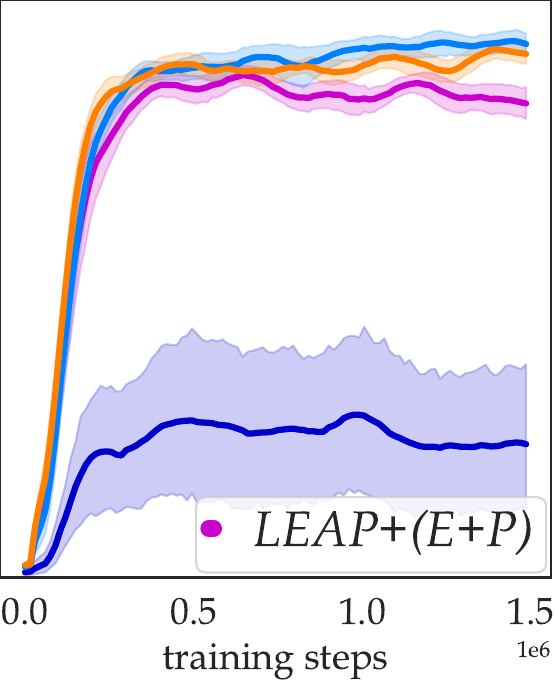}}
\hfill
\subfloat[OOD evaluation, $\delta=0.35$]{
\captionsetup{justification = centering}
\includegraphics[height=0.22\textwidth]{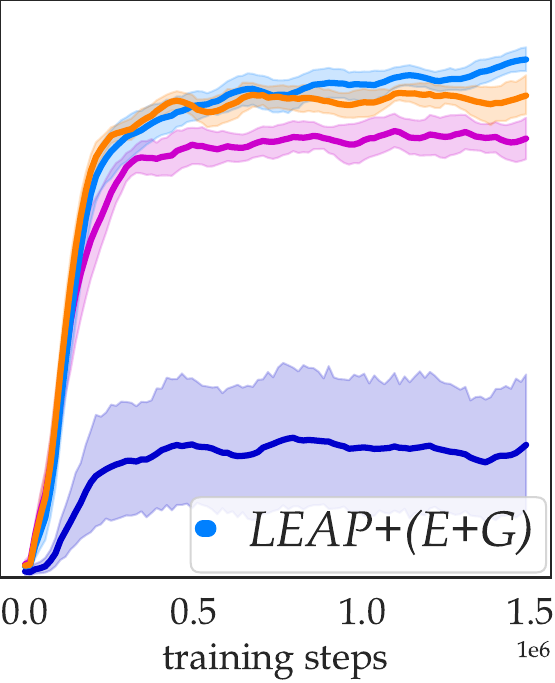}}
\hfill
\subfloat[OOD evaluation, $\delta=0.45$]{
\captionsetup{justification = centering}
\includegraphics[height=0.22\textwidth]{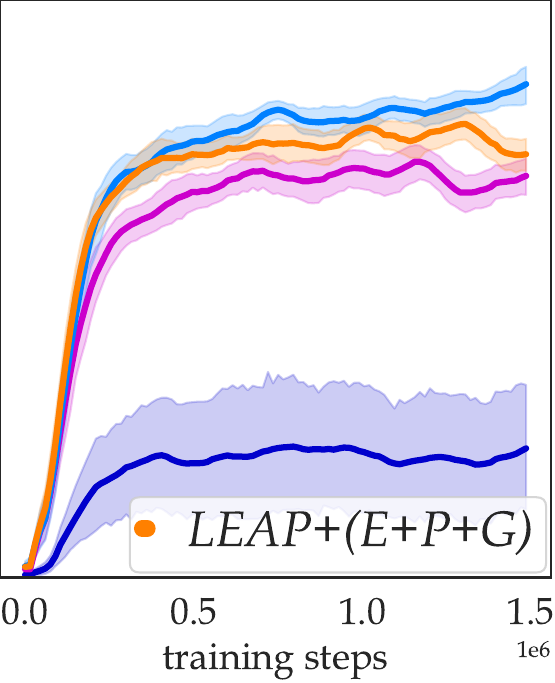}}
\hfill
\subfloat[OOD evaluation, $\delta=0.55$]{
\captionsetup{justification = centering}
\includegraphics[height=0.22\textwidth]{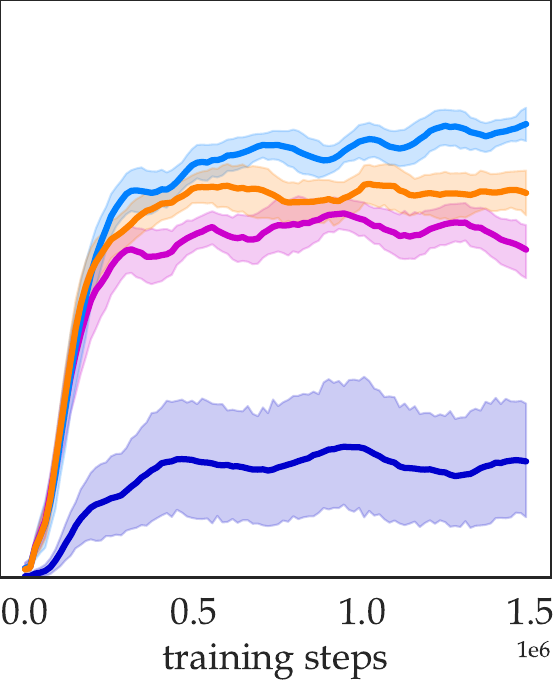}}

\caption[Evolution of OOD Performance of LEAP Variants on RDS]{\textbf{Evolution of OOD Performance of \LEAP{} Variants on \RDS{}}: All error bars (95\%-CI) are established over $20$ seed runs.
}

\label{fig:LEAP_RDS_perfevol}
\end{figure*}

\paragraph{Breakdown of Task Performance}
In Fig. \ref{fig:LEAP_RDS_perfevol}, we present the evolution of \LEAP{} variants' performance on the training tasks as well as the OOD evaluation tasks throughout the training process. Note that Fig. \ref{fig:LEAP_RDS} \textbf{d)} is an aggregation of all $4$ sources of OOD performance in Fig. \ref{fig:LEAP_RDS_perfevol} \textbf{b-e)}.

\begin{figure}[htbp]
\centering

\subfloat[\textit{Convergence to Optimal Value}]{
\captionsetup{justification = centering}
\includegraphics[height=0.32\textwidth]{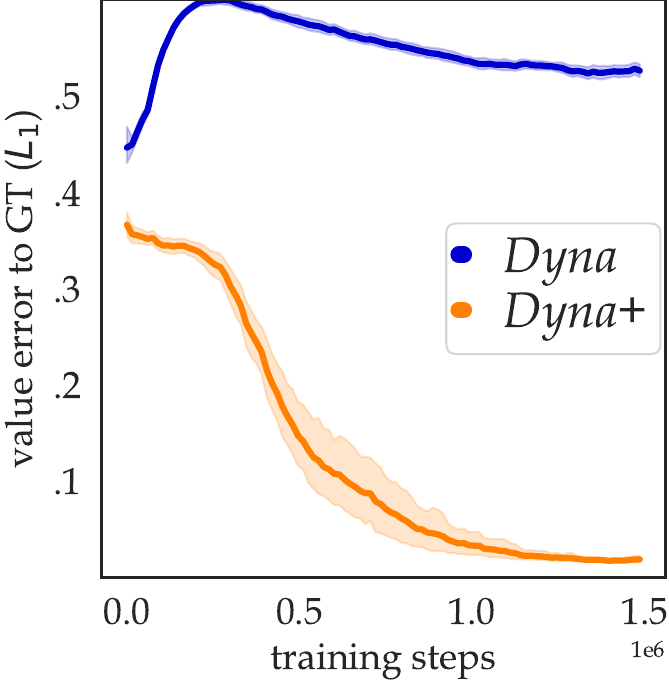}}
\hfill
\subfloat[\textit{Training Performance}]{
\captionsetup{justification = centering}
\includegraphics[height=0.32\textwidth]{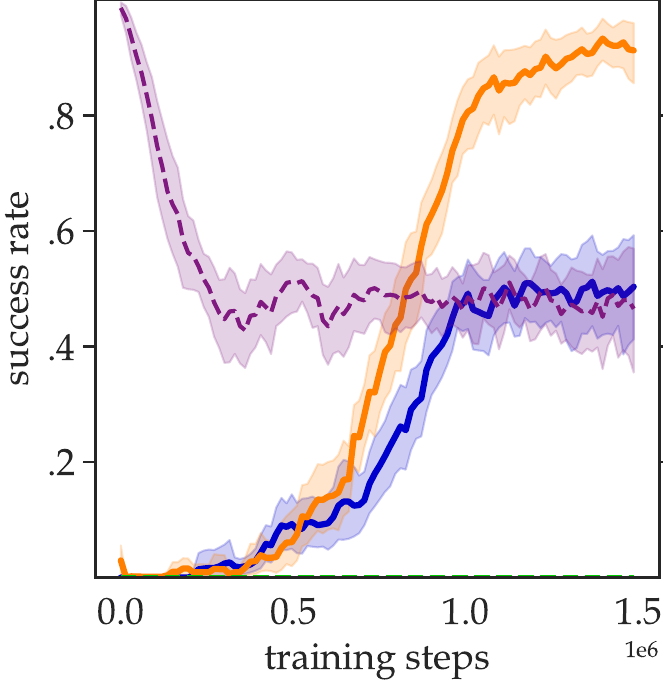}}

\caption[Dyna's Performance on SSM]{\textbf{\Dyna{}'s Performance on \SSM{}}: All error bars (95\%-CI) are established over $20$ seed runs. Compared to the baseline \Dyna{}, \DynaPlus{} rejects the updates toward $1$-infeasible generated states flagged by the evaluator, powered by \FEPG{}. \textbf{a)} Evolving mean $L_1$ distances between estimated $Q$ \& optimal values; \textbf{b)}: task performance on the $50$ training tasks \& \textcolor{purple}{rate of \DynaPlus{} rejecting updates}.
}
\label{fig:Dyna_SSM}
\end{figure}

\subsection{Rejecting Destabilizing Updates in Background Planning (Exp.~$\nicefrac{5}{8}$ - Exp.~$\nicefrac{6}{8}$)}
\label{sec:exp_main_dyna}

For background planning agents, since they are by design not as good as decision-time planning agents in OOD generalization, we only compare the difference in their training performance.

These two sets of experiments focus on a rollout-based background TAP agent - the classical 1-step \Dyna{} \citep{sutton1991dyna}, which uses its learned transition model to generate next states from existing states to construct simulated transitions that are used to update the value estimator, \ie{} a ``\Dyna{} update''. \citet{jafferjee2020hallucinating} demonstrated the benefit when the delusional \Dyna{} updates bootstrapped on hallucinated targets are rejected with an oracle. We replace the oracle using our learned evaluator.

With the same training setup, in Fig.~\ref{fig:Dyna_SSM}, we present the empirical results of how target rejection can significantly improve the performance of \Dyna{} on \SSM{}. The rejection rate stabilizes as both the generator and the evaluator learns. These observations are consistent with Exp.~$\nicefrac{6}{8}$, presented in Sec.~\ref{sec:exp_RDS_dyna}.\footnote{The implementation here can be extended to fixed-horizon rollout agents. In the Sec.~\ref{sec:appendix_dreamer}, we provide details on how we applied our \Dyna{} solution to \Dreamer{}V2 \citep{hafner2020mastering}.}

\phantomsection
\label{sec:exp_RDS_dyna}

\begin{figure*}[htbp]
\centering

\subfloat[\textit{Convergence to Optimal Value}]{
\captionsetup{justification = centering}
\includegraphics[height=0.32\textwidth]{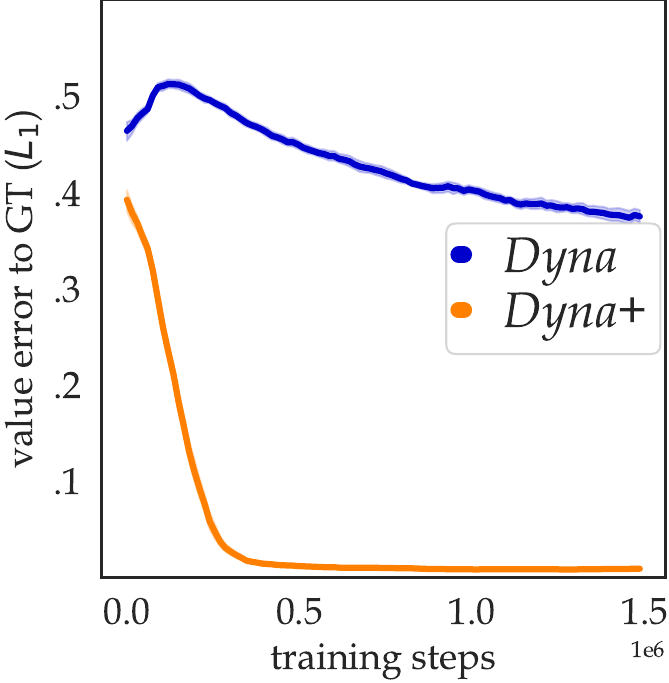}}
\hfill
\subfloat[\textit{Training Performance}]{
\captionsetup{justification = centering}
\includegraphics[height=0.32\textwidth]{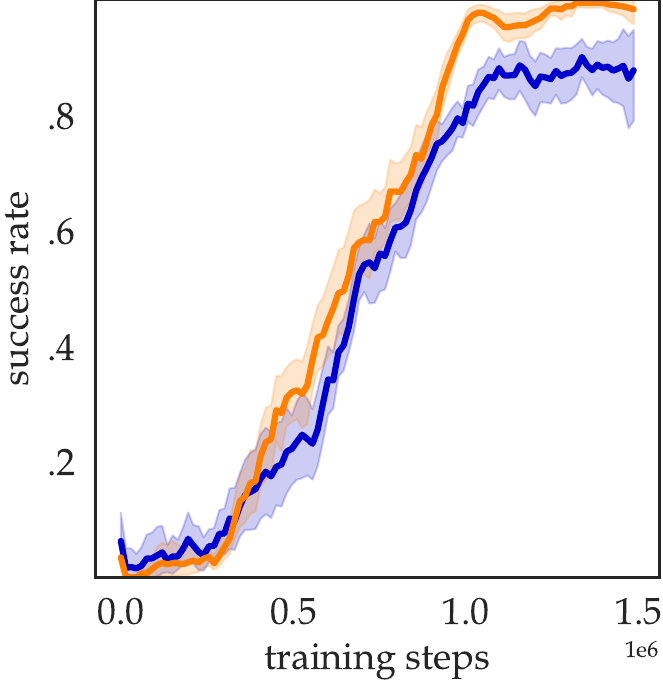}}
\hfill
\subfloat[\textit{Training Performance}]{
\captionsetup{justification = centering}
\includegraphics[height=0.32\textwidth]{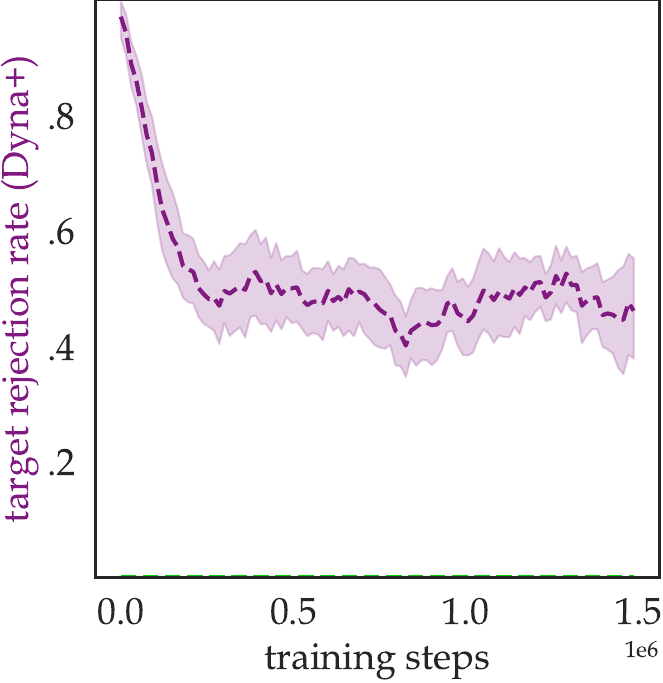}}

\caption[Dyna's Performance on RDS]{\textbf{\Dyna{}'s Performance on \RDS{}}: All error bars (95\%-CI) are established over $20$ seed runs. \textbf{a)}: Evolving mean $L_1$ distances between estimated $Q$ values \& ground truth optimals; \textbf{b)}: evaluation performance on the $50$ training tasks; \textbf{c)}: \textcolor{purple}{rate of rejecting \Dyna{} updates}.
}
\label{fig:Dyna_RDS}
\end{figure*}

In Fig.~\ref{fig:Dyna_RDS}, we present the empirical performance of a \Dyna{} variant with rejection enabled by \FEPG{}, which is significantly better than the baseline.

\subsection{Feasibility Convergence to Non-Singleton Targets \texorpdfstring{(Exp.~$\nicefrac{7}{8}$ \& $\nicefrac{8}{8}$)}{(Exp.~7/8 \& Exp.~8/8)}}
We test if our implemented feasibility evaluator for Exp.~$\nicefrac{1}{8}$ - Exp.~$\nicefrac{4}{8}$ could withstand targets that are non-singleton. In its previous implementation, we use $h$ to enforce the that the targets are singletons. In fact, each $\bm{g}^\odot$ takes the form of a state representation and $h$ is only activated if a state with exactly the same representation is reached. For the non-singleton experiments however, we let $h$ activate when a state is within distance one to the target state, effectively expanding each target set from size $1$ to maximally size $5$. Given the new termination mechanisms enforced by the new $h$, each target now, despite still taking the form of a state representation, has a new meaning. This setting mirrors the goal-conditioned path planning agents that seeks to reach certain neighborhoods of the planned waypoints.

With this setting, we can also intuitively analyze the composition of the target set. Specifically, if one of the member state is \Gtwo{}, then the whole target set are fully made of \Gtwo{}. If all the $5$ states are out of the state space, then the target is fully composed of \Gone{}. For \SSM{}, a target in the temporarily unreachable situation, \eg{}, $s \in \langle 1, 1 \rangle$ with target encoding $s^\odot \in \langle 0, 1 \rangle $, could be composed of not only \Gtwo{} states but also some \Gone{}.

\begin{figure*}[htbp]
\centering

\subfloat[\textit{Evolution of \Ezero{} Errors}]{
\captionsetup{justification = centering}
\includegraphics[height=0.31\textwidth]{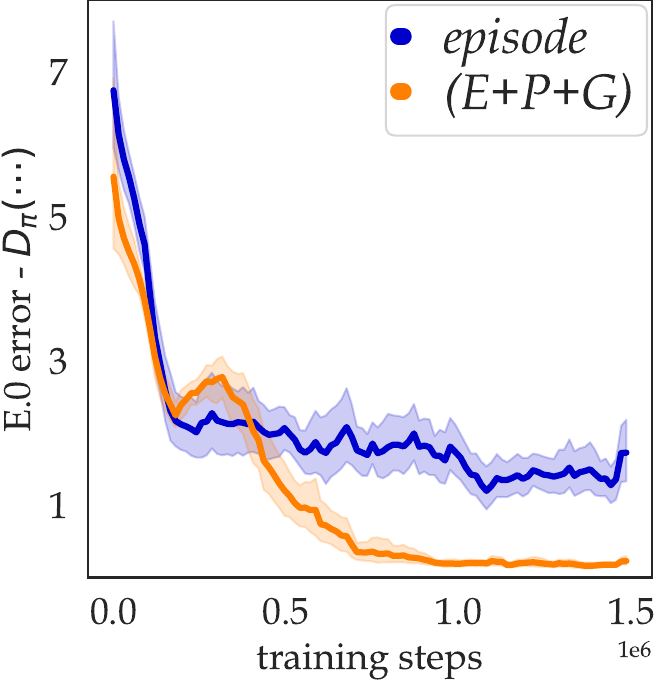}}
\hfill
\subfloat[\textit{Evolution of \Eone{} Errors}]{
\captionsetup{justification = centering}
\includegraphics[height=0.31\textwidth]{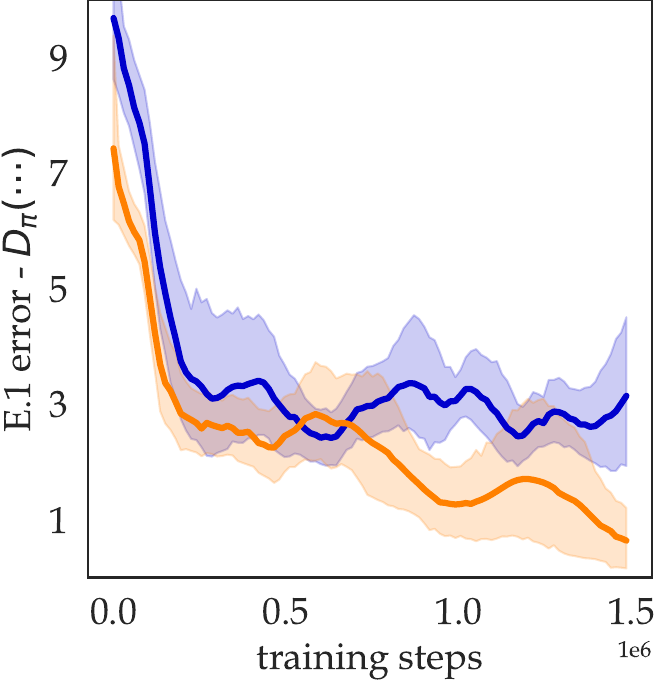}}
\hfill
\subfloat[\textit{Evolution of \Etwo{} Errors}]{
\captionsetup{justification = centering}
\includegraphics[height=0.31\textwidth]{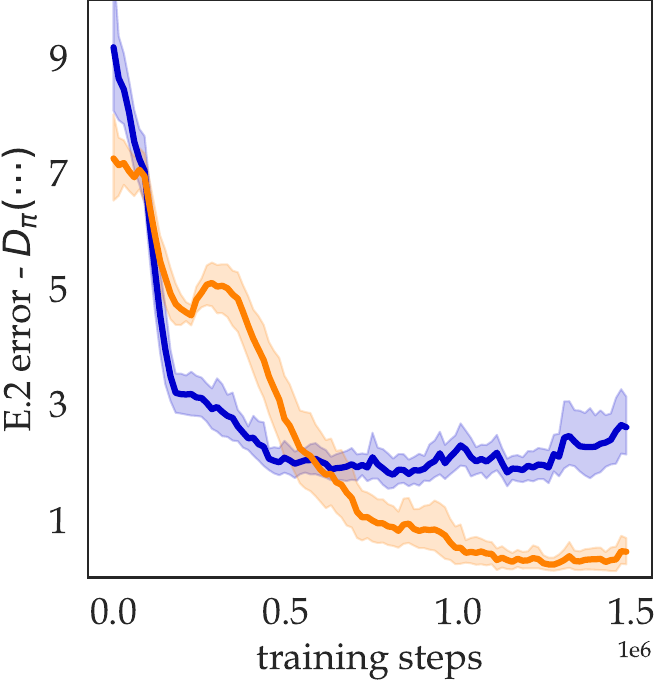}}

\caption[Feasibility of Non-Singleton Targets on SSM]{\textbf{Feasibility of Non-Singleton Targets on \SSM{}}: All error bars (95\%-CI) are established over $20$ seed runs. \textbf{a)} Evolution of \Ezero{} error; \textbf{b)} Evolution of \Eone{} error; \textbf{c)} Evolution of \Etwo{} error; The training data is acquired with random walk, since the introduced non-singleton targets do not lead to adequate performances.
}

\label{fig:nonsingleton_SSM}
\end{figure*}

We apply the new $h$ to evaluator training and to the ground truth DP solver, and then compare their differences. As we could observe from Fig.~\ref{fig:nonsingleton_SSM}, the proposed feasibility evaluator, with the help of the two assistive hindsight relabeling strategy, significantly reduces the feasibility errors in all categories.

\begin{figure*}[htbp]
\centering

\subfloat[\textit{Evolution of \Ezero{} Errors}]{
\captionsetup{justification = centering}
\includegraphics[height=0.32\textwidth]{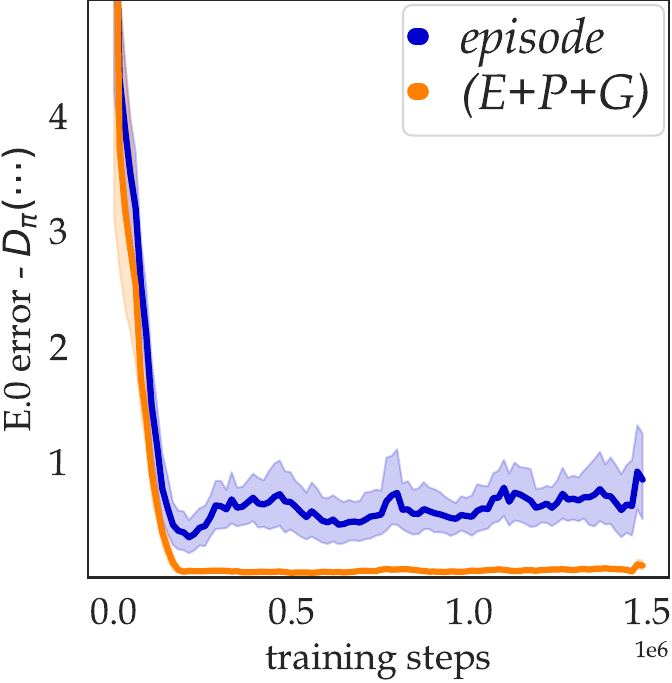}}
\hfill
\subfloat[\textit{Evolution of \Eone{} Errors}]{
\captionsetup{justification = centering}
\includegraphics[height=0.32\textwidth]{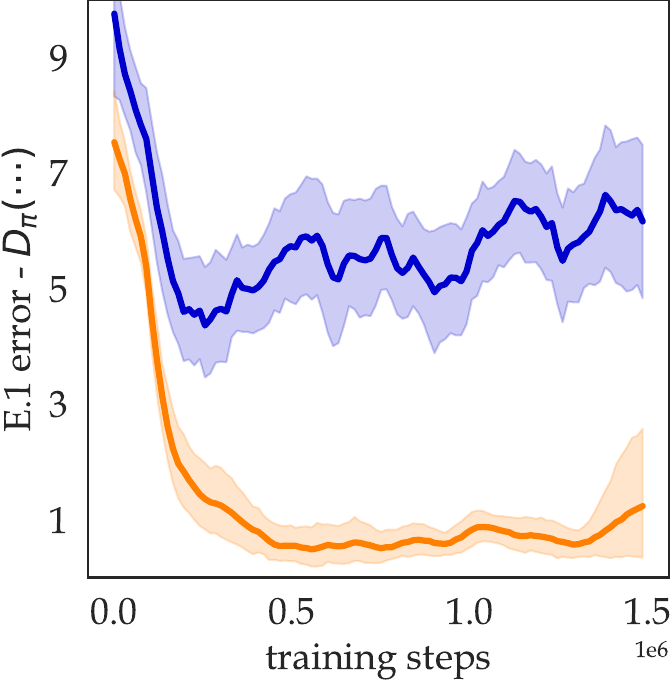}}

\caption[Feasibility of Non-Singleton Targets on RDS]{\textbf{Feasibility of Non-Singleton Targets on \RDS{}}: All error bars (95\%-CI) are established over $20$ seed runs. \textbf{a)} Evolution of \Ezero{} error; \textbf{b)} Evolution of \Eone{} error; The training data is acquired with random walk, since the introduced non-singleton targets do not lead to adequate performances.
}

\label{fig:nonsingleton_RDS}
\end{figure*}

We observe the similar results in \RDS{}, presented in Fig.~\ref{fig:nonsingleton_RDS}.

\section{Summary}

We characterized how generator hallucinations can cause trouble for TAP agents. Then, we proposed to evaluate the feasibility of targets \st{} the infeasible hallucinations can be properly rejected during planning. We proposed a combination of learning rules, architectures and data augmentation strategies that leads to robust and accurate output when the proposed evaluator is applied. In experiments, we showed that the evaluator can significantly address the harm of hallucinated targets in various kinds of planning agents. 

%% file: chapter_7_conclusion.tex
\chapter{Discussions: Conclusions, Limitations \& Future Work}
\label{cha:conclusion}
{\small
In this chapter we re-iterate the contributions of the thesis, discuss the limitations of the work and avenues for future research.
}

\minitoc

\section{Summary of Contributions}
This thesis explores brain-inspired model-based deep RL agents capable of effectively generalizing their learned skills to new, target application environments. Our research argues that the difficulties of generalization stem from the absence of appropriate reasoning abilities, which are necessary for agents to adapt to novel situations \citep{daniel2017thinking}. To address this, we introduced components inspired by higher cognitive functions involved in human conscious planning, significantly enhancing the agents' zero-shot OOD generalization capabilities \citep{sylvain2019locality}. Furthermore, we examined the formation of delusional behaviors in Target-Assisted Planning (TAP) agents, by leveraging insights from the human brain \citep{kiran2009understanding}.

The contributions of the thesis work clearly met the research objectives set before the start of my doctoral study.

We now revisit the main original contributions of this thesis and the implication of the findings.

\subsection{\nameref{cha:CP}}
\label{sec:contrib_CP}

Chap.~\ref{cha:CP} aimed to address the problem of unsatisfactory generalization in the skills learned by existing RL agents, which are often sensitive to environmental perturbations and OOD problem changes \citep{ada2024diffusion}. To tackle this, we designed a decision-time planning agent capable of dynamically focusing on the most relevant aspects of the state, thereby enhancing OOD generalization - making the first such contribution in the literature at the time when the work was carried out. The agent learns to dynamically focus on relevant (partial) aspects of the state during reasoning, while ignoring irrelevant environmental distractors that might hinder the generalization of its learned skills. Drawing inspiration from conscious decision-making behaviors in humans \citep{baars1993cognitive,baars2002conscious,dehane2017consciousness}, we implemented this bottleneck mechanism using top-down semi-hard attention (Sec.~\ref{sec:CP_bottleneck}), and developed a comprehensive system of DRL architectures supported by set-based representations, end-to-end learning, and tree search with Model Predictive Control (MPC). Finished in late 2020, this work was among the first to introduce transformer-based architectures for computational decision-making. In our generalization-focused experimental settings, the proposed bottleneck mechanism enables the agent to selectively focus its computational resources on relevant objects for planning, leading to a significant improvement in OOD performance. This work sparked discussions on how knowledge of human higher-level cognitive functions can be leveraged to enhance the generalization capabilities of computational decision-making systems. The benefits of the focused reasoning induced by the bottleneck algorithm allow CP to generalize better and this probably applies to other learning systems as well.

This work also made some specific original contributions:

\begin{itemize}[leftmargin=*]
\item \textbf{Set-Based / Object-Oriented Dynamics Model \& Other RL Components}: To the best of our knowledge, at the time the preprint of this project was published on \url{arxiv.org}, we were the first to successfully implement a set-based (object-oriented) latent space dynamics model for decision-time planning that was trained end-to-end without relying on a reconstruction signal. We developed a robust and effective design that has since been widely acknowledged and adopted, inspiring subsequent methods. Specifically, we addressed the ``alignment problem'', where positional encoding and object features can create confusion when matching estimates to training targets. Traditionally, this problem was tackled using matching losses, which often resulted in suboptimal performance while incurring significant computational cost. We overcame this challenge by designing an object representation that explicitly separates the dimensions of features and positional encodings. This design allows positional encodings to remain fixed during training, enabling them to be used for matching objects between the estimated set and the update target set. As a result, we could use simple losses, such as $L_2$, for effective training (Sec.~\ref{sec:CP_model_alignment}). Additionally, we developed an entire  architecture capable of conducting set-based RL end-to-end, including set-based state representation encoder, set-based value estimator, \etc{}.

\item \textbf{Spatial Abstraction - Top-Down Semi-Hard Attention Bottleneck}: The most significant contribution of this work is the successful implementation of a top-down attention-based bottleneck component that restricts the agent's reasoning abilities to a limited set of object slots (Sec.~\ref{sec:CP_bottleneck}). The dynamic selection of a subset of objects within the state set is conditioned on the agent's intention during the tree search of decision-time planning, as well as the relationships between the objects in the state. The use of semi-hard attention enabled backpropagation through a hard-selected continuous bottleneck, circumventing the numerical challenges associated with a fully hard attention bottleneck, where gradients cannot flow directly. Our insights into consciousness in the first sense guided the design of this bottleneck to facilitate generalization, and we successfully integrated this feature into RL \citep{dehane2017consciousness}.

\item \textbf{Using Multiple Predictive Losses to Enhance Set-based State Representations}: One key design of the agent is that all the training loss terms collectively shape the state representation (Sec.~\ref{sec:CP_training}), a concept traditionally applied to vectorized state representations but not to set-based representations. Our approach aimed to make the state representation capable of predicting values, reward transitions, terminal state status, and latent state dynamics, all at the same time, using all losses to numerically regulate each other. We validated this methodology and, in the process, discovered that KL-divergences, when used as loss terms, tend to balance each other due to their shared value ranges.
\end{itemize}

\subsection{\nameref{cha:skipper}}
\label{sec:contrib_skipper}

Chap.~\ref{cha:skipper} aimed to push the boundaries of existing decision-time planning methods, enabling them to plan in a way that is both spatially and temporally abstract — similar to how conscious planning in the human brain spans sparse decision points and focuses on relevant aspects of environmental states \citep{dehane2017consciousness,bengio2017consciousness}. To achieve this, we developed a framework called \Skipper{}, which automatically decomposes an overall task into smaller, more manageable steps by leveraging abstractions in both the temporal and spatial dimensions. \Skipper{} employs a constrained form of option-based planning, building on the consciousness-inspired spatial abstraction mechanisms discussed in Chap.~\ref{cha:CP}, particularly when considering each decomposed step (Sec.~\ref{sec:skipper_spatial_abstraction}). The option-based planning in \Skipper{} is conducted over proxy problems, a constrained form of SMDP for learning a problem decomposition of an overall MDP that aligns with the agent's capability to handle each decomposed step during the divide-and-conquer of the given task. Furthermore, we proved that the framework's performance on proxy problems is guaranteed under assumptions that are achievable in practice. This work demonstrates that spatially and temporally abstract planning is not only feasible, but also has performance guarantees, offering a promising direction for option-based planning. A clever way of blending spatio-temporal abstractions is probably a common need for learning systems that seek to tackle complex tasks.


Specific technical contributions are as follows:

\begin{itemize}[leftmargin=*]
\item \textbf{Proxy Problems to Divide-and-Conquer}: We proposed proxy problems, a constrained form of SMDP, tailored for goal-conditioned planning (Sec.~\ref{sec:skipper_proxy_problem}). We demonstrated that, given a set of checkpoints (a subset of the state space), proxy problems can be used to decompose a complex task into smaller, more manageable steps. Under appropriate assumptions, we proved that, as long as certain values can be estimated, proxy problems can enable optimal decision-making, providing a performance guarantee for planning agents that use them. Proxy problems anchor the \Skipper{} framework in principled theory, thus distinguishing it from heuristic-based methods in temporally-abstracted reasoning.

\item \textbf{Per-Sample TD-based Learning Rules Converging to Quantities Needed for Proxy Problems}: We designed learning rules that provably enable the agent to learn the essential values for optimal decision-making in proxy problems (Sec.~\ref{sec:skipper_update_rules}). Additionally, we introduced a simple and general technique that allows an agent to interchangeably estimate the distribution of distances and cumulative discount between states, using C51-style distributional learning \citep{bellemare2017distributional}. These learning rules obey the technical assumptions regarding estimation accuracy for decision-making with proxy problems.

\item \textbf{Extending Spatial Abstraction to Convolutional Features}: We extended the spatial abstraction mechanism proposed in Sec.~\ref{sec:CP_bottleneck} to feature maps learned with Convolutional Neural Networks (CNNs)~\citep{lecun1989backpropagation}, making the approach much more generally applicable (Sec.~\ref{sec:skipper_spatial_abstraction}).

\item \textbf{Context-Aware Conditional Goal Generator
}: We proposed a training loss for the goal generator that optimizes an alternative Evidence Lower BOund (ELBO) to produce goals that are likely to occur within the same episode, given the context of the current state (Sec.~\ref{sec:skipper_generator}). Compared to predictive models that are widely used in model-based RL, this generator directly proposes goals that be arbitrarily far away from the current state. The generator thus serves as the source for the vertices of proxy problems.
\end{itemize}

\subsection{\nameref{cha:delusions}}
\label{sec:contrib_delusions}

Chap.~\ref{cha:delusions} proposes a novel perspective from human psychiatry to analyze a common issue faced by TAP agents, planning agents that use generative models to sample state targets during planning: their tendency to be confused by hallucinated state targets, which undermines their performance and poses safety risks \citep{bengio2024managing}. This perspective allowed us to identify the causes of delusional behaviors shared by seemingly different kinds of planning agents and to unify them in one framework, enabling a solution that resembles the coordination between the belief formation and evaluation systems in human psychopathology \citep{kiran2009understanding}. We then proposed a combination of update rules, architecture, and hindsight relabeling strategy to enable learning an add-on feasibility evaluator that can reject proposed targets when they are not trustworthy, \ie{}, when they are a result of hallucination. After applying the proposed add-on feasibility evaluator to existing agents, we observed a significant reduction in delusional planning behaviors in several types of planning agents, leading to significant improvements in performance. Being able to understand that certain plans are off-limits may be a necessity for any agentic AI composed of learning systems.

More specifically, we make the following technical contributions:

\begin{itemize}[leftmargin=*]
\item \textbf{TAP Framework}: We abstract seemingly different kinds of existing planning methods that generate state targets into a single framework, Target-Assisted Planning (TAP), whose difficulties in dealing with hallucinations can be addressed with a unified solution. The two key components in TAP agents, the generator, and the evaluator, directly correspond to the belief formation and belief evaluation systems in the human brain, whose coordination is responsible to addressing delusions (Sec.~\ref{sec:target_directed_framework}). The TAP framework allows us to clearly highlight the problem of delusional planning behaviors in related RL methods, identify its causes, and propose an effective mitigation strategy: when a state target is a result of hallucination, its evaluation will not be trustworthy, and thus it should be rejected.

\item \textbf{Categorization of Hallucinations}:
Using the TAP framework, by incorporating a temporally-aware perspective, we differentiated two types of hallucinated states that TAP agents are prone to pursue: the permanently unreachable target states and the temporarily unreachable target states (Sec.~\ref{sec:problematic_targets}). We then used this insight to analyze the more general case, where targets corresponds to sets of states.

\item \textbf{Strategy of Rejecting Hallucinations}: 
Inspired by the belief evaluation system in the brain, we proposed to learn a feasibility evaluator, which acts as a firewall to reject  targets hallucinated by the generative model in TAP agents. We carefully examined the challenges of learning such an evaluator and proposed a solution that is robust against feasibility delusions. This solution can be directly applied to existing TAP agents as an add-on without the need of changing the baseline RL methods. We systematically validated that our proposed solution leads to a significant reduction in delusional planning behaviors and improved empirical performance of various planning agents (Sec.~\ref{sec:delusions_exp}).
\end{itemize}

Having discussed the contributions to original knowledge made in this thesis, we now turn to a discussion of the limitations of the  work.

\section{Limitations}
DRL research is still significantly hindered by the sensitivity of existing methods. This thesis aimed to improve generalization in DRL, while simultaneously being shaped by the very challenges it sought to address.

Although the research in this thesis primarily focuses on improving the generalization of RL agents to make them more applicable in the real world, a common limitation of the works presented here is that the experiments are somewhat limited in scope. For the sake of experimental rigor, we concentrated on demonstrating our claims using agents with simple neural network architectures in minimalist, carefully controlled environments. These results would be more compelling and impactful if we could extend our experiments to widely-used simulated benchmark suites, such as ProcGen \citep{cobbe2019procgen}, or, ideally, real-world tasks. However, our efforts in this direction have been hindered by the fact that this research area is still under-explored, making it difficult to identify appropriate baseline methods and experimental settings to validate our ideas convincingly and that such experiments would require computational resources not at our disposal. The primary object of study in the thesis is inductive bias, which is always a tradeoff of effectiveness in some problems for ineffectiveness in others. While our proposed components / framework could be useful for dynamics-consistent zero-shot OOD generalization tasks, OOD challenges are way more vast than being dynamics-consistent: our methods will not be applicable to those distributional shifts with inconsistent / unknown local dynamics. These more difficult OOD challenges would likely require online fusion of learned skills and these possibilities expose areas for future study. Additionally, the thesis argues that decision-time planning is more suited to task generalization than background planning. However, it is also arguably more susceptible to model error or misspecification. I acknowledge this because we did not claim to have found a cure-all.

I will now discuss the limitations of each project in detail.

\subsection{Limitations of Work on Conscious Planning (Chap.~\ref{cha:CP})}
\label{sec:limitations_CP}

The conscious planning work presented in Chap.~\ref{cha:CP} served as a proof-of-concept of an interesting research direction: System-2 DRL. Despite its significant novelty and the usefulness of the core top-down semi-hard attention bottleneck, the approach has several limitations:

\begin{itemize}[leftmargin=*]
\item
Although, in this project, a form of spatial abstraction is achieved with each search step in planning, the proposed agent still plans over the most atomic timesteps. Thus, as the search depth increases, not only does the number of search nodes grow exponentially, but so does the accumulation of errors due to imperfections in the learned models \citep{janner2019trust}. This approach has limited long-term potential because even a comprehensive model that predicts the environment in great detail would struggle, given the search demands required for longer-term planning, which can be further exacerbated by stochasticity in the environment. Incorporating temporally abstract actions, such as options, could help mitigate this problem. While promising, introducing temporal abstraction into model-based RL is a non-trivial task that requires careful investigation. We committed to exploring this challenge in Skipper (Chap.~\ref{cha:skipper}). We also investigated how to deal with imperfect models and hallucinated state targets, and proposed a generic solution in Chap.~\ref{cha:delusions}.

\item
During experimentation, we found that the constant per-step replanning used by the agents imposed significant computational burdens. This suggests that constant replanning may be prohibitive in environments that require rapid reaction, particularly when using a computationally expensive set-based transition model. A more efficient planning strategy could involve controlling when and where the agent performs replanning, potentially by estimating uncertainty. We addressed this challenge in Chap.~\ref{cha:skipper}, where temporally abstract planning requires only sparse replanning, triggered either by a timeout or when a target checkpoint is achieved.

\item
The set-based dynamics model is not yet capable of learning stochastic dynamics. Since the environments in Sec.~\ref{sec:CP_experiments} are fully deterministic, we did not extend the design to account for stochasticity. The challenge lies in implementing a latent sample space for an end-to-end trainable set-to-set framework. My intuition is that this can be done via variational approaches \citep{kingma2013auto}.
\end{itemize}

Recent years, the newly developed methods of learning a discrete bottleneck has given me more inspirations on approaching this project differently, which I would leave for my future research.

\subsection{Limitations of Work on \Skipper{} (Chap.~\ref{cha:skipper})}
\label{sec:limitations_skipper}

Although the spatio-temporal abstractions used in \Skipper{} offer clear advantages and represent a step closer to human-like planning, \Skipper{} is still undeniably distant from the level of conscious planning exhibited by humans. In my view, there are two major limitations of the current \Skipper{} framework.

\begin{itemize}[leftmargin=*]
\item
\textbf{Planning without Task-Awareness}: While the planning process is spatially and temporally abstract, it is not task-informed. Specifically, future checkpoints are generated randomly by sampling from the partial description space. Despite post-processing steps such as pruning, these checkpoints do not prioritize the most predictable or important states, which are critical for forming a meaningful long-term plan. To plan as efficiently as humans, reasoning agents need to be able to decompose tasks into abstract, task-specific steps.

\item
\textbf{Planning with States}: The framework relies on proxy problems that operate at the state level. In contrast, human planning does not focus on detailed states but rather on high-level transitions or changes in state, often over partial or evolving states. We will discuss this point more in Sec.~\ref{sec:future_work}.
\end{itemize}

Additionally, there are several technical limitations in the proposed framework:

\begin{itemize}[leftmargin=*]
\item
\textbf{Continuous State Spaces \& Partial Observability}: The current implementation of \Skipper{} is designed for fully observable tasks with discrete state and action spaces. We adopted this minimalist approach to isolate challenges unrelated to the core idea of this work. \Skipper{} is naturally compatible with continuous action spaces, and the only modification required is replacing the baseline agent with one that supports continuous actions, such as TD3 \citep{fujimoto2018addressing}. However, when it comes to continuous state spaces, identifying when a checkpoint has been reached becomes more challenging. Specifically, the agent may never exactly fulfill a target checkpoint, which requires us to either use a distance metric to approximate state equivalence or to rely on the equivalence of partial descriptions (as implemented in the current version). Our current implementation sets the partial descriptions as bundles of binary variables, allowing for quick and straightforward comparison across any state space. However, we know that the overwhelming amount of states will pose a tremendous challenge for this design, because each instantiation of a partial description can correspond to a huge set of states. Regarding partial observability, although the current implementation does not incorporate a recurrent mechanism, the framework is compatible with such an approach. Handling partial observability would require augmenting the state encoder with recurrent or memory-based mechanisms, and ensuring the checkpoint generator works directly with the learned state representations. We acknowledge that future work is needed to evaluate \Skipper{}'s performance on popular partially observable benchmark suites, which would require incorporating components to handle partial observability and scaling up the architecture for greater expressive power.

\item
\textbf{More Intuitive Theory Boundary}: We do not know the precise boundaries of the  proxy problems theory, since it only indicates performance guarantees based on a condition of estimation accuracy, which in turn does not correspond trivially to a set of well-defined problems. We should explore, outside the scope of sparse-reward navigation, how this approach can be used to facilitate better generalization, and at the same time, try to find more powerful theoretical results that can guide us better.
\end{itemize}

\subsection{Limitations of Work on Rejecting Hallucinations (Chap.~\ref{cha:delusions})}
\label{sec:limitations_delusions}

The limitations of the work presented in Chap.~\ref{cha:delusions} primarily lie in the experiments. If possible, we would like to test our proposed feasibility evaluator on a wider range of TAP methods in more diverse  environments, which could demonstrate the benefits of addressing delusional planning behaviors more convincingly. Additionally, our experiments focused on gridworld environments, while we would like to verify the performance gain on environments with more complex features.

Some other planning agents propose ``targets'' that do not directly correspond to reaching sets of states, but instead,  maximize certain signals without verifying if they reached the proposed targets, \ie{}, without providing $h$ (defined in Sec.~\ref{sec:target_directed_framework}). We would like to investigate such agents in future work, to understand how they are impacted by hallucinations.

\section{Future Work}
\label{sec:future_work}

Abstractions are essential  for effective and efficient reasoning. This means that an agent must learn and reason at appropriate levels of detail, recognizing which aspects of the task are important and which are not, both spatially and temporally, as discussed in Chap.~\ref{cha:skipper}.

Humans can decompose complex tasks into abstract steps that are topologically sorted, encompassing spatio-temporal abstractions that are inherently task-aware. My goal is to enable RL agents to reason similarly.

From a traditional RL perspective, understanding how a computational agent could achieve abstract planning behavior is critical for studying decision-making, as it could help avoid the challenges associated with reasoning over excessive details, among other issues. In essence, this approach shifts the difficulty of learning accurate models to discovering crucial events, \ie{}, the robustness of the agent’s models can be improved if the abstract space in which the agent plans is well-constructed.

For future work, I aim to develop decision-making agents that make abstract plans based on partial changes in states, mimicking human reasoning. When planning a trip to Paris, for instance, experienced humans can easily break down the task into abstract steps without needing to reconstruct all the environmental details. A rough decomposition of the task might include abstract steps such as buying flight tickets, booking hotels, \etc{}. Each of these steps represents a partial change to the state, without the need to consider irrelevant details. For example, one does not need to think about what to have for dinner when planning these steps, and the states corresponding to completing each step can be infinite. The ability to discover and reason with partial information within an efficient abstract planning space allows humans to plan effectively for the future and generalize OOD.

In pursuit of this behavior, I propose that MBRL agents should reason in a space of abstract situations and events, focusing on sets of states defined by partial descriptions and their changes. This space should be discovered by the agent autonomously by trial-and-error.

For over a year now, I have been working on implementing this formulation, which includes abstract planning behaviors based on partial changes in states over extended periods of time. I am currently investigating how a generic credit-assignment mechanism could guide abstract planning.

\textbf{\textit{This is the end of the main parts of this thesis. Thank you very much for reading.}}

%% file: chapter_8_appendices.tex
\chapter{APPENDICES}\label{cha:appendix}

In this chapter, the assistive details of the thesis will be provided.

\section[Auxiliaries—CP]{Technical Auxiliaries for Chap.~\ref{cha:CP}}
\label{app:cp_aux}
\label{sec:CP_aux}

The source code of our experiments of this chapter can be found at \url{https://github.com/mila-iqia/conscious-planning}.

\subsection{Additional Experimental Insights}
\label{app:cp_insights}

\subsubsection{Integer Observations}
\label{app:cp_integer}
For MiniGrid environments \citep{chevalierboisvert2018minigrid}, the observations consist of integers encoding the object and the status of the grids. We found that for the UP models with these integer observations, the transformer layers are not sufficiently capable to capture the dynamics. Such problem can be resolved after increasing the depth of the FC layer depth by another hidden layer. This is one of the reasons why we prioritized on using CP models for the observation-level learning of \Dyna{}, \ie{}, CP models can handle integer features without deepening.

Similarly, we have tested the effect of increasing the depth of the linear transformations in SA layers. We did not observe significance in the enhancement of the performance, in terms of model learning or RL performance.

\subsubsection{Addressing Memorization with Noisy Shift}
\label{app:cp_memorization}
We discovered a generic trick to enforce better generalization based on our state-set encoding: if we use fixed integer-based positional tails which correspond to the absolute coordinates of the objects, we can add a global noise to all the $x$ and $y$ components in a set whenever one is encoded. By doing so, the coordinate systems would be randomly shifted every time the agent updates itself. Such shifts would prevent the agent from memorizing based on absolute positions. This trick could potentially enhance the agents' understanding of the dynamics even if in a classical static RL setting, under which the environments are fixed.

\subsection{Experiment Configurations}
\label{app:cp_configs}
The source code for this project is implemented with TensorFlow 2.x and open-source at \url{https://github.com/mila-iqia/Conscious-Planning}.

Multi-Processing: we implement a multiprocess configuration similar to that of Ape-X \cite{horgan2018distributed}, where $8$ explorers collect and sends batches of $64$ training transitions to the central buffer, with which the trainer trains. A pause signal is introduced when the trainer cannot consume fast enough \st{}, the uni-process and the multiprocess implementation have approximately the same performance, excluding the wall time.

Feature Extractor: We used the Bag-Of-Word (BOW) encoder suggested in \cite{hui2020babyai}. Since the experiments employ a fully-observable setting, we did not use frame stack. In MiniGrid-BabyAI environments, a grid is represented by three integers, and three trainable embeddings are created for the BOW representation. For each object (grid), each integer feature would be first independently transformed into embeddings, which are then mean-pooled to produce the final feature. The three embeddings are learnable and linear (with biases).

Stop criterion: Each run stops after $2.5 \times 10^{6}$ agent-environment interactions.

Replay Buffer: We used Prioritized Experience Replay (PER) of size $10^{6}$ \citep{schaul2016prioritized}, the same as in \cite{hessel2017rainbow}. We do not use the weights on the model updates, only the TD updates.

\textbf{Optimization}: We have used Adam \cite{kingma2014adam} with learning rate $2.5 \times 10^{-4}$ and epsilon $1.5\times {10}^{-4}$. The learning rate is the same as in \cite{mnih2015human}. Our tests show that using $6.25 \times 10^{-5}$, as suggested in \cite{hessel2017rainbow}, would be too slow. The batch size is the same for both value estimator training and model training, $64$. The training frequency is the same as in \cite{hessel2017rainbow}: every $4$ agent-environment interactions.

$\gamma$: Same as in \cite{hessel2017rainbow}. $0.99$.

In the experiments, we wanted functional architectures with minimal sizes for all the components. Thus, globally for the set-input architectures, we have limited the depth of the transformer layers to be $N=1$ wherever possible. The FC components are MLPs with $1$-hidden layer of width $64$. Exceptionally, we find that the effectiveness of the value estimator needs to be guaranteed with at least $3$-transformer layers. For the distributional output, while the value estimator has an output of $4$ atoms, the reward estimator has only $2$.

\textbf{Transformers}: For the SA sub-layers, we have used $8$ heads globally. For the FC sub-layers, we have used 2-layer MLP with $64$ hidden units globally. All the transformer related components have only one transformer layer except for that of the value estimator, which has three transformer layers before the pooling. We found that the shallower value estimators exhibit unstable training behaviors when used in the non-static settings.

\textbf{Set Representation}: The length of an object in the state set has length $32$, where the feature is of length $24$ and the positional embedding has length $8$. Note that the length of objects must be divisible by the number of heads in the attentions. The positional embeddings are trainable, however their initial values are constructed by the absolute $xy$ coordinates from each corner of the gridworld ($4 \times 2 = 8$). We found that without such initialization, the positional embedding would collapse.

\textbf{Action Embedding}: Actions are embedded as one-hot vectors with length $8$.

\textbf{Planning Steps}: for each planning session, the maximum number of simulations based on the learned transition model is 5.

\textbf{Exploration}: $\epsilon$ takes value from a linear schedule that decreases from $0.95$ to $0.01$ in the course of $10^{6}$ agent-environment interactions, same as in \cite{hessel2017rainbow}. For evaluation, $\epsilon$ is fixed to be $10^{-3}$.

\textbf{Distributional Outputs}: We have used distributional outputs \cite{bellemare2017distributional} for the reward and value estimators. $2$ atoms for reward estimation (mapping the interval of $[0, 1]$) and $4$ atoms for value estimation (mapping the interval of $[0, 1]$).

\textbf{Regularization}: We find that layer norm is crucial to guarantee the reproducibility of the performance with set-representations. We apply layer normalization \cite{ba2016layer} in the sub-layers of transformers as well as at the end of the encoder and model dynamics outputs. This applies for the NOSET baseline as well.

\textbf{modelfree baseline}: We did not use the full Rainbow agent \cite{hessel2017rainbow} as the baseline, because we want to keep our agent as minimalist as possible. The agent does not need the dueling head and the noisy net components to perform well, according to our preliminary ablation tests.

\section[Auxiliaries—Skipper]{Technical Auxiliaries for Chap.~\ref{cha:skipper}}
\label{app:skipper_aux}
\label{sec:skipper_aux}

The source code of our experiments of this chapter can be found at \url{https://github.com/mila-iqia/skipper}.

\subsection{\texorpdfstring{\Skipper{}}{Skipper}}
\label{sec:skipper_exp_details}

\subsubsection{Training}
The agent is based on a distributional prioritized double DQN. All the trainable parameters are optimized with Adam at a rate of $2.5\times10^{-4}$ \citep{kingma2014adam}, with a gradient clipping by value (maximum absolute value $1.0$). The priorities for experience replay sampling are equal to the per-sample training loss.

\subsubsection{Full State Encoder}
The full-state encoder is a $2$-layer residual block (with kernel size 3 and doubled intermediate channels) combined with the $16$-dimensional bag-of-words embedder of BabyAI \citep{hui2020babyai}.

\subsubsection{Partial State Selector (Spatial Abstraction)}
The selector $\sigma$ is implemented with single-head (not multihead, thus the output linear transformation of the default multihead attention implementation in PyTorch is disabled.) top-$4$ attention, with each local perceptive field of size $8\times8$ cells. Layer normalization \citep{ba2016layer} is used before and after the spatial abstraction.

\subsubsection{Estimators}
The estimators, which operate on the partial states, are $3$-layered MLPs with $256$ hidden units.

An additional estimator for termination is learned, which instead of taking a pair of partial states as input, takes only one, and is learned to classify terminal states with cross-entropy loss. The estimated distance from terminal states to other states would be overwritten with $\infty$. The internal $\gamma$ for intrinsic reward of \red{$\pi$} is $0.95$, while the task $\gamma$ is $0.99$

The estimators use C51 distributional TD learning \citep{dabney2018distributional}. That is, the estimators output histograms (\softmax{} over vector outputs) instead of scalars. We regress the histogram towards the targets, where these targets are skewed histograms of scalar values, towards which KL-divergence is used to train. At the output, there are $16$ bins for each histogram estimation (value for policy, reward, distance).




\subsubsection{Checkpoint Generator}
Although \Skipper{} is designed to have the generator work on state level, that is, it should take learned state representations as inputs and have state representations as outputs, in our experiments, the generator actually operates on observation inputs and outputs. This is because of the preferred compactness of the observations and the equivalence to full states under full observability in our experiments.

The context extractor $\scriptE_c$ is a $32$-dimensional BabyAI BOW embedder. It encodes an input observation into a representation of the episodic context.

The partial description extractor $\scriptE_z$ is made of a $32$-dimensional BabyAI BOW embedder, followed by $3$ aforementioned residual blocks with $3\times3$ convolutions (doubling the feature dimension every time) in between, ended by global maxpool and a final linear projection to the latent weights. The partial descriptions are bundles of $6$ binary latents, which could represent at most $64$ ``kinds'' of checkpoints. Inspired by VQ-VAE \citep{van2017neural}, we use the argmax of the latent weights as partial descriptions, instead of sampling according to the \softmax{}-ed weights. This enables easy comparison of current state to the checkpoints in the partial description space, because each state deterministically corresponds to one partial description. We identify reaching a target checkpoint if the partial description of the current state matches that of the target.

The fusing function first projects linearly the partial descriptions to a $128$-dimensional space and then uses deconvolution to recover an output which shares the same size as the encoded context. Finally, a residual block is used, followed by a final $1x1$ convolution that downscales the concatenation of context together with the deconv'ed partial description into a 2D weight map. The agent's location is taken to be the {\fontfamily{qcr}\selectfont argmax} of this weight map.

The whole checkpoint generator is trained end-to-end with a standard VAE loss. That is the sum of a KL-divergence for the agent's location, and the entropy of partial descriptions, weighted by $2.5 \times 10^{-4}$, as suggested in \url{https://github.com/AntixK/PyTorch-VAE}. Note that the per-sample losses in the batches are not weighted for training according to priority from the experience replay.

We want to mention that if one does not want to generate non-goal terminal states as checkpoints, we could also seek to train on reversed $\langle S^\odot, S_t \rangle$ pairs. In this case, the checkpoints to reconstruct will never be terminal.

\subsubsection{HER}
Each experienced transition is further duplicated into $4$ hindsight transitions at the end of each episode. Each of these transitions is combined with a randomly sampled observation from the same trajectory as the relabelled ``goal''. The size of the hindsight buffer is extended to $4$ times that of the baseline that does not learn from hindsight accordingly, that is, $4\times 10^{6}$.

\subsubsection{Planning}
As introduced, we use value iteration over options \citep{sutton1999between} to plan over the proxy problem represented as an SMDP. We use the matrix form $Q = R_{S \times S} + \Gamma V$, where $R$ and $\Gamma$ are the estimated edge matrices for cumulative rewards, respectively. Note that this notation is different from the ones we used in the manuscript. The checkpoint value $V$, initialized as all-zero, is taken on the maximum of $Q$ along the checkpoint target (the actions for $\mu$) dimension. When planning is initiated during decision time, the value iteration step is called $5$ times. We do not run until convergence, since with low-quality estimates during the early stages of the learning, this would be a waste of time. The edges from the current state towards other states are always set to be one-directional, and the self-loops are also removed. This means the first column as well as the diagonal elements of $R$ and $\Gamma$ are all zeros. Besides pruning edges based on the distance threshold, as introduced in the main paper, the terminal estimator is also used to prune the matrices $R$ and $\Gamma$: the rows corresponding to the terminal states are all zeros.

The only difference between the two variants, \ie{}, \Skipper{}-once and \Skipper{}-regen is that the latter variant would discard the previously constructed proxy problem and construct a new one every time the planning is triggered. This introduces more computational effort while lowering the chance that the agent gets ``trapped'' in a bad proxy problem that cannot form effective plans to achieve the goal. If such a situation occurs with \Skipper{}-regen, as long as the agent does not terminate the episode prematurely, a new proxy problem will be generated to hopefully address the issue. Empirically, as we have demonstrated in the experiments, such variant in the planning behavior results in generally significant improvements in terms of generalization abilities at the cost of extra computation.

\subsubsection{Hyperparameter Tuning}
Some hyperparameters introduced by \Skipper{} can be located in the pseudocode in Alg.~\ref{alg:skipper}.

\paragraph{Timeout and Pruning Threshold} 
Intuitively, we tied the timeout to be equal to the distance pruning threshold. The timeout kicks in when the agent thinks a checkpoint can be achieved within \eg{}, $8$ steps, but already spent $8$ steps yet still could not achieve it. 

This leads to how we tuned the pruning (distance) threshold: we fully used the advantage of our experiments on DP-solvable tasks: with a snapshot of the agent during its training, we can sample many $\langle$ starting state, target state $\rangle$ pairs and calculate the ground truth distance between the pair, as well as the failure rate of reaching from the starting state to the target state given the current policy $\pi$, then plot them as the $x$ and $y$ values respectively for visualization. We found such curves to evolve from high failure rate at the beginning, to a monotonically increasing curve, where at small true distances, the failure rates are near zero. We picked $8$ because the curve starts to grow explosively when the true distances are more than $9$. 

\paragraph{$k$ for $k$-medoids}
We tuned this by running a sensitivity analysis on \Skipper{} agents with different $k$’s, whose results are presented previously in this Appendix.

Additionally, we prune from $32$ checkpoints because $32$ checkpoints could achieve (visually) a good coverage of the state space as well as its friendliness to NVIDIA accelerators. 

\paragraph{Size of local Perception Field}
We used a local perception field of size 8 because our baseline model-free agent would be able to solve and generalize well within $8\times8$ tasks, but not larger. Roughly speaking, our spatial abstraction breaks down the overall tasks into $8\times8$ sub-tasks, which the policy could comfortably solve.

\paragraph{Model-free Baseline Architecture}
The baseline architecture (distributional, Double DQN) was heavily influenced by the architecture used in the previous work \citep{zhao2021consciousness}, which demonstrated success on similar but smaller-scale experiments ($8\times8$). The difference is that while then we used computationally heavy components such as transformer layers on a set-based representation, we replaced them with a simpler and effective local perception component. We validated our model-free baseline performance on the tasks proposed in \citet{zhao2021consciousness}.

\subsection{\texorpdfstring{\LEAP{}}{LEAP}}
\label{sec:leap_exp_details}

\subsubsection{Adaptation for Discrete Action Spaces}
The \LEAP{} baseline has been implemented from scratch for our experiments, since the original open-sourced implementation\footnote{\url{https://github.com/snasiriany/leap}} was not compatible with environments with discrete action spaces. \LEAP{}'s training involves two pretraining stages, that are, generator pretraining and distance estimator pretraining, which were originally named the VAE and RL pretrainings. Despite our best effort, that is to be covered in detail, we found that \LEAP{} was unable to get a reasonable performance in its original form after rebasing it on a discrete model-free RL baseline.

\subsubsection{Replacing the Model}
We tried to identify the reasons why the generalization performance of the adapted \LEAP{} was unsatisfactory: we found that the original VAE used in \LEAP{} is not capable to handle even few training tasks, let alone generalize well to the evaluation tasks. Even by combining the idea of the context / partial description split (still with continuous latents), during decision time, the planning results given by the evolutionary algorithm (Cross Entropy Method, CEM, \citet{rubinstein1997optimization}) almost always produce delusional plans that are catastrophic in terms of performance. This was why we switched into \LEAP{} the same conditional generator we proposed in the paper, and adapted CEM accordingly, due to the change from continuous latents to discrete.

We also did not find that using the pretrained VAE representation as the state representation during the second stage helped the agent's performance, as the paper claimed. In fact, the adapted \LEAP{} variant could only achieve decent performance after learning a state representation from scratch in the RL pretraining phase. Adopting \Skipper{}'s splitting generator also disables such choice.

\subsubsection{Replacing TDM}
The original distance estimator based on Temporal Difference Models (TDM) also does not show capable performance in estimating the length of trajectories, even with the help of a ground truth distance function (calculated with DP). Therefore, we switched to learning the distance estimates with our proposed method. Our distance estimator is not sensitive to the sub-goal time budget as TDM and is hence more versatile in environments like that was used in the main paper, where the trajectory length of each checkpoint transition could highly vary. Like for \Skipper{}, an additional terminal estimator has been learned to make \LEAP{} planning compatible with the terminal lava states. Note that this \LEAP{} variant was trained on the same sampling scheme with HER as in \Skipper{}.

The introduced distance estimator, as well as the accompanying full-state encoder, are of the same architecture, hyperparameters, and training method as those used in\Skipper{}. The number of intermediate subgoals for \LEAP{} planning is tuned to be $3$, which close to how many intermediate checkpoints \Skipper{} typically needs to reach before finishing the tasks. The CEM is called with $5$ iterations for each plan construction, with a population size of $128$ and an elite population of size $16$. We found no significant improvement in enlarging the search budget other than additional wall time. The new initialization of the new population is by sampling a $\epsilon$-mean of the elite population (the binary partial descriptions), where $\epsilon = 0.01$ to prevent the loss of diversity. Because of the very expensive cost of using CEM at decision time and its low return of investment in terms of generalization performance, during the RL pretraining phase, the agent performs random walks over uniformly random initial states to collect experience.

\subsection{\texorpdfstring{\Director{}}{Director}}
\label{sec:director_exp_details}

\begin{SCfigure}[][htbp]
\includegraphics[width=0.25\textwidth]{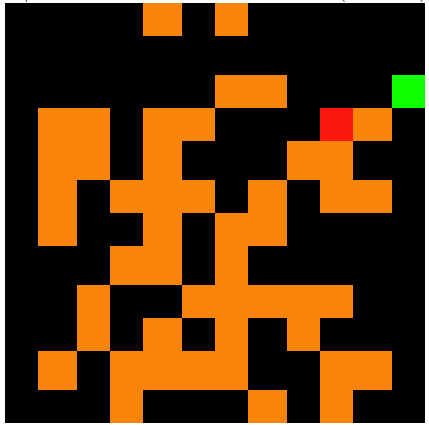}
\caption[An Example for Simplified Observations for \Director{}]{\textbf{An Example for Simplified Observations for \Director{}}}
  \label{fig:rds_director}
\end{SCfigure}

\subsubsection{Adaptation}
We based our experiments of Director \citep{hafner2022deep} on the publicly available code (\url{https://github.com/danijar/director}) released by the authors. Except for a few changes in the parameters, which are depicted in Tab. \ref{tab:director_configs}, we have used the default configuration provided for Atari environments. Note that as the Director version in which the worker receives no task rewards performed worse in our tasks, we have used the version in which the worker receives scaled task rewards (referred to as ``Director (worker task reward)'' in \citet{hafner2022deep}). This agent has also been shown to perform better across various domains in \citet{hafner2022deep}.

\paragraph{Encoder} Unlike \Skipper{} and \LEAP{} agents, the Director agent receives as input a simplified RGB image of the current state of the environment (see Fig.~\ref{fig:rds_director}). This is because we found that Director performed better with its original architecture, which was designed for image-based observations. We removed all textures to simplify the RGB observations.

\begin{table}[htbp]
    \centering
    \caption[Changed Hyperparameters of \Director{}]{\textbf{Changed Hyperparameters of \Director{}}}
    
    \begin{tabular}{|p{4cm}|p{5cm}|}
    \hline
    Parameter & Value \\
    \hline
    replay\_size & 2M \\
    replay\_chunk & 12 \\
    imag\_horizon & 8 \\
    env\_skill\_duration & 4 \\
    train\_skill\_duration & 4 \\
    worker\_rews & \{extr: 0.5, expl: 0.0, goal: 1.0\} \\
    sticky & False \\ 
    gray & False \\
    \hline
    \end{tabular}
    \label{tab:director_configs}
\end{table}

\subsubsection{Failure Modes: Bad Generalization, Sensitive to Short Trajectories}

\begin{figure}[htbp]
\centering
\subfloat[training, $\delta = 0.4$]{
\captionsetup{justification = centering}
\includegraphics[height=0.22\textwidth]{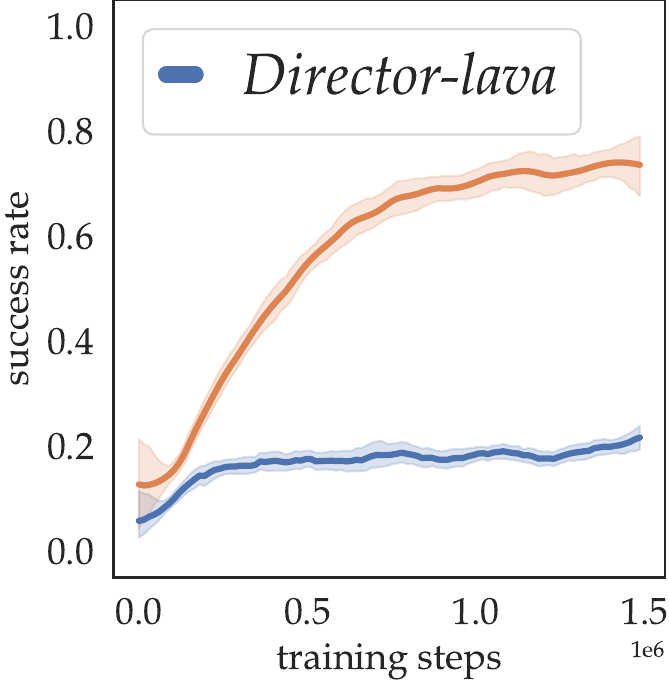}}
\hfill
\subfloat[OOD, $\delta = 0.25$]{
\captionsetup{justification = centering}
\includegraphics[height=0.22\textwidth]{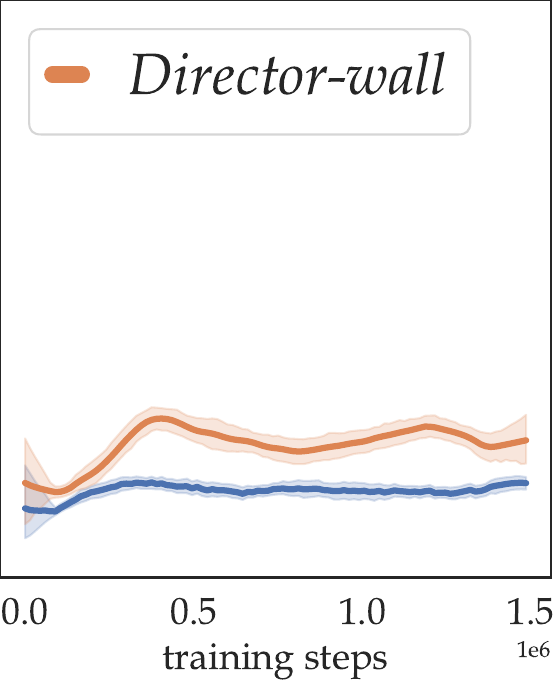}}
\hfill
\subfloat[OOD, $\delta = 0.35$]{
\captionsetup{justification = centering}
\includegraphics[height=0.22\textwidth]{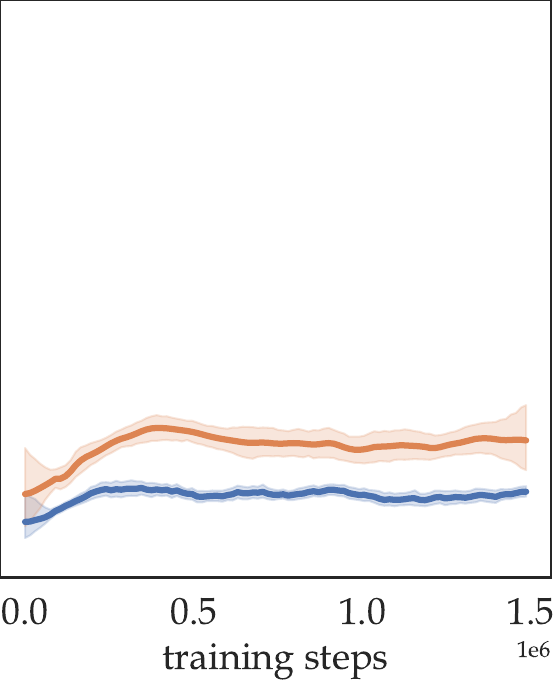}}
\hfill
\subfloat[OOD, $\delta = 0.45$]{
\captionsetup{justification = centering}
\includegraphics[height=0.22\textwidth]{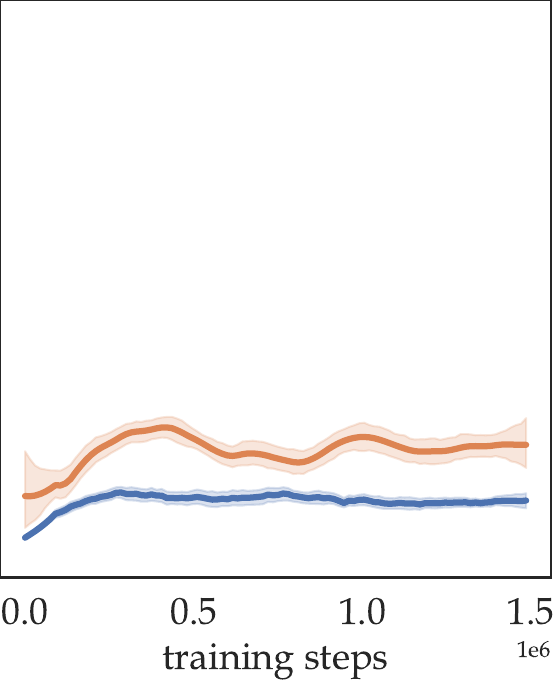}}
\hfill
\subfloat[OOD, $\delta = 0.55$]{
\captionsetup{justification = centering}
\includegraphics[height=0.22\textwidth]{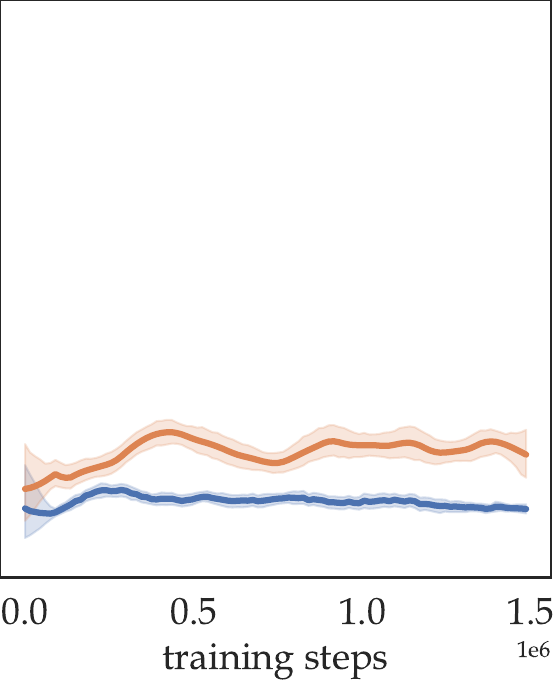}}

\caption[Results of Director on Tasks with Lavas \vs{} on Tasks with Walls]{\textbf{Results of Director on Tasks with Lavas \vs{} on Tasks with Walls}: the results are obtained with $50$ training tasks. The results for Director-lava (same as in the main paper) are obtained from $20$ independent seed runs.}
\label{fig:50_envs_director_wall}
\end{figure}

\paragraph{Training Performance} We investigated why Director is unable to achieve good training performance (Fig.~\ref{fig:50_envs}). 
As Director was designed to be trained solely on environments where it is able to collect long trajectories to train a good enough recurrent world model \citep{hafner2022deep}, we hypothesized that Director may perform better in domains where it is able to interact with the environment through longer trajectories by having better recurrent world models (\ie{}, the agent does not immediately die as a result of interacting with specific objects in the environment). To test this, we experimented with variants of the used tasks, where the lava cells are replaced with wall cells, so the agent does not die upon trying to move towards them (we refer to this environment as the ``walled'' environment). The corresponding results on $50$ training tasks are depicted in Fig.~\ref{fig:50_envs_director_wall}. As can be seen, the Director agent indeed performs better within the training tasks than in the environments with lava. 

\begin{figure}[htbp]
\centering
\subfloat[training, $\delta = 0.4$]{
\captionsetup{justification = centering}
\includegraphics[height=0.22\textwidth]{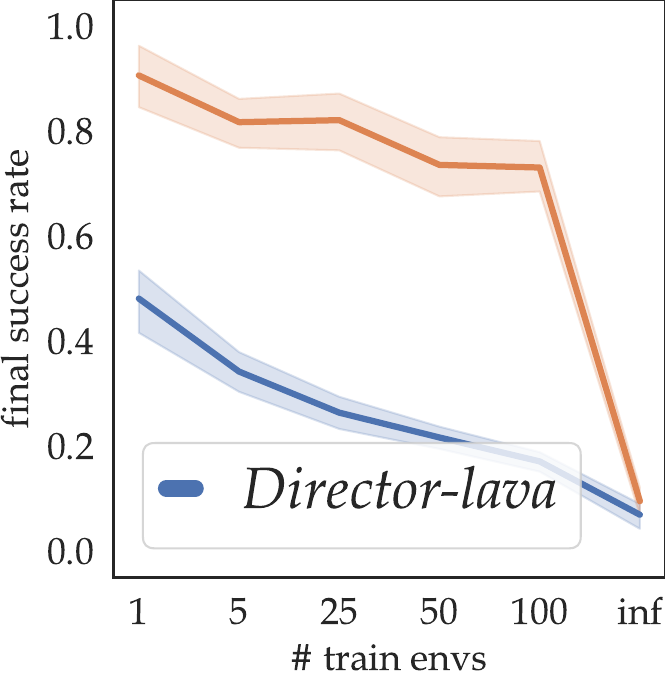}}
\hfill
\subfloat[OOD, $\delta = 0.25$]{
\captionsetup{justification = centering}
\includegraphics[height=0.22\textwidth]{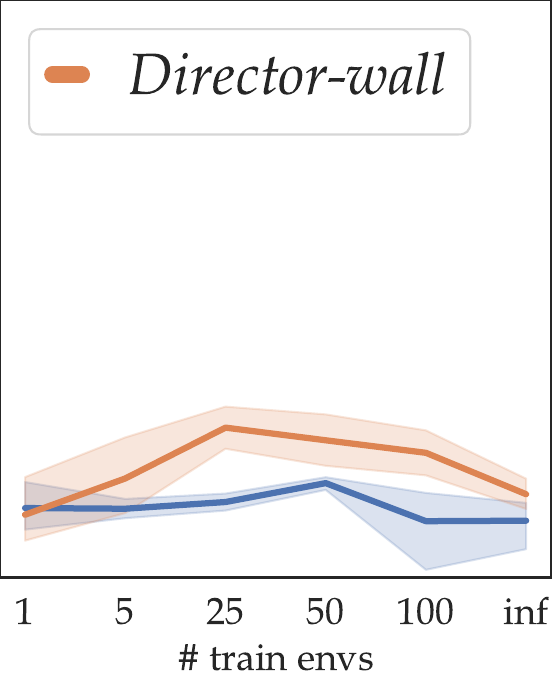}}
\hfill
\subfloat[OOD, $\delta = 0.35$]{
\captionsetup{justification = centering}
\includegraphics[height=0.22\textwidth]{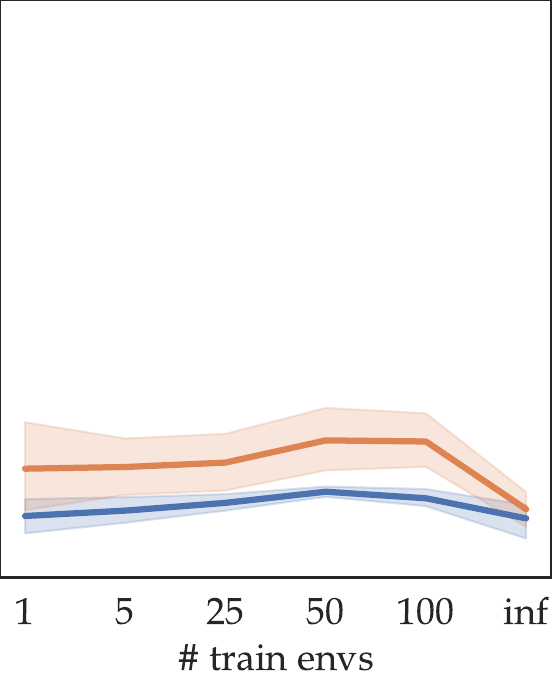}}
\hfill
\subfloat[OOD, $\delta = 0.45$]{
\captionsetup{justification = centering}
\includegraphics[height=0.22\textwidth]{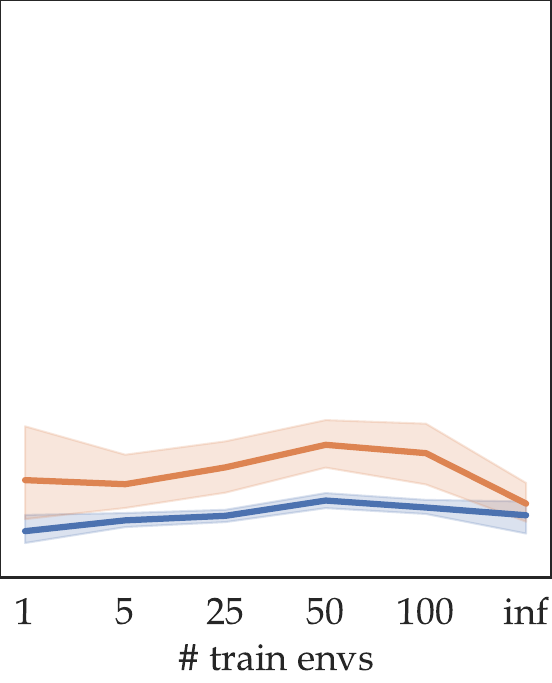}}
\hfill
\subfloat[OOD, $\delta = 0.55$]{
\captionsetup{justification = centering}
\includegraphics[height=0.22\textwidth]{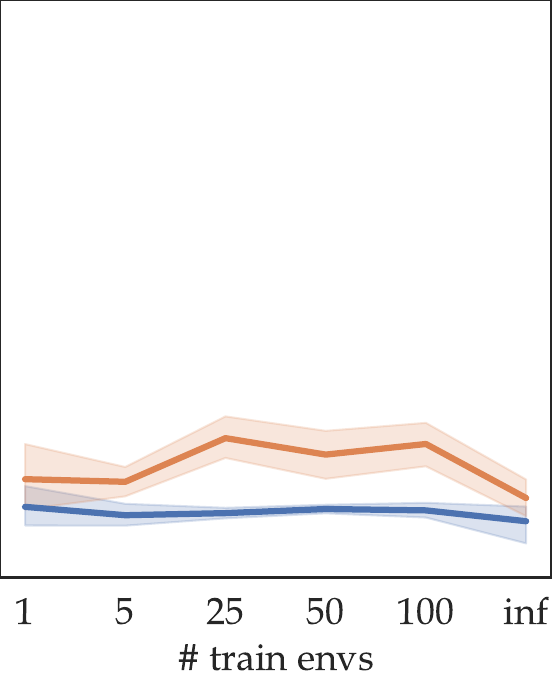}}

\caption[Generalization Performance of \Director{} on Different Numbers of ``Walled'' Training Tasks]{\textbf{Generalization Performance of Agents on Different Numbers of Training Tasks (while \Director{} runs on the walled environments)}: besides \Director{}, each data point and corresponding error bar (95\% confidence interval) are processed from the final performance from $20$ independent seed runs. \Director{}-wall's results are obtained from $20$ runs.}
\label{fig:num_envs_all_director_wall}
\end{figure}

\paragraph{Generalization Performance} We also investigated why Director is unable to achieve good generalization (Fig.~\ref{fig:50_envs}). As Director trains its policies solely from the imagined trajectories predicted by its learned world model, we believe that the low generalization performance is due to Director being unable to learn a good enough world model that generalizes to the evaluation tasks. The generalization performances in both the ``walled'' and regular environments, depicted in Fig.~\ref{fig:50_envs_director_wall}, indeed support this argument. Similar to what we did in the main paper, we also present experimental results for how the generalization performance changes with the number of training environments. Results in Fig.~\ref{fig:num_envs_all_director_wall} show that the number of training environments has little effect on its poor generalization performance.

\section[Auxiliaries—Delusions]{Technical Auxiliaries for Chap.~\ref{cha:delusions}}
\label{app:delusions_aux}

The source code of our experiments of this chapter can be found at \url{https://github.com/mila-iqia/delusions}.

\subsection{Implementation of \pertaskstr{}}
\pertaskstr{} takes the advantage of the fact that training is done on limited number of fixed task instances. We give each task a unique task identifier. At relabeling time, \pertaskstr{} samples observations among all the transitions marked with the same identifier as the current training task instance. This can be trivially implemented with individual auxiliary experience replays that store only the experienced states with memory-efficient pointers to the buffered $x_t$'s in the main HER.

\subsection{Implementation Details for Experiments}
\label{sec:details_implement}

\subsubsection{\Skipper{}}
\label{sec:skipper_exp_details_delusions}

Our adaptation of \Skipper{} over the original implementation\footnote{\url{https://github.com/mila-iqia/Skipper}} in \citet{zhao2024consciousness} is minimal. We have additionally added two simple vertex pruning procedures before the vertex pruning based on $k$-medoids. These two procedures include: 1) prune vertices that are duplicated, and 2) prune vertices that cannot be reached from the current state with the estimated connectivity.

We implemented a version of generator that can reliably handle both \RDS{} and \SSM{} with the same architecture. Please consult {\fontfamily{qcr}\selectfont{models.py}} in the submitted source code for its detailed architecture.

For \SSM{} instances, since the state spaces are $4$-times bigger than those of \RDS{}, we ask that \Skipper{} generate twice the number of candidates (both before and after pruning) for the proxy problems.

All other architectures and hyperparameters are identical to the original implementation.

For better adaptability during evaluation and faster training, \Skipper{} variants in Chap.~\ref{cha:skipper} keeps the constructed proxy problem for the whole episode during training and replanning only triggers a re-selection, while during evaluation, the proxy problems are always erased and re-constructed.

The quality of our adaptation of the original implementation can be assured by the fact the \FE{} variant's performance matches the original on \RDS{}.

\subsubsection{\LEAP{}}
\label{sec:leap_exp_details_delusions}

\LEAP{}'s training involves two pretraining stages, that are, generator pretraining and distance estimator training.

We improved upon the adopted discrete-action space compatible implementation of \LEAP{} \citep{nasiriany2019planning} from \citet{zhao2024consciousness}. We gave \LEAP{} additional flexibility to use fewer subgoals along the way to the task goal if necessary. Also, we improved upon the Cross-Entropy Method (CEM), such that elite sequences would be kept intact in the next population during the optimization process. We increased the base population size of each generation to $512$ and lengthened the number of iterations to $10$.

For \RDS{} $12 \times 12$ and \SSM{} $8 \times 8$, at most $3$ subgoals are used in each planned path. We find that employing more subgoals greatly increases the burden of CEM and lower the quality of the evolved subgoal sequences, leading to bad performance that cannot be effectively analyzed.

We used the same generator architecture and hyperparameters as in \Skipper{}. All other architectures and hyperparameters remain unchanged.

Similarly for \LEAP{}, for better adaptability during evaluation, the planned sequences of subgoals are always reconstructed whenever planning is triggered. While in training, the sequence is reused and only a subgoal selection is conducted.

The quality of our adaptation of the original implementation can be assured by the fact the \FE{} variant's performance matches the original on \RDS{}.

\subsection{Applying the Evaluator on \Dreamer{}v2} 
\label{sec:appendix_dreamer}

To demonstrate that our approach functions effectively in more generalist settings, such as those with continuous state and action spaces and partial observability, and to illustrate its application to a modern TAP agent, we integrated our proposed evaluator into \Dreamer{}v2 \citep{hafner2020mastering}. The evaluator filters out potentially delusional values from infeasible states that might distort the $\lambda$-returns derived from imagined trajectories. Given the technical complexity ahead, we suggest readers familiarize themselves with \Dreamer{}v2 before continuing \citep{hafner2020mastering}.

Although \Dreamer{}v2's stochastic states are discrete and could theoretically support similarity assessments, their design ensures they rarely repeat due to numerous possibilities, making them too random for our similarity function $h$. Consequently, we rely on the deterministic state representations $\bm{s}$, which also prompt us to more thought-provoking discussions.

\Dreamer{}v2 operates as a \Dyna{}-like method, employing fixed-horizon rollouts with autoregressively imagined states as targets. Lacking a built-in similarity function $h$, it provides an opportunity to showcase how we construct $h$ in our approach. Our method incorporates various realism aspects to assess state similarity between the next state and the target state, influencing the branching in Eq.~\ref{eq:rule_feasibility_gamma} during evaluator updates \citep{russell2025gaia2}.

\subsubsection{How to craft $h$: Observational Realism}
Observational realism, \ie{}, the similarity in terms of state representations is the first obvious criteria for $h$.

Theoretically, one might simplistically assume state equivalence by defining an $\epsilon$-ball around the target state. However, in practice, an $\epsilon$-ball based on $L_2$ distances proves inadequate due to varying representation scales. Instead, we employ Mahalanobis distances, which better accommodate the representations' distributional variations.

To be more precise, we use an Exponential Moving Average (EMA) of the covariance of concatenated current-next deterministic state pairs $[\bm{s}_t, \bm{s}_{t+1}]$ to calculate the Mahalanobis distances between the next state pair $[\bm{s}_t, \bm{s}_{t+1}]$ and target state pair $[\bm{s}_t, \hat{\bm{s}}_{t+1}]$.

\subsubsection{How to craft $h$: Behavioral Realism}
The second focus is behavioral realism: does the agent exhibit similar behavior (\eg{}, in value, reward, and discount estimations) across the states ($\bm{s}_{t+1}$ \& $\hat{\bm{s}}_{t+1}$)?

Here, we apply Mahalanobis distances to pairs of current and future values, rewards, and discounts, ensuring the states appear similar from the agent's perspective.

Caution is required with action-realism. Naively applying our method to one-hot encoded discrete actions could result in a singular covariance matrix for the $\epsilon$-ball computation.

Preliminary Atari experiments suggest setting distinct $\epsilon$ values for different components—state representations, value estimations, reward estimations, and discount estimations.

\subsubsection{How to Relabel: Just-In-Time (JIT) Construction}

Since \Dreamer{}v2 samples sub-trajectories and computes state representations autoregressively, we forgo a separate HER for storing source-target pairs, opting instead for Just-In-Time (JIT) construction. Designed for single-environment training and evaluation, \Dreamer{}v2 allows us to implement a \FEG{} variant on agent-sampled sub-trajectories. Initial tests indicate a balanced mix of \episodestr{} (within sub-trajectories) and \generatestr{} performs effectively.

\subsubsection{How to Reject: Three Criteria for $\lambda$-returns}

\Dreamer{}v2 leverages its model to imagine future states and values, using these, along with intermediate rewards and discounts, to compute $\lambda$-returns for each origin state.

For such strategy, we implemented the following $3$ criteria for rejecting the imagined states:

\begin{enumerate}[leftmargin=*]
\item \textbf{Transition-wise Rejection}: If a next state seems unlikely to follow from the current state, its value is deemed untrustworthy. This process is repeated for all imagined transitions. Notably, a state rejected as infeasible in one transition might still be reachable elsewhere, so subsequent states are not automatically discarded.

\item \textbf{Point-to-Point (P2P) Rejection for Targets}: Starting from a replay-sampled base state, we assess whether each imagined state is reachable, regardless of steps taken. This counters hallucinated targets from accumulated errors over the imagination horizon \citep{talvitie2017self}, excluding such states from value estimation targets.

\item \textbf{P2P Rejection for Current States}: Entirely unreachable states are excluded as current states in multi-step value updates, though subsequent states may remain viable.
\end{enumerate}

\begin{figure*}[htbp]
\centering
\captionsetup{justification = centering}
\includegraphics[width=1.00\textwidth]{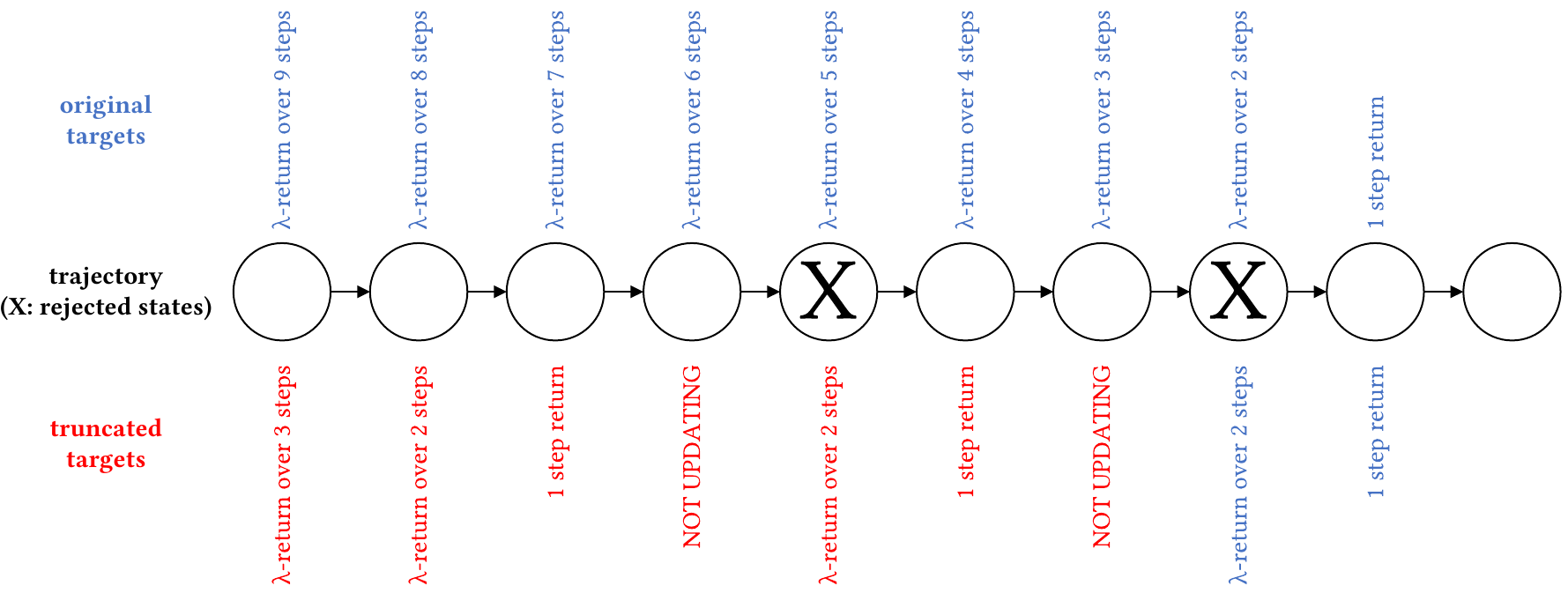}
\caption[Truncated $\lambda$-Returns with Rejected States for \Dreamer{}]{\textbf{Truncated $\lambda$-Returns with Rejected States for \Dreamer{}}: the original $\lambda$-returns are illustrated in the top row, while the truncated returns are illustrated in the bottom row with the differences marked in \textcolor{red}{red}. Our strategy ensures that the critic targets in the trajectories can be maximally preserved for updates. The states right before the rejected ones will have no trustworthy critic targets and are thus not updated. Starting from the last rejected state, all critic targets remain the same as the originals.
}
\label{fig:truncated_lambda_returns}
\end{figure*}

The first two criteria yield a binary mask to truncate $\lambda$-returns in sampled sub-trajectories, excluding untrustworthy values while preserving horizons for reliable ones. Our repository offers an efficient implementation, maintaining the complexity as the original, un-truncated $\lambda$-returns. The behavior of the (critic) target-based rejection is presented in Fig.~\ref{fig:truncated_lambda_returns}.

The third criterion masks updates to wholly infeasible imagined states. By examining the rejection rate by the horizon index, the evaluator can also be used to understand how long the imagined trajectories are likely to be trustworthy and thus adjust the associated hyperparameters.

We developed a standalone, user-friendly evaluator (implemented in PyTorch) that integrates seamlessly into TAP agents like \Dreamer{}v2, employing its own optimizer and target networks for robust learning when activated. Please check \codeword{evaluator.py} in the source code repository.

We tuned the hyperparameters using the \codeword{Atari} environments and found that both the autoregressive estimations of distances and the P2P distances (towards the target states) in the sampled and imagined trajectories roughly converge to the estimated ground truth values, which are deduced from their time indices. This is the best we can do for environments without ground truth access.

Regrettably, our Atari100k preliminary results with $10^5$ interactions show negligible performance gains over the baseline \citep{kaiser2019model}. This is likely because the state representations of \Dreamer{} usually takes a significant portion of training to stabilize and for the evaluator to adapt to. The differences are expected to show with prolonged experiments where a significant number of updates will be made after the state representations stabilize. Limited computational resources prevented our extended experiments, and we invite those with greater capacity to investigate further.

Our \Dreamer{} implementations can be found in the source code repository.